\pdfoutput=1 
\documentclass{article}

    \usepackage[final,nonatbib]{neurips_2022}

\usepackage[utf8]{inputenc} 
\usepackage[T1]{fontenc}    
\usepackage{hyperref}       
\usepackage{url}            
\usepackage{booktabs}       
\usepackage{amsfonts}       
\usepackage{nicefrac}       
\usepackage{microtype}      
\usepackage{xcolor}         

\usepackage{todonotes}
\usepackage{import}
\usepackage{mathrsfs}
\usepackage{amsmath}
\usepackage{float}
\usepackage{caption}
\usepackage{subcaption}
\usepackage{dblfloatfix}
\usepackage{algorithm}
\usepackage{algpseudocode}
\usepackage{multicol}
\usepackage{wrapfig}

\usepackage{amsmath,amsthm,amssymb,amsfonts}
\usepackage{stmaryrd}
\usepackage{mathtools,marvosym}
\usepackage{soul}
\usepackage{xparse, xspace}
\usepackage{MnSymbol}
\usepackage[shortlabels]{enumitem}
\usepackage{tikz-cd}
\usepackage{tikz}
\usepackage{wasysym}
\usepackage{pst-node, multido}
\usetikzlibrary{automata, positioning, arrows, trees, decorations.pathreplacing, tikzmark}
\usepackage{float}
\usepackage{graphicx}

\setlist{topsep=0pt}

\makeatletter
\renewcommand{\subset}{\subseteq}

\DeclareMathOperator*{\dprime}{\prime \prime}

\DeclareMathOperator*{\argmin}{arg\,min}

\newcommand{\N}{\mathbb{N}}

\newcommand{\R}{\mathbb{R}}

\newcommand{\tspace}{\allowbreak\ }
\NewDocumentCommand\given{s}{
    \IfBooleanTF{#1}{
        \,\middle|\,
    }{
        \,|\,
    }
}

\NewDocumentCommand\p{s m}{
    \IfBooleanTF{#1}{
        \left [ #2 \right ]
    }{
        \left ( #2 \right )
    }
}

\newcommand{\createwordcommand}[2]{
    \NewDocumentCommand{#1}{s s s}{\IfBooleanTF{##1}{
            \IfBooleanTF{##2}{
                \IfBooleanTF{##3}
                {#2}{#2\tspace}
            }{\tspace#2}
        }{\tspace#2\tspace}
    }
}

\NewDocumentCommand\prob{s m}{
    \IfBooleanTF{#1}{
        P\left(#2\right)
    }{
        P(#2)
    }
}

\newcommand\Dist{\mathrm{Dist}}

\newtheorem{thm}{Theorem}

\theoremstyle{definition}
\newtheorem{definition}{Definition}

\newtheorem*{claim}{Claim}

\theoremstyle{remark}

\DeclareMathAlphabet{\mathpzc}{OT1}{pzc}{m}{it}

\newcommand{\rebuttal}[1]{#1}

\newcommand{\x}{\times}

\usepackage{xr}
\makeatletter
\newcommand*{\addFileDependency}[1]{
  \typeout{(#1)}
  \@addtofilelist{#1}
  \IfFileExists{#1}{}{\typeout{No file #1.}}
}
\makeatother

\title{Continuous MDP Homomorphisms and\\ Homomorphic Policy Gradient}

\author{%
  Sahand Rezaei-Shoshtari\\
  McGill University and Mila\\
  \And
  Rosie Zhao \\
  McGill University and Mila\\
  \And
  Prakash Panangaden\\
  McGill University and Mila\\
  \AND
  David Meger \\
  McGill University and Mila
  \And
 Doina Precup \\
 McGill University, Mila, and DeepMind
}

\begin{document}

\maketitle

\begin{abstract}
Abstraction has been widely studied as a way to improve the efficiency and generalization of reinforcement learning algorithms. In this paper, we study abstraction in the continuous-control setting. We extend the definition of MDP homomorphisms to encompass continuous actions in continuous state spaces.  We derive a policy gradient theorem on the abstract MDP, which allows us to leverage approximate symmetries of the environment for policy optimization. Based on this theorem, we propose an actor-critic algorithm that is able to learn the policy and the MDP homomorphism map simultaneously, using the lax bisimulation metric.  We demonstrate the effectiveness of our method on benchmark tasks in the DeepMind Control Suite.  Our method's ability to utilize MDP homomorphisms for representation learning leads to improved performance when learning from pixel observations.

\end{abstract}

\section{Introduction}
\label{sec:intro}

\begin{wrapfigure}{R}{0.42\textwidth}
\vspace{-2em}
\begin{center}
    \includegraphics[width=0.42\textwidth]{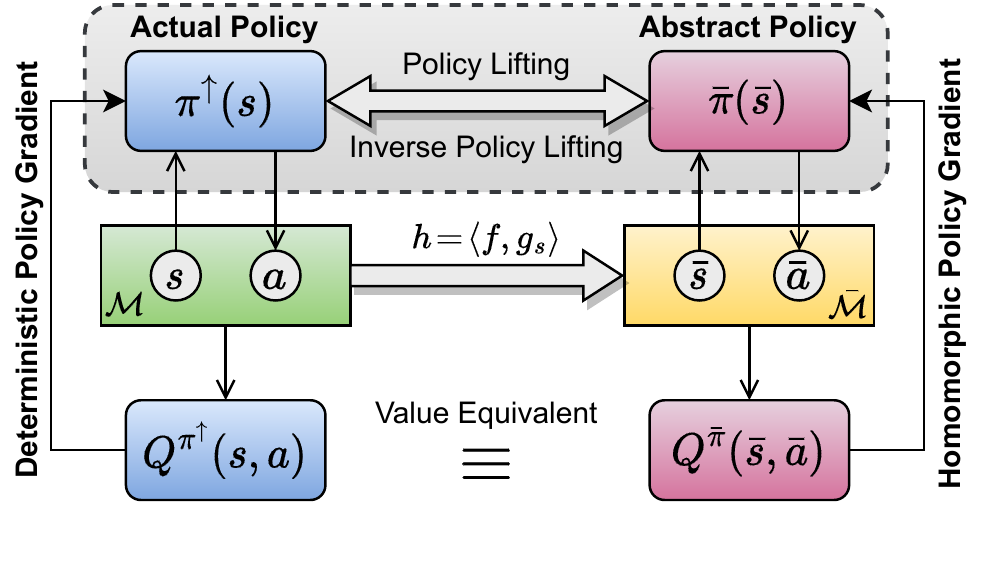}
    \caption{Schematics of our method.  The actual MDP $\mathcal{M}$ is used to train $Q^{\pi^\uparrow}$ and update $\pi^\uparrow$ with DPG, while the abstract MDP ${\overline{M}}$ is used to train $Q^{\overline{\pi}}$ and update $\overline{\pi}$ with HPG.  ${\overline{\mathcal{M}}}$ is the MDP homomorphic image of $\mathcal{M}$ obtained by learning the homomorphism map $h \!=\! ( f, g_s )$. Policies $\pi^\uparrow$ and $\overline{\pi}$ can be derived from each other.}
    \label{fig:hpg_diagram}
\end{center}
\vspace{-2em}
\end{wrapfigure}

For reinforcement learning from high-dimensional observations, such as images,
learning a simpler problem by abstraction from the original problem can be
critical \rebuttal{\cite{abel2016near, li2006towards}}.  The coupling between states, actions and rewards
complicates learning RL abstractions.  MDP homomorphisms \rebuttal{\cite{ravindran2004algebraic, ravindran2001symmetries, ravindran2004approximate, narayanamurthy2008hardness}} 
define a concept that allows one to exploit symmetries, yielding behavioral equivalence
and preserving values, while giving the potential to arrive at a substantially
smaller MDP. \rebuttal{Recent works \cite{van2020plannable, van2020mdp, biza2019online}} have shown
that using MDP homomorphisms is effective to guide learning in discrete
problems.  This paper is one of the first to consider MDP homomorphisms in the
continuous-control setting and to develop actor-critic algorithms with a tightly integrated state-action abstraction.
To that end, we identify and answer a series of key challenges:

\textbf{Can MDP homomorphisms be defined on continuous state and action spaces?}
Our first contribution is to define continuous MDP homomorphisms on continuous state and action spaces, which requires more intricate proofs that do not follow in any direct way from the finite case and requires tools from measure theory and differential geometry.

\textbf{Can MDP homomorphisms be tightly integrated into the policy gradient?}
Our second contribution is the derivation of the \emph{homomorphic policy
  gradient} (HPG) theorem to closely integrate the abstract MDP into the policy
gradient.  Importantly, we rigorously prove  that performing HPG on the abstract
MDP is equivalent to performing the deterministic policy gradient (DPG) on the
actual MDP. Therefore,
HPG can act as an additional gradient estimator capable of utilizing approximate symmetries for improved sample efficiency. 
To derive these results, we prove that continuous MDP homomorphisms preserve value functions \cite{grimm2020value}, which in turn enables their use for policy evaluation. 
 
\textbf{Can MDP homomorphisms be learned simultaneously with the optimal policy
 in a practical deep reinforcement learning algorithm?} We propose a deep actor-critic algorithm, depicted in Figure \ref{fig:hpg_diagram}, based on HPG,
referred to as \emph{Deep Homomorphic Policy Gradient} (DHPG), that unifies
state and action abstractions.  DHPG is able to simultaneously learn
the policy and the homomorphism map using the lax bisimulation metric
\cite{taylor2008bounding}, a metric for measuring the equivalence of state-action pairs under an MDP homomorphism relation.
We empirically show that state-action abstractions
learned through MDP homomorphisms provide a natural inductive bias for
representation learning.  

Despite the existence of well-studied abstraction notions, learning state
abstractions in a scalable fashion for continuous control remains a key
challenge.  In contrast to previous works on learning MDP homomorphisms
\rebuttal{\cite{van2020plannable, van2020mdp, biza2019online}}, our algorithm is readily applicable to
continuous actions, and compared to previous works guided by bisimulations
\cite{zhang2020learning, gelada2019deepmdp, kemertas2021towards}, our algorithm
leads to more robust solutions, as suggested by our empirical results. \rebuttal{The bisimulation relation \cite{milner1989communication, larsen1991bisimulation, blute1997bisimulation, Desharnais02, givan2003equivalence} and bisimulation metrics \cite{Desharnais99b, ferns2005metrics, ferns2006methods, ferns2011bisimulation} do not allow abstracting actions as they require exact matching of actions, whereas MDP homomorphisms and equivalently the lax bisimulation metric \cite{taylor2008bounding} remove this strong limitation, giving them greater modeling flexibility.} Most
importantly, the key difference between prior works \rebuttal{\cite{van2020plannable, van2020mdp, zhang2020learning, gelada2019deepmdp, kemertas2021towards}} and our method is the
homomorphic policy gradient theorem which allows for a tight integration of the abstraction notion into the policy gradient \rebuttal{that theoretically motivates using the abstract MDP for policy optimization in the actual MDP.} Our contributions are: 
\begin{enumerate}[noitemsep,nosep]
    \item Defining continuous MDP homomorphisms on continuous state and action spaces, using tools from measure theory and differential geometry. \item Proving that continuous MDP homomorphisms preserve value and optimal value functions.
    \item Deriving the homomorphic policy gradient theorem. 
    \item Developing a deep actor-critic algorithm for learning the optimal policy simultaneously with the MDP homomorphism map in challenging continuous control problems. 
\end{enumerate}
DHPG improves upon strong baselines on pixel observations \cite{yarats2021mastering, zhang2020learning} on DM Control, and our visualizations demonstrate the potential of MDP homomorphisms
in learning structured representations that can preserve values and represent
the minimal MDP image.  To the best of our knowledge, this is the first homomorphic policy
gradient derivation and the first work to define and scale up MDP
homomorphisms to continuous visual control problems.
Our code is publicly available at \href{https://github.com/sahandrez/homomorphic_policy_gradient}{https://github.com/sahandrez/homomorphic\_policy\_gradient}.

\section{Background}
\label{sec:background}
\subsection{Markov Decision Processes }
We consider the standard MDP that is defined by a 5-tuple $( {\mathcal{S}}, {\mathcal{A}}, {\tau_a}, {R}, \gamma )$, with \emph{state space} ${\mathcal{S}}$, \emph{action space} ${\mathcal{A}}$, \emph{transition dynamics} ${\tau_a \!:\! \mathcal{S} \!\times\! \mathcal{A} \!\to\! \Dist(\mathcal{S})}$, \emph{reward function} ${R\!:\! \mathcal{S} \!\times\! \mathcal{A}} \!\to\! \mathbb{R}$, and \emph{discount factor} $\gamma \!\in\! (0, 1]$. The goal is to find a policy $\pi: \mathcal{S} \to \Dist(\mathcal{A})$ that maximizes the expected sum of discounted rewards, the \emph{expected return}, defined as $\mathbb{E}_\pi\![ R_t ] \!=\! \mathbb{E}_\pi \![ \sum_{k=0}^T \!\gamma^{k} r_{t + k + 1} ]$. \emph{Value function} $V^\pi(s)$ denotes the expected return from $s$ under policy $\pi$, and \emph{action-value function} $Q^\pi(s, a)$ denotes the expected return from $s$ after taking action $a$ under $\pi$.
Value functions are fixed points of the Bellman equation \cite{bellman1957dynamic} and can be computed iteratively through a process referred to as \emph{policy evaluation} \cite{sutton2018reinforcement}. Similarly, optimal value functions $V^*(s)$ and $Q^*(s, a)$ are fixed points of the Bellman optimality equation \cite{bellman1957dynamic}.

\subsection{MDP Homomorphisms}
MDP homomorphisms are formally defined for finite MDPs by Ravindran and Barto \cite{ravindran2001symmetries} as: 
\begin{definition}[MDP Homomorphism]
\label{def:mdp_homo}
An \emph{MDP homomorphism} $h \!=\! ( f, g_s ) \!:\! {\mathcal{M}} \!\to\! {\overline{\mathcal{M}}}$ is a surjective map from a finite MDP ${\mathcal{M}} \!=\! ( {\mathcal{S}}, {\mathcal{A}}, {R}, {\tau_a}, \gamma )$ onto an abstract finite MDP ${\overline{\mathcal{M}}} \!=\! ( {\overline{\mathcal{S}}, \overline{\mathcal{A}}, \overline{R}, \overline{\tau}_{\overline{a}}}, \gamma ) $ where $f\!:\! {\mathcal{S}} \!\to\! {\overline{\mathcal{S}}}$ and $g_s\!:\! {\mathcal{A}} \!\to\! {\overline{\mathcal{A}}}$ are surjective maps onto the abstract state and action spaces:
\begin{align}
    \label{eq:reward_invariance}
    \text{Invariance of reward: }& {\overline{R}}(f(s), g_s(a)) = {R}(s, a) \quad \forall s \in \mathcal{S}, a \in \mathcal{A} \\
    \label{eq:transition_equivariance}
    \text{Equivariance of transitions: }& \overline{\tau}_{g_s(a)}(f(s^\prime)| f(s)) = \sum_{s^{\prime\prime} \in [s^\prime]_{B_h | {\mathcal{S}}}}\tau_{a}(s^{\dprime} | s) \quad \forall s \in \mathcal{S}, a \in \mathcal{A} 
\end{align}
\end{definition}
\vspace{-\parskip}
where $B_h$ is the partition of ${\mathcal{S}}$ induced by the equivalence relation of homomorphism $h$, $B_h | {\mathcal{S}}$ is the projection of $B_h$ onto $\mathcal{S}$, and $[s^\prime]_{B_h | {\mathcal{S}}}$ denotes the block of $B_h | {\mathcal{S}}$ to which $s^\prime$ belongs. Thus, when applying action $a$ in state $s$, the right-hand side is the probability that the resulting state is in $[s^\prime]_{B_h | {\mathcal{S}}}$. The abstract MDP ${\overline{\mathcal{M}}}$ is in fact the quotient MDP ${\mathcal{M}} / B_h$ based on the homomorphism map $h \!:\! {\mathcal{M}} \!\to\! {\mathcal{M}} / B_h$. 
As MDP homomorphisms are sensitive with respect to changes in rewards or transitions, \emph{approximate} MDP homomorphisms \cite{ravindran2004approximate} allow equations (\ref{eq:reward_invariance}-\ref{eq:transition_equivariance}) to hold approximately. The significance of MDP homomorphisms is the \emph{optimal value equivalence} between $\mathcal{M}$ and $\mathcal{\overline{M}}$ \cite{ravindran2001symmetries}:
\begin{equation}
    \label{eq:optimal_value_equivalence}
    V^*(s) = \overline{V}^*(f(s)) \quad \forall s \in \mathcal{S},  \qquad Q^*(s, a) = \overline{Q}^*(f(s), g_s(a)) \quad \forall s \in \mathcal{S}, a \in \mathcal{A}
\end{equation}
which in turn allows for learning the optimal policy $\overline{\pi}^*$ in the abstract MDP and consequently \emph{lifting} it to obtain the optimal policy in the actual MDP, using:
\begin{equation*}
    \pi^\uparrow(a | s) = \dfrac{\overline{\pi}(\overline{a} | f(s))}{|\{a \in g_s^{-1}(\overline{a}) \}|}, \qquad \forall s \in {\mathcal{S}}, a \in g_s^{-1}(\overline{a})
\end{equation*}
where $g_s^{-1}(\overline{a})$ denotes the set of actions that have the same image $\overline{a}$ under $g_s$. Equivalently, the two policies must satisfy $\sum_{a \in g_s^{-1}(\overline{a})} \pi^\uparrow(a | s) = \overline{\pi}(\overline{a} | f(s))$ for all $s \in {\mathcal{S}}$ and $\overline{a} \in \overline{\mathcal{A}}$.

\subsection{Bisimulation and Lax Bisimulation Metrics}
\label{sec:bisim}
\emph{Bisimulation} for finite MDPs \cite{Desharnais02, givan2003equivalence} defines an equivalence relation on $\mathcal{S}$ where two states $s_i$ and $s_j$ are equivalent or \emph{bisimilar} if ${R}(s_i, a) \!=\! {R}(s_j, a)$ and ${\tau_a}(C | s_i) \!=\! {\tau_a}(C | s_j)$ for all $a \!\in\! {\mathcal{A}}$ and every equivalence class $C$ defined by the equivalence relation. The rigidity of bisimulation limits its applications. \emph{Bisimulation metrics} \cite{ferns2005metrics, ferns2006methods, ferns2011bisimulation} measure the equivalence as an approximation:
\begin{align}
    \label{eq:bisim_metric}
    d_\text{bisim}\big(s_i, s_j \big) = \max_{a \in {\mathcal{A}}} c_r \big|R(s_i, a) - R (s_j, a) \big| + c_t K \big(\tau_a (\cdot | s_i), \tau_a(\cdot | s_j) \big),
\end{align}
where the first term measures reward similarity and $K$ is the Kantorovich (Wasserstein) metric measuring the distance between the transition probabilities. However, bisimulation metrics can still be brittle as they require the behaviour to match for all actions. This may be problematic particularly in the case of continuous actions in which small changes to actions may not drastically change the outcome. Additionally, bisimulation metrics are not able to represent environment symmetries. Instead, \emph{lax bisimulation} \cite{taylor2008bounding} waives the requirement on action matching in favor of extending the state equivalence relation to state-action equivalence. Taylor et al. \cite{taylor2008bounding} show that lax bisimulation is precisely the same relation as the MDP homomorphism and define the \emph{lax bisimulation metric} as:
\begin{align}
    \label{eq:lax_bisim_metric}
    d_\text{lax} \big(( s_i, a_i ), ( s_j, a_j ) \big) = c_r \big|R(s_i, a_i) - R (s_j, a_j) \big| + c_t K \big(\tau_{a_i}(\cdot | s_i), \tau_{a_j}(\cdot | s_j) \big).
\end{align}
Furthermore, Taylor et al. \cite{taylor2008bounding} show that minimizing the lax bisimulation metric corresponds to finding approximate MDP homomorphisms and bound the value error.
\vspace{-0.8\parskip}

\section{Value Equivalence Property}
\label{sec:value_equiv}
To motivate the use of MDP homomorphisms for policy evaluation and consequently policy optimization, we first prove their \emph{value equivalence property} in the finite case as the generalization of the prior result of the \emph{optimal} value equivalence \cite{ravindran2001symmetries}, stated in Equation \eqref{eq:optimal_value_equivalence}. The proof is in Appendix \ref{supp:proof_value_equiv_finite}. 
\begin{thm}[Value Equivalence]
\label{thrm:value_equiv_finite_main}
Let $\overline{\mathcal{M}}$ be the image of an MDP homomorphism $h$ from a finite $\mathcal{M}$. Then any two corresponding policies $\pi^\uparrow = \text{lift}(\overline{\pi})$ have equivalent values: 
\begin{align*}
    V^{\pi^\uparrow}(s) = V^{\overline{\pi}}(f(s)) \quad \forall s \in \mathcal{S}, \qquad Q^{\pi^\uparrow}\!(s, a) = Q^{\overline{\pi}}(f(s), g_s(a)) \quad \forall s \in \mathcal{S}, a \in \mathcal{A}
\end{align*}
\end{thm}
\vspace{-1.5\parskip}

\section{Continuous MDP Homomorphisms}
\label{sec:cont_mdp_hom}
To concretely lay out the foundations of using MDP homomorphisms for continuous control, we extend their definition to continuous state and action spaces, and derive results analogous to the finite case. First, we define continuous MDPs and state our underlying assumptions. Importantly, the correct definitions of continuous MDPs and continuous MDP homomorphisms require care regarding measurability and differentiability of spaces, and our formulation is chosen to fit the HPG derivation; see Appendix \ref{supp:math_tools} for an overview of the tools we used from measure theory and differential geometry.
\begin{definition}[Continuous MDP]
    \label{def:cont_mdp}
  A \emph{continuous Markov decision process (MDP)} is a $6$-tuple:
  \[\mathcal{M} = (\mathcal{S},\Sigma,\mathcal{A},\forall a\in \mathcal{A}\; \tau_a:\mathcal{S}\x\Sigma\to[0,1],R:\mathcal{S}\x \mathcal{A}\to \R, \gamma)\] where $\mathcal{S}$, the state space is assumed to be a Polish space, $\Sigma$ is a $\sigma$-algebra on $\mathcal{S}$\footnote{Usually the Borel algebra.}, $\mathcal{A}$, the space of \emph{actions}, is a
  locally compact metric space, usually taken to be a subset of $\R^n$, $\tau_a$ is
  the transition probability kernel for each possible action $a$, for each fixed $s$,
  $\tau_a(\cdot|s)$ is a probability distribution on $\Sigma$ while $R$ is the reward
  function, and $\gamma$ is the discount factor. Furthermore, for all $s \in \mathcal{S}$ and $B \in \Sigma$ the map $a \mapsto \tau_a(B|s)$ is smooth. The last assumption is required for differentiability with respect to actions $a$, which is needed in Section \ref{sec:hpg} for deriving the HPG theorem.
\end{definition}
\vspace{-\parskip}
Given the continuous MDPs described above, we define continuous MDP homomorphisms. The equivariance condition on the transition dynamics, Equation~\eqref{eq:transition_equivariance}, can no longer be expressed in terms of a discrete sum over partitions, and instead we use the $\sigma$-algebra structure on the different state spaces.

\begin{definition}[Continuous MDP Homomorphism]
\label{def:cont_mdp_homo}
    A \textit{continuous MDP homomorphism} is a map $h = ( f, g_s ): \mathcal{M} \to \overline{\mathcal{M}}$ where $f: \mathcal{S} \to \overline{\mathcal{S}}$ and for every $s$ in $\mathcal{S}$, $g_s: \mathcal{A} \to \overline{\mathcal{A}}$ are measurable, surjective maps such that the following hold:
    \begin{align}
        \text{Invariance of reward: }& \overline{R}(f(s), g_s(a)) = R(s,a) \qquad \forall s \in \mathcal{S}, a \in \mathcal{A} \\
        \text{Equivariance of transitions: }& \overline{\tau}_{g_s(a)}(\overline{B}| f(s)) = \tau_a(f^{-1}(\overline{B})| s) \qquad \forall \; s \in \mathcal{S}, a \in \mathcal{A}, \overline{B} \in \overline{\Sigma}
    \end{align}
\end{definition}
\vspace{-\parskip}
Note that if $g_s$ is the identity map, the second condition reduces to $\overline{\tau}_a(\overline{B} | f(s)) \!=\! \tau_a(f^{-1}(\overline{B}) | s)$ which is simply the condition for preservation of transition probabilities as used in bisimulation \cite{Desharnais02}.

\subsection{Optimal Value Equivalence}
Assuming the conditions given in Definition~\ref{def:cont_mdp_homo}, we prove that optimal value functions are preserved by the continuous MDP homomorphism as in the finite case.
\begin{thm}[Optimal Value Equivalence]
\label{thm:opt_equiv_continuous}
Let $\overline{\mathcal{M}} = ( \overline{\mathcal{S}}, \overline{\Sigma}, \overline{\mathcal{A}}, \overline{\tau}_{\overline{a}}, \overline{R} )$ be the image of a continuous MDP homomorphism $h = (f, g_s)$ from $\mathcal{M} = ( \mathcal{S}, \Sigma, \mathcal{A}, \tau_a, R )$. Then:
\begin{equation}
\label{eq: opt_equiv_continuous}
    V^*(s) = \overline{V}^*(f(s)) \quad \forall s \in \mathcal{S}, \qquad  Q^*(s,a) = {\overline{Q}^*}(f(s), g_s(a)) \quad \forall (s, a) \in \mathcal{S} \times \mathcal{A}
\end{equation}
\end{thm}
\vspace{-\parskip}
The proof, given in Appendix~\ref{sec:opt_equiv_continuous}, uses the change of variable formula of the pushforward measure of $\tau_a(\cdot | s)$ with respect to $f$ to change the integration space from $\mathcal{S}$ to $\mathcal{\overline{S}}$.
\subsection{Value Equivalence for Lifting Deterministic Policies}
\label{sec:value_equiv_continuous}
As in the finite case, we also require a lifting process to define $\pi^\uparrow \! = \! \text{\emph{lift}}(\overline{\pi})$ given a policy $\overline{\pi}$ on the abstract MDP. In general, the lifted policy needs to satisfy the relation $\pi^\uparrow(g_s^{-1}(\beta) | s) = \overline{\pi}(\beta | f(s))$ for every Borel set $\beta \subset \mathcal{\overline{A}}$ and $s \in \mathcal{S}$. While our initial progress shows that lifted stochastic policies exist based on the disintegration theorem, the full proof and design of a computationally tractable algorithm for this process is left for future work. Therefore, here and in the subsequent sections we assume the policy is deterministic in which case the lifted policy can be simply obtained by choosing one representative for the preimage $g_s^{-1}\big(\overline{\pi}(f(s))\big)$. If we select $g_s$ to be a bijection, the lifted policy can be uniquely defined as $\pi^\uparrow(s) \!=\! g_s^{-1}\big(\overline{\pi}(f(s))\big)$. The assumption on deterministic policies is not limiting, as in general the optimal policy of a given MDP is deterministic \cite{bertsekas2012dynamic}. With this lifting definition, we state and prove the following value equivalence result:




\begin{thm}[Value Equivalence for Deterministic Policies]
\label{thrm:value_equiv_continuous}
Let $\overline{\mathcal{M}}$ be the image of a continuous MDP homomorphism $h=(f, g_s)$ from $\mathcal{M}$, then any two deterministic policies $\pi^\uparrow: \mathcal{S} \to \mathcal{A}$ and $\overline{\pi}:\overline{\mathcal{S}} \to \overline{\mathcal{A}}$ where $\pi^\uparrow = \text{lift}(\overline{\pi})$ have equivalent value functions on their domain:
\begin{align*}
    V^{\pi^\uparrow}(s) = V^{\overline{\pi}}(f(s)) \quad \forall s \in \mathcal{S}, \qquad Q^{\pi^\uparrow}\!(s, a) = Q^{\overline{\pi}}(f(s), g_s(a)) \quad \forall (s, a) \in \mathcal{S} \times \mathcal{A}
\end{align*}
\end{thm}
\vspace{-\parskip}
The proof, given in Appendix~\ref{supp:value_equiv_continuous}, uses the change of variable formula of the pushforward measure of $\tau_a(\cdot | s)$ with respect to $f$  to change the integration space from $\mathcal{S}$ to $\mathcal{\overline{S}}$ and assumes $g_s$ to be bijective. 

\section{Homomorphic Policy Gradient}
\label{sec:hpg}
The next goal of this work is to derive a policy gradient estimator using samples obtained from the abstract MDP. Intuitively, this allows for direct incorporation of state-action abstraction as an inductive bias for policy optimization, thereby reducing the variance of actor updates and improving sample efficiency. Equipped with continuous MDP homomorphisms from Definition \ref{def:cont_mdp_homo} and their value equivalence property, we now derive the \emph{homomorphic policy gradient} (HPG) theorem.

In this section, we assume having access to an MDP homomorphism map $h \!=\! (f, g_s)$, parameterized by differentiable functions. The problem of learning such mapping from samples is addressed in Section \ref{sec:hac}. Additionally, we assume the MDP and the homomorphism map adhere to the conditions of Definition \ref{def:cont_mdp} and Appendix \ref{sec:assumptions}. Similarly to prior works on policy gradients \cite{sutton2000policy, silver2014deterministic}, we define the performance measure as $J(\theta) = \mathbb{E}_{\pi}[V^\pi (s)]$ where the expectation is over the uncertainty in transitions, rewards, and initial states. Finally, as detailed in Section \ref{sec:value_equiv_continuous}, our results are derived for deterministic policies and a bijective $g_s$. Notably, this choice allows us to parameterize one of the policies and to uniquely derive the other policy. In practice, we parameterize the actual policy as $\pi^\uparrow_\theta$ and obtain the abstract policy as $\overline{\pi}_\theta = g_s(\pi^\uparrow_\theta(s))$.
First, we show the \emph{equivalence of policy gradients}:
\begin{thm}[Equivalence of Deterministic Policy Gradients]
    \label{thm:det_grad_equiv}
    Let $\overline{\mathcal{M}}$ be the image of a continuous MDP homomorphism $h$ from $\mathcal{M}$, and let $\pi^\uparrow_\theta : \mathcal{S} \to \mathcal{A}$ be the lifted deterministic policy corresponding to the abstract deterministic policy $\overline{\pi}_\theta : \overline{\mathcal{S}} \to \overline{\mathcal{A}}$. Then for any $(s, a) \in \mathcal{S} \times \mathcal{A}$ we have:
    $$
    \nabla_a Q^{\pi^\uparrow_\theta}(s, a) \Big|_{a = \pi^\uparrow_\theta(s)} \nabla_\theta \pi^\uparrow_\theta(s) = \nabla_{\overline{a}} Q^{\overline{\pi}_\theta}(\overline{s}, \overline{a}) \Big|_{\overline{a} = \overline{\pi}_\theta(\overline{s})} \nabla_\theta\overline{\pi}_\theta(\overline{s}).
    $$
\end{thm}
\vspace{-\parskip}
The proof is given in Appendix \ref{app:det_grad_equiv} and uses the chain rule and the inverse function theorem on manifolds, which in turn raises the need for $g_s$ to be a bijection and local diffeomorphism. Theorem \ref{thm:det_grad_equiv} highlights that the gradient of the abstract MDP is equivalent to that of the original, despite the underlying spaces being abstracted. This implies that performing HPG on the abstract MDP is equivalent to performing DPG on the actual MDP, allowing us to use them synergistically to update the same parameters $\theta$, as shown in Figure \ref{fig:hpg_diagram}. 

While one can naively use Theorem \ref{thm:det_grad_equiv} to substitute gradients of the standard DPG, theoretically this does not produce any useful result as the expectation remains estimated with respect to the stationary state distribution of the actual MDP $\mathcal{M}$ under $\pi_\theta^\uparrow(s)$. However, using properties of continuous MDP homomorphisms, we can change the integration space from $\mathcal{\mathcal{S}}$ to $\mathcal{\overline{\mathcal{S}}}$, and consequently estimate the policy gradient with respect to the stationary distribution of the abstract MDP $\mathcal{\overline{M}}$ under $\overline{\pi}_\theta(\overline{s})$:
\begin{thm} [Homomorphic Policy Gradient Theorem]
\label{thrm:det_hpg}
Let $\overline{\mathcal{M}}$ be the image of a continuous MDP homomorphism $h$ from $\mathcal{M}$, and let $\overline{\pi}_\theta : \overline{\mathcal{S}} \to \overline{\mathcal{A}}$ be a deterministic abstract policy defined on $\overline{\mathcal{M}}$. Then the gradient of the performance measure $J(\theta)$, defined on the actual MDP $\mathcal{M}$, w.r.t. $\theta$ is:  
$$
    \nabla_\theta J(\theta) = \int_{\overline{s} \in \overline{\mathcal{S}}} \rho^{\overline{\pi}_\theta}(\overline{s})  \nabla_{\overline{a}} Q^{\overline{\pi}_\theta}(\overline{s}, \overline{a}) \Big|_{\overline{a} = \overline{\pi}_\theta(\overline{s})} \nabla_\theta \overline{\pi}_\theta(\overline{s}) d\overline{s}.
$$
where $\rho^{\overline{\pi}_\theta}(\overline{s})$ is the discounted state distribution of $\overline{\mathcal{M}}$ following the deterministic policy $\overline{\pi}_\theta(\overline{s})$.
\end{thm}
\vspace{-\parskip}
The proof is given in Appendix \ref{app:deterministic_hpg} and applies the result of Theorem \ref{thm:det_grad_equiv} and the change of variables formula of the pushforward measure on the state space. The significance of Theorem \ref{thrm:det_hpg}, which forms the basis of our proposed homomorphic actor-critic algorithm, is twofold. First, we can get another estimate for the policy gradient based on the approximate MDP homomorphic image in addition to DPG. Although the two policy gradient estimates are not statistically independent from one another as they are tied through the homomorphism map, HPG will potentially have less variance at the expense of some bias due to the approximation of the MDP homomorphism.

Second, since the minimal image of an MDP is the MDP homomorphic image \cite{ravindran2001symmetries}, the abstract critic $Q^{\overline{\pi}_\theta}$ is trained on a simplified problem. In other words, each abstract state-action pair $(\overline{s}, \overline{a})$ used to train $Q^{\overline{\pi}_\theta}$ represents all $(\! s, a\! )$ pairs that are equivalent under the MDP homomorphism relation, thus improving sample efficiency. However, the amount of complexity reduction is dependent on the approximate symmetries of the environment, as also supported by our empirical results.

\section{Homomorphic Actor-Critic Algorithms}
\label{sec:hac}
We propose a deep actor-critic algorithm based on HPG by adapting DDPG \cite{lillicrap2015continuous}. We refer to our method as \emph{Deep Homomorphic Policy Gradient} (DHPG). Although HPG is applicable to any other deterministic actor-critic, we chose DDPG as its simplicity compared to modern choices \cite{barth2018distributed, horgan2018distributed} allows for a better study on the impact of MDP homomorphisms. 
Notably, a strong advantage of DHPG is that it is readily applicable to pixel observations without the need for extra mechanisms such as image reconstruction \cite{gelada2019deepmdp, ha2018world, yarats2021improving, hafner2019dream}, as the notion of MDP homomorphism provides a natural inductive bias for learning representations that preserve values and optimal values. 

Since DHPG is learning the MDP homomorphism map $h$ online and concurrently with the policy, using the actual MDP for training the critic, specifically at the early stages of training, is helpful. Therefore, we utilize a separate critic for each MDP ${\mathcal{M}}$ and ${\overline{\mathcal{M}}}$. Ultimately, critics are used to update a single set of parameters, as shown in Figure \ref{fig:hpg_diagram}; see Appendix \ref{sec:ablation_dhpg_variants} for the ablation study on this.

Thus, the components of DHPG are: actual critic $Q_\psi(\!s, a\!)$, abstract critic $\overline{Q}_{\overline{\psi}}(\!\overline{s}, \overline{a}\!)$, deterministic actor $a \!\! =\!\! \pi_\theta(s)$, homomorphism map $h_{\phi, \eta} \!\!=\! ( f_\phi(s), g_\eta(s, a) )$, reward predictor $\overline{R}_\rho(\overline{s})$, and probabilistic transition model $\overline{\tau}_\nu(\overline{s}^\prime | \overline{s}, \overline{a})$ which outputs a Gaussian distribution. We use target networks and a vanilla replay buffer \cite{mnih2013playing, lillicrap2015continuous}. As discussed in Section \ref{sec:hpg}, an abstract actor is obtained as $\overline{a} \!=\! g_{s}(\pi_\theta(s))$. In case of pixel observations, a single image encoder $E_\mu$ is shared among all components.

\begin{figure}[b!]
    \centering
    \begin{subfigure}[b]{0.24\textwidth}
         \centering
         \includegraphics[width=\textwidth]{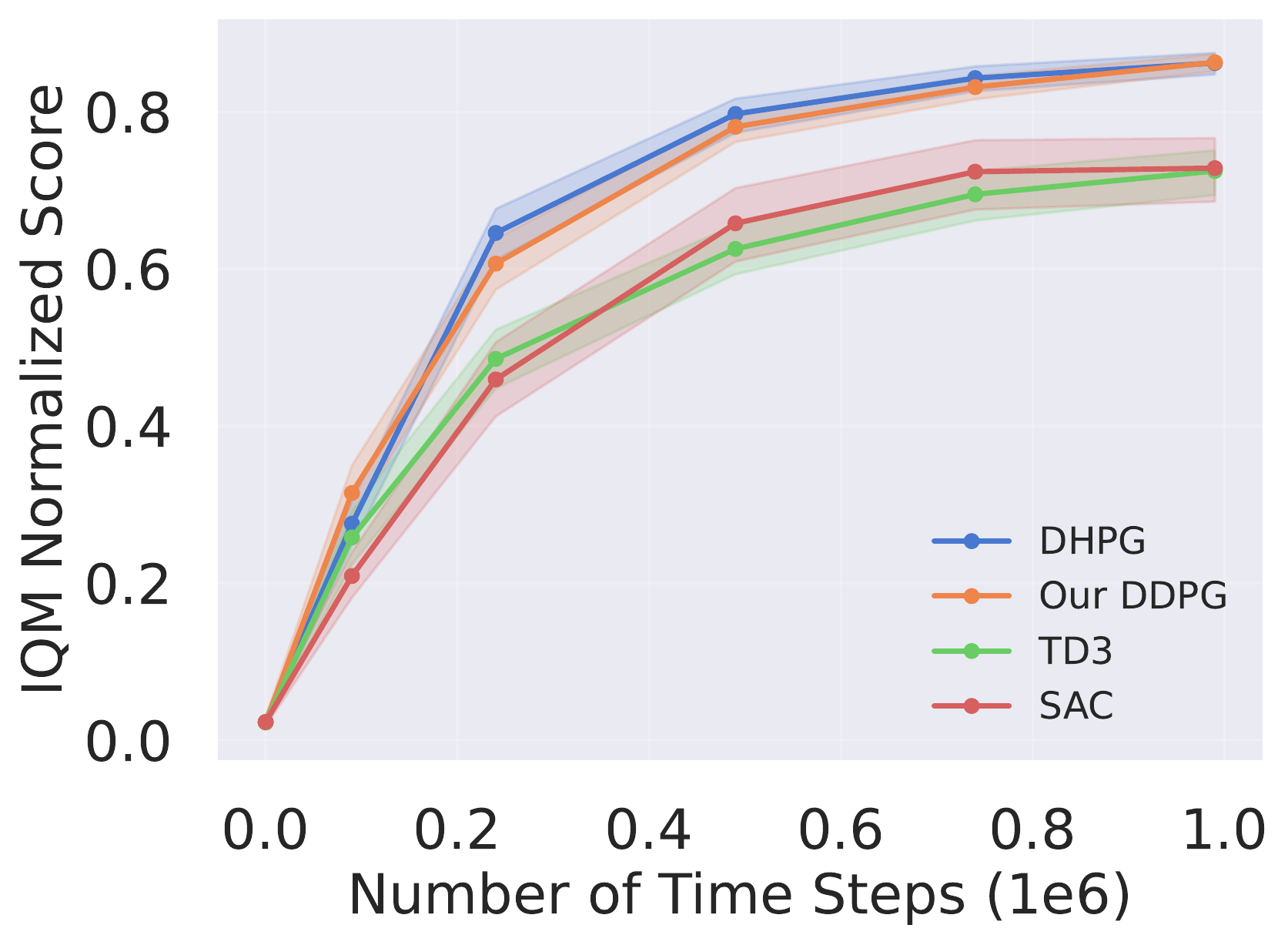}
         \caption{Sample efficiency.}
         \label{fig:states_sample_efficiency}
    \end{subfigure}
    \begin{subfigure}[b]{0.24\textwidth}
         \centering
         \includegraphics[width=\textwidth]{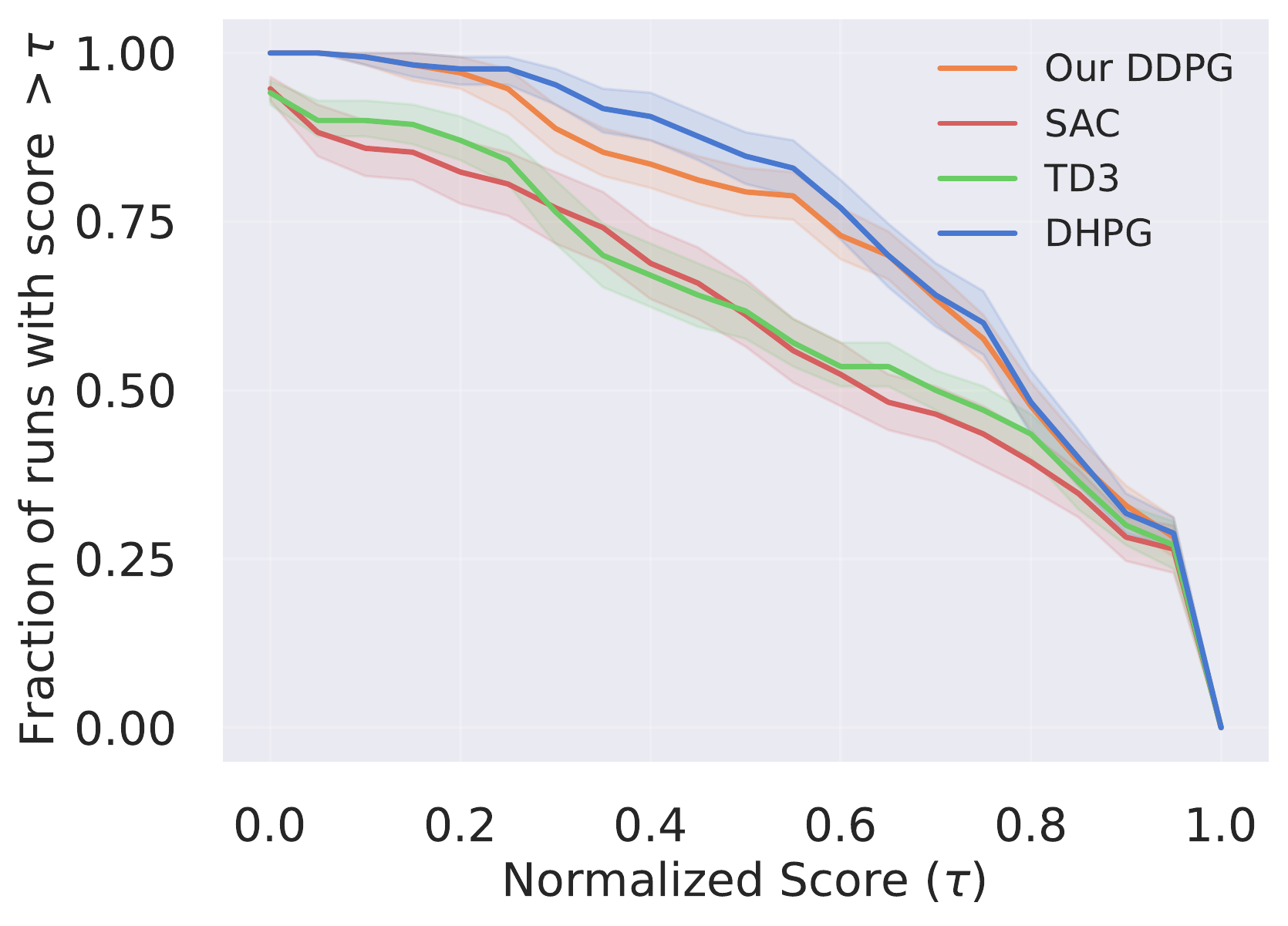}
         \caption{Performance profiles.}
         \label{fig:states_perf_profile_main}
    \end{subfigure}
    \begin{subfigure}[b]{0.24\textwidth}
         \centering
         \includegraphics[width=\textwidth]{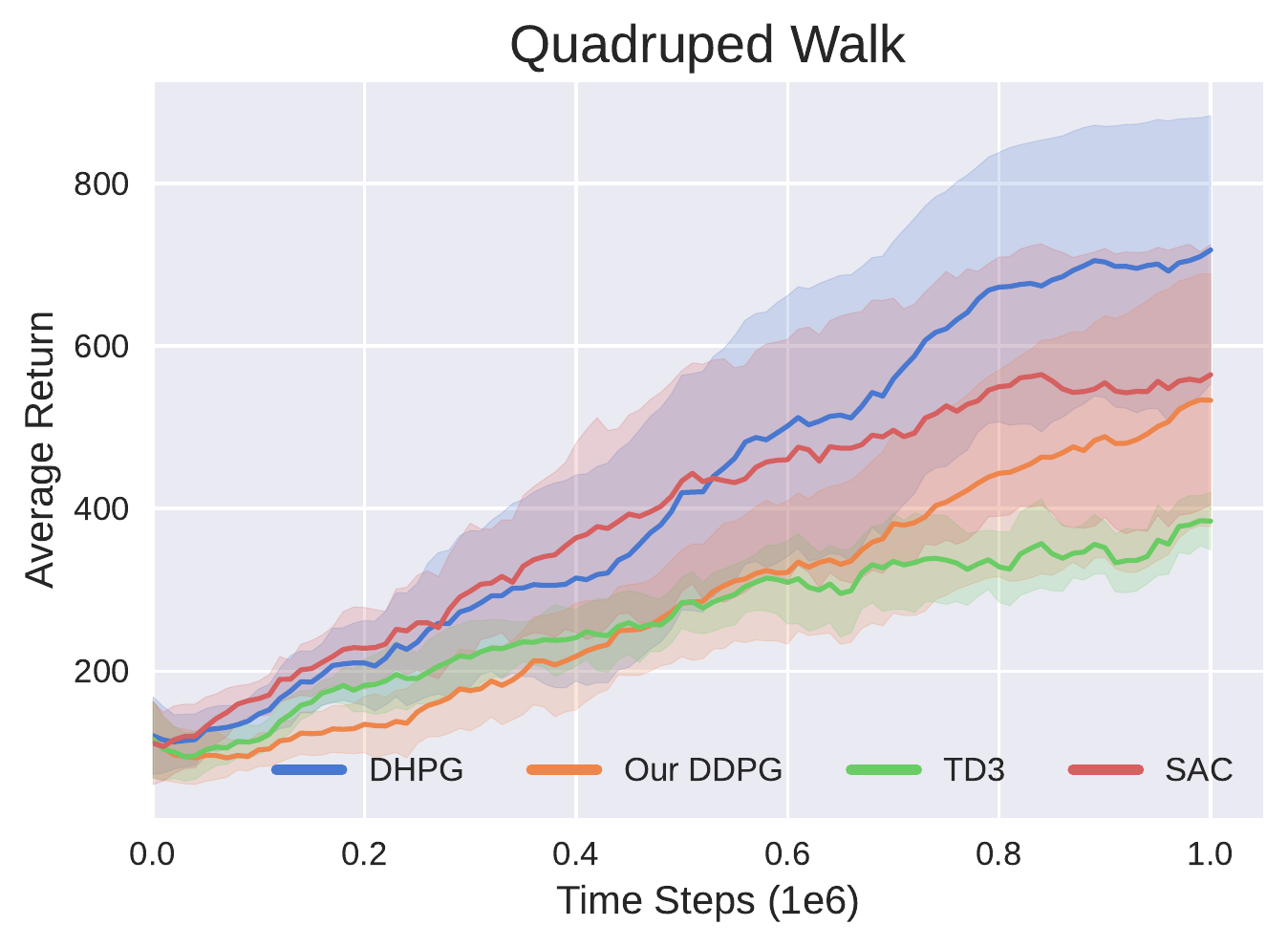}
         \caption{Learning curves.}
         \label{fig:states_res_1}
    \end{subfigure}
    \begin{subfigure}[b]{0.24\textwidth}
         \centering
         \includegraphics[width=\textwidth]{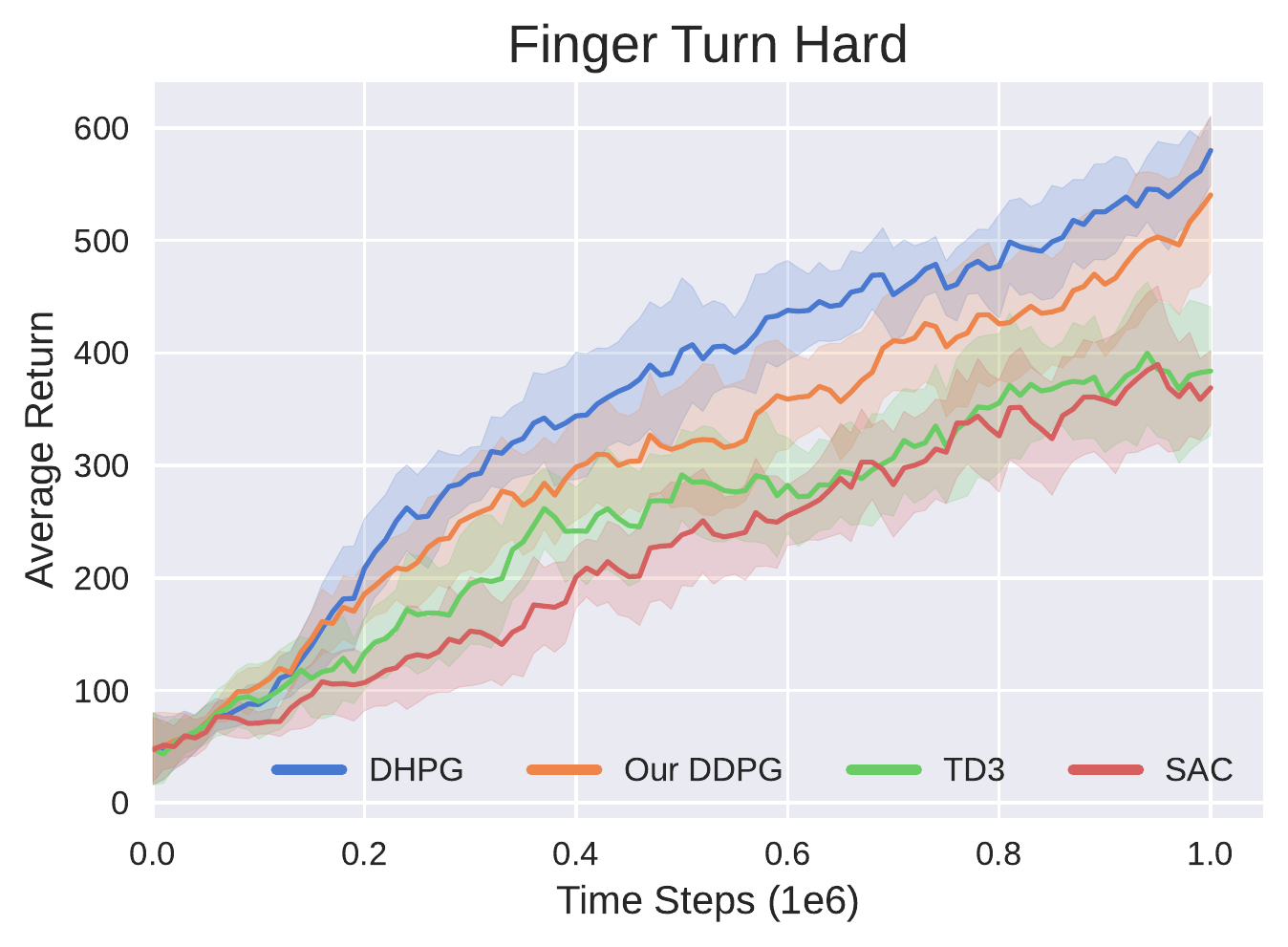}
         \caption{Learning curves.}
         \label{fig:states_res_2}
    \end{subfigure}
    \caption{Results of DM Control tasks with \textbf{state observations} obtained on 10 seeds. RLiable metrics are aggregated over 17 tasks. \textbf{(a)} RLiable IQM scores as a function of number of steps for comparing sample efficiency, \textbf{(b)} RLiable performance profiles at 500k steps, \textbf{(c)}-\textbf{(d)} examples of learning curves. Full results are in Appendix \ref{sec:additional_results_states}. Shaded regions represent $95\%$ confidence intervals.}
    \label{fig:states_results}
\end{figure}

\paragraph{Training the Policy and Critic.} Actual and abstract critics are trained using $n$-step TD error for a faster reward propagation \cite{barth2018distributed}. The loss function for each critic is therefore defined as the expectation of the $n$-step Bellman error estimated over transitions samples from the replay buffer $\mathcal{B}$:
\begin{align}
    \label{eq:critic_loss_1}
    \mathcal{L}_\text{actual critic} (\psi) &= \mathbb{E}_{(s, a, s^\prime, r) \sim \mathcal{B}} \big[ \big( R_t^{(n)} + \gamma^n Q_{\psi^\prime}({s}_{t+n}, {a}_{t+n}) - Q_\psi({s}_t, {a}_t) \big)^2 \big] \\
    \label{eq:critic_loss_2}
    \mathcal{L}_\text{abstract critic} (\overline{\psi}, \phi, \eta) &= \mathbb{E}_{(s, a, s^\prime, r) \sim  \mathcal{B}} \big[ \big(R_t^{(n)} + \gamma^n \overline{Q}_{\overline{\psi}^\prime}({\overline{s}}_{t+n}, {\overline{a}}_{t+n}) - \overline{Q}_{\overline{\psi}}({\overline{s}}_t, {\overline{a}}_t) \big)^2 \big],
\end{align}
where ${\overline{s}_t} \!=\! f_\phi({s}_t)$ and ${\overline{a}}_t \!=\! g_\eta({s}_t, {a}_t)$ are computed using the learned MDP homomorphism, $\psi^\prime$ and $\overline{\psi}^\prime$ denote parameters of target networks, and $R_t^{(n)} \!\!=\!\! \sum_{i=0}^{n-1} \gamma^i r_{t+i}$ is the $n$-step return. Consequently, we train the policy using DPG \cite{silver2014deterministic} and HPG from Theorem \ref{thrm:det_hpg} by backpropagating the following loss:
\begin{align}
    \mathcal{L}_\text{actor}(\theta) \approx - \mathbb{E}_{s \sim \mathcal{B}} \Big[ Q_\psi \big(s, \pi_\theta(s) \big) + \overline{Q}_{\overline{\psi}} \big(f_\phi(s), g_\eta\big(s, \pi_\theta(s)\big)\big) \Big].
    \label{eq:hpg_actor_update}
\end{align}
Here, the two gradients are added together and a single policy update is conducted; see Appendix \ref{sec:ablation_dhpg_variants} for the ablation study on other combinations of HPG and DPG. 
Finally, we utilize target policy smoothing and delayed actor updates \cite{fujimoto2018addressing}. The pseudo-code of DHPG is presented in Appendix \ref{sec:pseudocode}.

\paragraph{Learning Continuous MDP Homomorphisms.}
We now address the problem of learning continuous MDP homomorphisms. While few methods have been proposed for learning finite MDP homomorphisms \cite{van2020plannable, van2020mdp}, these are not readily extendable to continuous actions. In this work, we use the lax bisimulation metric \cite{taylor2008bounding}, Equation \eqref{eq:lax_bisim_metric}, to propose a loss function that encodes lax bisimilar states closer together in the abstract space. The lax bisimulation metric is applicable to continuous actions and as a (pseudo-)metric, it can naturally represent approximate MDP homomorphisms.

Following the same intuition as prior works on bisimulations \cite{zhang2020learning}, we define our proposed lax bisimulation loss over pairs of transition tuples sampled from the replay buffer. We permute samples to compute their pairwise distance in the abstract space and their pairwise lax bisimilarity distance. Consequently, we minimize the distance between these two terms:
{
\small
\begin{equation}
    \mathcal{L}_\text{lax}(\phi, \eta) = \mathbb{E}_{\mathcal{B}} \big[ \| f_\phi({s}_i) - f_\phi({s}_j) \|_1 - \|r_i-r_j\|_1 - \alpha W_2 \big( \overline{\tau}_\nu(\cdot | f_\phi({s}_i), g_\eta(s_i, {a}_i)), \overline{\tau}_\nu(\cdot | f_\phi({s}_j), g_\eta(s_j, a_j)) \big) \big]
    \label{eq:lax_bisim_loss}
\end{equation}
}
Similarly to Zhang et al. \cite{zhang2020learning}, we replaced the Kantorovich ($W_1$) metric in Equation \eqref{eq:lax_bisim_metric} with the $W_2$ metric as there is an explicit formula for it for Gaussian distributions. 
Finally, we apply the conditions of a continuous MDP homomorphism map from Definition \ref{def:cont_mdp_homo} via the loss function of:
\begin{equation}
    \label{eq:hom_loss}
    \mathcal{L}_\text{h}(\phi, \eta, \nu, \rho) = \mathbb{E}_{(s_i, a_i, s_i^\prime, r_i) \sim \mathcal{B}} \big[ \big( f_\phi({s}_i^\prime) - {\overline{s}}_i^\prime \big)^2 + \big( r_i \!-\! \overline{R}_\rho(f_\phi({s}_i)) \big)^2  \big],
\end{equation}
where ${\overline{s}}_i^\prime \!\sim\!  \overline{\tau}_\nu(\cdot | f_\phi({s}_i), g_\eta({s}_i, {a}_i))$. The final loss function is obtained as  $\mathcal{L}_\text{lax}(\phi, \eta) + \mathcal{L}_\text{h}(\phi, \eta, \rho, \nu)$.

\section{Experiments}
\label{sec:exp}
In our experiments, we aim to answer the following key questions:
\begin{enumerate}[noitemsep,nosep]
    \item Does the homomorphic policy gradient improve policy optimization?
    \item What are the qualitative properties of the learned representations and the abstract MDP?
    \item Can DHPG learn and recover the minimal MDP image from raw pixel observations?
\end{enumerate}

We evaluate DHPG on continuous control tasks from DM Control on state and pixel observations. Importantly, to reliably evaluate our algorithm against the baselines and to correctly capture the distribution of results, we follow the best practices proposed by Agarwal et al. \cite{agarwal2021deep} and report the interquartile mean (IQM) and performance profiles aggregated on all tasks over 10 random seeds. While our baseline results are obtained using the official code, when possible 
\footnote{We use the official implementations of DrQv2, DBC, and SAC-AE, while we re-implement DeepMDP due to the unavailability of the official code. See Appendix \ref{sec:baseline_impl} for full details.}, some of the results may differ from the originally reported ones due to the difference in the seed numbers and our goal to present a faithful representation of the true performance distribution \cite{agarwal2021deep}.

\subsection{State Observations}
\label{sec:results_states}
\begin{wrapfigure}{R}{0.5\textwidth}
\vspace{-2em}
\begin{center}
    \centering
    \hfill
    \begin{subfigure}[b]{0.245\textwidth}
         \centering
         \includegraphics[width=\textwidth]{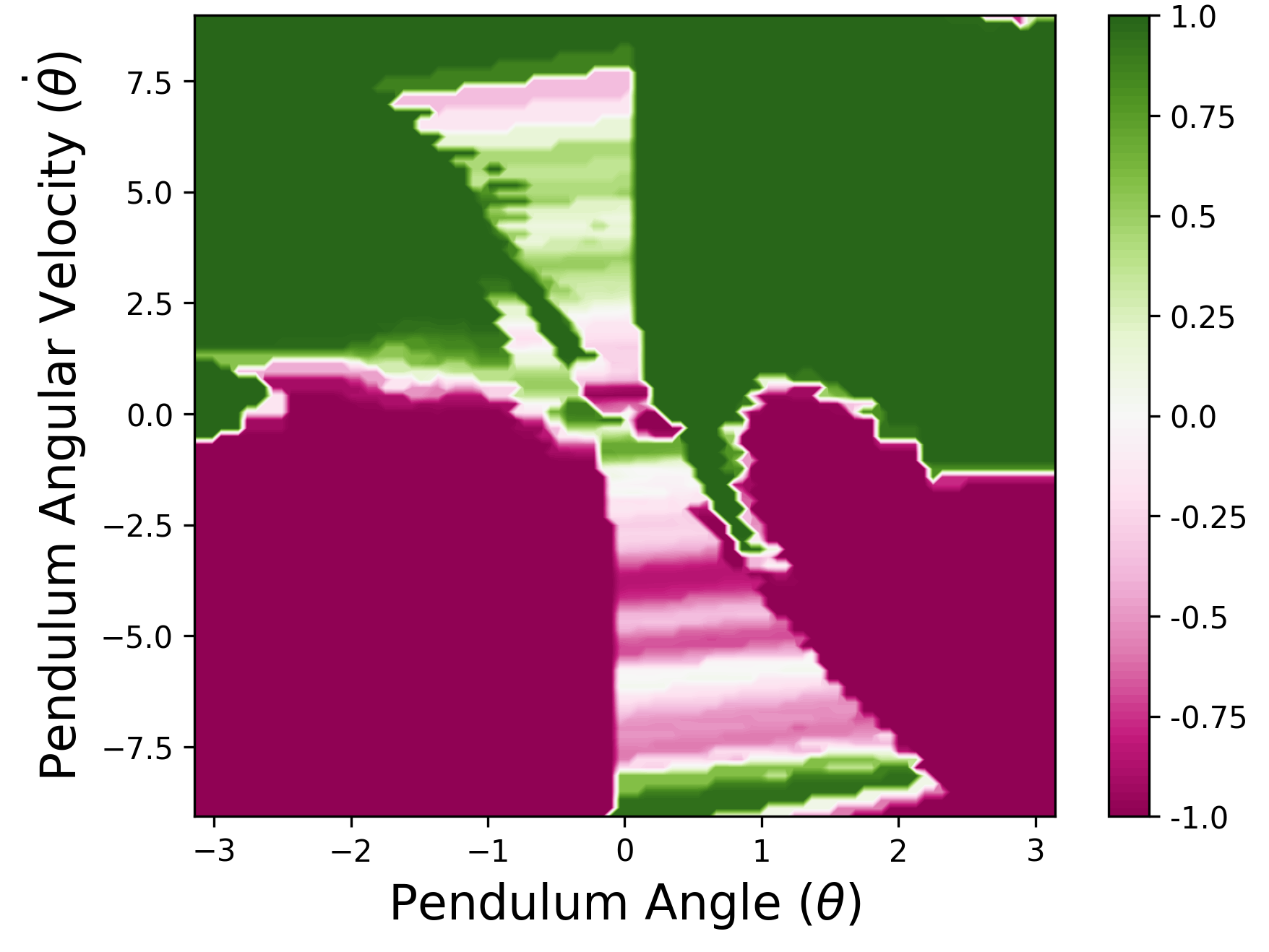}
         \caption{\rebuttal{Actual optimal policy.}}
         \label{fig:states_pendulum_action_contour_a}
    \end{subfigure}
    \hfill
    \begin{subfigure}[b]{0.245\textwidth}
         \centering
         \includegraphics[width=\textwidth]{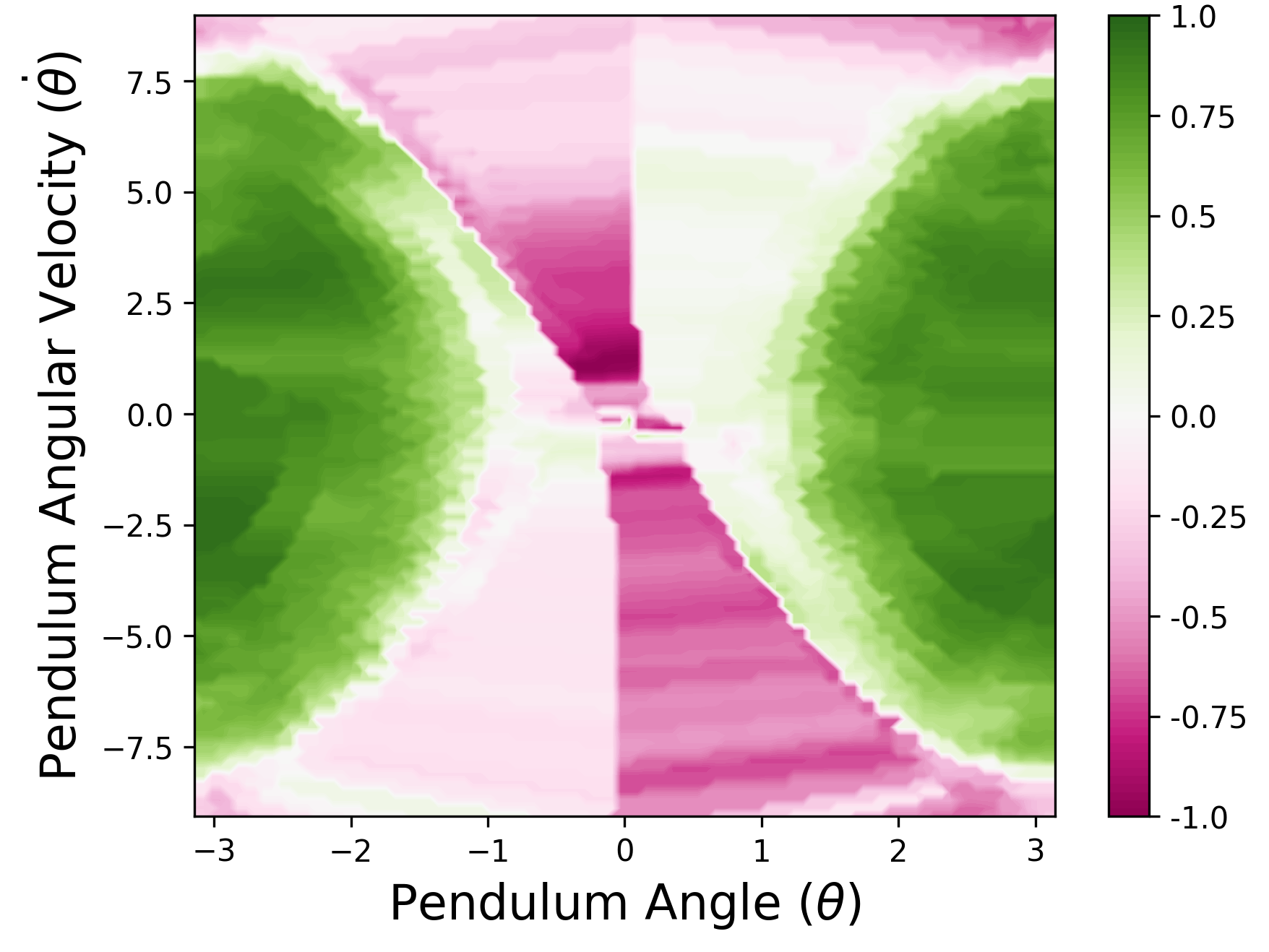}
         \caption{\rebuttal{Abstract optimal policy.}}
         \label{fig:states_pendulum_action_contour_b}
    \end{subfigure}
    \hfill
    \vspace{-1em}
    \caption{Contours of actual and abstract \rebuttal{optimal} actions over the state space of the pendulum-swingup task. Colors represent action values, and states are $s \!=\! (\theta, \dot{\theta})$. \textbf{(a)} \rebuttal{Actual optimal policy;  contours of optimal actions $a^* \!=\! \pi^{\uparrow^*}\!(s)$}. \textbf{(b)} \rebuttal{Abstract optimal policy; contours of abstract optimal actions $\overline{a}^* \!=\! g_s(a^*) \!=\! \overline{\pi}^*(\overline{s})$}. The relation $g_{s_1}\!(a_1) \!=\! g_{s_2}\!(a_2)$ holds for equivalent state-action pairs, \rebuttal{and the abstract optimal policy is symmetric.}}
    \label{fig:states_pendulum_action_contour}
    \vspace{-2em}
\end{center}
\end{wrapfigure}
We compare DHPG on state observations against three commonly-used off-policy model-free algorithms: DDPG \cite{lillicrap2015continuous}, TD3 \cite{fujimoto2018addressing}, and SAC \cite{haarnoja2018soft}. All methods use $n$-step returns, and share the same hyperparameters presented in Appendix \ref{sec:hyperparams}. 
For a fair comparison with DDPG and a better study of the impact of HPG, we have improved our DDPG by adding delayed policy updates and target policy \cite{fujimoto2018addressing}. Thus, the only difference between our DDPG and TD3 is the clipped double Q-learning present in TD3, which appears to be hurting the performance in some tasks of DMC as also observed in \cite{pardo2020tonic, QingLi2021continuousbenchmark}. 

\textbf{DHPG outperforms or matches other algorithms on state observations and has a better sample efficiency.} Results are presented in Figure \ref{fig:states_results}, and \emph{full results are in Appendix \ref{sec:additional_results_states}}. Expectedly, performance gains are larger on tasks with symmetries, as DHPG is able to learn a compressed abstract MDP.

\textbf{The learned mapping ${h \!\!=\!\! (f, g_s\!)}$ demonstrates properties of an MDP homomorphism.} We use the pendulum swingup task to visualize its learned MDP homomorphism, as its symmetries are perfectly intelligible. Two state-action pairs $(s_1\!=\!(\theta_1, \dot{\theta}_1), a_1)$ and $(s_2\!=\! (\theta_2, \dot{\theta}_2), a_2)$ are equivalent if $a_1 \!=\! -a_2$, $\theta_1 \!=\! -\theta_2$, and $\dot{\theta}_1 \!=\! -\dot{\theta}_2$. Therefore, the learned action representations are expected to reflect this by setting $g_{s_1}(a_1) \!=\! g_{s_2}(a_2)$. Figure \ref{fig:states_pendulum_action_contour_a} shows contours of optimal actions over $\mathcal{S}$, while Figure \ref{fig:states_pendulum_action_contour_b} shows action representations $\overline{a} \!=\! g_s(a)$ of optimal actions over $\mathcal{S}$. Clearly, abstract actions adhere to the aforementioned relation for equivalent state-action pairs, indicating $g_s(a)$ is in fact representing the action encoder of an MDP homomorphism mapping. 

\subsection{Pixel Observations}
\label{sec:results_pixels}
We compare the effectiveness of DHPG on pixel observations against DBC \cite{zhang2020learning}, DeepMDP \cite{gelada2019deepmdp}, SAC-AE \cite{yarats2021improving}, and state-of-the-art performing DrQ-v2 \cite{yarats2021mastering}. All methods use $n$-step returns, share the same hyperparameters in Appendix \ref{sec:hyperparams} and all hyperparameters are adapted from DrQ-v2 \emph{without any further tuning}. Importantly, for a fair comparison with DrQ-v2 which uses image augmentation, we present two variations of DHPG and other baselines, \emph{with and without image augmentation}.

\begin{figure}[t!]
    \centering
    \begin{subfigure}[b]{0.24\textwidth}
         \centering
         \includegraphics[width=\textwidth]{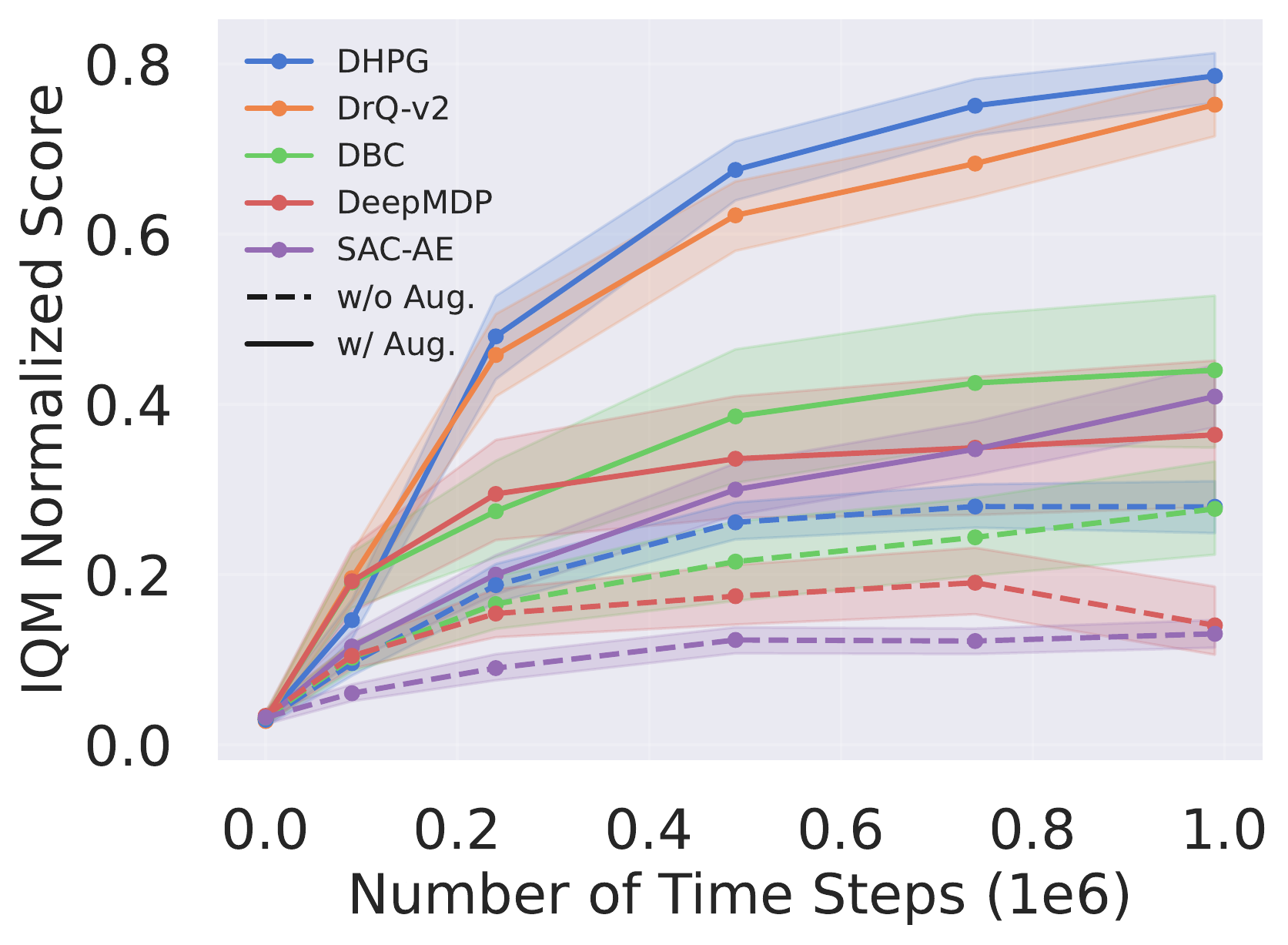}
         \caption{Sample efficiency.}
         \label{fig:pixels_sample_efficiency}
    \end{subfigure}
    \begin{subfigure}[b]{0.24\textwidth}
         \centering
         \includegraphics[width=\textwidth]{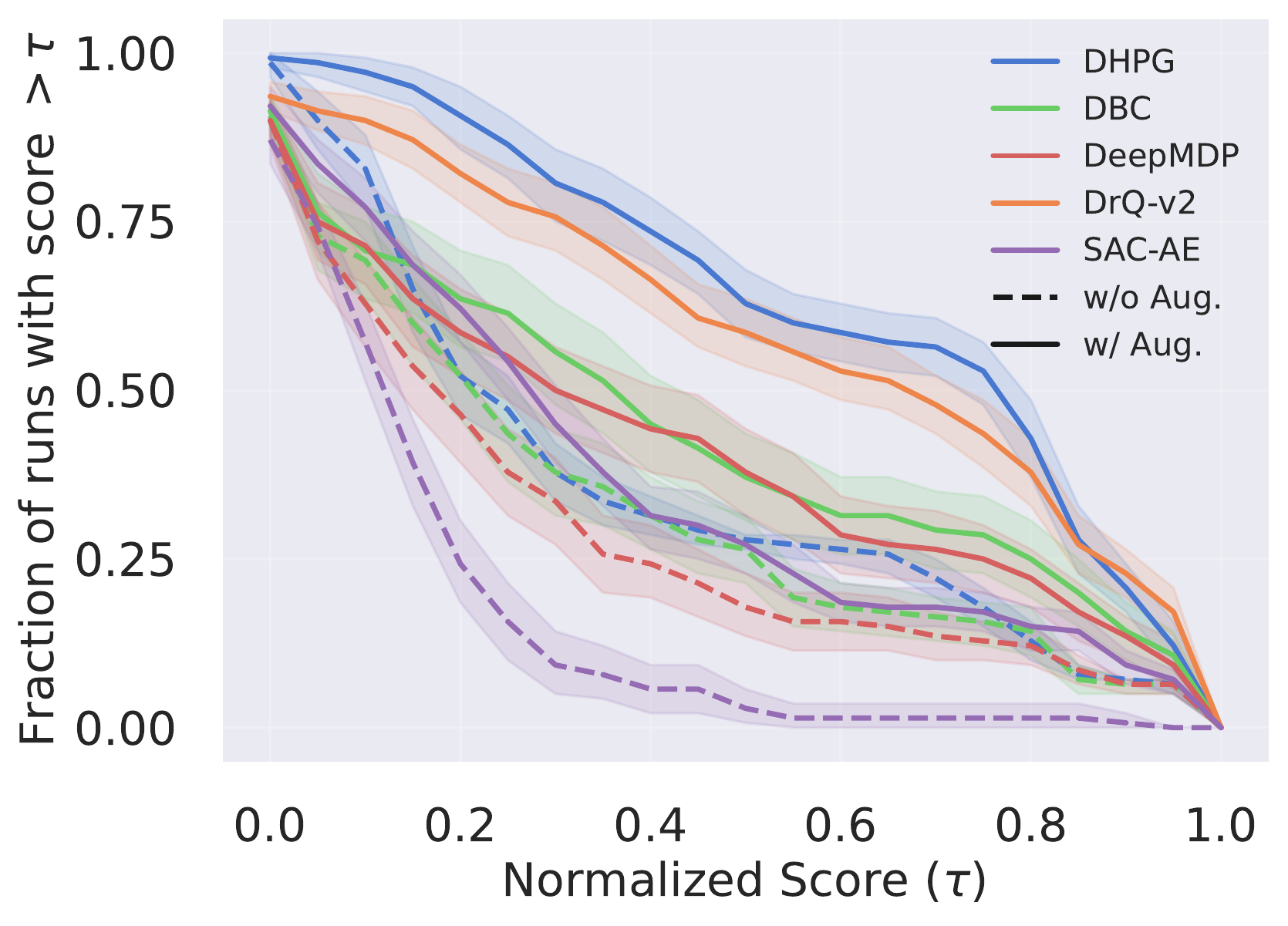}
         \caption{Performance profiles.}
         \label{fig:spixels_perf_profile_main}
    \end{subfigure}
    \begin{subfigure}[b]{0.24\textwidth}
         \centering
         \includegraphics[width=\textwidth]{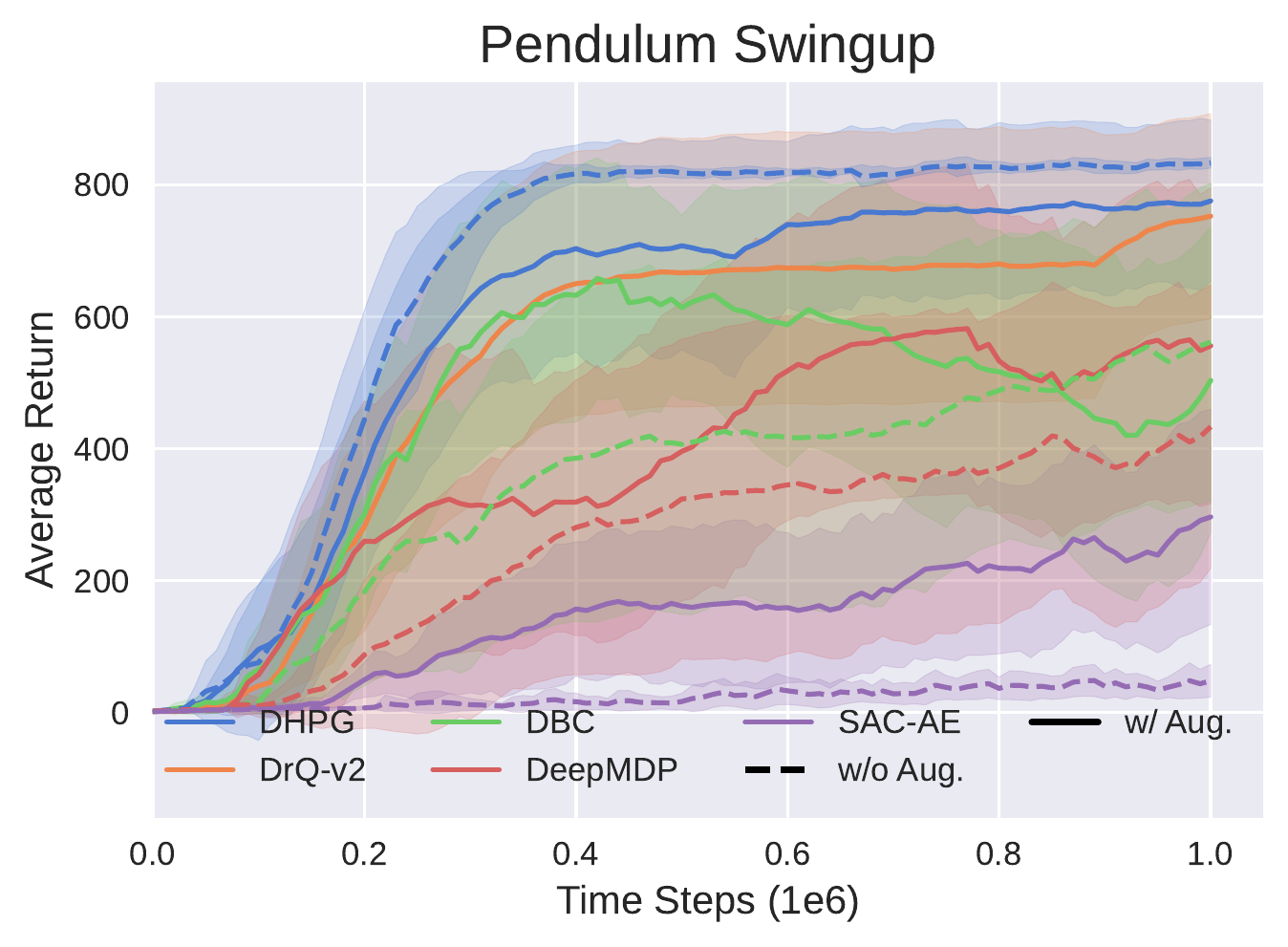}
         \caption{Learning curves.}
         \label{fig:pixels_res_1}
    \end{subfigure}
    \begin{subfigure}[b]{0.24\textwidth}
         \centering
         \includegraphics[width=\textwidth]{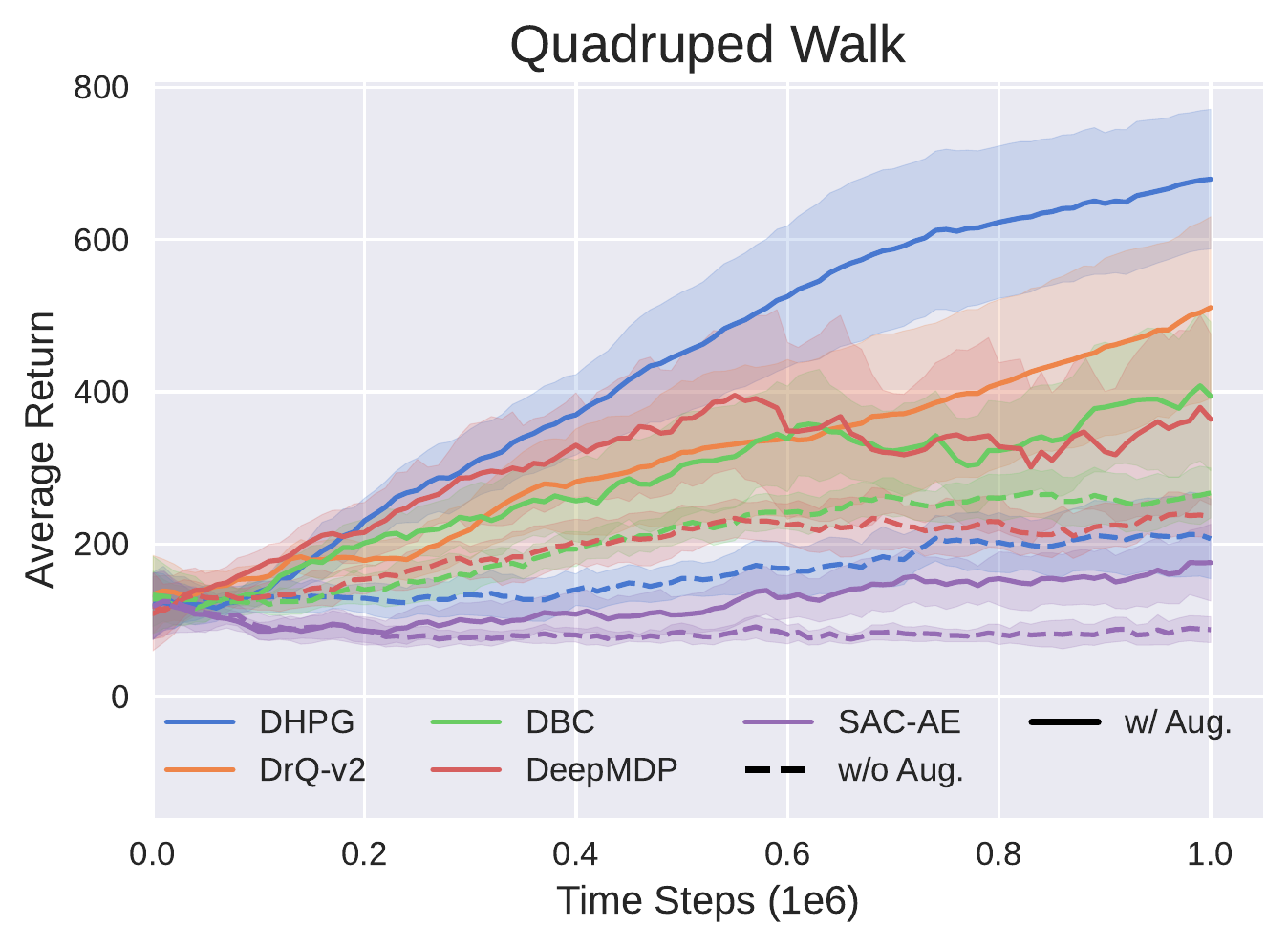}
         \caption{Learning curves.}
         \label{fig:pixels_res_2}
    \end{subfigure}
    \caption{Results of DM Control tasks with \textbf{pixel observations} obtained on 10 seeds. RLiable metrics are aggregated over 14 tasks. \textbf{(a)} RLiable IQM scores as a function of number of steps for comparing sample efficiency, \textbf{(b)} RLiable performance profiles at 500k steps, \textbf{(c)}-\textbf{(d)} examples of learning curves. Full results are in Appendix \ref{sec:additional_results_pixels}. Shaded regions represent $95\%$ confidence intervals.}
    \label{fig:pixels_results}
\end{figure}

\begin{wrapfigure}{R}{0.5\textwidth}
\vspace{-2em}
    \begin{center}
    \begin{subfigure}[b]{0.175\textwidth}
        \centering
        \includegraphics[width=\textwidth]{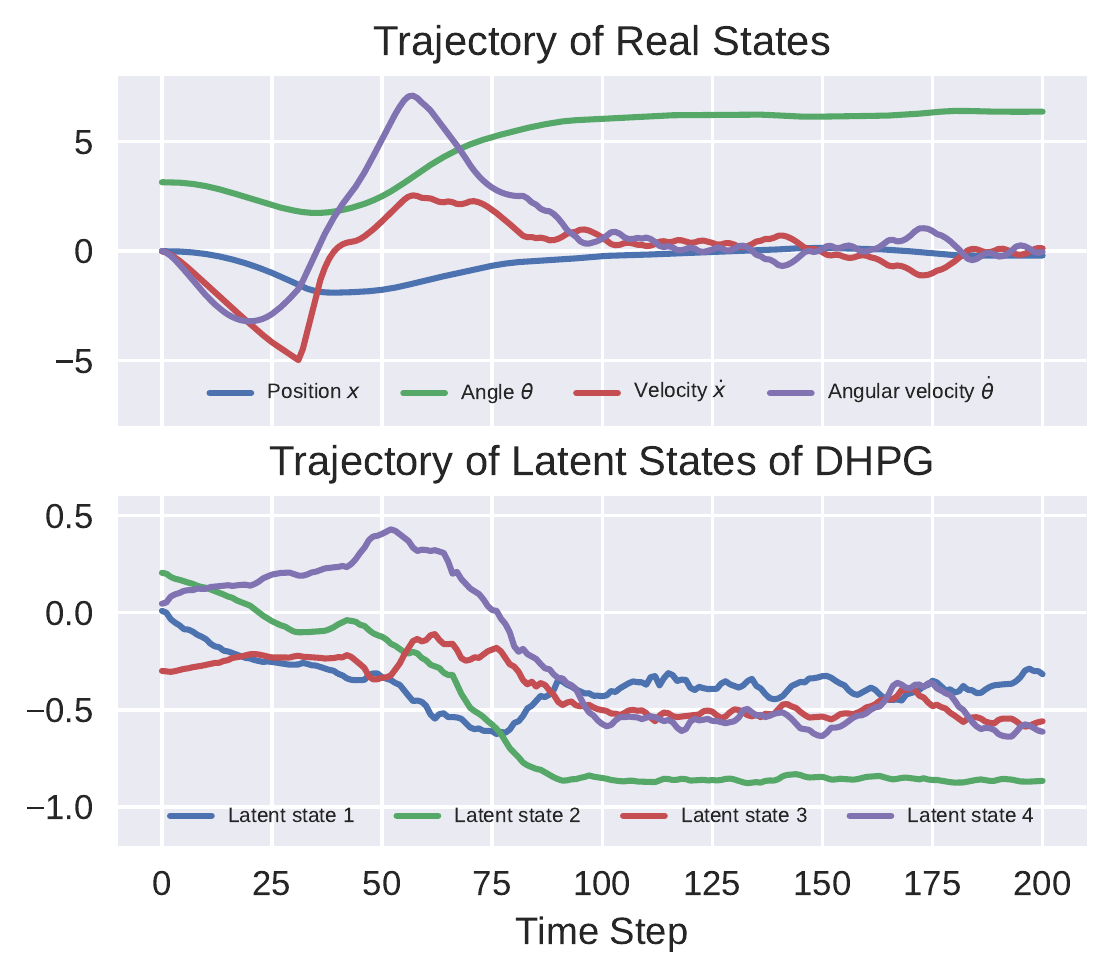}
        \caption{Trajectories.}
        \label{fig:low_dim_results_traj}
    \end{subfigure}
    \hfill
    \begin{subfigure}[b]{0.145\textwidth}
        \centering
        \includegraphics[width=\textwidth]{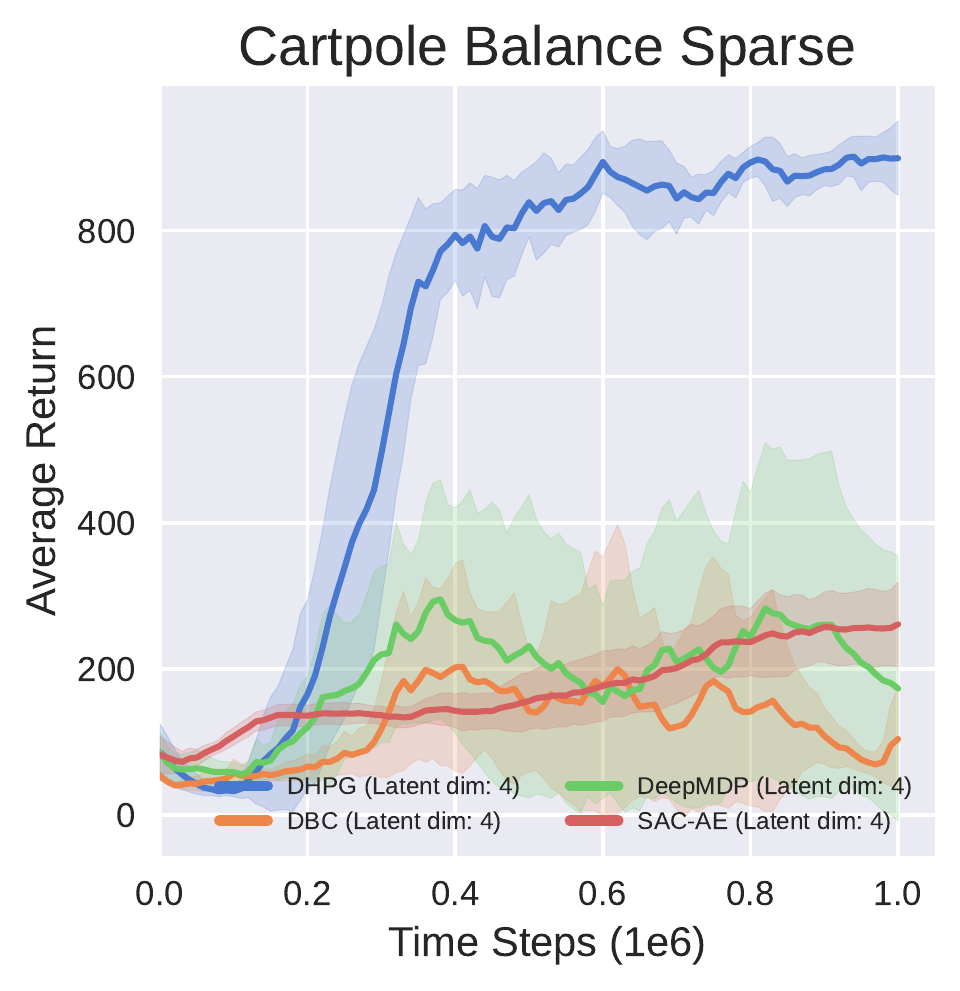}
        \caption{Curves.}
    \end{subfigure}
    \hfill
    \begin{subfigure}[b]{0.145\textwidth}
        \centering
        \includegraphics[width=\textwidth]{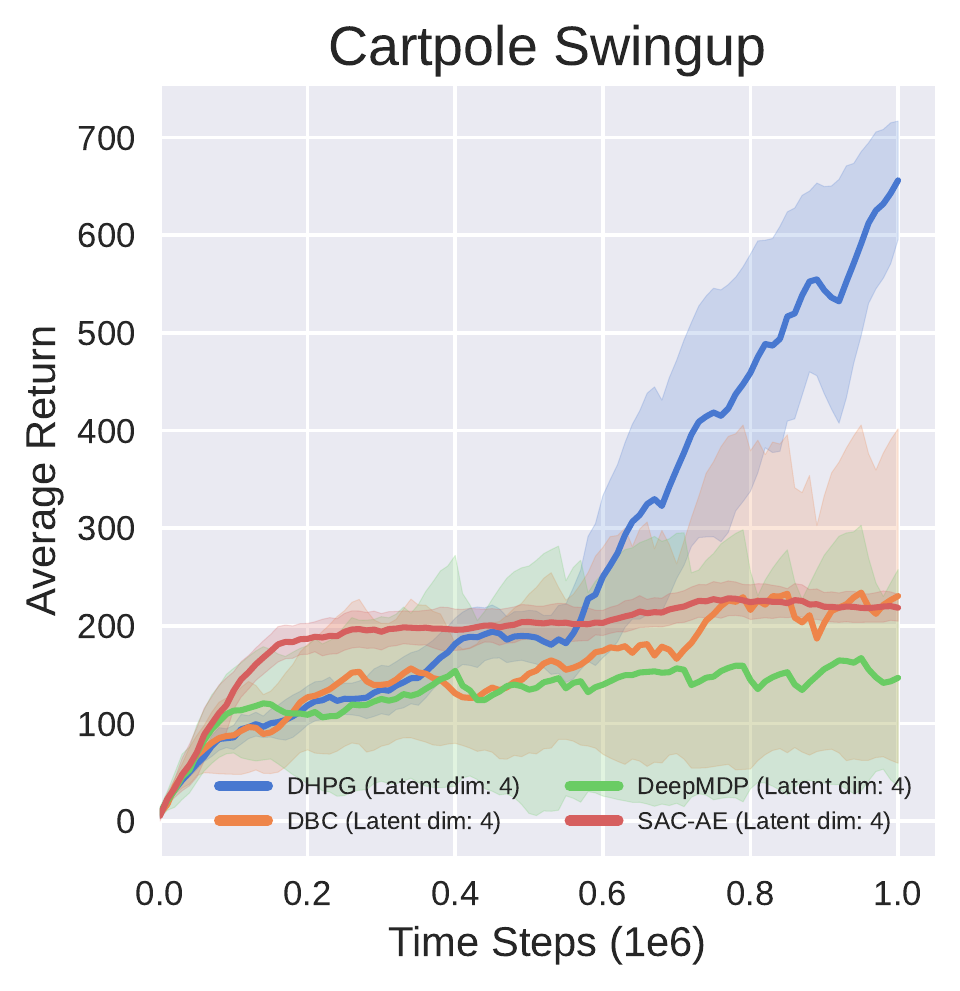}
        \caption{Curves.}
    \end{subfigure}
    \caption{Effectiveness of DHPG in recovering the minimal MDP from \textbf{pixels}. All methods are limited to a 4-dimensional latent space which is equal to the dimensions of the real state space of cartpole. \textbf{(a)} Trajectories of real states obtained from Mujoco and trajectories of latent states of DHPG. \textbf{(b, c)} Learning curves averaged on 10 seeds.}
    \label{fig:low_dim_results}
\end{center}
\vspace{-1.5em}
\end{wrapfigure}

\textbf{DHPG outperforms or matches other algorithms on pixel observations, demonstrating its effectiveness in representation learning.} Results are presented in Figure \ref{fig:pixels_results} and \emph{full results are in Appendix \ref{sec:additional_results_pixels}}. Interestingly, DHPG without image augmentation outperforms DrQ-v2 on domains with easily learnable MDP homomorphism maps, such as cartpole and pendulum, showing its power of representation learning.

\textbf{DHPG can learn and recover a low-dimensional MDP image.} A key strength of MDP homomorphisms is their ability to represent the minimal MDP image \cite{ravindran2001symmetries}, which is particularly important when learning from pixel observations. To demonstrate this ability, we have limited the latent space dimensions to the dimension of the real system and compared DHPG (without image augmentation) with baselines in Figure \ref{fig:low_dim_results}. While other methods are not able to learn the tasks, DHPG can successfully learn the policy and the minimal low-dimensional latent space. Surprisingly, trajectories of the latent states resemble that of the real states as shown in Figure \ref{fig:low_dim_results_traj}.

\begin{figure}[b!]
    \centering
    \hfill
    \begin{subfigure}[b]{0.24\textwidth}
        \centering
        \includegraphics[width=\textwidth]{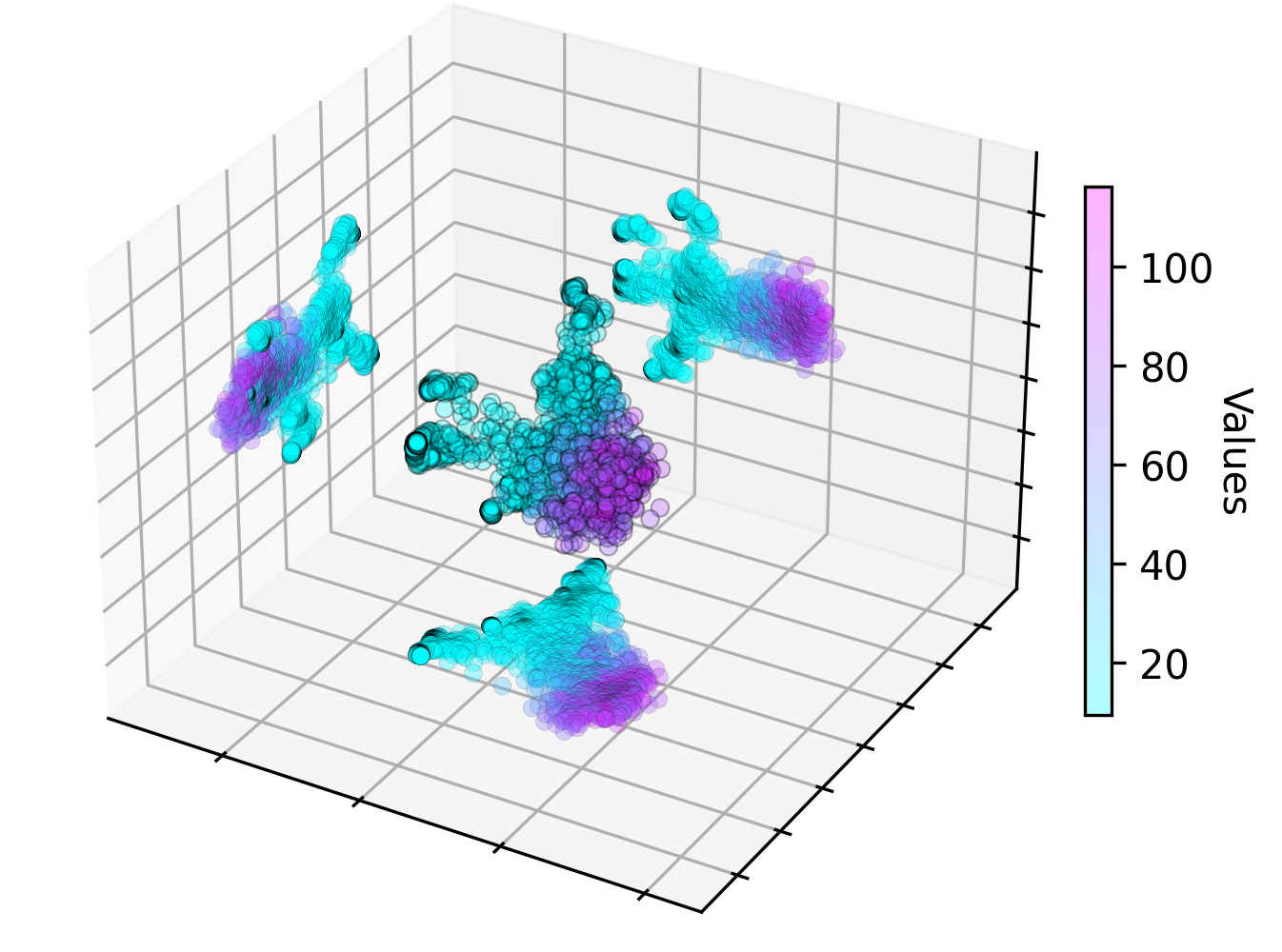}
        \caption{Latent states, DHPG.}
        \label{fig:pixels_quadruped_mdp_vis_a}
    \end{subfigure}
    \hfill
    \begin{subfigure}[b]{0.24\textwidth}
        \centering
        \includegraphics[width=\textwidth]{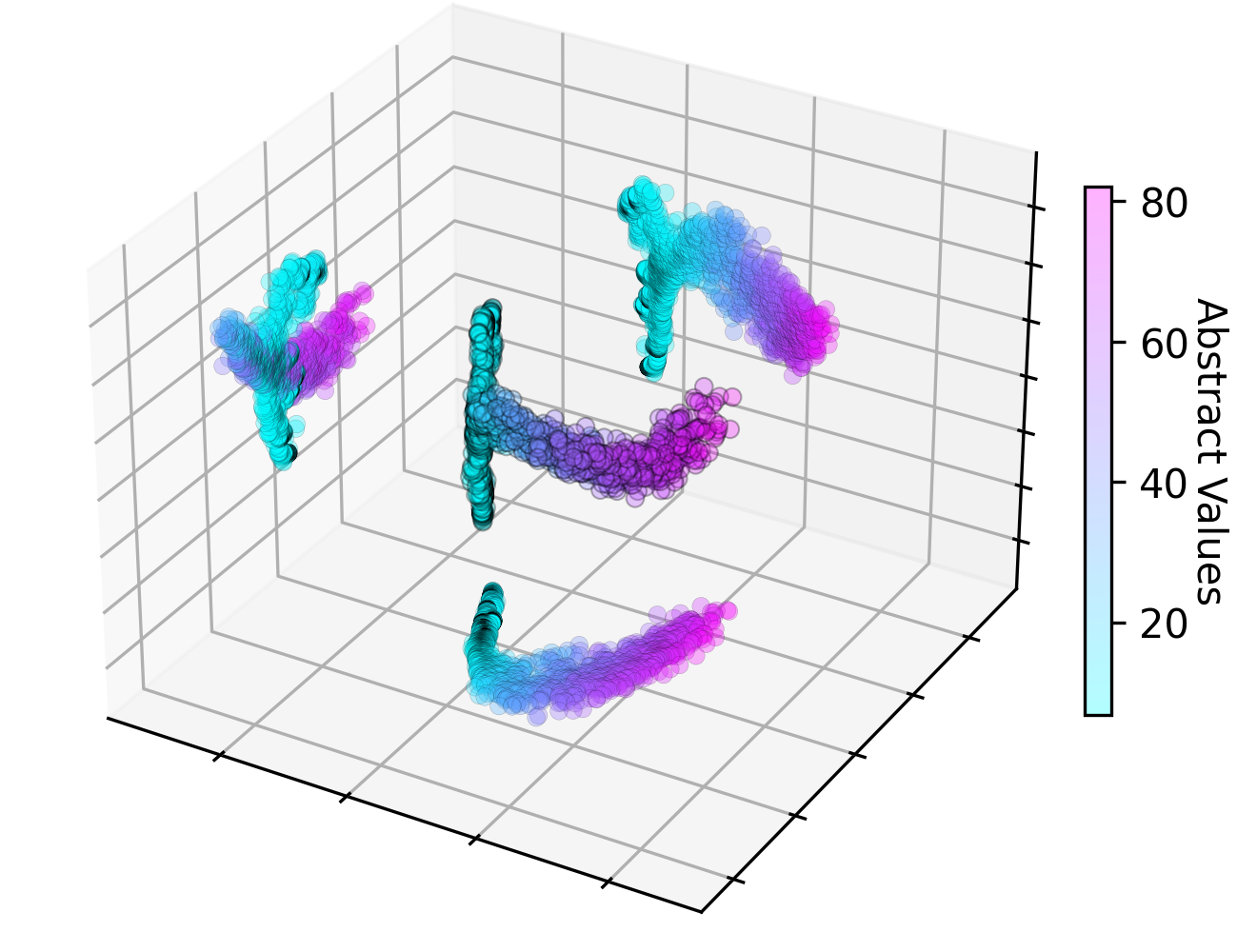}
        \caption{Abstract states, DHPG.}
        \label{fig:pixels_quadruped_mdp_vis_b}
    \end{subfigure}
    \hfill
    \begin{subfigure}[b]{0.24\textwidth}
        \centering
        \includegraphics[width=\textwidth]{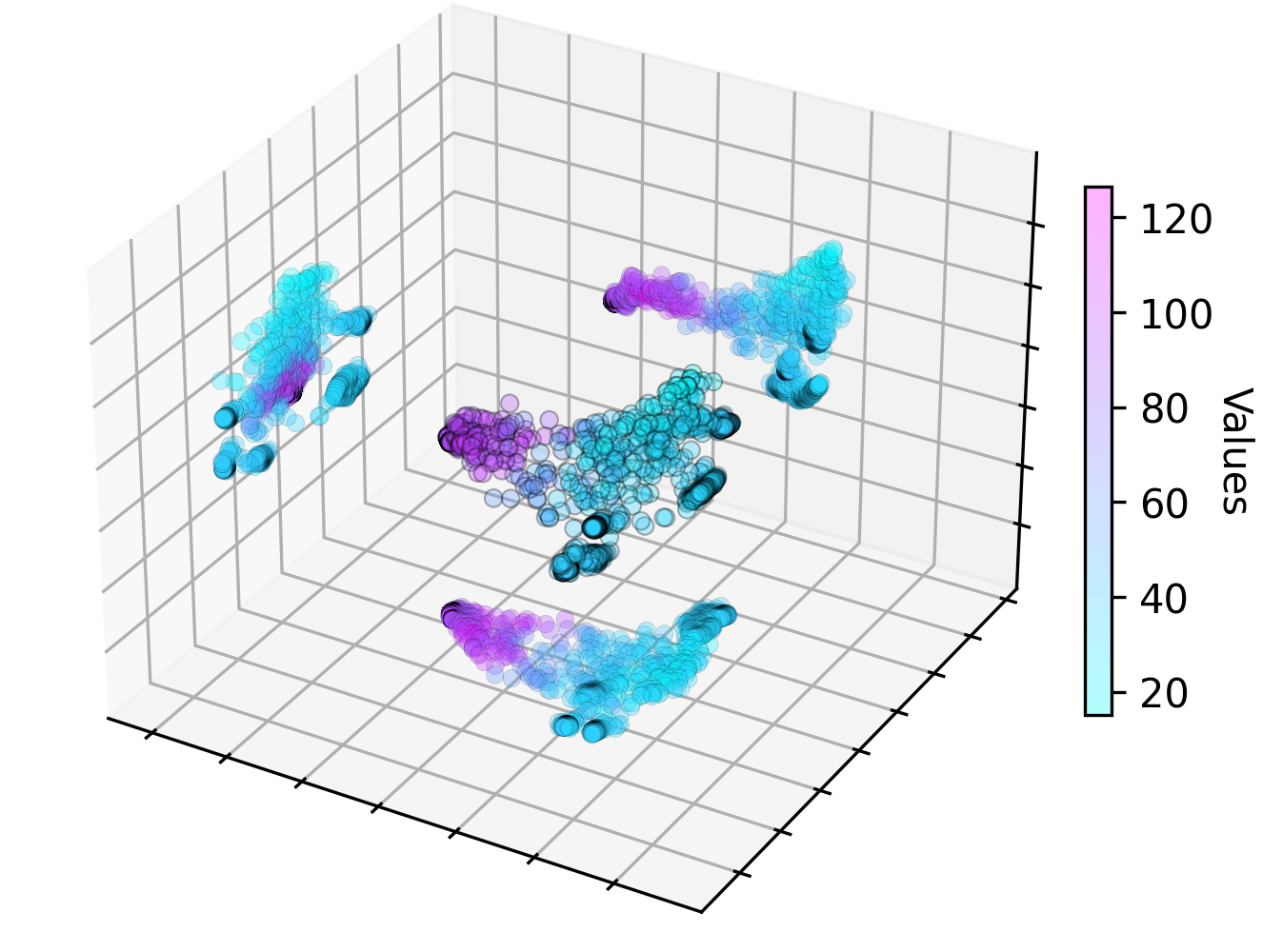}
        \caption{Latent states, DBC.}
    \end{subfigure}
    \hfill
    \begin{subfigure}[b]{0.24\textwidth}
        \centering
        \includegraphics[width=\textwidth]{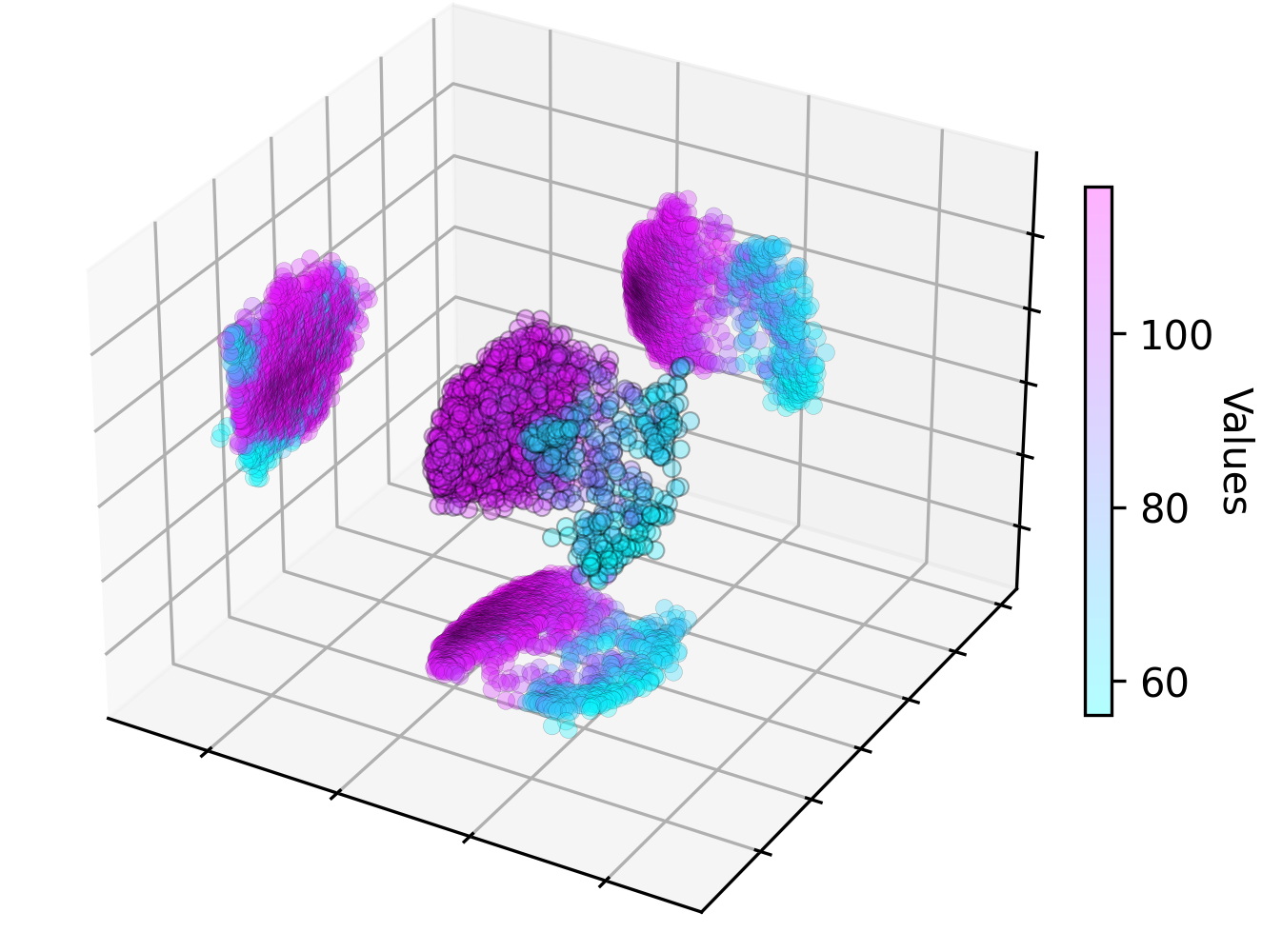}
        \caption{Latent States, DrQ-v2.}
    \end{subfigure}
    \hfill
    \caption{PCA projection of learned representations for quadruped-walk with \textbf{pixel observations}. \textbf{(a)} Latent states $s \!=\! E_\mu(o)$, \textbf{(b)} abstract latent states $\overline{s} \!=\! f_\phi(E_\mu(o))$ for DHPG, \textbf{(c)} latent states $s \!=\! E_\mu(o)$ for DBC, and \textbf{(d)} DrQ-v2. Color of each point denotes its value learned by $Q(s, a)$ or $\overline{Q}(\overline{s}, \overline{a})$. Points are also projected onto each main plane. The homomorphism map of DHPG has mapped \rebuttal{the latent states of corresponding legs (e.g., left forward leg and right backward leg)} \textbf{(a)} on to the same abstract latent states \textbf{(b)}, indicating a clear structure in ${\overline{\mathcal{S}}}$.}
    \label{fig:pixels_quadruped_mdp_vis}
\end{figure}

\textbf{The abstract MDP demonstrates properties of an MDP homomorphic image.} To qualitatively demonstrate the significance of learning joint state-action representations, Figure \ref{fig:pixels_quadruped_mdp_vis} shows visualizations of latent states for quadruped-walk, a task with symmetries around movements of its four legs. Interestingly, while the latent space of DHPG (Figure \ref{fig:pixels_quadruped_mdp_vis_a}) shows distinct states for each leg, abstract state encoder $f_\phi$ has mapped \rebuttal{corresponding legs (e.g., left forward leg and right backward leg)} to the same abstract latent state (Figure \ref{fig:pixels_quadruped_mdp_vis_b}) as they are some homomorphic image of one another. Clearly, DBC and DrQ-v2 are not able to achieve this.

\textbf{The learned representations and the MDP homomorphism map transfer to new tasks within the same domain.} Importantly, one consideration with representation learning methods relying on rewards is the transferability of the learned representations to \rebuttal{a new reward setting within the same domain}. To ensure that our method does not hinder such transfer, we have carried out experiments in which the actor, critics, and the learned MDP homomorphism map are transferred to another task from the same domain. Results, given in Appendix \ref{sec:transfer_supp} show that our method has not compromised transfer abilities. 

\begin{figure}[t!]
    \centering
    \begin{subfigure}[b]{0.24\textwidth}
         \centering
         \includegraphics[width=\textwidth]{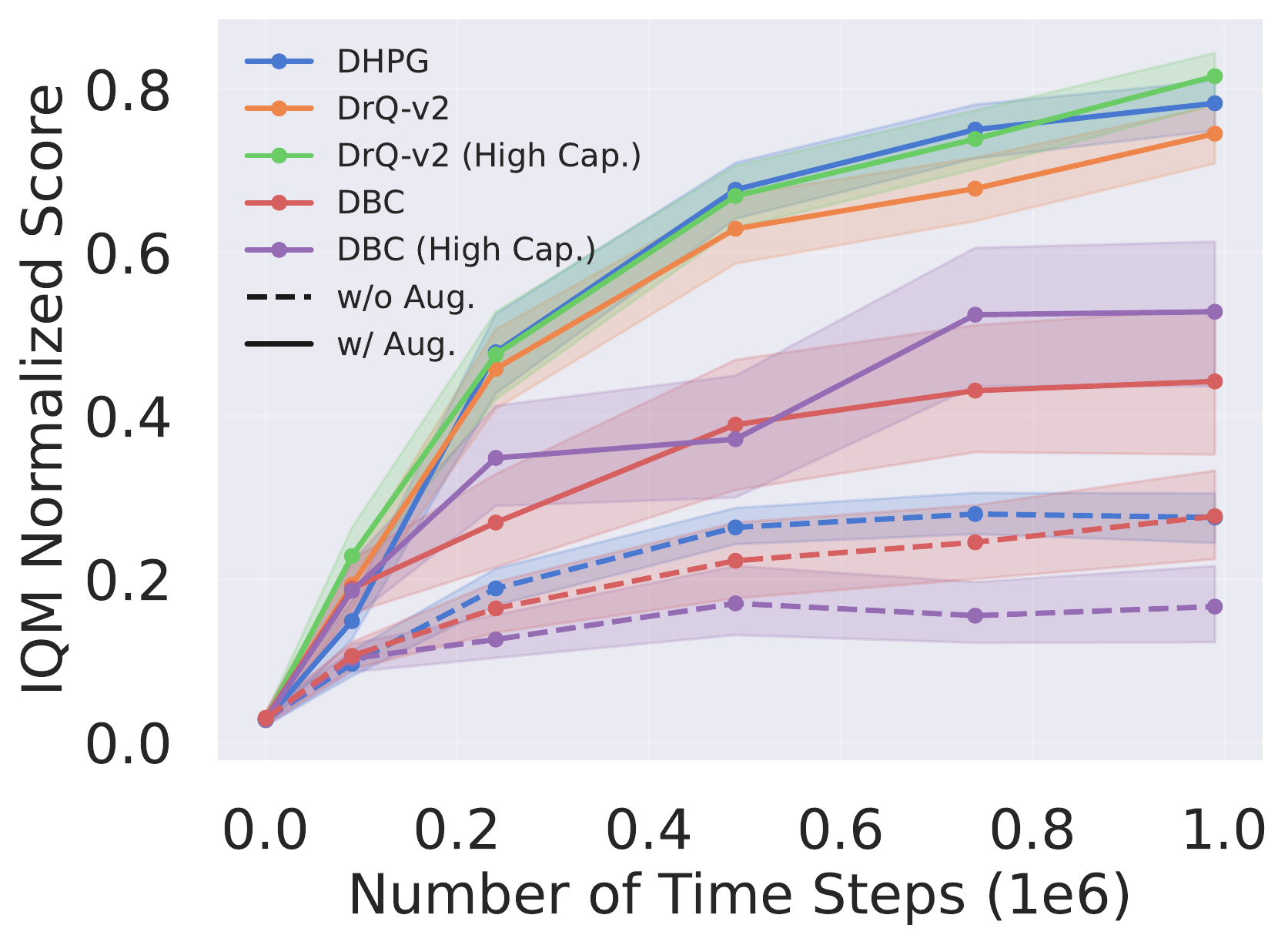}
         \caption{Sample efficiency.}
         \label{fig:high_cap_sample_efficiency}
    \end{subfigure}
    \begin{subfigure}[b]{0.24\textwidth}
         \centering
         \includegraphics[width=\textwidth]{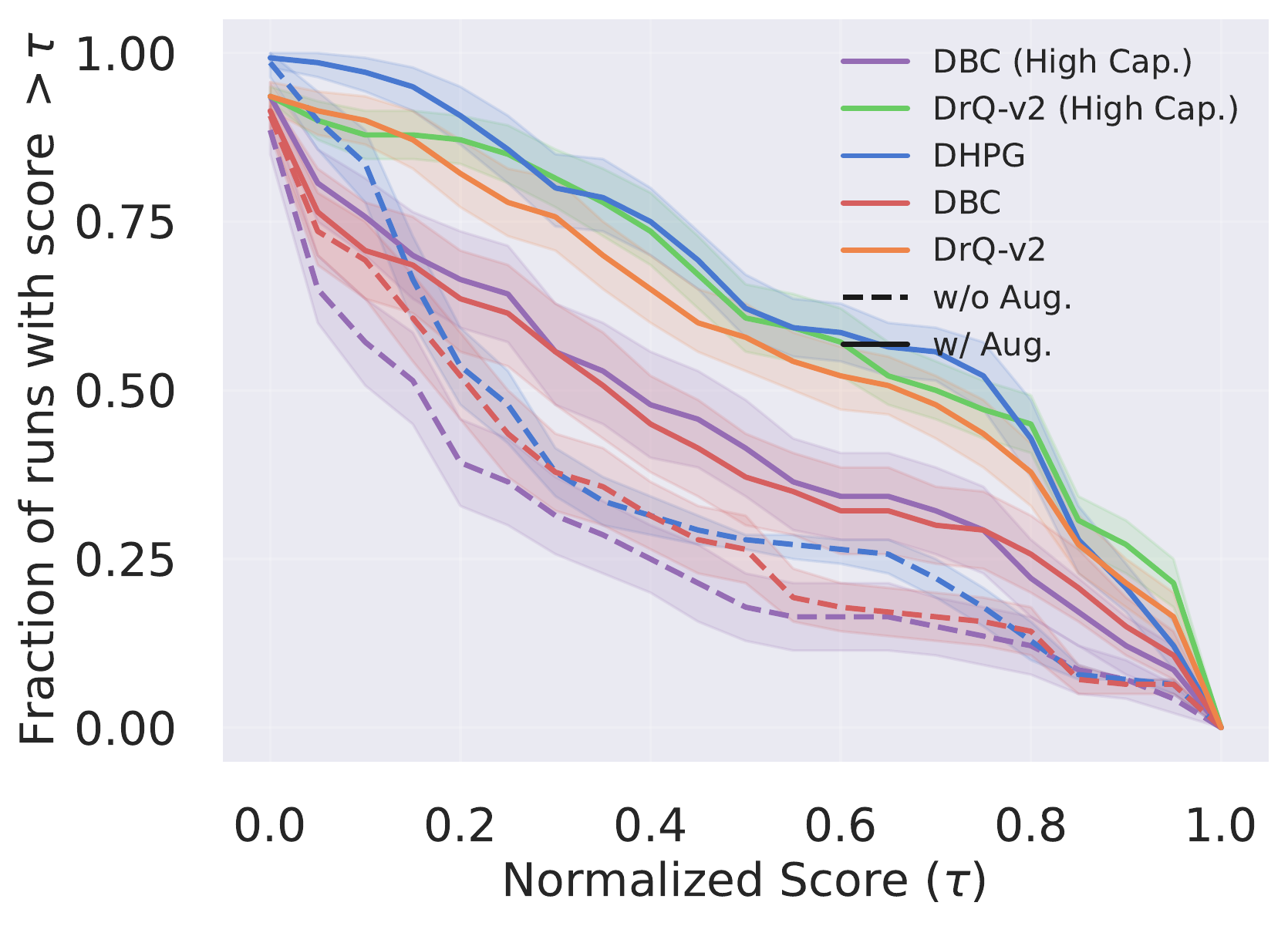}
         \caption{Performance profiles.}
         \label{fig:high_cap_perf_profile_main}
    \end{subfigure}
    \begin{subfigure}[b]{0.24\textwidth}
         \centering
         \includegraphics[width=\textwidth]{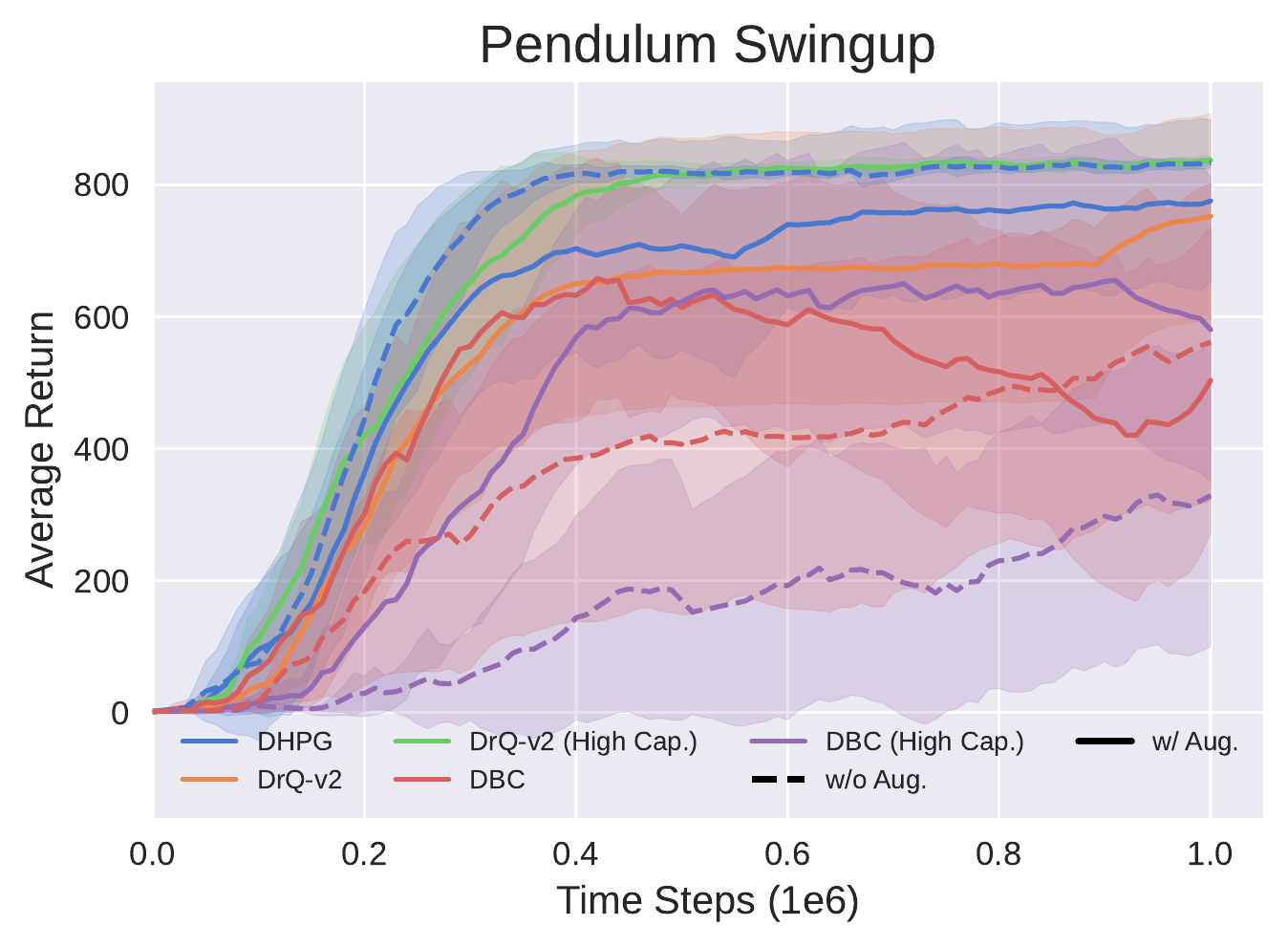}
         \caption{Learning curves.}
         \label{fig:high_cap_res_1}
    \end{subfigure}
    \begin{subfigure}[b]{0.24\textwidth}
         \centering
         \includegraphics[width=\textwidth]{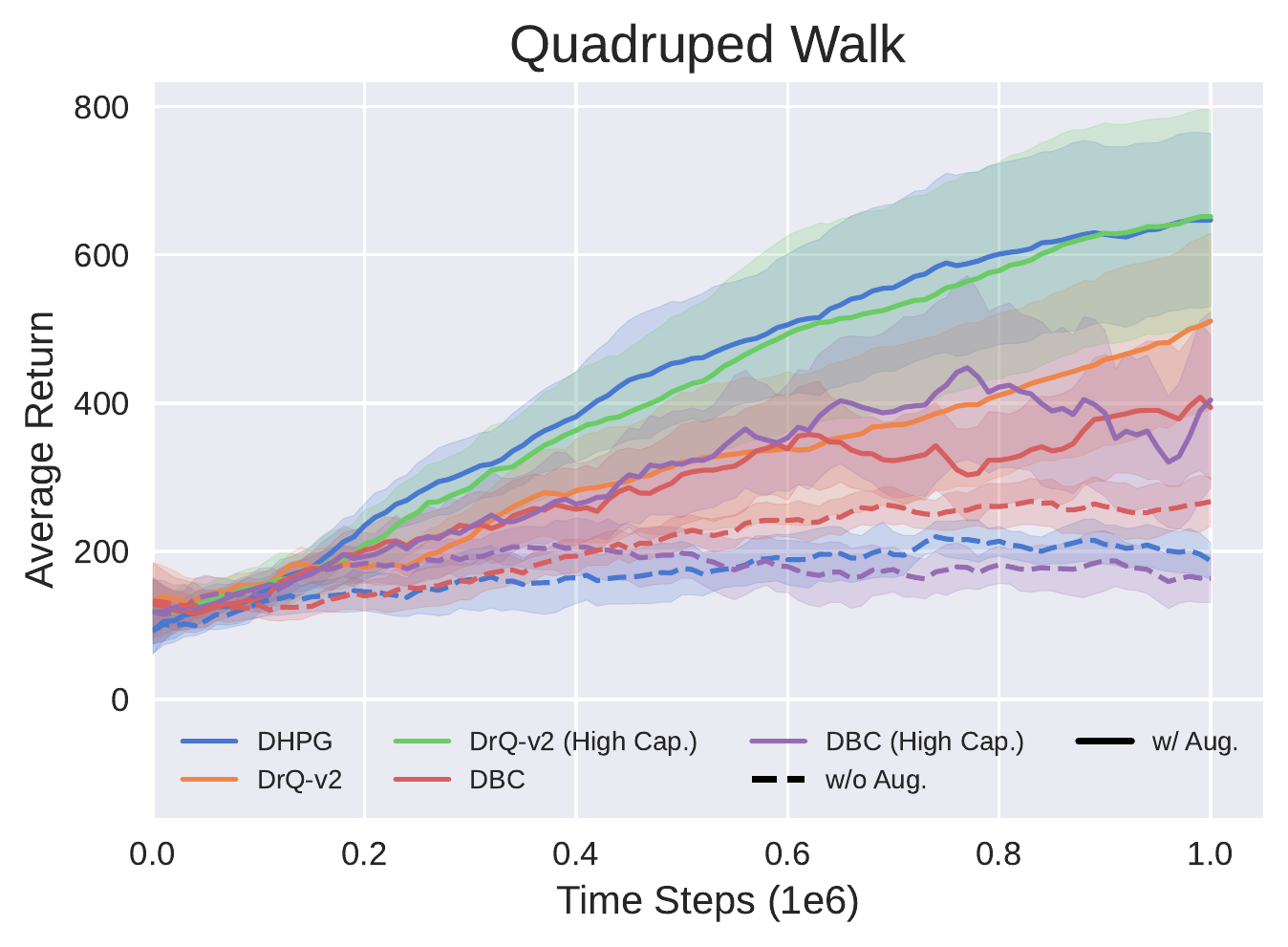}
         \caption{Learning curves.}
         \label{fig:high_cap_res_2}
    \end{subfigure}
    \caption{\rebuttal{Results of DM Control tasks with \textbf{pixel observations} for \textbf{higher-capacity variants} of DBC and DrQ-v2 obtained on 10 seeds. RLiable metrics are aggregated over 14 tasks. \textbf{(a)} RLiable IQM scores as a function of number of steps for comparing sample efficiency, \textbf{(b)} RLiable performance profiles at 500k steps, \textbf{(c)}-\textbf{(d)} examples of learning curves. Full results are in Appendix \ref{sec:high_capacity_supp}. Shaded regions represent $95\%$ confidence intervals.}}
    \label{fig:high_capacity_results}
\end{figure}

\rebuttal{
\textbf{Accounting for the larger network capacity of DHPG compared to the baselines.} Since our DHPG algorithm contains additional networks, such as the parameterized MDP homomorphism map and the abstract critic, it may have a higher network capacity compared to the baselines. To control for the effect of the network capacity and for a fair evaluation, we compare DHPG with higher-capacity variants of DBC and DrQ-v2 that have a larger critic networks, selected such that the total number of parameters are considerably more than that of DHPG. Results are presented in Figure \ref{fig:high_capacity_results}, while \emph{full results and a detailed description of the total number of parameters are in Appendix \ref{sec:high_capacity_supp}}. As suggested by the results, DHPG outperforms or matches the performance of the higher-capacity baselines, demonstrating the improved performance is rather due to the use of the abstract MDP homomorphic image for representation learning and performing HPG updates. 
}

\textbf{Additional Experiments.}
We study the value equivalence property as a measure for the quality of the learned MDP homomorphisms in Appendix \ref{sec:value_equiv_supp}, and we present ablation studies on DHPG variants, and the impact of $n$-step return on our method in Appendices \ref{sec:ablation_dhpg_variants} and \ref{sec:ablation_n_step}, respectively.

\section{Related Work}
\label{sec:related}
\textbf{State Abstraction.} Bisimulation \cite{milner1989communication,
  larsen1991bisimulation} is a notion of behavioral equivalence between
systems. It was extended to continuous state spaces by Blute et
al. \cite{blute1997bisimulation,Desharnais02} and extended to MDPs by Givan et
al. \cite{givan2003equivalence}.  Bisimulation metrics \cite{Desharnais99b,
  ferns2005metrics, ferns2006methods, ferns2011bisimulation} define a
pseudometric to quantify the degree of behavioural similarity.  Recently, Zhang et al. \cite{zhang2020learning} defined a loss function for learning representations via bisimilarity of latent states, and Kemertas et al. \cite{kemertas2021towards} have further improved its robustness. Castro \cite{castro2020scalable} has proposed a method to approximate the bisimulation metric for deterministic MDPs with continuous states but discrete actions. van der Pol et al. \cite{van2020plannable} have defined a contrastive loss based on MDP homomorphisms for learning an abstract MDP for planning, however, their method is only applicable to finite MDPs. Another approach is to directly embed the MDP homomorphic relation in the network architecture \cite{van2020mdp, van2021multi}. Other recently proposed methods seek to learn representations that preserve values \cite{grimm2020value, grimm2021proper} or policies \cite{agarwal2020contrastive}, or via a sampling-based similarity metric \cite{castro2021mico}. \rebuttal{Finally, state abstractions can in principle help improve transferring of policies \cite{abel2019state, castro2010using, soni2006using, sorg2009transfer, rajendran2009learning}, or learning temporally extended actions \cite{castro2011automatic, wolfe2006defining, wolfe2006decision, sutton1999between}.}

\textbf{Action Abstraction.} Action representations are often studied in the context of large discrete action spaces \cite{sallans2004reinforcement} as a form of a look-up embedding that is known \emph{a-priori} \cite{dulac2015deep}, factored representations \cite{sharma2017learning}, or policy decomposition \cite{chandak2019learning}. Action representations can also be learned from expert demonstrations \cite{tennenholtz2019natural}. More related to our work is dynamics-aware embeddings \cite{whitney2019dynamics} where a combined state-action embedding for continuous control is learned. In contrast, we use the notion of homomorphisms to learn the state-dependent action representations, while preserving values. Action representations can be combined with temporal abstraction \cite{sutton1999between} for discovering extended actions \cite{ravindran2003relativized, abel2020value, castro2010using, castro2011automatic}.

\textbf{State Representation Learning.} Extant methods for learning the
underlying state space from raw observations often use latent models
\rebuttal{\cite{gelada2019deepmdp, hafner2019dream, hafner2019learning, ha2018world, biza2021learning}},
auxiliary prediction tasks \cite{jaderberg2016reinforcement, liu2019self,
  lyle2021effect}, physics-inspired inductive biases
\cite{jonschkowski2015learning, cranmer2020lagrangian,
  greydanus2019hamiltonian}, unsupervised learning \cite{hjelm2018learning,
  liu2021aps}, or self-supervised learning \cite{anand2019unsupervised,
  sinha2021s4rl, hansen2020self, hansen2021generalization,
  fan2021secant}. From another point of view, representation learning can be
effectively decoupled from the RL problem \cite{eslami2018neural,
  stooke2021decoupling}.  Symmetries of the environment can also be used for
representation learning \rebuttal{\cite{mondal2022eqr, mahajan2017symmetry, park2022learning, wang2021so2, higgins2018towards,
  higgins2021symetric, quessard2020learning, caselles2019symmetry}}. In fact, MDP
homomorphisms are specializations of such approaches for RL.  A key distinguishing factor of MDP homomorphisms is their ability to take actions into account for representation learning in the same premises as Thomas et al. \cite{thomas2017independently}. Recently, simple image augmentation methods have shown significant improvements in RL performance \cite{yarats2020image, lee2019network}. Since these approaches are in general orthogonal to state abstractions, they can be combined together.

\section{Conclusion}
\label{sec:conclusion}
In this paper, we developed the novel theory of continuous MDP homomorphisms using measure theory, and we rigorously proved their value and optimal value equivalence properties. We derived the homomorphic PG in order to directly use a joint state-action abstraction for policy optimization. Importantly, we rigorously proved that applying our homomorphic PG on the abstract MDP is equivalent to applying the standard DPG on the actual MDP. Based on our novel theoretical results, we developed a deep actor-critic algorithm that can simultaneously learn the policy and the MDP homomorphism map using the lax bisimulation metric. Our algorithm improves upon strong baselines in both learning from state and pixel observations. The visualization of the latent space demonstrates the strong potential of MDP homomorphisms in learning structured representations that can  preserve value functions. 
We believe that our work will open-up future possibilities for the application of MDP homomorphisms in challenging continuous control problems.

\section*{Acknowledgements}
SRS is supported by an NSERC CGS-D scholarship. RZ was supported by an NSERC CGS-M scholarship at the time this work was completed. We would like to thank Juan Camilo Gamboa Higuera, Harley Wiltzer, and Scott Fujimoto for insightful discussions.

\bibliography{refs}
\bibliographystyle{plain}

\clearpage

\section*{Checklist}


\begin{enumerate}

\item For all authors...
\begin{enumerate}
  \item Do the main claims made in the abstract and introduction accurately reflect the paper's contributions and scope?
    \answerYes{All the theoretical claims are substantiated in Sections \ref{sec:value_equiv}, \ref{sec:cont_mdp_hom}, and \ref{sec:hpg}. All the empirical claims are substantiated in Section \ref{sec:exp} and Appendix \ref{sec:additional_results}.}  
    \item Did you describe the limitations of your work?
    \answerYes{Theoretical limitations and assumptions are discussed in Section \ref{sec:cont_mdp_hom}, and empirical limitations are discussed in Section \ref{sec:exp}.}
  \item Did you discuss any potential negative societal impacts of your work?
    \answerNo{Our work is foundational research on reinforcement learning and state abstraction, hence we do not foresee any substantive societal and ethical implications.}
  \item Have you read the ethics review guidelines and ensured that your paper conforms to them?
    \answerYes{}
\end{enumerate}

\item If you are including theoretical results...
\begin{enumerate}
  \item Did you state the full set of assumptions of all theoretical results?
    \answerYes{The full set of assumptions are detailed in Section \ref{sec:cont_mdp_hom} and Appendix \ref{sec:assumptions}.}
        \item Did you include complete proofs of all theoretical results?
    \answerYes{The proofs of all theoretical results are given in Appendix \ref{sec:proofs}.}
\end{enumerate}

\item If you ran experiments...
\begin{enumerate}
  \item Did you include the code, data, and instructions needed to reproduce the main experimental results (either in the supplemental material or as a URL)?
    \answerYes{Our code, including the instructions, are submitted in the supplemental material.}
  \item Did you specify all the training details (e.g., data splits, hyperparameters, how they were chosen)?
    \answerYes{Hyperparameters, implementation, and training details are given in Appendix \ref{sec:implementation}}
  \item Did you report error bars (e.g., with respect to the random seed after running experiments multiple times)?
    \answerYes{All experiments are obtained on 10 seeds, and we report confidence intervals, interquartile mean, median, mean, and performance profiles as suggested by Agarwal et al. \cite{agarwal2021deep}.}
  \item Did you include the total amount of compute and the type of resources used (e.g., type of GPUs, internal cluster, or cloud provider)?
    \answerYes{We describe our hardware setup in Appendix \ref{sec:implementation}, but we do not include the amount of compute used.}
\end{enumerate}

\item If you are using existing assets (e.g., code, data, models) or curating/releasing new assets...
\begin{enumerate}
  \item If your work uses existing assets, did you cite the creators?
    \answerYes{The creators are cited appropriately and information of each asset is included in Appendix \ref{sec:baseline_impl}.}
  \item Did you mention the license of the assets?
    \answerNA{All assets are open-source and under permissive open-source licenses.}
  \item Did you include any new assets either in the supplemental material or as a URL?
    \answerYes{We include the code of our algorithm in the supplemental material.}
  \item Did you discuss whether and how consent was obtained from people whose data you're using/curating?
    \answerNA{We are using open-source simulators and reinforcement learning environments.}
  \item Did you discuss whether the data you are using/curating contains personally identifiable information or offensive content?
    \answerNA{We are using open-source simulators and reinforcement learning environments.}
\end{enumerate}

\item If you used crowdsourcing or conducted research with human subjects...
\begin{enumerate}
  \item Did you include the full text of instructions given to participants and screenshots, if applicable?
    \answerNA{}
  \item Did you describe any potential participant risks, with links to Institutional Review Board (IRB) approvals, if applicable?
    \answerNA{}
  \item Did you include the estimated hourly wage paid to participants and the total amount spent on participant compensation?
    \answerNA{}
\end{enumerate}

\end{enumerate}

\clearpage


\newpage
\appendix

\section{Additional Background}
\label{sec:additional_bg}

\subsection{Background on the Policy Gradient Theorem}
\label{sec:background_pg}
RL algorithms can be broadly divided into \emph{value-based} and \emph{policy gradient} (PG) methods. While value-based methods select actions via a greedy maximization step  based on the learned action-values, PG methods directly optimize a parameterized policy $\pi_\theta$ based on the performance gradient $\nabla_\theta J(\theta)$. Thus, unlike value-based methods, PG algorithms inherit the strong, albeit local, convergence guarantees of the gradient descent and are naturally extendable to continuous actions. The fundamental theorem underlying PG methods is the \emph{policy gradient theorem} \cite{sutton2000policy}:
\begin{equation}
    \label{eq:standard_pg}
    \nabla_\theta J(\pi_\theta) = \int_{s \in {\mathcal{S}}} \rho^{\pi_\theta}(s) \int_{a \in {\mathcal{A}}} \nabla_\theta \pi_\theta(a | s) Q^{\pi_\theta}(s, a)
\end{equation}
where $\rho^{\pi_\theta} (s) = \lim_{t \rightarrow \infty} \gamma^t P(s_t=s | s_0, a_{0:t} \sim \pi_\theta)$ is the discounted stationary distribution of states under $\pi_\theta$ which is assumed to exist and to be independent of the initial state distribution (ergodicity assumption). The significance of the PG theorem is that the effect of policy changes on the state distribution does not appear in its expression, allowing for a sample-based estimate of the gradient \cite{williams1992simple}.

The deterministic policy gradient (DPG) is derived for deterministic policies by Silver et al. \cite{silver2014deterministic} as:
\begin{equation}
    \label{eq:standard_dpg}
    \nabla_\theta J(\pi_\theta) = \int_{s \in {\mathcal{S}}} \rho^{\pi_\theta}(s) \nabla_\theta \pi_\theta(s) \nabla_a Q^{\pi_\theta}(s, a) \big|_{a=\pi_\theta(s)}
\end{equation}
Since DPG does not need to integrate over the action space, it is often more sample-efficient than the stochastic policy gradient \cite{silver2014deterministic}. However, a noise needs to be manually injected during exploration as the deterministic policy does not have any inherent means of exploration. Finally, it is worth noting that due to the differentiation of the value function with respect to $a$, DPG is only applicable to continuous actions. 

\subsection{Mathematical Tools}
\label{supp:math_tools}
Various mathematical concepts from measure theory and differential geometry are presented in this section. We only explicitly introduce concepts which are directly mentioned or relevant to the proofs presented in section~\ref{sec:proofs}; for a more comprehensive overview, we direct the reader to textbooks such as \cite{bogachev2007measure, lang2012differential, spivak2018calculus}.

\begin{definition}[$\sigma$-algebra]
    Given a set $X$, a $\sigma$-algebra on $X$ is a family $\Sigma$ of subsets of $X$ such that 1) $X \in \Sigma$, 2) $A \in \Sigma$ implies $A^c \in \Sigma$ (closure under complements), and 3) if $(A_i)_{i \in \N}$ satisfies $A_i \in \Sigma$ for all $i \in \N$, then $\cup_{i \in \N}A_i \in \Sigma$ (closure under countable union). The tuple $(X, \Sigma)$ is a measurable space.
\end{definition}

The $\sigma$-algebra of a space specifies the sets in which a measure is defined; in probability theory--- and in our use case--- a $\sigma$-algebra represents a collection of events which can be assigned probabilities.

\begin{definition}[Pushforward measure]
Let $(X_1, \Sigma_1)$ and $(X_2, \Sigma_2)$ be two measurable spaces, $f: X_1 \to X_2$ a measurable map and $\mu: \Sigma_1 \to [0, \infty]$ a measure on $X_1$. Then the pushforward measure of $\mu$ with respect to $f$, denoted $f_*(\mu): \Sigma_2 \to [0, \infty]$ is defined as:
$$
    (f_*(\mu))(B) = \mu(f^{-1}(B)) \; \forall \; B \in \Sigma_2.
$$
\end{definition}

\begin{thm}[Change of variables]
\label{thm:cov}
A measurable function $g$ on $X_2$ is integrable with respect to $f_*(\mu)$ if and only if the function $g \circ f$ is integrable with respect to $\mu$, in which case the integrals are equal:
$$
    \int_{X_2} g d(f_*(\mu)) = \int_{X_1}g \circ f d\mu.
$$
\end{thm}

\begin{definition}[Local diffeomorphism]
\label{def:local_diffeo}
Let $M$ and $N$ be differentiable manifolds. A function $f : M \to N$ is a \emph{local diffeomorphism}, if for each point $x \in M$ there exists an open set $U$ containing $x$ such that $f(U)$ is open in $N$ and $f|_U : U \to f(U)$ is a diffeomorphism.
\end{definition}

\begin{thm}[Inverse function theorem for manifolds]
\label{thm:inv_function}
If $f : M \to N$ is a smooth map whose differential $df_x : T_x M \to T_{f(x)}N$ is an isomorphism at a point $x \in M$. Then $f$ is a local diffeomorphism at $x$.
\end{thm}

\begin{thm}[Chain rule for manifolds]
If $f : M \to N$ and $g : N \to O$ are smooth maps of manifolds, then:
$$d
    (g \circ f)_{x} = dg_{f(x)} \circ df_x.
$$
\end{thm}

\section{Assumptions and Conditions}
\label{sec:assumptions}
The derivation of our homomorphic policy gradient theorem is for continuous state and action spaces. Therefore, we have assumed the following regularity conditions on the actual MDP ${\mathcal{M}}$ and its MDP homomorphic image ${\overline{\mathcal{M}}}$ under the MDP homomorphism map $h$. The conditions are largely based on the regularity conditions of the deterministic policy gradient theorem \cite{silver2014deterministic}:

\textbf{Regularity conditions 1:} $\tau_a(s' | s)$, $\nabla_a \tau_a(s' | s)$, $\overline{\tau}_{\overline{a}}(\overline{s}' | \overline{s})$, $\nabla_{\overline{a}} \overline{\tau}_{\overline{a}}(\overline{s}' | \overline{s})$, $R(s, a), \nabla_a R(s, a)$, $\overline{R}(\overline{s}, \overline{a}), \nabla_{\overline{a}} \overline{R}(\overline{s},  \overline{a})$, $\pi^\uparrow_\theta(s), \nabla_\theta \pi^\uparrow_\theta(s), \overline{\pi}_\theta(\overline{s})$, $\nabla_{\theta} \overline{\pi}_\theta(\overline{s})$, $p_1(s)$, and $\overline{p}_1(\overline{s})$ are continuous with respect to all parameters and variables $s, \overline{s}, a, \overline{a}, s'$, and $\overline{s}'$.

\textbf{Regularity conditions 2:} There exists a $b$ and $L$ such that $\sup_s p_1(s) \!<\! b$, $\sup_{\overline{s}} \overline{p}_1(\overline{s}) < b$, $\sup_{a, s, s'} \tau_a(s'|s) < b$, $\sup_{\overline{a}, \overline{s}, \overline{s}'} \overline{\tau}_{\overline{a}}(\overline{s}'|\overline{s}) < b$, $\sup_{a, s} R(s, a) < b$, $\sup_{\overline{a}, \overline{s}} \overline{R}(\overline{s}, \overline{a}) < b$, $\sup_{a, s, s'}\|\nabla_a \tau_a(s'|s) \| < L, \sup_{\overline{a}, \overline{s}, \overline{s}'}\|\nabla_{\overline{a}} \overline{\tau}_{\overline{a}}(\overline{s}'|\overline{s}) \| < L$, $\sup_{s, a} \| \nabla_a R(s, a)\| < L, \sup_{\overline{s}, \overline{a}} \| \nabla_{\overline{a}} \overline{R}(\overline{s}, \overline{a})\| < L$.

We also assume the following conditions on the continuous MDP homomorphism map $h = (f, g_s)$, as discussed in Definition \ref{def:cont_mdp_homo}:

\textbf{Regularity conditions 3:} The action mapping $g_s(a)$ is a local diffeomorphism (Definition \ref{def:local_diffeo}). Hence it is continuous with respect to $a$ and locally bijective with respect to $a$. Additionally, $\nabla_a g_s(a)$ is continuous with respect to the parameter $a$, and there exists a $L$ such that $\sup_{s, a} \| \nabla_a g_{s}(a)\| < L$.

\clearpage
\section{Proofs}
\label{sec:proofs}
Below are the proofs accompanying Sections \ref{sec:value_equiv}, \ref{sec:cont_mdp_hom} and \ref{sec:hpg}.

\subsection{Proof of Theorem \ref{thrm:value_equiv_finite_main}: Value Equivalence}
\label{supp:proof_value_equiv_finite}

\begin{proof}
The proof is along the lines of the \emph{optimal value equivalence} theorem of Ravindran and Barto \cite{ravindran2001symmetries}. We define the $m$-step discounted action value function $Q_m^{\pi^\uparrow}(s, a)$ recursively for all $(s, a) \in {\mathcal{S} \times \mathcal{A}}$ and for all integers $m \geq 1$ as:
\begin{equation*}
    Q_m^{\pi^\uparrow}(s, a) = R(s, a) + \gamma \sum_{s' \in {\mathcal{S}}} \tau_a(s' | s) \sum_{a' \in {\mathcal{A}}} \pi^\uparrow(a' | s') Q_{m-1}^{\pi^\uparrow}(s', a'),
\end{equation*}
with $Q_{0}^{\pi^\uparrow}(s, a) = R(s,a)$. The proof is by induction on $m$; the base case of $m=0$ is true because:
\begin{equation*}
    Q_0^{\pi^\uparrow}(s, a) = R(s, a) = \overline{R}(f(s), g_s(a)) = Q_0^{\overline{\pi}}(f(s), g_s(a)).
\end{equation*}
Now suppose towards induction that $Q_k^{\pi^\uparrow}(s, a) = Q_k^{\overline{\pi}}(f(s), g_s(a))$ for all values of $k$ less than $m$ and all state action pairs $(s, a) \in {\mathcal{S} \times \mathcal{A}}$. Using the fact that $h = (f, g_s)$ is an MDP homomorphism, we have:
\begingroup
\allowdisplaybreaks
\begin{align}
    Q_m^{\pi^\uparrow}(s, a) &= R(s, a) + \gamma \sum_{s' \in {\mathcal{S}}} \tau_a(s' | s) \sum_{a' \in {\mathcal{A}}} \pi^\uparrow(a' | s') Q_{m-1}^{\pi^\uparrow}(s', a') \nonumber \\ 
    &= R(s, a) + \gamma \sum_{[s']_{B_h | {\mathcal{S}}} \in B_h | {\mathcal{S}}} \; \sum_{s'' \in [s']_{B_h | {\mathcal{S}}}} \tau_a(s'' | s) \sum_{a' \in {\mathcal{A}}} \pi^\uparrow(a' | s') Q_{m-1}^{\overline{\pi}}(f(s'), g_{s'}(a')) \label{eq:equiv_finite_1} \\
    &= R(s, a) + \gamma \sum_{[s']_{B_h | {\mathcal{S}}} \in B_h | {\mathcal{S}}} \; \sum_{s'' \in [s']_{B_h | {\mathcal{S}}}} \tau_a(s'' | s) \sum_{\overline{a}' \in {\overline{\mathcal{A}}}} \; \sum_{a'' \in g_{s'}^{-1}(\overline{a}')} \pi^\uparrow(a'' | s') Q_{m-1}^{\overline{\pi}}(f(s'), \overline{a}') \nonumber \\
    &= \overline{R}(f(s), g_s(a)) + \gamma \sum_{[s']_{B_h | {\mathcal{S}}} \in B_h | {\mathcal{S}}} \overline{\tau}_{g_s(a)}(f(s') | f(s)) \sum_{\overline{a}' \in {\overline{\mathcal{A}}}} \overline{\pi}(\overline{a}' | f(s')) Q_{m-1}^{\overline{\pi}} (f(s'), \overline{a}') \label{eq:equiv_finite_2} \\
    &= \overline{R}(f(s), g_s(a)) + \gamma \sum_{\overline{s}' \in {\overline{\mathcal{S}}}} \overline{\tau}_{g_s(a)}(\overline{s}' | f(s))   \sum_{\overline{a}' \in {\overline{\mathcal{A}}}} \overline{\pi}(\overline{a}' | \overline{s}') Q_{m-1}^{\overline{\pi}} (\overline{s}', \overline{a}') \nonumber \\
    &= Q_m^{\overline{\pi}} (f(s), g_s(a)) \nonumber.
\end{align}
\endgroup
Where in equation \eqref{eq:equiv_finite_1} we used the fact that $Q^{\pi^\uparrow}_{m-1}(s, a) = Q^{\overline{\pi}}_{m-1}(f(s), g_s(a))$ from  the induction assumption. In equation \eqref{eq:equiv_finite_2} we used  $\sum_{s'' \in [s']_{B_h | {\mathcal{S}}}} \tau_a(s'' | s) = \overline{\tau}_{g_{s}}(f(s') | f(s))$ and $\sum_{a'' \in g_{s'}^{-1}(\overline{a}')} \pi^\uparrow(a'' | s') = \overline{\pi}(\overline{a}' | f(s'))$ from the definition of MDP homomorphism \cite{ravindran2001symmetries}. Since $R$ and $\overline{R}$ are bounded, it follows by induction that $Q^{\pi^\uparrow}(s, a) = Q^{\overline{\pi}}(f(s), g_s(a))$ for all $(s, a) \in {\mathcal{S} \times \mathcal{A}}$. 

The proof for $V^{\pi^\uparrow}(s) = V^{\overline{\pi}}(f(s))$ follows directly from the equivalence of action value functions and the fact that the two policies are tied together through the lifting process because in general we have: $V^\pi(s) = \sum_{a \in A}\pi(a|s) Q^\pi(s, a)$.
\end{proof}

\subsection{Proof of Theorem~\ref{thm:opt_equiv_continuous}: Optimal Value Equivalence for Continuous MDP Homomorphisms}
\label{sec:opt_equiv_continuous}
\begin{proof}
The proof follows along the same lines as Ravindran and Barto \cite{ravindran2001symmetries}. We will first prove the following claim:
\begin{claim}
    For $m \geq 1$, define the sequence $Q_m: \mathcal{S} \times \mathcal{A} \to \R$ as
    $$Q_m(s,a) = R(s,a) + \gamma \int_{s' \in \mathcal{S}}\tau_a(ds' | s) \sup_{a' \in \mathcal{A}}Q_{m-1}(s', a')$$
    and $Q_0(s,a) = R(s,a)$. Define the sequence $\overline{Q}_m: \overline{\mathcal{S}} \times \overline{\mathcal{A}} \to \R$ analogously. Then for any $(s, a) \in \mathcal{S} \times \mathcal{A}$ we have
    $$Q_m(s,a) = \overline{Q}_m(f(s), g_s(a)).$$
\end{claim}
We will prove this claim by induction on $m$. The base case $m= 0$ follows from the reward invariance property of continuous MDP homomorphisms:
$$Q_0(s,a) = R(s,a) = \overline{R}(f(s), g_s(a)) = \overline{Q}_0(f(s), g_s(a)).$$
For the inductive case, note that
\begin{align}
    Q_m(s,a) &= R(s,a) + \gamma \int_{s' \in \mathcal{S}}\tau_a(ds' | s) \sup_{a' \in \mathcal{A}}Q_{m-1}(s', a') \nonumber \\
    &= \overline{R}(f(s), g_s(a)) + \gamma \int_{s' \in \mathcal{S}}\tau_a(ds' | s) \sup_{a' \in \mathcal{A}}\overline{Q}_{m-1}(f(s'), g_{s'}(a')) \label{thm:opt2}\\
    &= \overline{R}(f(s), g_s(a)) + \gamma \int_{s' \in \mathcal{S}}\tau_a(ds' | s) \sup_{\overline{a}' \in \overline{\mathcal{A}}}\overline{Q}_{m-1}(f(s'), \overline{a}') \label{thm:opt3}\\
    &= \overline{R}(f(s), g_s(a)) + \gamma \int_{\overline{s}' \in \overline{\mathcal{S}}}\overline{\tau}_{g_s(a)}(d\overline{s}' | f(s)) \sup_{\overline{a}' \in \overline{\mathcal{A}}}\overline{Q}_{m-1}(\overline{s}', \overline{a}') \label{thm:opt4}\\
    &= Q_{m-1}(f(s), g_s(a)),
\end{align}
where Equation~\ref{thm:opt2} follows from the inductive hypothesis, Equation~\ref{thm:opt3} follows from $g_s$ being surjective, and Equation~\ref{thm:opt4} follows from the change of variables formula (Theorem \ref{thm:cov}); indeed, from Definition~\ref{def:cont_mdp_homo} we have the pushforward measure of $\tau_a(\cdot | s)$ with respect to $f$ equals $\tau_{g_s(a)}( \cdot | f(s))$ and here $g: \overline{\mathcal{S}} \to \R$ is defined as $g(\overline{s}) = \sup_{\overline{a}' \in \overline{\mathcal{A}}} \overline{Q}_{m-1}(\overline{s}, \overline{a}')$. This concludes the induction proof. Since $\lim_{m \to \infty} Q_{m}(s,a) = Q^*(s,a)$, it follows that $Q^*(s,a) = \overline{Q}^*(f(s), g_s(a))$. 

The proof for $V^*(s) = \overline{V}^*(f(s))$ follows directly from the equivalence of optimal action value functions as $V^*(s) = \max_a Q^*(s, a)$ in general.
\end{proof}

\subsection{Proof of Theorem~\ref{thrm:value_equiv_continuous}: Value Equivalence for Deterministic Policies and Continuous MDP Homomorphisms}
\label{supp:value_equiv_continuous}
\begin{proof}[Proof] Unlike the proofs of Theorems \ref{thrm:value_equiv_finite_main} and \ref{thm:opt_equiv_continuous}, here we assume the policy is deterministic due to the complications of lifting stochastic policies discussed in Section \ref{sec:value_equiv_continuous}. Therefore, the lifting process can be simply obtained as $\pi^\uparrow(s) \!=\! g_s^{-1} (\overline{\pi}(f(s))$ and the inverse of the lifting process is $\overline{\pi}(f(s)) = g_s(\pi^\uparrow(s))$, as the mapping $g_s$ is assumed to be an invertible continuous map (Appendix \ref{sec:assumptions}).

Similarly to Ravindran and Barto \cite{ravindran2001symmetries}, the proof is by induction. We define the $m$-step discounted action value function $Q^{\pi^\uparrow}_m (s, a)$ for the domain ${\mathcal{S} \times \mathcal{A}}$ and for all integers $m \geq$ as:
\begin{equation*}
    Q_m^{\pi^\uparrow}(s, a) = R(s, a) + \gamma \int_{s' \in {\mathcal{S}}} \tau_a(ds' | s)  Q_{m-1}^{\pi^\uparrow}(s', \pi^\uparrow(s')),
\end{equation*}
with $Q_{0}^{\pi^\uparrow}(s, a) = R(s,a)$ for all pairs $(s,a) \in \mathcal{S} \times \mathcal{A}$. The proof is by induction on $m$, the base case of $m=0$ is true because:
\begin{equation*}
    Q_0^{\pi^\uparrow}(s, a) = R(s, a) = \overline{R}(f(s), g_s(a)) = Q_0^{\overline{\pi}}(f(s), g_s(a)).
\end{equation*}
Now suppose towards induction that $Q_k^{\pi^\uparrow}(s, a) = Q_k^{\overline{\pi}}(f(s), g_s(a))$ for all values of $k$ less than $m$ on the domain ${\mathcal{S} \times \mathcal{A}}$. Using the fact that $h = (f, g_s)$ is a continuous MDP homomorphism, we have:
\begingroup
\begin{align}
    Q_m^{\pi^\uparrow}(s, a) &= R(s, a) + \gamma \int_{s' \in {\mathcal{S}}} \tau(ds' | s) Q_{m-1}^{\pi^\uparrow}(s', \pi^\uparrow(s')) \nonumber \\
    &= R(s, a) + \gamma \int_{s' \in \mathcal{S}} \tau_a(ds' | s)  Q_{m-1}^{\overline{\pi}}(f(s'), g_{s'}(\pi^\uparrow(s'))) \label{eq:cont_value_equiv_1} \\
    &= \overline{R}(f(s), g_s(a)) + \gamma \int_{s' \in \mathcal{S}} \tau_a(ds' | s) Q_{m-1}^{\overline{\pi}}(f(s'), \overline{\pi}(f(s'))) \label{eq:cont_value_equiv_2} \\
    &= \overline{R}(f(s), g_s(a)) + \gamma \int_{\overline{s} \in \overline{\mathcal{S}}} \overline{\tau}_{g_s(a)}( d\overline{s}| f(s))  Q_{m-1}^{\overline{\pi}}(\overline{s}', \overline{\pi}(\overline{s}')) \label{eq:cont_value_equiv_3}\\
    &= Q^{\overline{\pi}}_m (f(s), g_s(a)).
\end{align}
\endgroup
Where in equation \eqref{eq:cont_value_equiv_1}, we used the induction assumption,in equation \eqref{eq:cont_value_equiv_2} we used the definition the inverse of policy lifting as defined above, and in equation \eqref{eq:cont_value_equiv_3} we applied the change of variables formula (Theorem \ref{thm:cov}) using the fact that $\overline{\tau}_{g_s(a)}(\cdot | f(s))$ is the pushforward measure of $\tau_a(\cdot| s)$ under $f$ by definition. Since $R$ and $\overline{R}$ are bounded, it follows by induction that $Q^{\pi^\uparrow}(s, a) = Q^{\overline{\pi}}(f(s), g_s(a))$.

The proof for $V^{\pi^\uparrow}(s) = V^{\overline{\pi}}(f(s))$ follows directly from the equivalence of action value functions and the fact that the two policies are tied together through the lifting process because $V^\pi(s) = Q^\pi(s, \pi(s))$ for deterministic policies.

\end{proof}

\subsection{Proof of Theorem \ref{thm:det_grad_equiv}: Equivalence of Deterministic Policy Gradients}
\label{app:det_grad_equiv}
\begin{proof} Assuming the conditions described in Appendix \ref{sec:assumptions}, we first take the derivative of the deterministic policy lifting relation w.r.t. the policy parameters $\theta$ using the chain rule:
\begin{align}
    (g_s \circ \pi^\uparrow)(s) &= (\overline{\pi} \circ f)(s) \nonumber \\
    d(g_s \circ \pi^\uparrow)_\theta (s) &= d(\overline{\pi} \circ f)_\theta (s) \nonumber \\
    d(g_s)_{\pi^\uparrow(s)} \circ d(\pi^\uparrow)_\theta (s) &= d(\overline{\pi} \circ f)_\theta (s) \nonumber \\
    \underbrace{\nabla_a g_s(a) \big|_{a = \pi^\uparrow(s)}}_{P} \nabla_\theta \pi^\uparrow(s) &= \nabla_\theta \overline{\pi}(f(s)),
    \label{eq:grad_equiv_1}
\end{align}
where $\circ$ is the composition operator and the dimensions of the matrices are $P \in \R^{|\overline{\mathcal{A}}| \times |A|}$, $\nabla_\theta \pi^\uparrow(s) \in \R^{|A| \times |\theta|}$, and $\nabla_\theta \overline{\pi}(\overline{s}) \in \R^{|\overline{\mathcal{A}}| \times |\theta|}$. Second, we take the derivative of the value equivalence theorem w.r.t. the actions $a$ using the chain rule: 
\begin{align}
    Q^{\pi^\uparrow}(s, a) &= Q^{\overline{\pi}} (f(s), g_s(a)) \nonumber \\
    dQ^{\pi^\uparrow}(s, a)_a &= dQ^{\overline{\pi}}(f(s), g_s(a))_a \nonumber \\
    \nabla_a Q^{\pi^\uparrow}(s, a) \big|_{a = \pi^\uparrow(s)} &= \nabla_{\overline{a}} Q^{\overline{\pi}}(f(s), \overline{a}) \big|_{\overline{a} = \overline{\pi}(f(s))} \underbrace{\nabla_a g_s(a) \Big|_{a = g_s^{-1}(\overline{\pi}(f(s)))}}_{P},
    \label{eq:grad_equiv_2}
\end{align}
where the dimensions of the matrices are $\nabla_a Q^{\pi^\uparrow}(s, a) \in \R^{|A|}$, $\nabla_{\overline{a}} Q^{\overline{\pi}}(\overline{s}, \overline{a}) \in \R^{|\overline{\mathcal{A}}|}$, and similarly as before $P \in \R^{|\overline{\mathcal{A}}| \times |A|}$. As we assumed the $g_s$ to be a local diffeomorphism, the inverse function theorem (Theorem \ref{thm:inv_function}) states that the matrix $P$ is invertible, thus we right-multiply both sides of equation \eqref{eq:grad_equiv_2} by $P^{-1}$ and left-multiply the resulting equation by equation \eqref{eq:grad_equiv_1} to obtain the desired result:
\begin{align}
    \nabla_a Q^{\pi^\uparrow}(s, a) \big|_{a = \pi^\uparrow(s)} P^{-1} P  \nabla_\theta \pi^\uparrow(s) &= \nabla_{\overline{a}} Q^{\overline{\pi}}(f(s), \overline{a}) \big|_{\overline{a} = \overline{\pi}(f(s))} \nabla_\theta \overline{\pi}(f(s)) \nonumber \\ 
    \nabla_a Q^{\pi^\uparrow}(s, a) \big|_{a = \pi^\uparrow(s)} \nabla_\theta \pi^\uparrow(s) &= \nabla_{\overline{a}} Q^{\overline{\pi}}(f(s), \overline{a}) \big|_{\overline{a} = \overline{\pi}(f(s))} \nabla_\theta \overline{\pi}(f(s)).
\end{align}
\end{proof}

\subsection{Proof of Theorem \ref{thrm:det_hpg}: Homomorphic Policy Gradient}
\label{app:deterministic_hpg}
\begin{proof}
The proof follows along the same lines of the deterministic policy gradient theorem \cite{silver2014deterministic}, but with additional steps for changing the integration space from $\mathcal{S}$ to $\mathcal{\overline{S}}$. First, we derive a recursive expression for $\nabla_\theta V^{{\pi^\uparrow_\theta}}(s)$ as:
\begingroup
\allowdisplaybreaks
\begin{align}
    \nabla_\theta V^{\pi^\uparrow_\theta} (s) &= \nabla_\theta Q^{\pi^\uparrow_\theta}\Big(s, \pi^\uparrow_\theta(s) \Big) \nonumber \\
        &= \nabla_\theta \Big[R (s, \pi^\uparrow_\theta(s)) + \gamma \int_{\mathcal{S}} \tau_{\pi^\uparrow_\theta(s)} (s, ds') V^{\pi^\uparrow_\theta}(s') \Big] \nonumber \\
        &=\nabla_\theta \pi^\uparrow_\theta(s) \nabla_a R(s, a) \Big|_{a = \pi^\uparrow_\theta(s)} \nonumber \\ &\qquad + \gamma \int_{\mathcal{S}} \Big[ \tau_{\pi^\uparrow_\theta(s)} (s, ds') \nabla_\theta V^{\pi^\uparrow_\theta}(s') + \nabla_\theta \pi^\uparrow_\theta(s) \nabla_a \tau_{a} (s, ds') \Big|_{a = \pi^\uparrow_\theta(s)} V^{\pi^\uparrow_\theta}(s') \Big] \label{eq:hpg_eq_1} \\
        &= \nabla_\theta \pi^\uparrow_\theta(s) \nabla_a \Big[ R \big(s, a \big) \!+\! \gamma \!\! \int_{\mathcal{S}} \tau_{a} (s, ds') V^{\pi^\uparrow_\theta}(s') \Big] \Big|_{a = \pi^\uparrow_\theta(s)} \!\!\!+\! \gamma\!\! \int_{\mathcal{S}} \tau_{\pi^\uparrow_\theta(s)} (s, ds') \nabla_\theta V^{\pi^\uparrow_\theta}(s') \nonumber \\
        &= \nabla_\theta \pi^\uparrow_\theta(s) \nabla_a Q^{\pi^\uparrow_\theta}(s, a)\Big|_{a = \pi^\uparrow_\theta(s)} + \gamma \int_{\mathcal{S}} \tau_{\pi^\uparrow_\theta(s)} (s, ds') \nabla_\theta V^{\overline{\pi}_\theta}(f(s')) \label{eq:hpg_eq_2} \\
        &= \nabla_\theta \pi^\uparrow_\theta(s) \nabla_a Q^{\pi^\uparrow_\theta}(s, a)\Big|_{a = \pi^\uparrow_\theta(s)} + \gamma \int_{\overline{\mathcal{S}}} \overline{\tau}_{g_s(\pi^\uparrow_\theta(s))} (f(s), d\overline{s}') \nabla_\theta V^{\overline{\pi}_\theta}(f(s')) \label{eq:hpg_eq_3} \\
        &=  \nabla_\theta \overline{\pi}_\theta(f(s)) \nabla_{\overline{a}} Q^{\overline{\pi}_\theta}(f(s), \overline{a})  \Big|_{\overline{a} = \overline{\pi}_\theta(f(s))} + \gamma \int_{\overline{\mathcal{S}}} \overline{\tau}_{\overline{\pi}_\theta(\overline{s})}(\overline{s}, d\overline{s}') \nabla_\theta V^{\overline{\pi}_\theta}(\overline{s}') \label{eq:hpg_eq_4} \\
        &=  \nabla_\theta \overline{\pi}_\theta(f(s)) \nabla_{\overline{a}} Q^{\overline{\pi}_\theta}(f(s), \overline{a})  \Big|_{\overline{a} = \overline{\pi}_\theta(f(s))} + \gamma \int_{\overline{\mathcal{S}}}  p(\overline{s} \rightarrow \overline{s}', 1, \overline{\pi}_\theta ) \nabla_\theta V^{\overline{\pi}_\theta}(\overline{s}') d\overline{s}'. \nonumber
\end{align}
\endgroup
Where $p(\overline{s} \rightarrow \overline{s}', t, \overline{\pi}_\theta)$ is the probability of going from $\overline{s}$ to $\overline{s}'$ under the policy $\overline{\pi}_\theta(\overline{s})$ in $t$ time steps. In equation \eqref{eq:hpg_eq_1} we were able to apply the Leibniz integral rule to exchange the order of derivative and integration because of the regularity conditions on the continuity of the functions. In equation \eqref{eq:hpg_eq_2} we used the value equivalence property, and in equation \eqref{eq:hpg_eq_3} we used the change of variables formula based on the pushforward measure (\ref{thm:cov}) of $\tau_a(s, .)$ with respect to $f$. Finally, in equation \eqref{eq:hpg_eq_4} we used the equivalence of policy gradients from Theorem \ref{thm:det_grad_equiv}. By recursively rolling out the formula above, we obtain:
\begin{align}
    \nabla_\theta V^{\pi^\uparrow_\theta} (s) &= \nabla_\theta \overline{\pi}_\theta(f(s)) \nabla_{\overline{a}} Q^{\overline{\pi}_\theta}(f(s), \overline{a}) \Big|_{\overline{a} = \overline{\pi}_\theta(f(s))} \nonumber \\ 
    &\quad + \gamma \int_{\overline{\mathcal{S}}}  p(\overline{s} \rightarrow \overline{s}', 1, \overline{\pi}_\theta ) \nabla_\theta \overline{\pi}_\theta(f(s')) \nabla_{\overline{a}} Q^{\overline{\pi}_\theta}(f(s'), \overline{a}) \Big|_{\overline{a} = \overline{\pi}_\theta(f(s'))}  d\overline{s}' \nonumber \\
    &\quad + \gamma^2 \int_{\overline{\mathcal{S}}}  p(\overline{s} \rightarrow \overline{s}', 1, \overline{\pi}_\theta ) \int_{\overline{\mathcal{S}}}  p(\overline{s}' \rightarrow \overline{s}'', 1, \overline{\pi}_\theta ) \nabla_\theta V^{\pi^\uparrow_\theta}(f(s'')) d\overline{s}'' d\overline{s}' \nonumber \\
    &= \nabla_\theta \overline{\pi}_\theta(f(s)) \nabla_{\overline{a}} Q^{\overline{\pi}_\theta}(f(s), \overline{a}) \Big|_{\overline{a} = \overline{\pi}_\theta(f(s))} \nonumber \\ 
    &\quad + \gamma \int_{\overline{\mathcal{S}}}  p(\overline{s} \rightarrow \overline{s}', 1, \overline{\pi}_\theta ) \nabla_\theta \overline{\pi}_\theta(f(s')) \nabla_{\overline{a}} Q^{\overline{\pi}_\theta}(f(s'), \overline{a}) \Big|_{\overline{a} = \overline{\pi}_\theta(f(s'))}  d\overline{s}' \nonumber \\
    &\quad + \gamma^2 \int_{\overline{\mathcal{S}}}  p(\overline{s} \rightarrow \overline{s}'', 2, \overline{\pi}_\theta ) \nabla_\theta V^{\overline{\pi}_\theta}(f(s'')) d\overline{s}'' \label{eq:hpg_eq_5}\\
    &\vdots \nonumber \\
    &= \int_{\overline{\mathcal{S}}} \sum_{t=0}^\infty \gamma^t p(\overline{s} \rightarrow \overline{s}', t, \overline{\pi}_\theta ) \nabla_\theta \overline{\pi}_\theta(f(s)) \nabla_{\overline{a}} Q^{\overline{\pi}_\theta}(f(s), \overline{a}) \Big|_{\overline{a} = \overline{\pi}_\theta(f(s))} d\overline{s}'.
\end{align}
Where in equation \eqref{eq:hpg_eq_5} we exchanged the order of integration using the Fubini's theorem that requires the boundedness of $\| \nabla_\theta V^{\overline{\pi}_\theta} (s) \|$ as described in the regularity conditions. Finally, we take the expectation of $\nabla_\theta V^{\pi^\uparrow_\theta}(s)$ over the initial state distribution:
\begin{align}
    \nabla_\theta J(\theta) &= \nabla_\theta \int_{\mathcal{S}} p_1(s) V^{\pi^\uparrow_\theta}(s) ds \nonumber \\
    &= \int_{\mathcal{S}} p_1(s) \nabla_\theta V^{\pi^\uparrow_\theta}(s) ds \nonumber \\
    &= \int_{\mathcal{S}} p_1(s) \int_{\overline{\mathcal{S}}} \sum_{t=0}^\infty \gamma^t p(\overline{s} \rightarrow \overline{s}', t, \overline{\pi}_\theta ) \nabla_\theta \overline{\pi}_\theta(f(s)) \nabla_{\overline{a}} Q^{\overline{\pi}_\theta}(f(s), \overline{a}) \Big|_{\overline{a} = \overline{\pi}_\theta(f(s))} d\overline{s}' ds \nonumber \\
    &= \int_{\overline{\mathcal{S}}} \overline{p}_1(\overline{s}) \int_{\overline{\mathcal{S}}} \sum_{t=0}^\infty \gamma^t p(\overline{s} \rightarrow \overline{s}', t, \overline{\pi}_\theta ) \nabla_\theta \overline{\pi}_\theta(f(s)) \nabla_{\overline{a}} Q^{\overline{\pi}_\theta}(f(s), \overline{a}) \Big|_{\overline{a} = \overline{\pi}_\theta(f(s))} d\overline{s}' d\overline{s} \label{eq:hpg_eq_6} \\
    &= \int_{\mathcal{S}} \rho^{\overline{\pi}_\theta}(\overline{s}) \nabla_\theta \overline{\pi}_\theta(\overline{s}) \nabla_{\overline{a}} Q^{\overline{\pi}_\theta}(\overline{s}, \overline{a}) \Big|_{\overline{a} = \overline{\pi}_\theta(\overline{s})} d\overline{s}.
\end{align}
Where $\rho^{\overline{\pi}_\theta}(\overline{s})$ is the discounted stationary distribution induced by the policy $\overline{\pi}_\theta$. In equation \eqref{eq:hpg_eq_6} we used the change of variable formula. Similar to the steps before, we have used the Leibniz integral rule to exchange the order of integration and derivative, used Fubini's theorem to exchange the order of integration.
\end{proof}

\clearpage
\section{Full Results}
\label{sec:additional_results}

As discussed in Section \ref{sec:exp}, we evaluate DHPG on continuous control tasks from DM Control on state and pixel observations. Importantly, to reliably evaluate our algorithm against the baselines and to correctly capture the distribution of results, we follow the best practices proposed by Agarwal et al. \cite{agarwal2021deep} and report the interquartile mean (IQM) and performance profiles aggregated on all tasks over 10 random seeds. While our baseline results are obtained using the official code, when possible, some of the results may differ from the originally reported ones due to the difference in the seed numbers and our goal to present a faithful representation of the true performance distribution \cite{agarwal2021deep}. 

We use the official implementations of DrQv2, DBC, and SAC-AE, while we re-implement DeepMDP due to the unavailability of the official code; See Appendix \ref{sec:baseline_impl} for full details on the baselines.

\subsection{State Observations}
\label{sec:additional_results_states}
\begin{figure}[h!]
     \centering
     \begin{subfigure}[b]{0.24\textwidth}
         \centering
         \includegraphics[width=\textwidth]{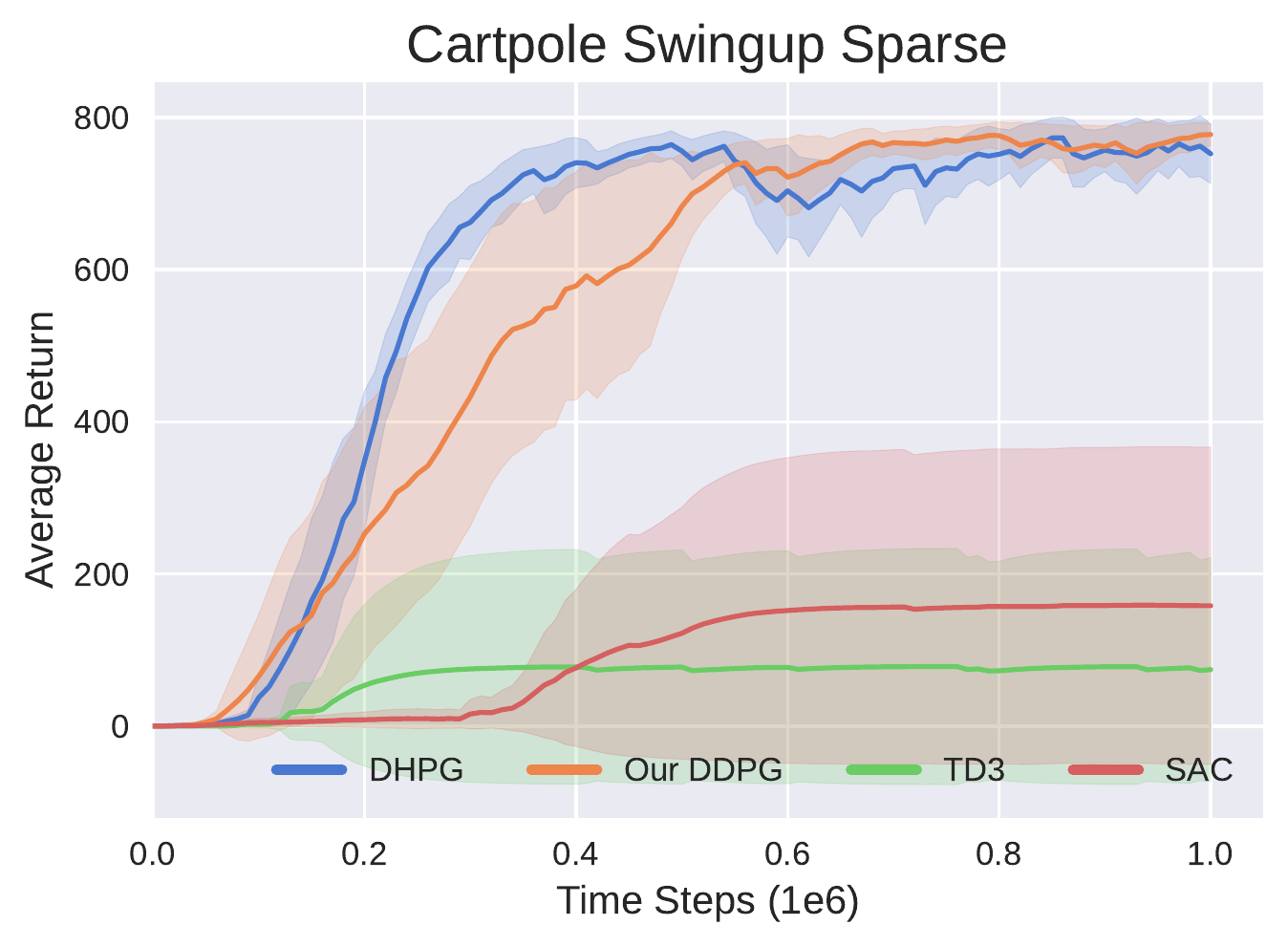}
     \end{subfigure}
     \hfill
     \begin{subfigure}[b]{0.24\textwidth}
         \centering
         \includegraphics[width=\textwidth]{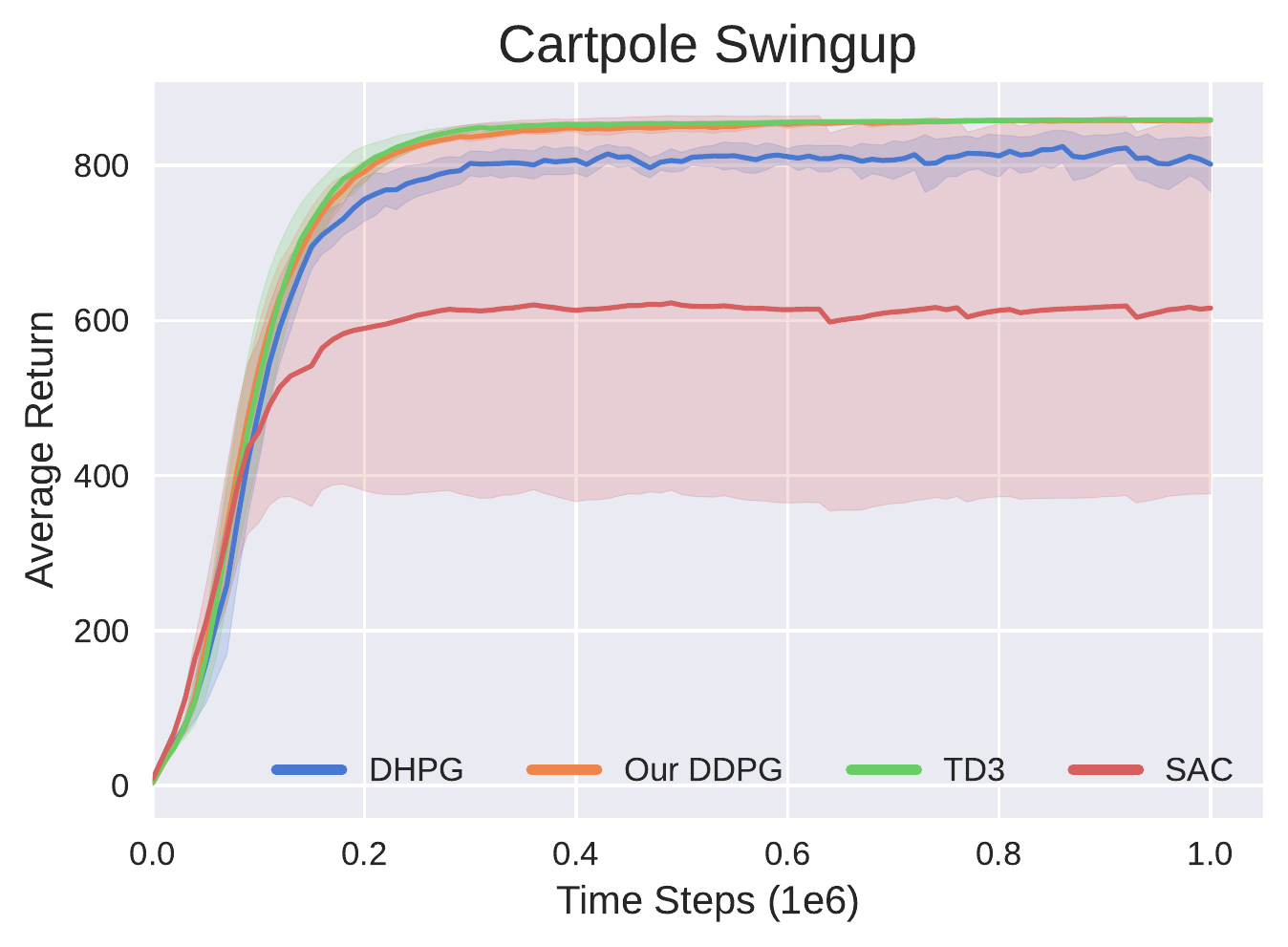}
     \end{subfigure}
     \hfill
     \begin{subfigure}[b]{0.24\textwidth}
         \centering
         \includegraphics[width=\textwidth]{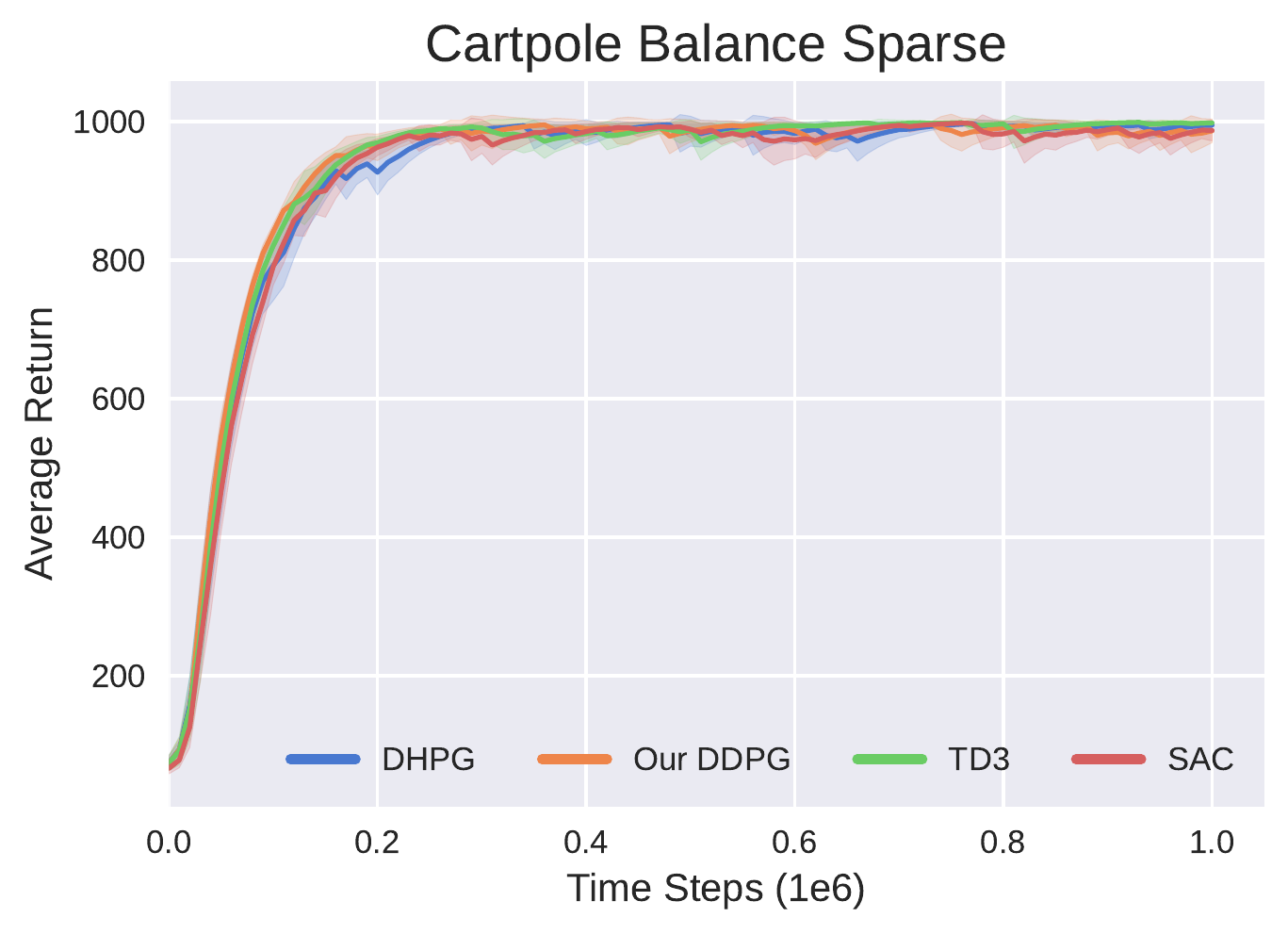}
     \end{subfigure}
     \hfill
     \begin{subfigure}[b]{0.24\textwidth}
         \centering
         \includegraphics[width=\textwidth]{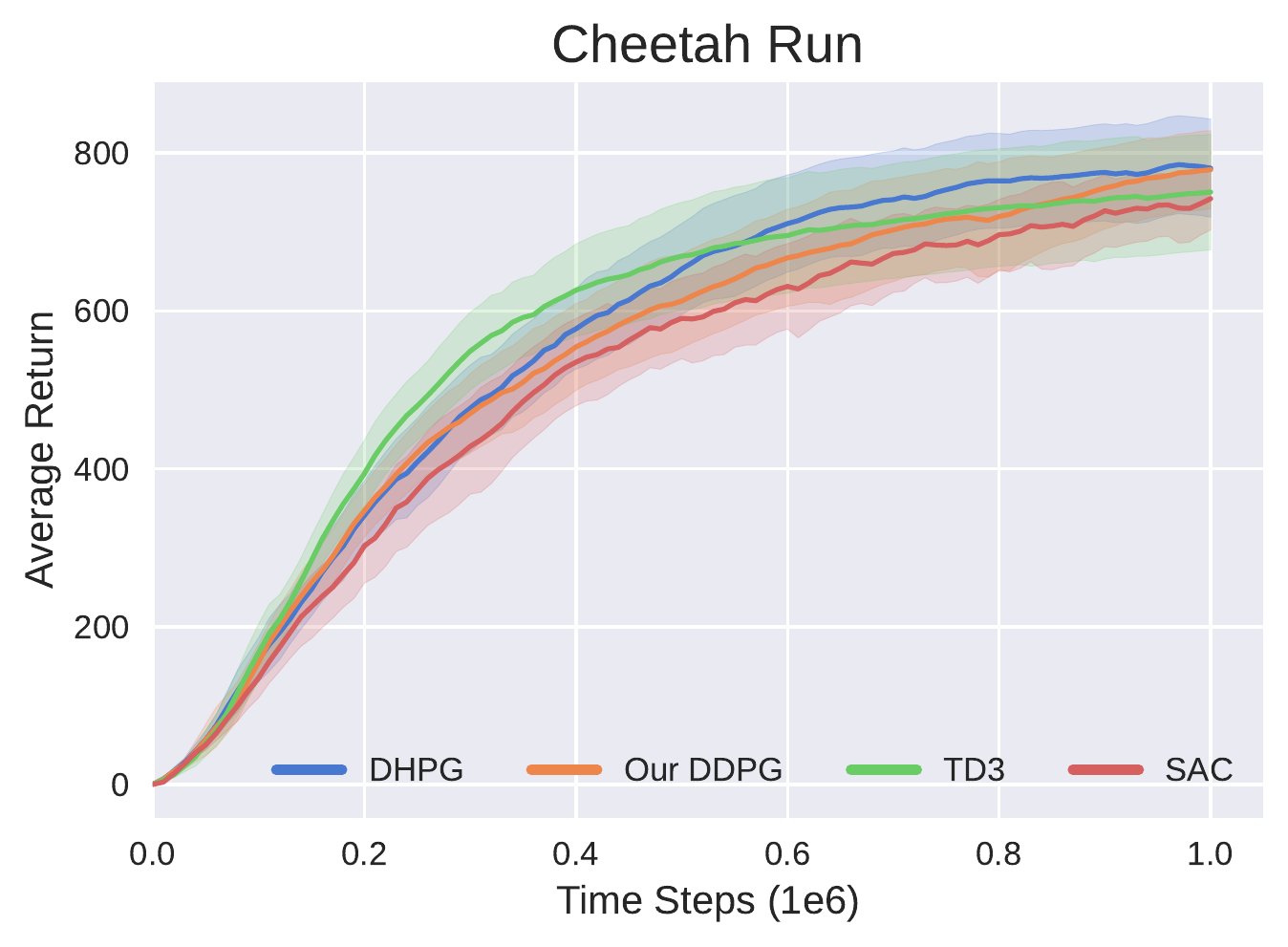}
     \end{subfigure}
     \hfill
     
     \begin{subfigure}[b]{0.24\textwidth}
         \centering
         \includegraphics[width=\textwidth]{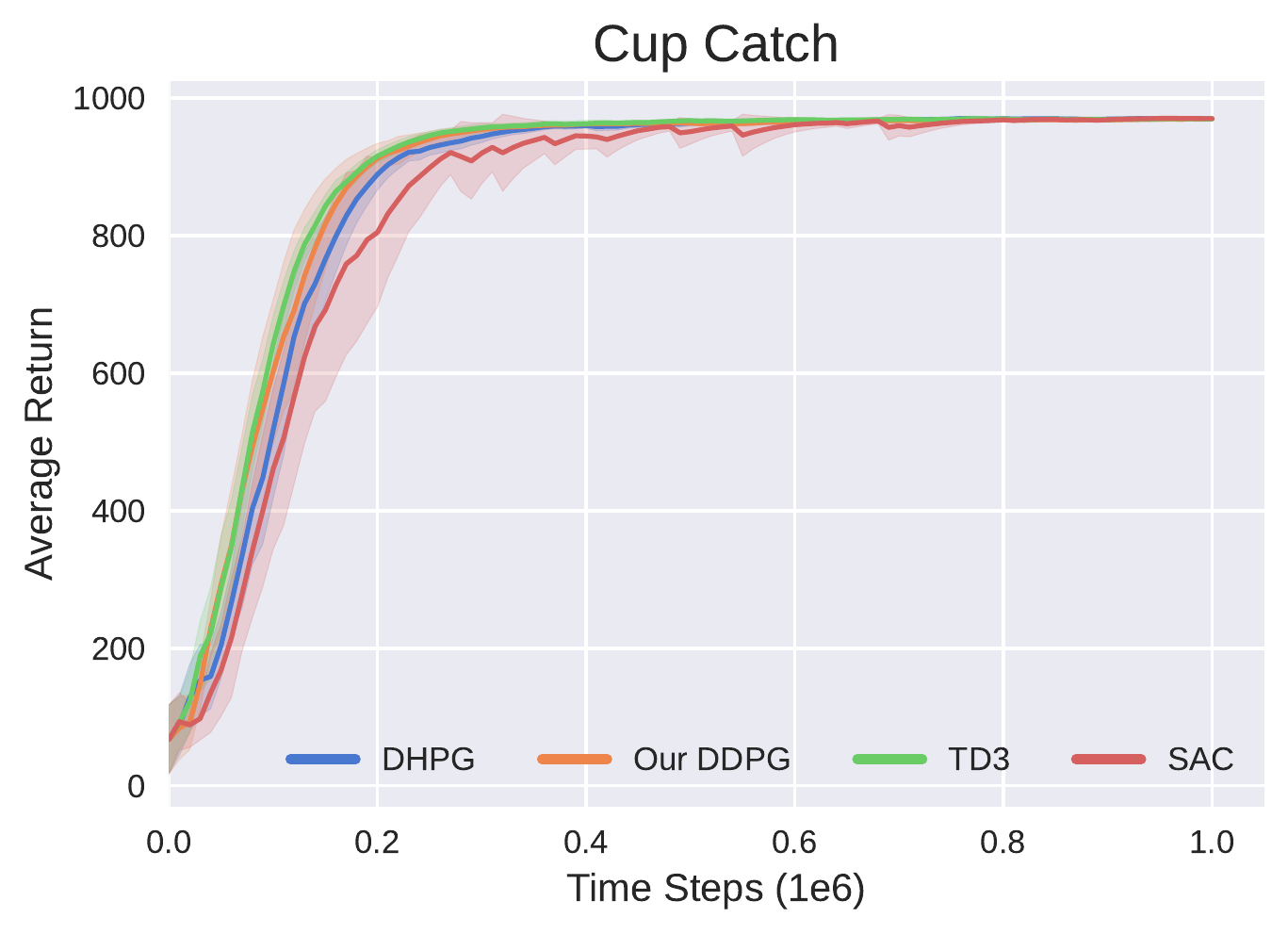}
     \end{subfigure}
     \hfill
     \begin{subfigure}[b]{0.24\textwidth}
         \centering
         \includegraphics[width=\textwidth]{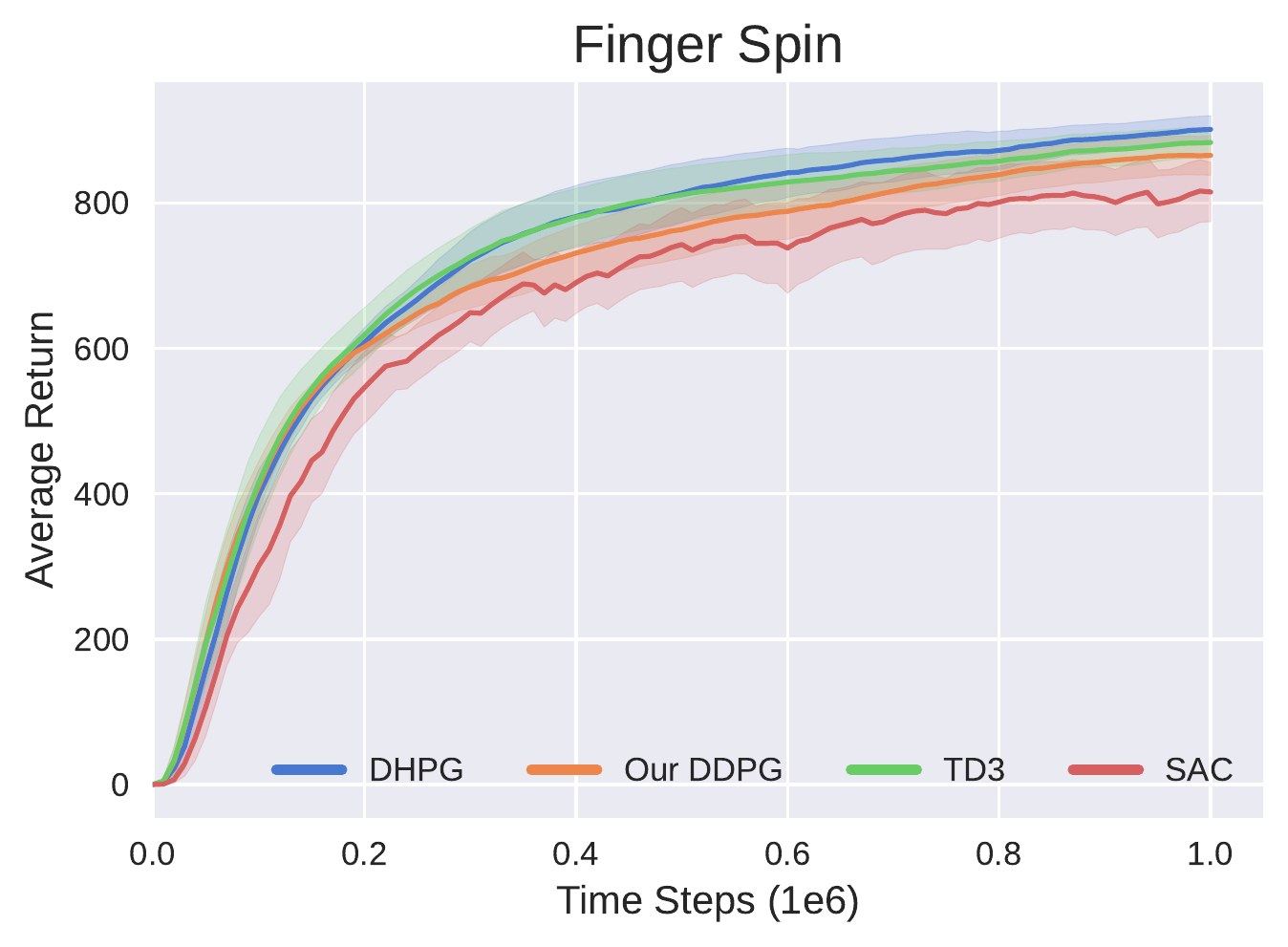}
     \end{subfigure}
     \hfill
     \begin{subfigure}[b]{0.24\textwidth}
         \centering
         \includegraphics[width=\textwidth]{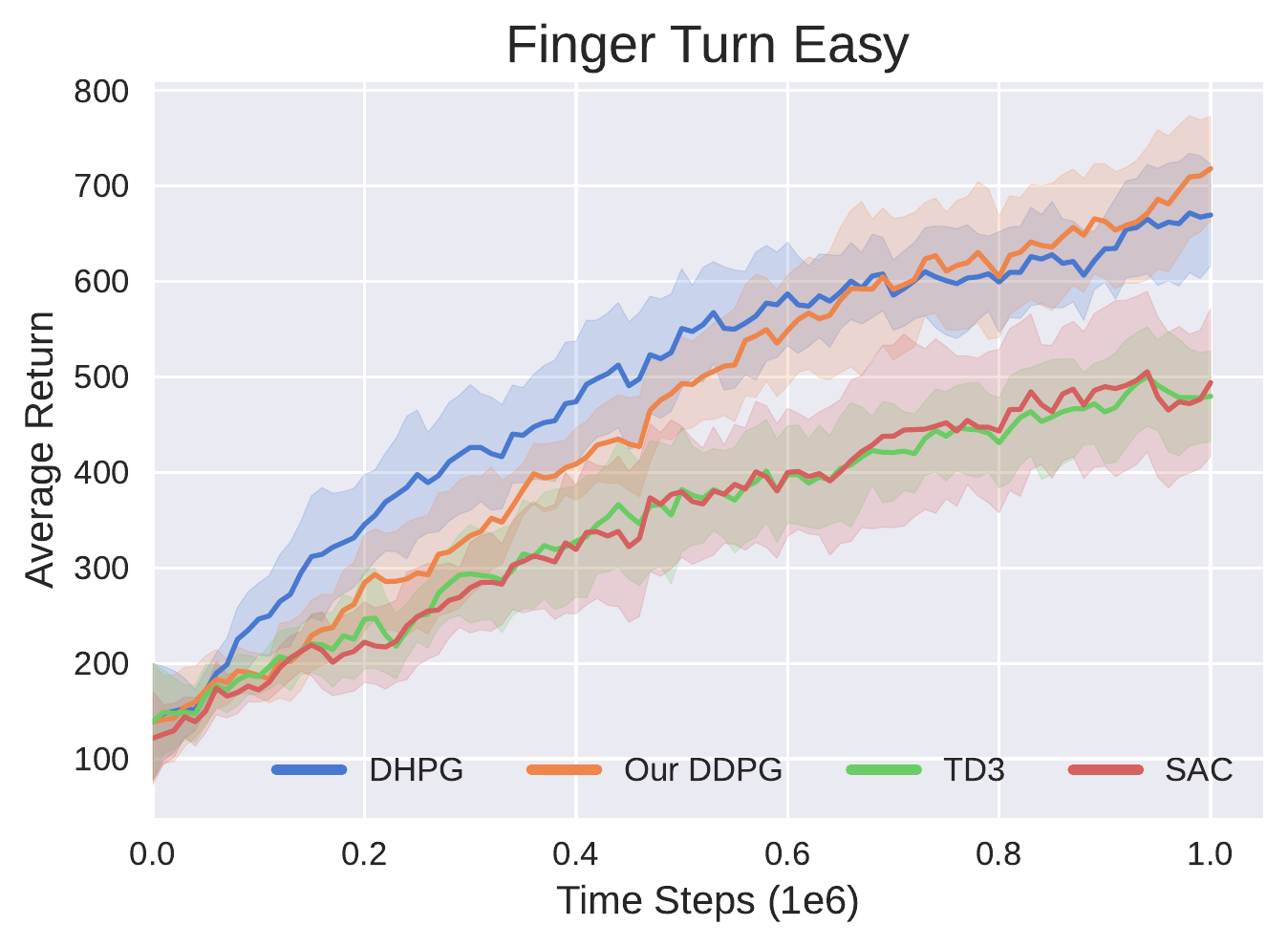}
     \end{subfigure}
     \hfill
     \begin{subfigure}[b]{0.24\textwidth}
         \centering
         \includegraphics[width=\textwidth]{figures/states_main/states_finger_turn_hard_episode_reward_eval.pdf}
     \end{subfigure}
     \hfill
     
     \begin{subfigure}[b]{0.24\textwidth}
         \centering
         \includegraphics[width=\textwidth]{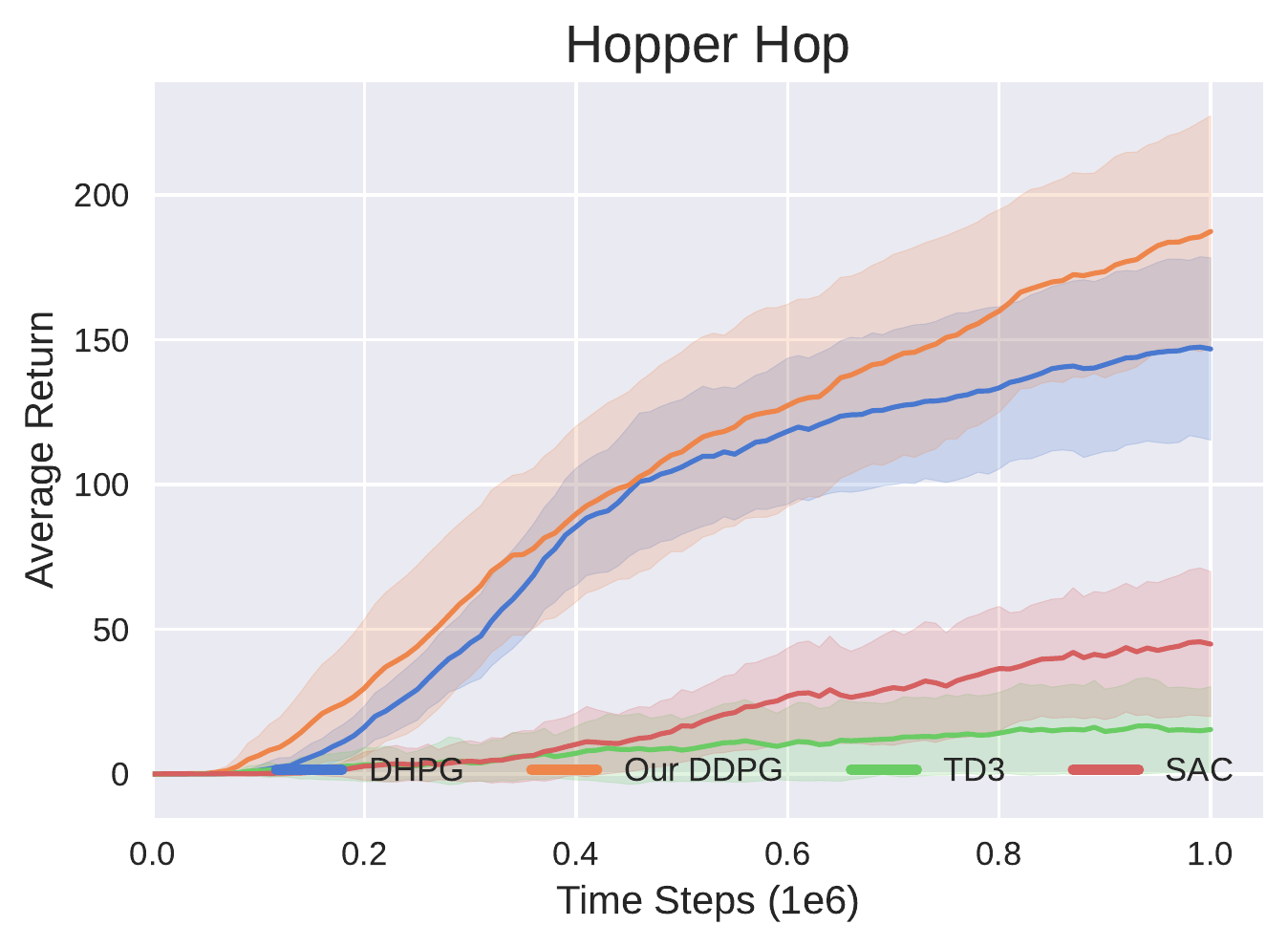}
     \end{subfigure}
     \hfill
     \begin{subfigure}[b]{0.24\textwidth}
         \centering
         \includegraphics[width=\textwidth]{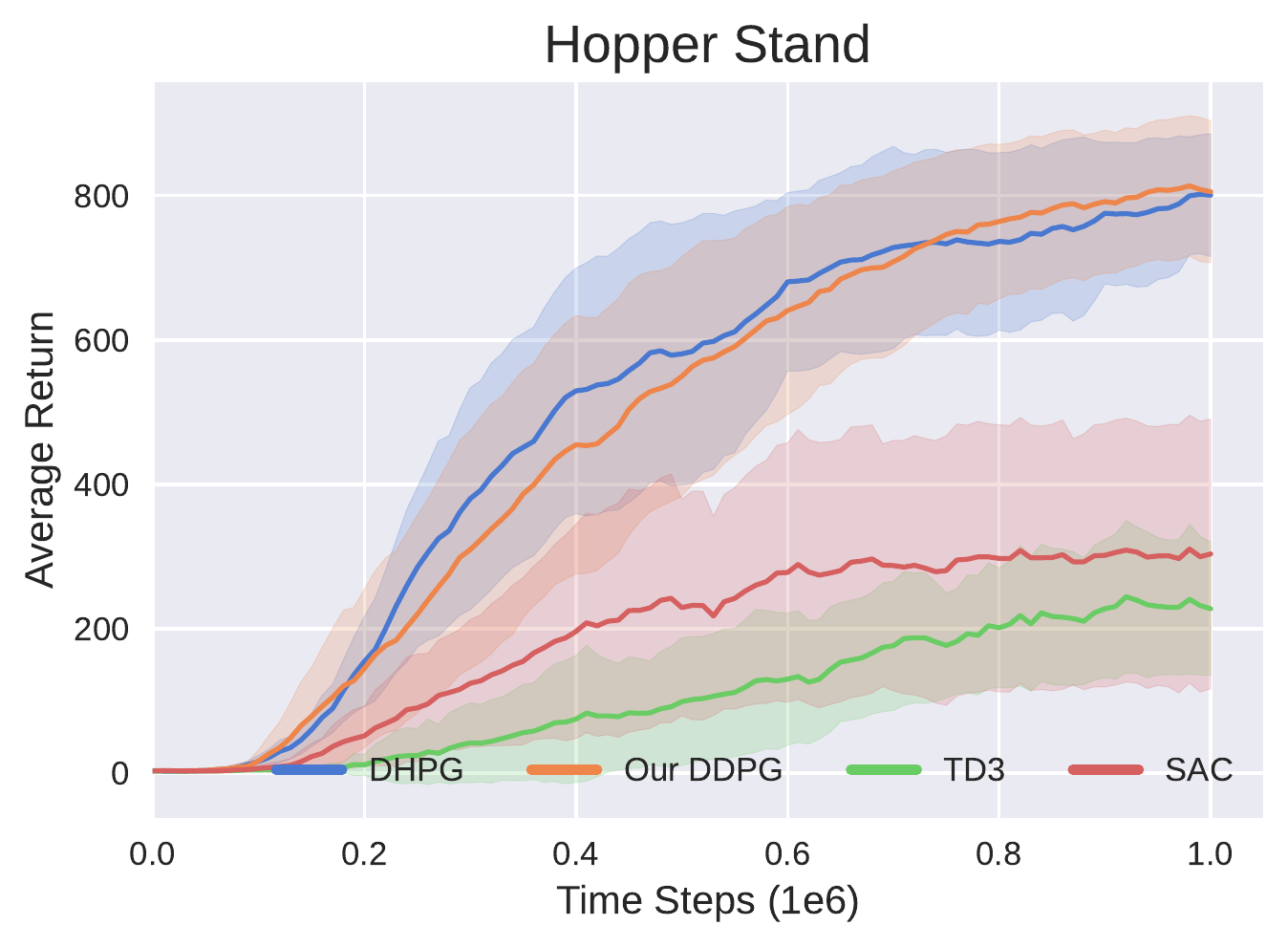}
     \end{subfigure}
     \hfill
     \begin{subfigure}[b]{0.24\textwidth}
         \centering
         \includegraphics[width=\textwidth]{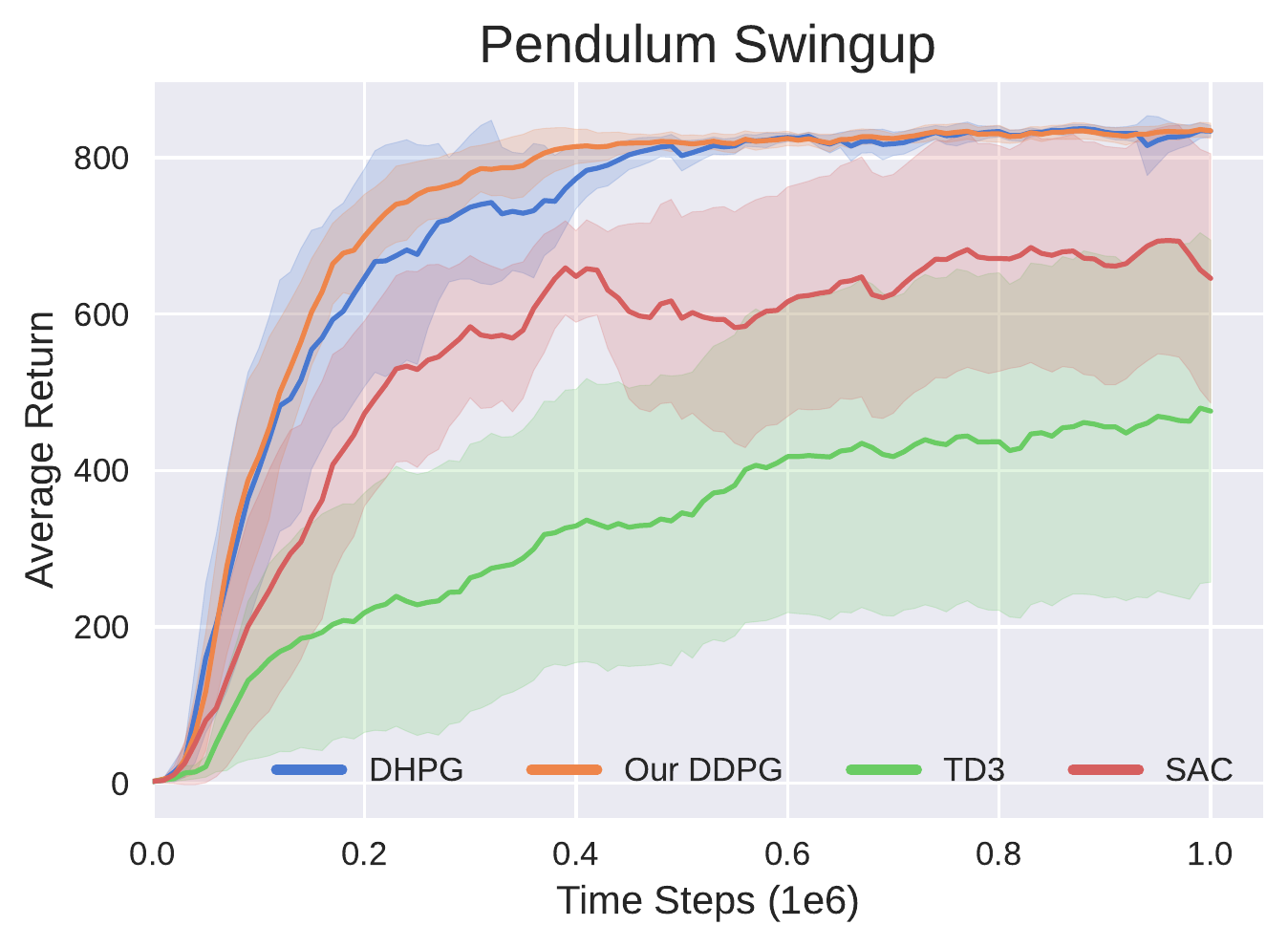}
     \end{subfigure}
        \hfill     
     \begin{subfigure}[b]{0.24\textwidth}
         \centering
         \includegraphics[width=\textwidth]{figures/states_main/states_quadruped_walk_episode_reward_eval.pdf}
     \end{subfigure}
     \hfill
     
     \hfill
     \begin{subfigure}[b]{0.24\textwidth}
         \centering
         \includegraphics[width=\textwidth]{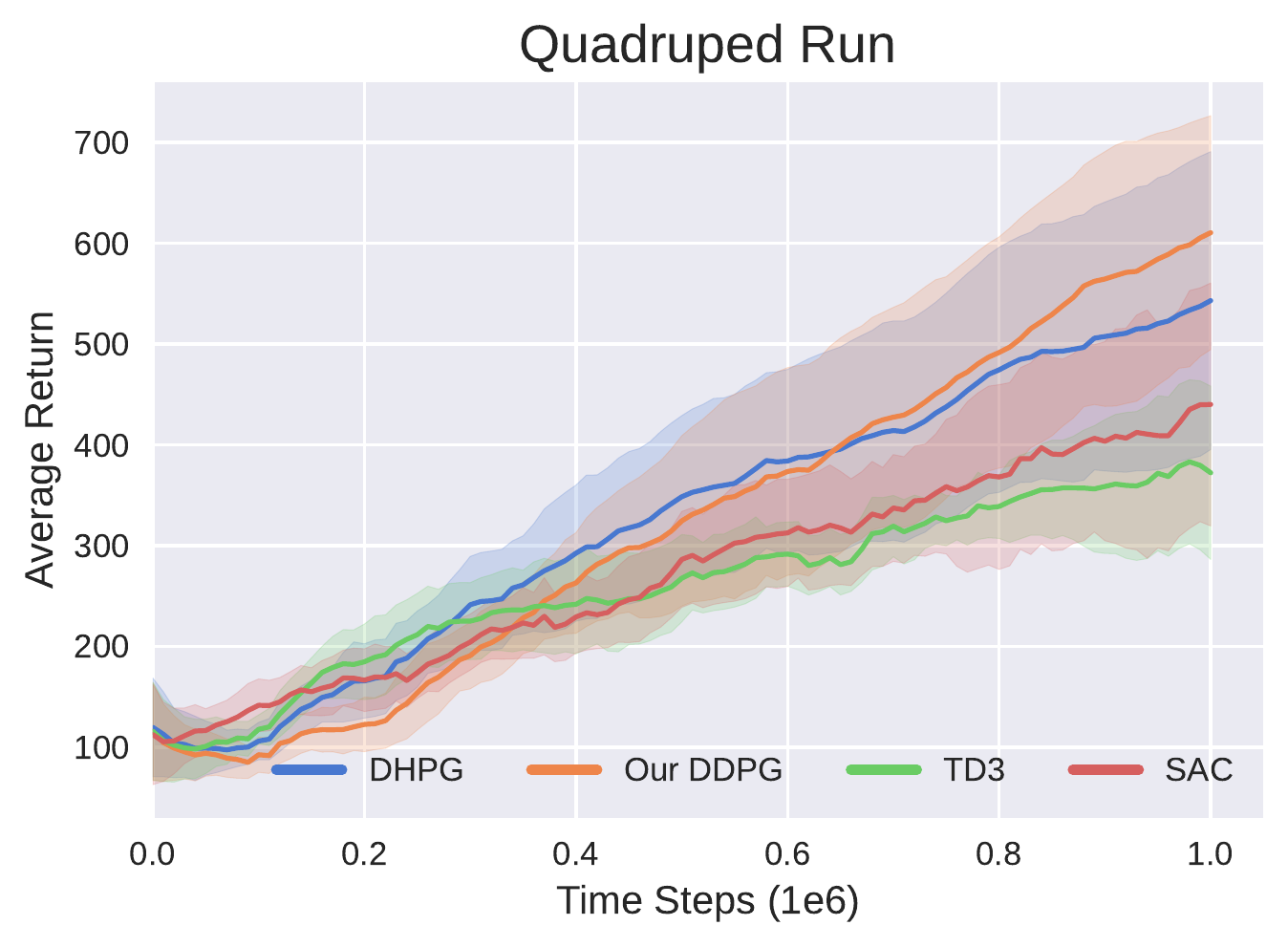}
     \end{subfigure}
     \hfill
     \begin{subfigure}[b]{0.24\textwidth}
         \centering
         \includegraphics[width=\textwidth]{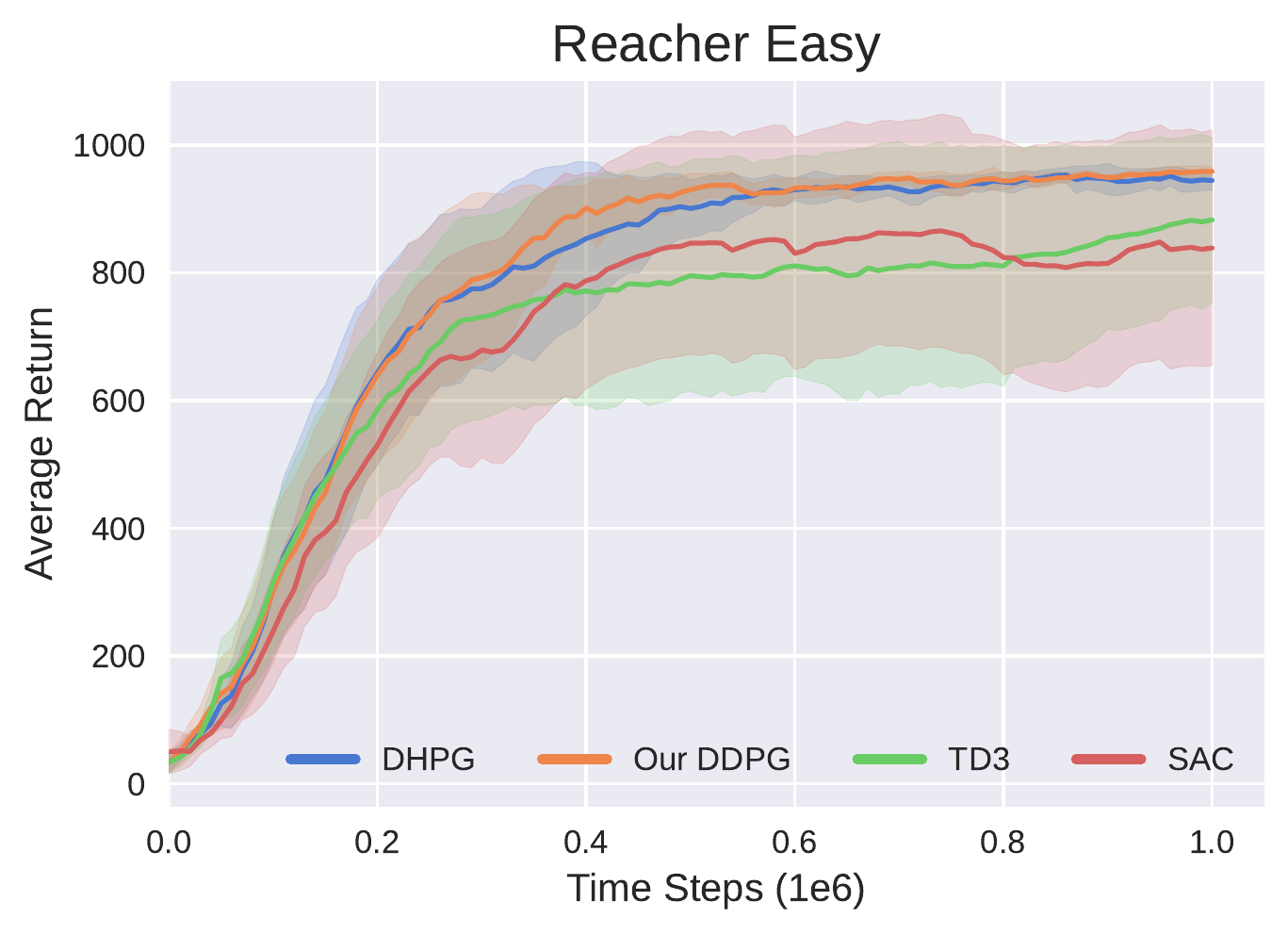}
     \end{subfigure}
     \hfill
     \begin{subfigure}[b]{0.24\textwidth}
         \centering
         \includegraphics[width=\textwidth]{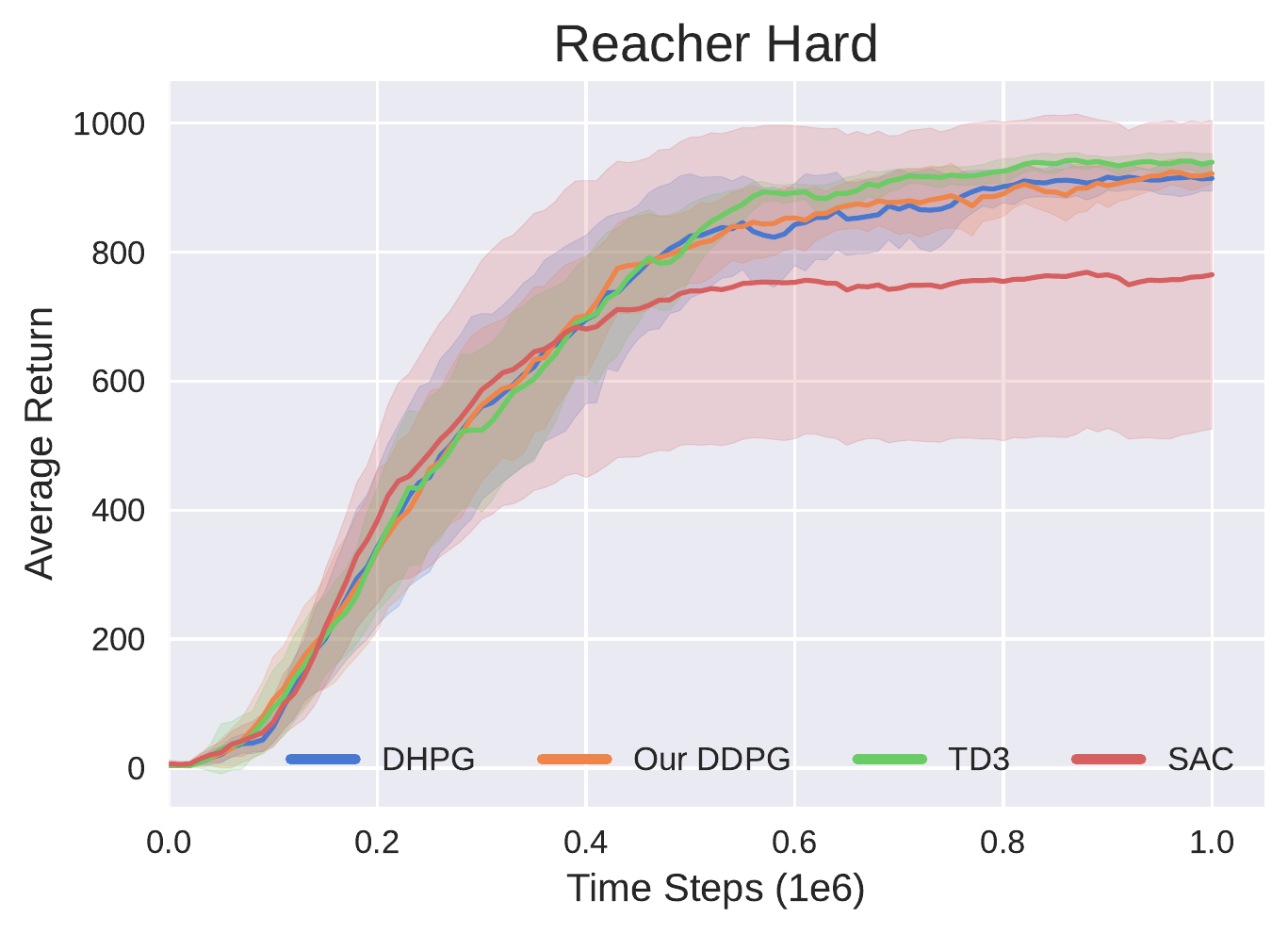}
     \end{subfigure}
     \hfill
     \begin{subfigure}[b]{0.24\textwidth}
         \centering
         \includegraphics[width=\textwidth]{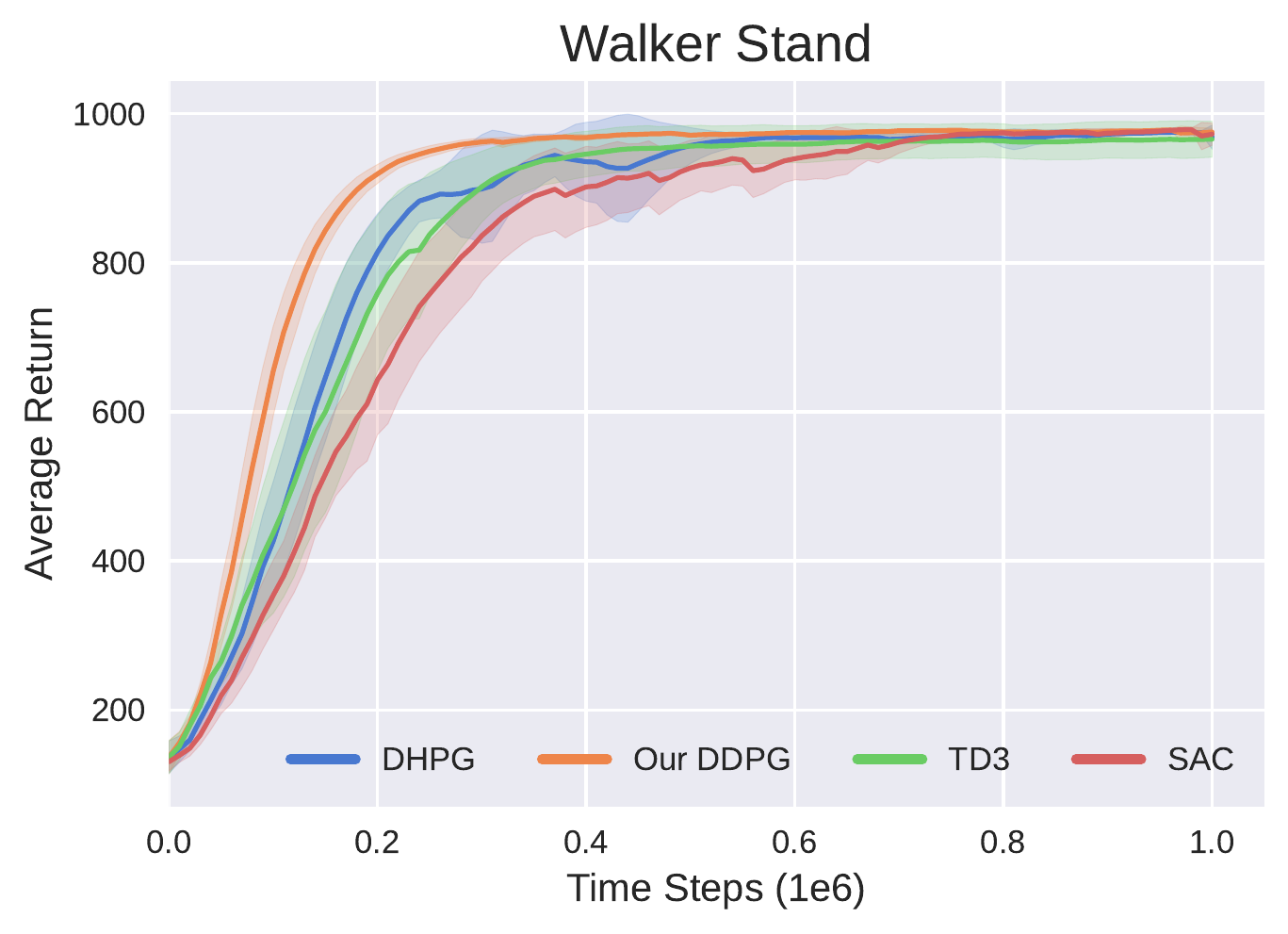}
     \end{subfigure}

     \begin{subfigure}[b]{0.24\textwidth}
         \centering
         \includegraphics[width=\textwidth]{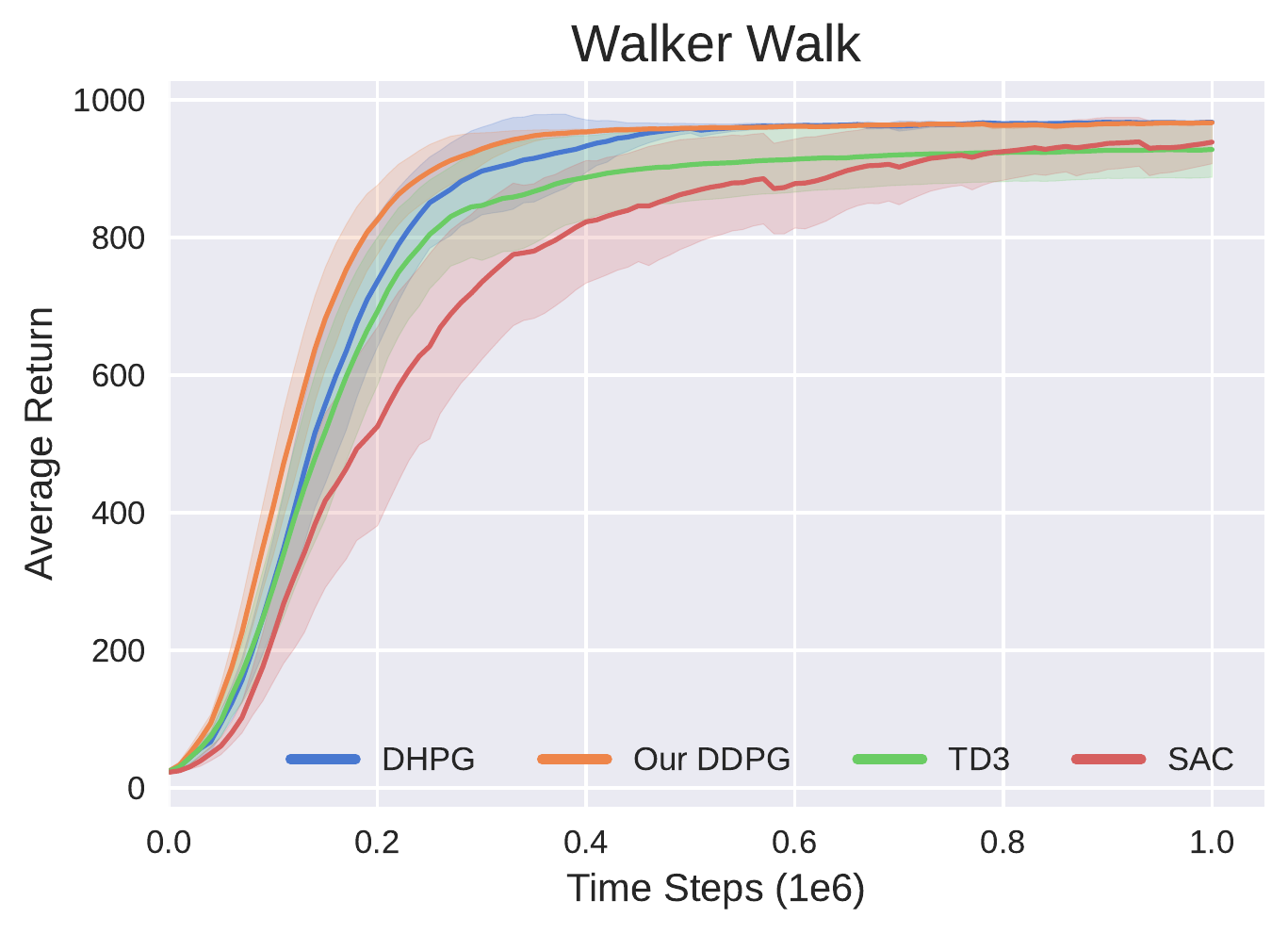}
     \end{subfigure}
     \begin{subfigure}[b]{0.24\textwidth}
         \centering
         \includegraphics[width=\textwidth]{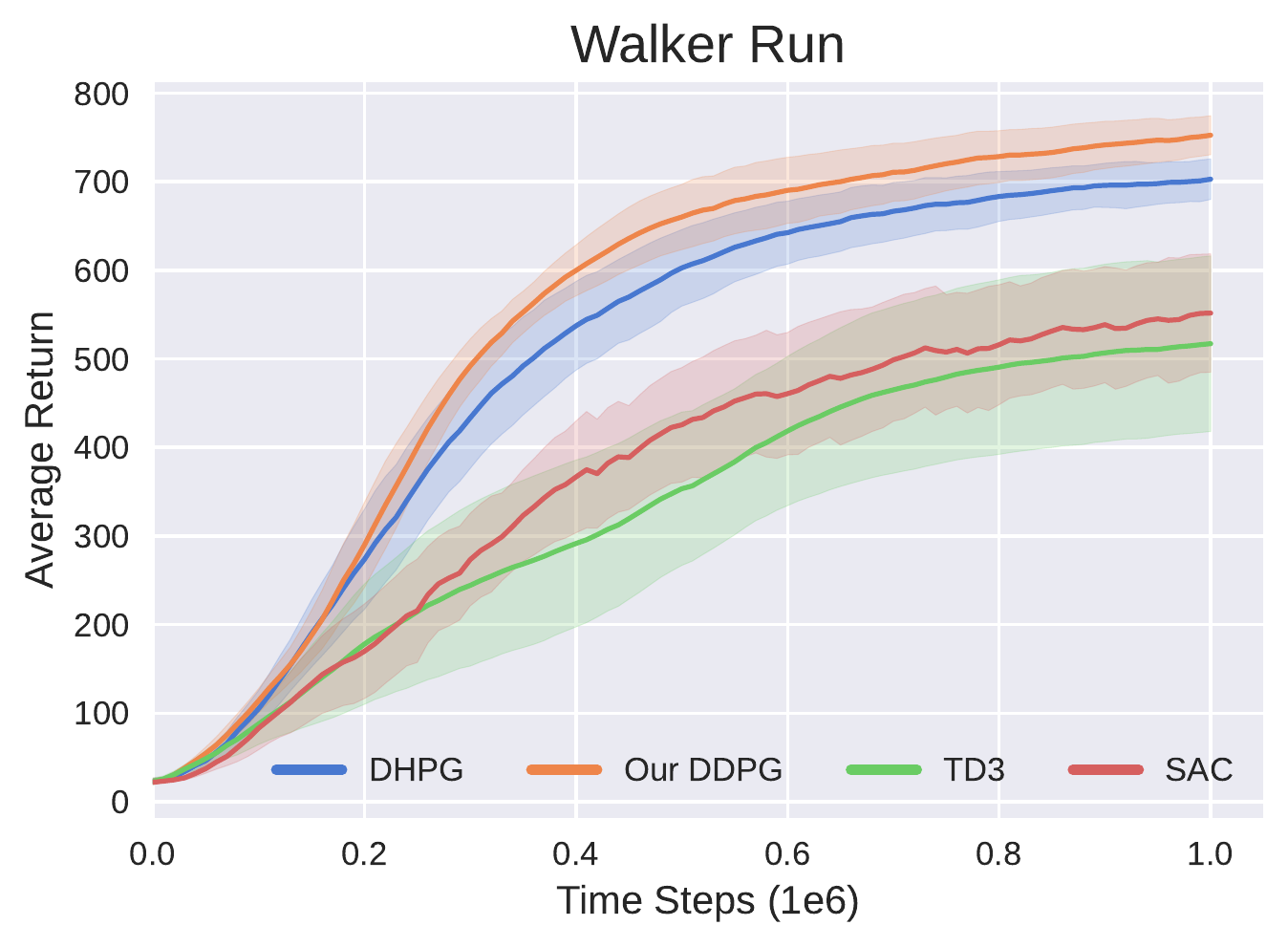}
     \end{subfigure}
    \caption{Learning curves for 18 DM control tasks with \textbf{state observations}. Mean performance is obtained over 10 seeds and shaded regions represent $95\%$ confidence intervals. Plots are smoothed uniformly for visual clarity.}
    \label{fig:state_results_supp}
\end{figure}

Figure \ref{fig:state_results_supp} shows full results obtained on 18 DeepMind Control Suite tasks with state observations to supplement results of Section \ref{sec:results_states}. Domains that require excessive exploration and large number of time steps (e.g., acrobot, swimmer, and humanoid) are not included in this benchmark.

\begin{figure}[h!]
    \centering
    \begin{subfigure}[b]{0.45\textwidth}
        \includegraphics[width=\textwidth]{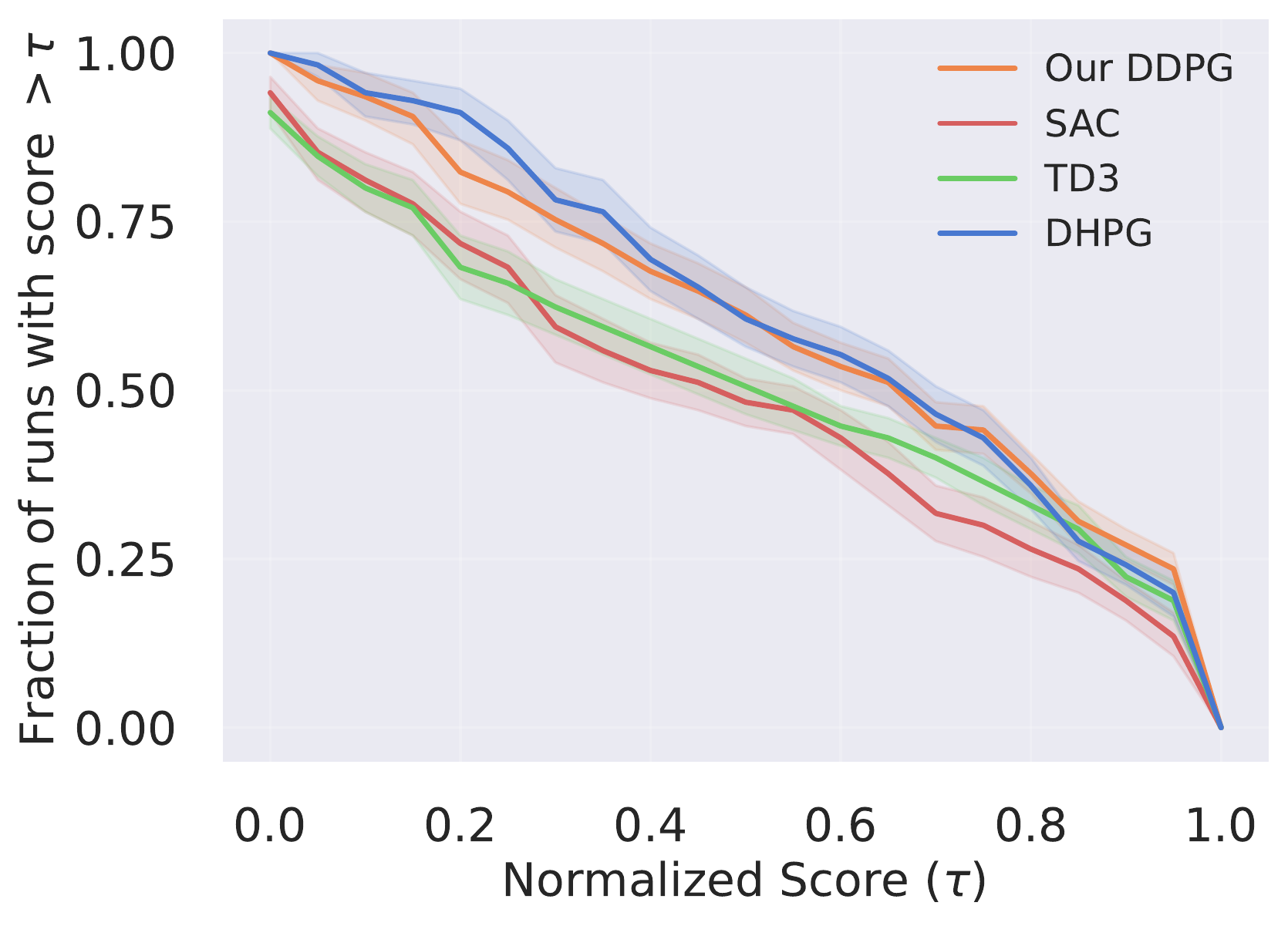}
        \caption{250k step benchmark.}
    \end{subfigure}
    \hfill
    \begin{subfigure}[b]{0.45\textwidth}
        \includegraphics[width=\textwidth]{figures/states_rliable/states_performance_profiles_500k.pdf}
        \caption{500k step benchmark.}
    \end{subfigure}
    \caption{Performance profiles for \textbf{state observations} based on 17 tasks over 10 seeds, at 250k steps \textbf{(a)}, and at 500k steps \textbf{(b)}. Shaded regions represent $95\%$ confidence intervals.}    
    \label{fig:state_results_performance_profiles}
\end{figure}

\begin{figure}[h!]
    \centering
    \begin{subfigure}[b]{0.95\textwidth}
        \includegraphics[width=\textwidth]{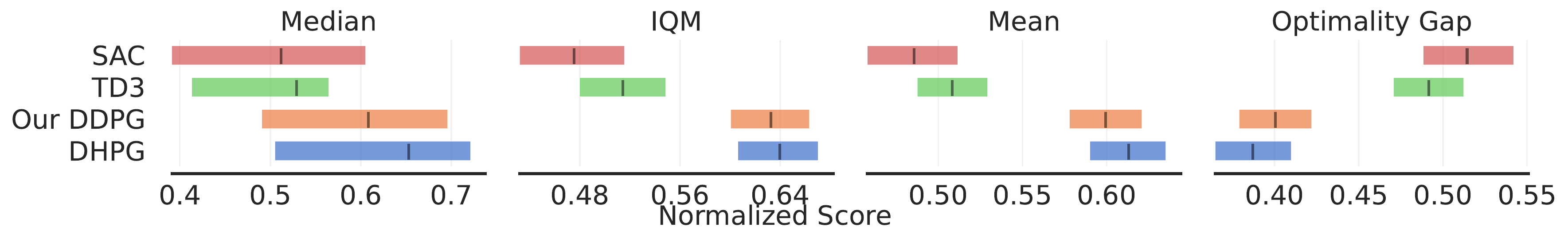}
        \caption{250k step benchmark.}
    \end{subfigure}
    
    \begin{subfigure}[b]{0.95\textwidth}
        \includegraphics[width=\textwidth]{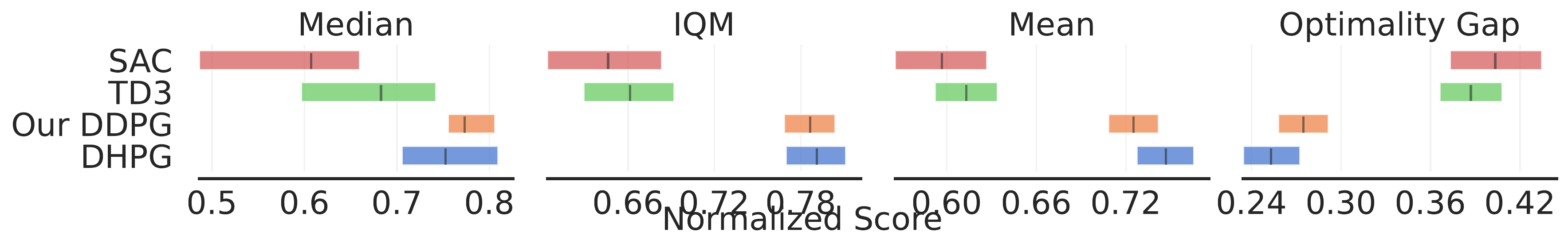}
        \caption{500k step benchmark.}
    \end{subfigure}
    \caption{Aggregate metrics for \textbf{state observations} with $95\%$ confidence intervals based on 17 tasks over 10 seeds, at 250k steps \textbf{(a)}, and at 500k steps \textbf{(b)}.}    
    \label{fig:state_results_aggregate_metrics}
\end{figure}

Figures \ref{fig:state_results_performance_profiles} and \ref{fig:state_results_aggregate_metrics} respectively show performance profiles and aggregate metrics \cite{agarwal2021deep} on 17 tasks; hopper hop is removed from RLiable evaluation as none of the algorithms have acquired reasonable performance in 1 million steps. 


\clearpage

\clearpage

\subsection{Pixel Observations}
\begin{figure}[h!]
     \centering
     \begin{subfigure}[b]{0.24\textwidth}
         \centering
         \includegraphics[width=\textwidth]{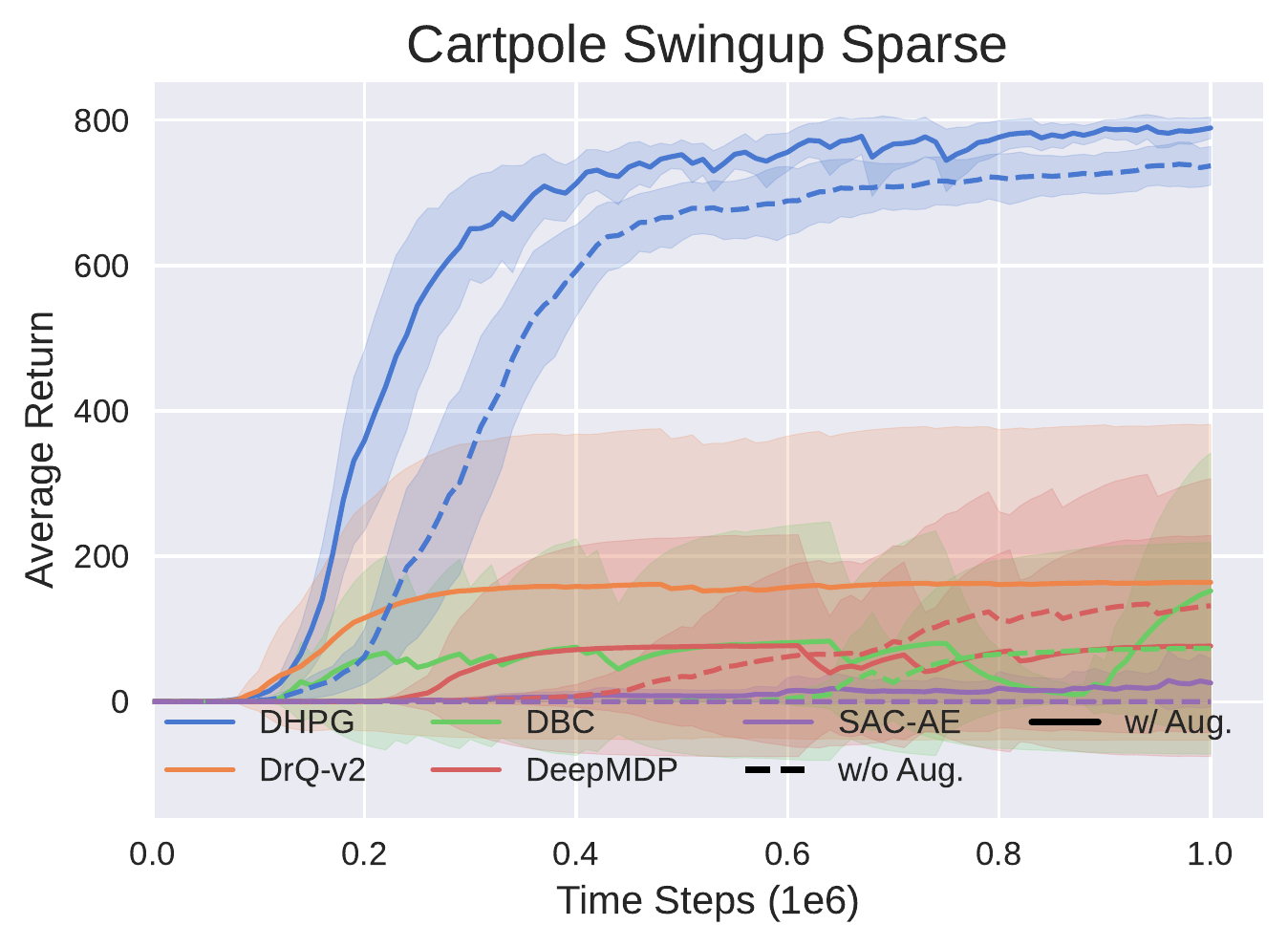}
     \end{subfigure}
     \hfill
     \begin{subfigure}[b]{0.24\textwidth}
         \centering
         \includegraphics[width=\textwidth]{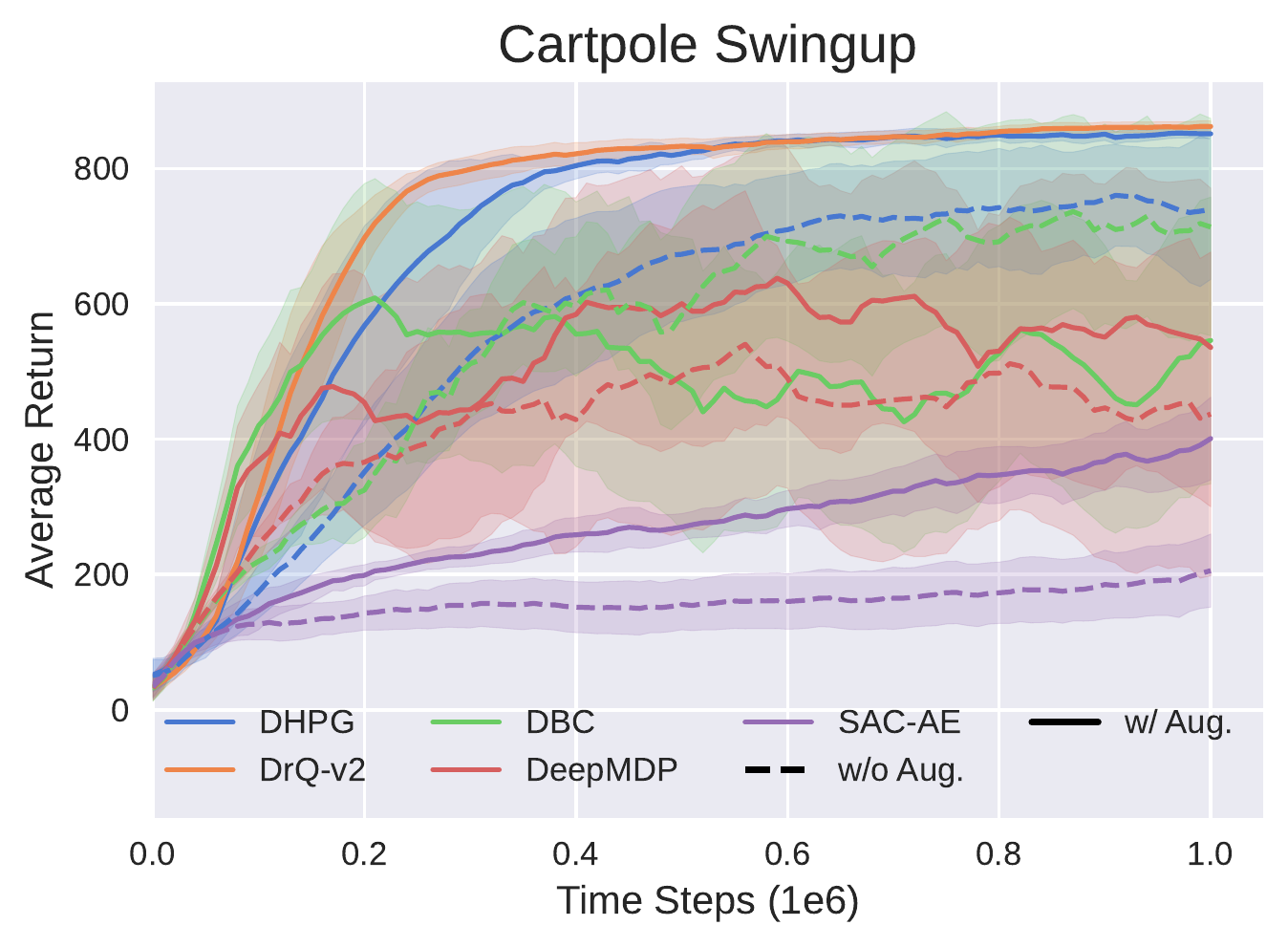}
     \end{subfigure}
     \hfill
     \begin{subfigure}[b]{0.24\textwidth}
         \centering
         \includegraphics[width=\textwidth]{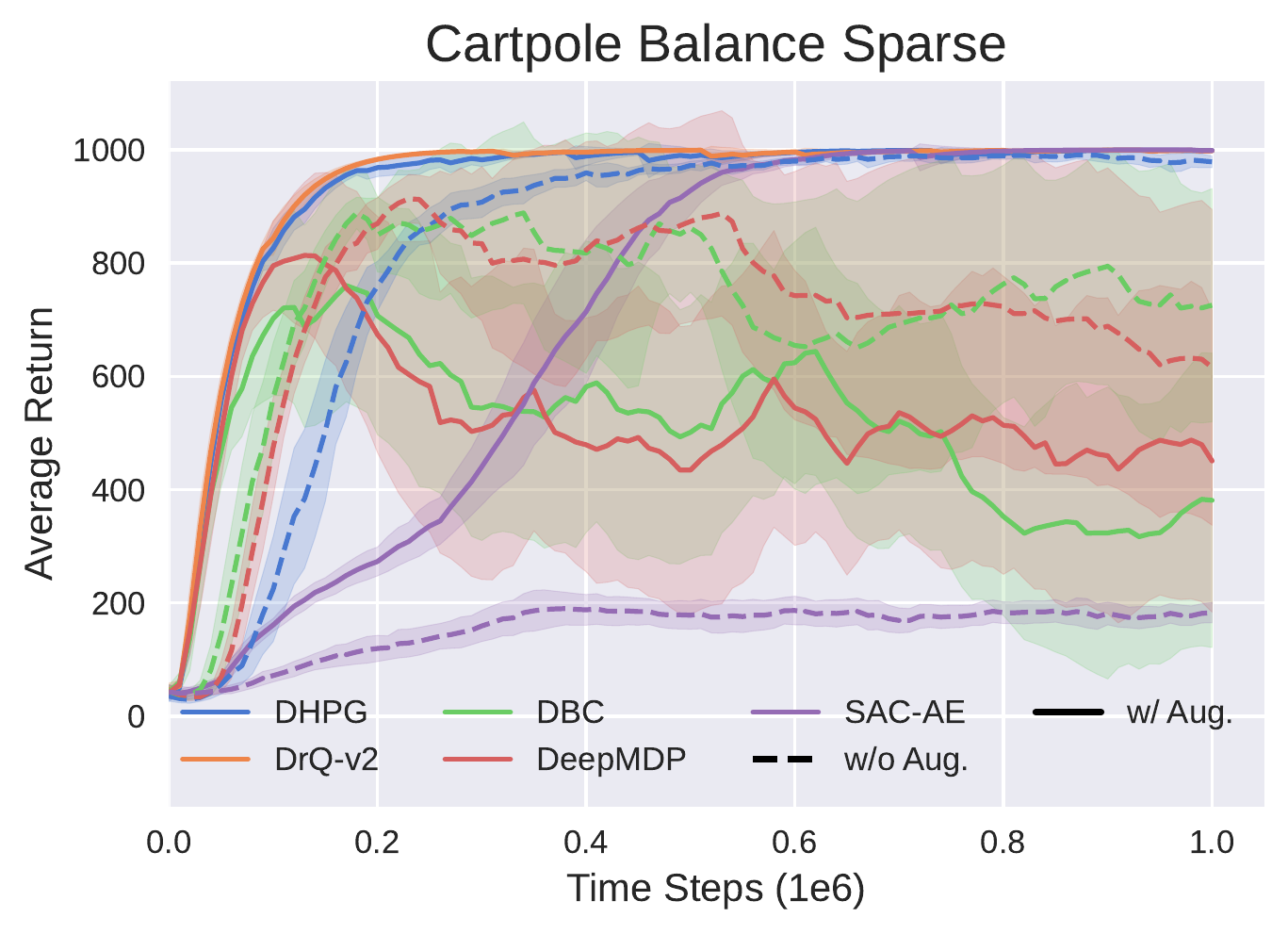}
     \end{subfigure}
     \hfill
     \begin{subfigure}[b]{0.24\textwidth}
         \centering
         \includegraphics[width=\textwidth]{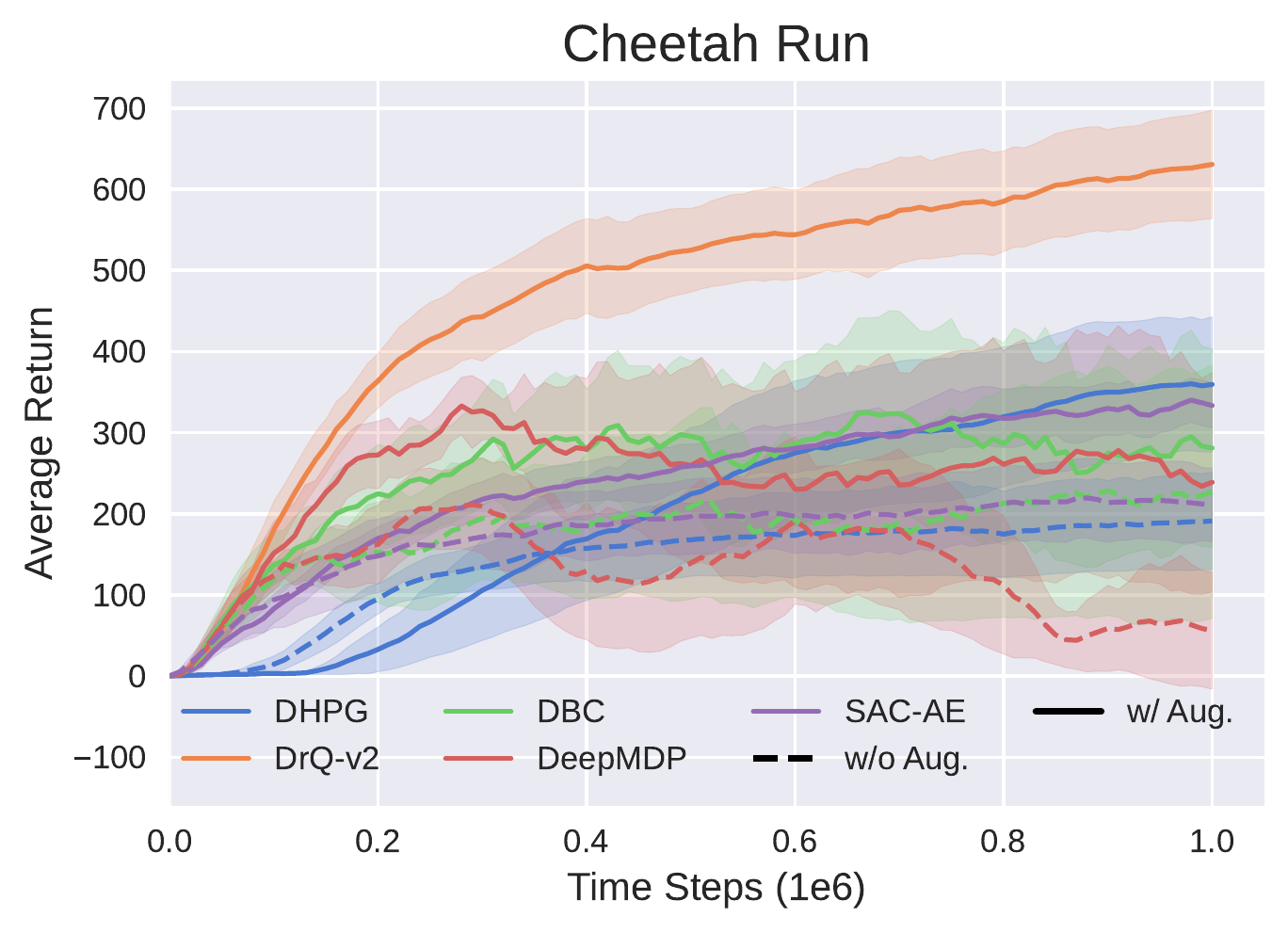}
     \end{subfigure}
     \hfill
     
     \begin{subfigure}[b]{0.24\textwidth}
         \centering
         \includegraphics[width=\textwidth]{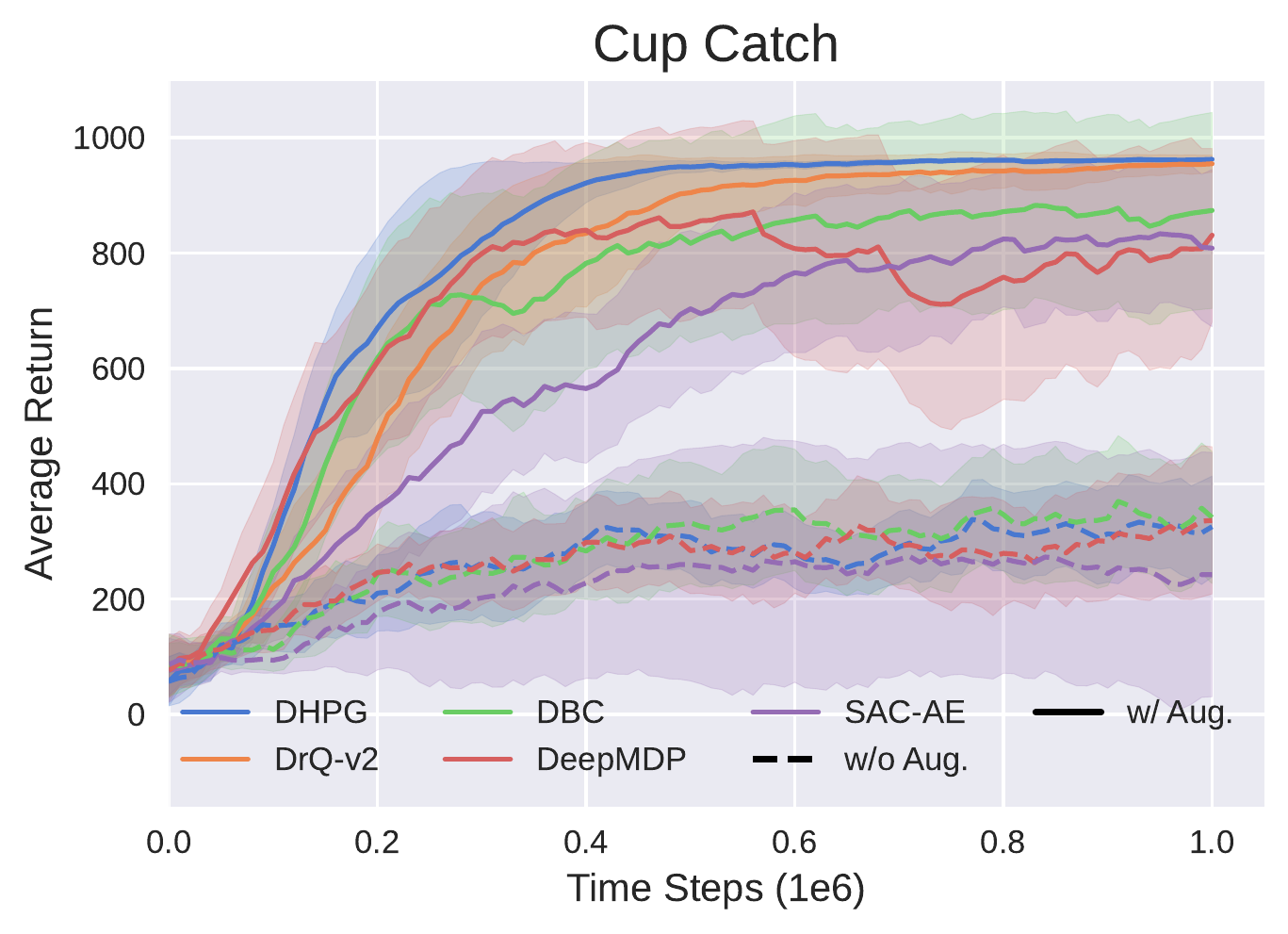}
     \end{subfigure}
     \hfill
     \begin{subfigure}[b]{0.24\textwidth}
         \centering
         \includegraphics[width=\textwidth]{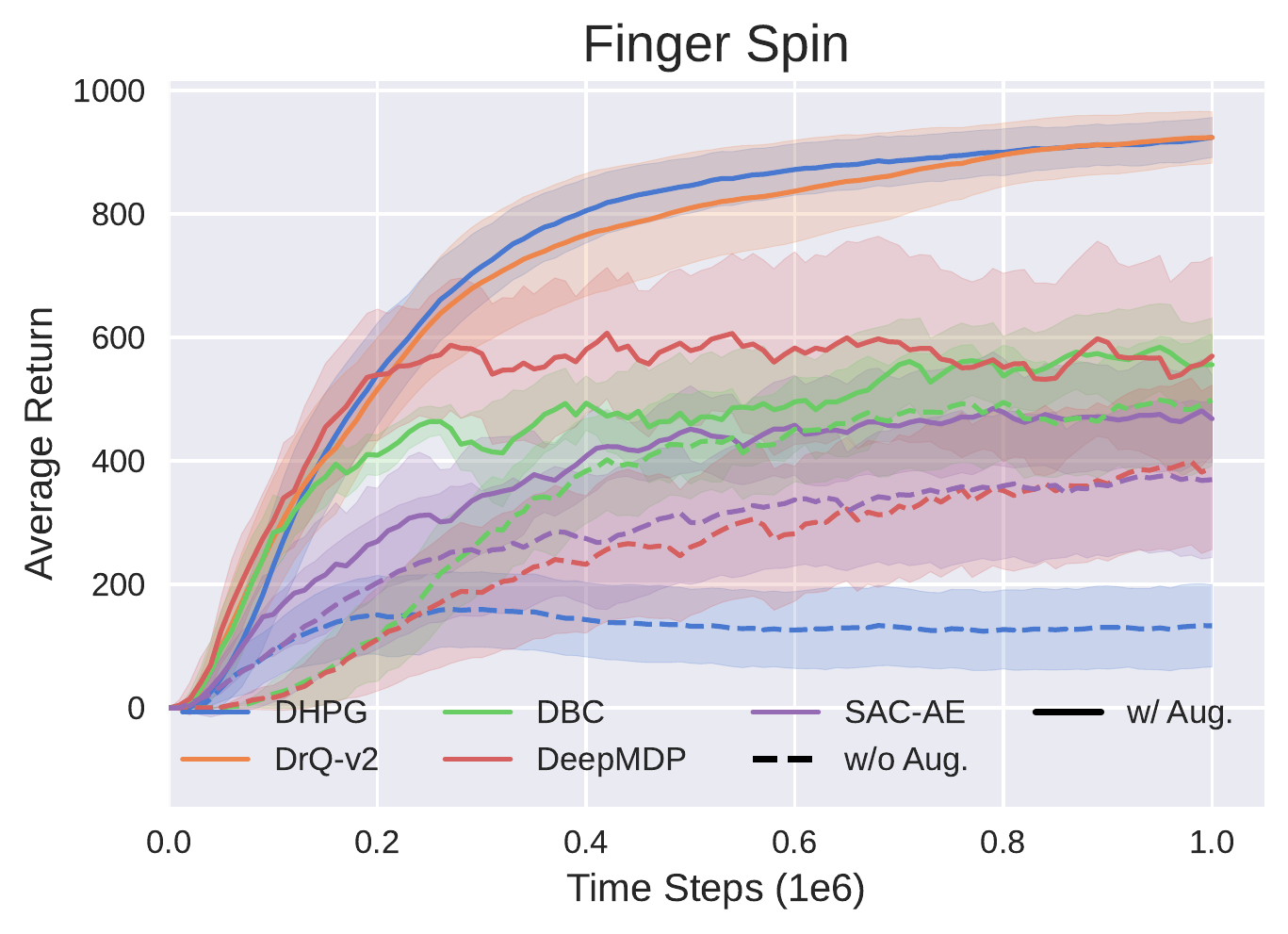}
     \end{subfigure}
     \hfill
     \begin{subfigure}[b]{0.24\textwidth}
         \centering
         \includegraphics[width=\textwidth]{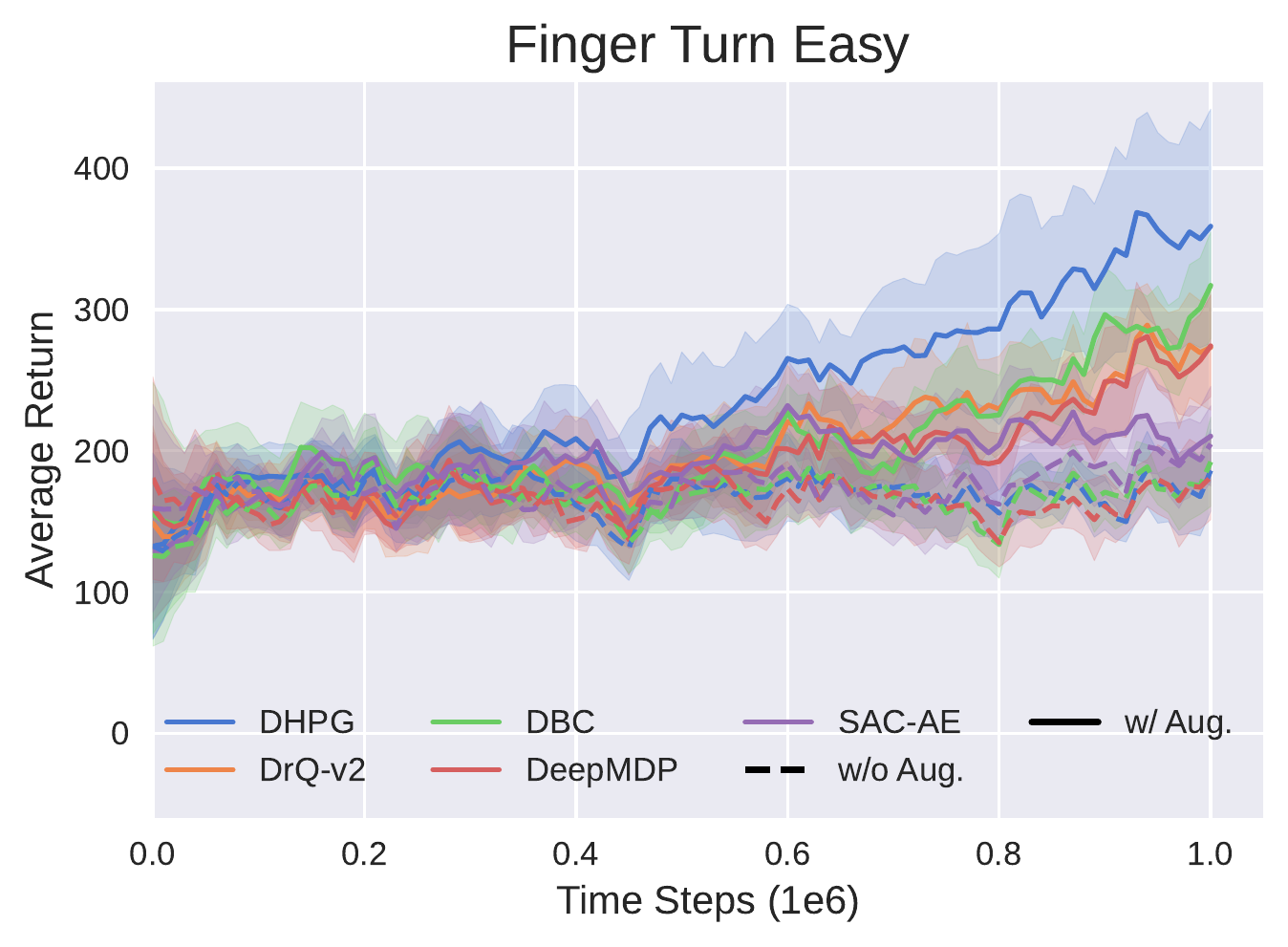}
     \end{subfigure}
     \hfill
     \begin{subfigure}[b]{0.24\textwidth}
         \centering
         \includegraphics[width=\textwidth]{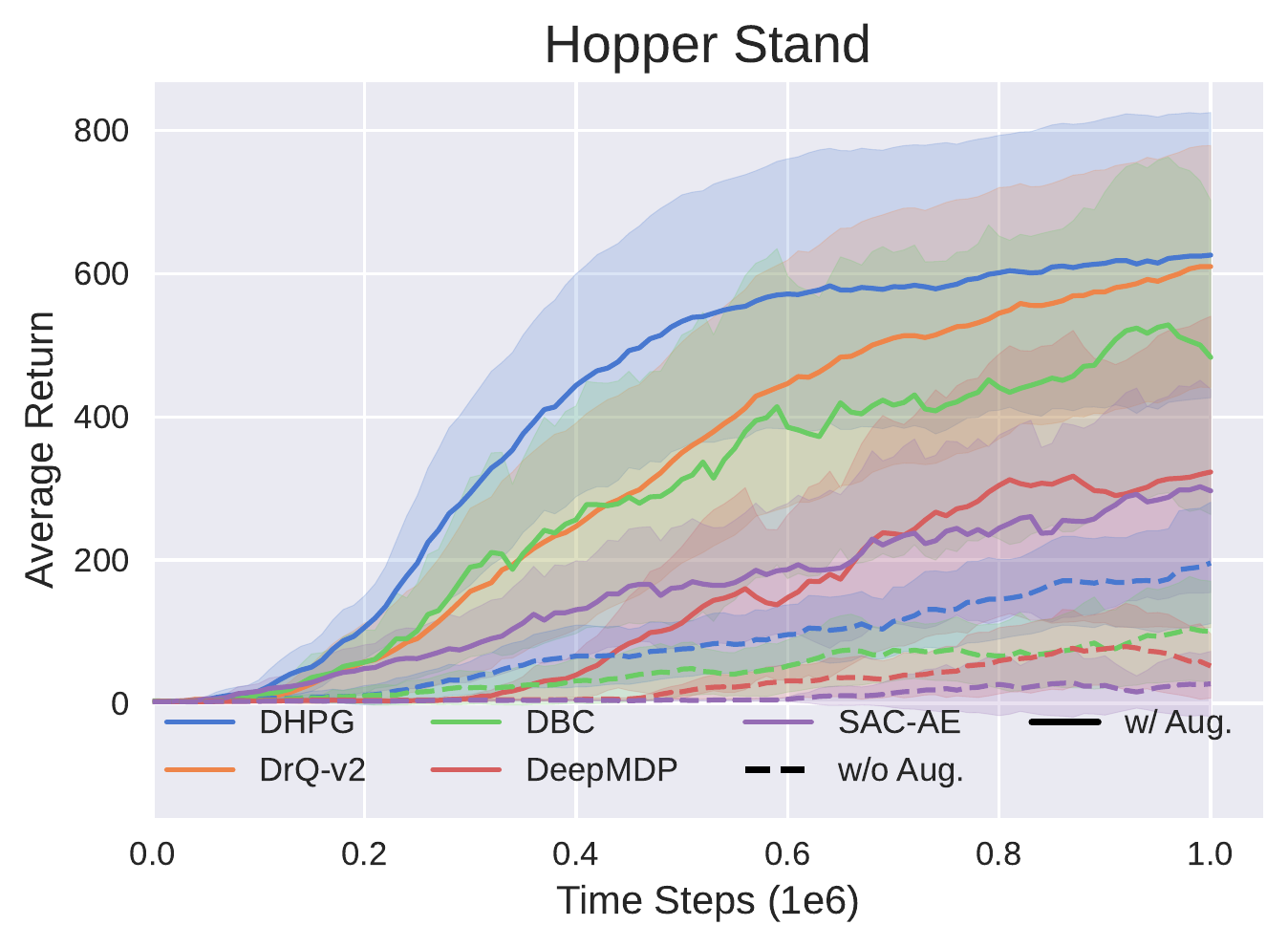}
     \end{subfigure}
     \hfill
     
     \begin{subfigure}[b]{0.24\textwidth}
         \centering
         \includegraphics[width=\textwidth]{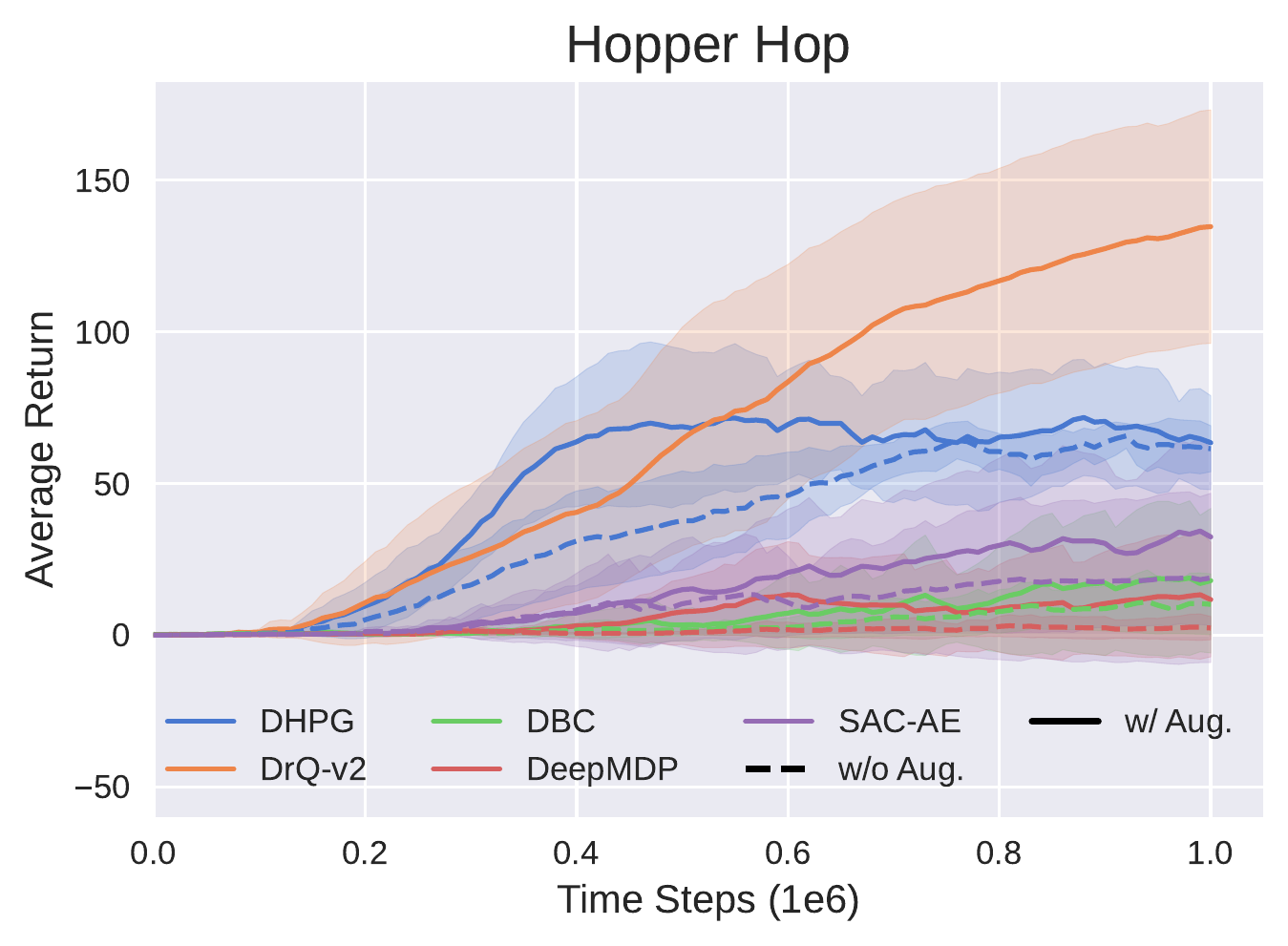}
     \end{subfigure}
     \hfill
     \begin{subfigure}[b]{0.24\textwidth}
         \centering
         \includegraphics[width=\textwidth]{figures/pixels_main/pixels_pendulum_swingup_episode_reward_eval.pdf}
     \end{subfigure}
     \hfill
     \begin{subfigure}[b]{0.24\textwidth}
         \centering
         \includegraphics[width=\textwidth]{figures/pixels_main/pixels_quadruped_walk_episode_reward_eval.pdf}
     \end{subfigure}
     \hfill
     \begin{subfigure}[b]{0.24\textwidth}
         \centering
         \includegraphics[width=\textwidth]{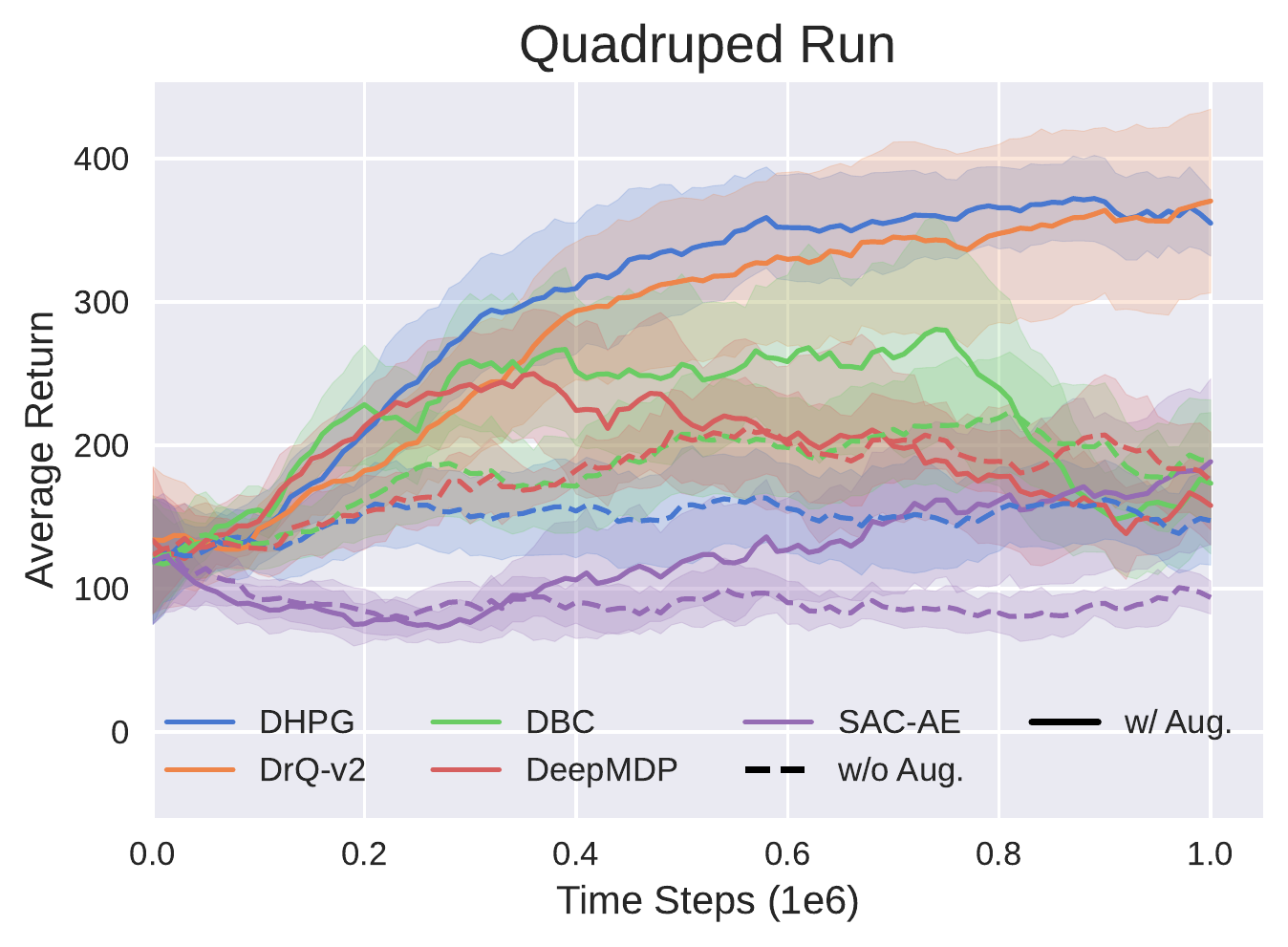}
     \end{subfigure}
     \hfill
     
     \begin{subfigure}[b]{0.24\textwidth}
         \centering
         \includegraphics[width=\textwidth]{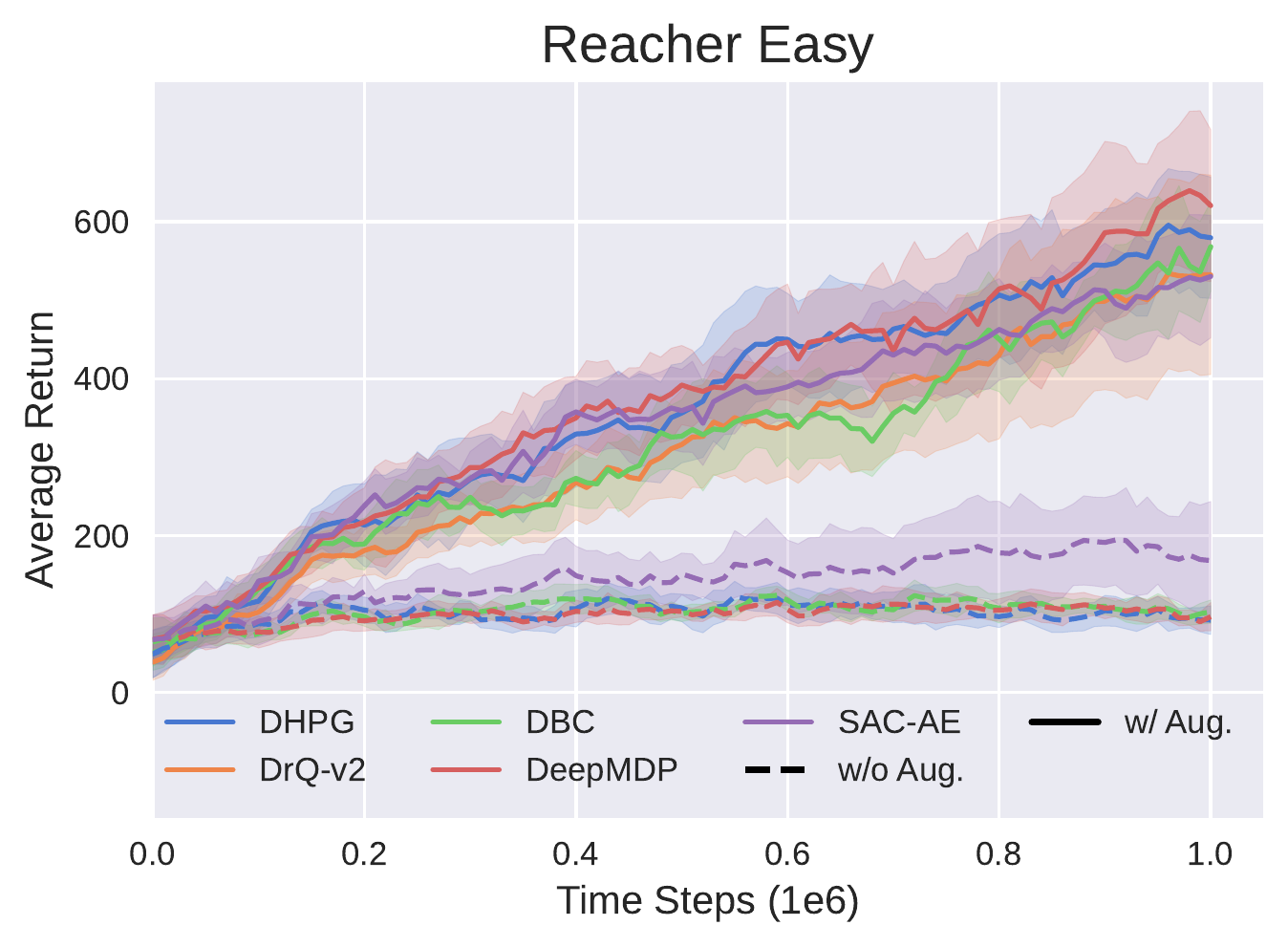}
     \end{subfigure}
     \hfill
     \begin{subfigure}[b]{0.24\textwidth}
         \centering
         \includegraphics[width=\textwidth]{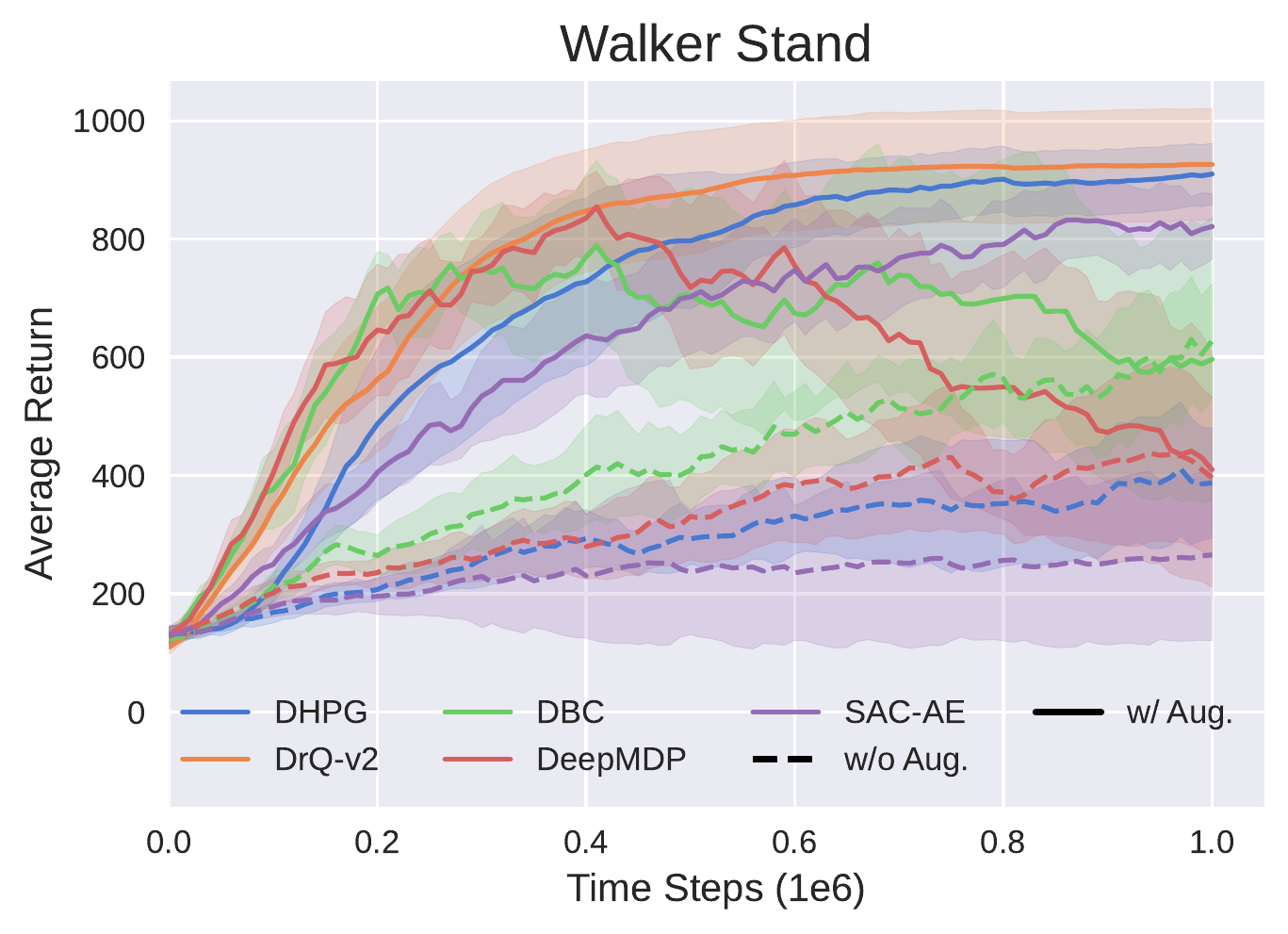}
     \end{subfigure}
     \begin{subfigure}[b]{0.24\textwidth}
         \centering
         \includegraphics[width=\textwidth]{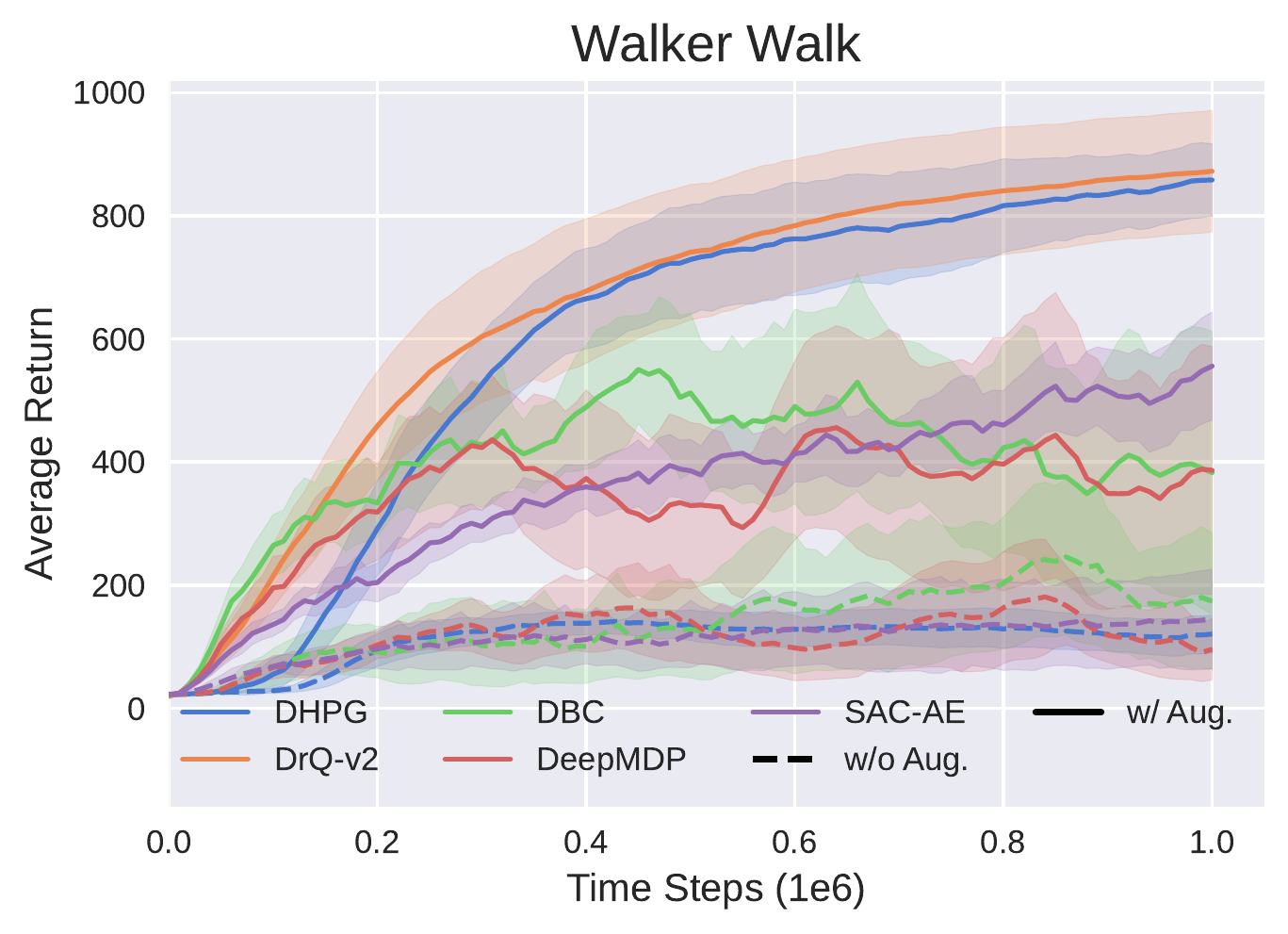}
     \end{subfigure}
     \hfill
     \begin{subfigure}[b]{0.24\textwidth}
         \centering
         \includegraphics[width=\textwidth]{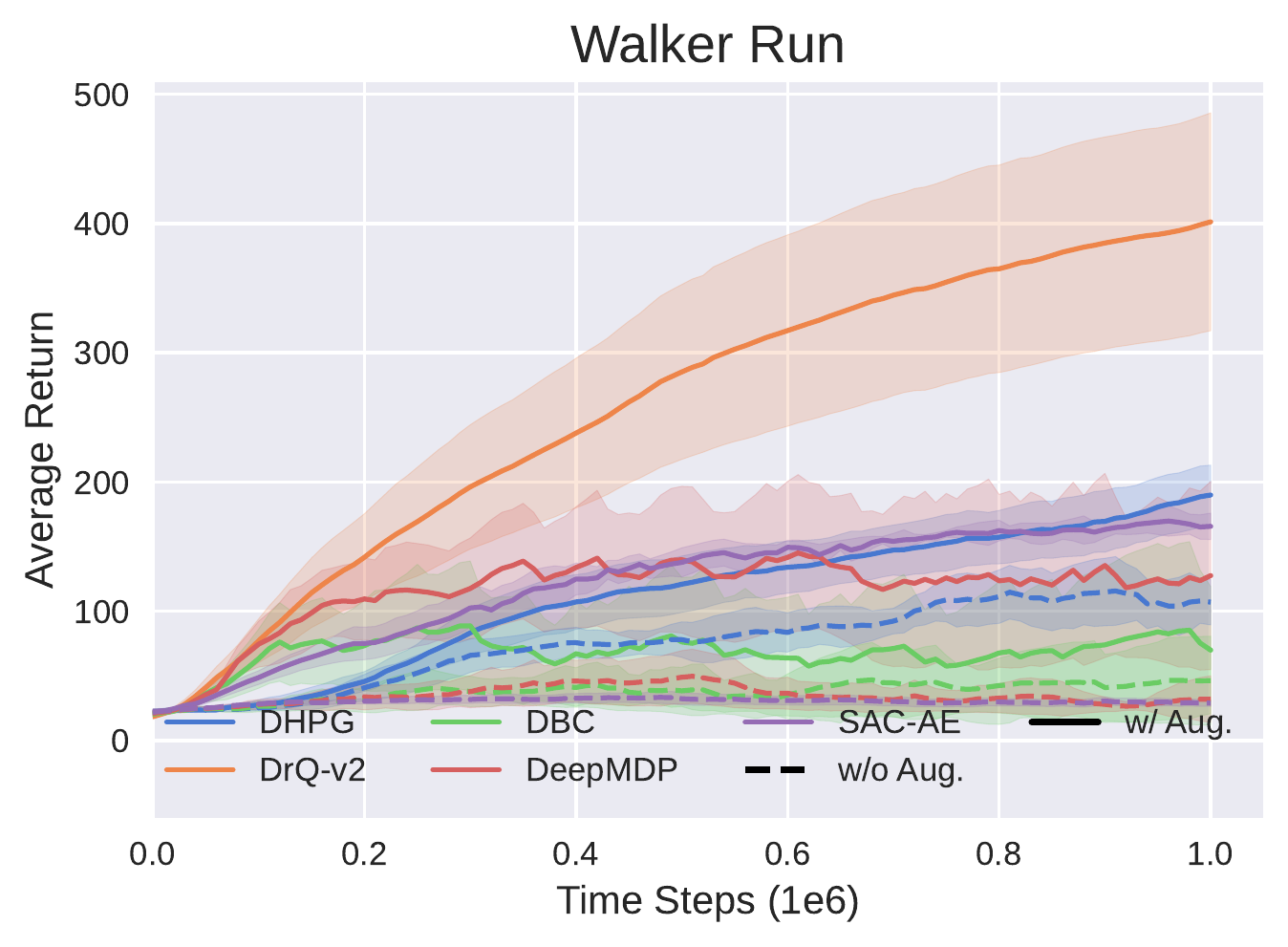}
     \end{subfigure}
     \hfill
    \caption{Learning curves for 16 DM control tasks with \textbf{pixel observations}. Mean performance is obtained over 10 seeds and shaded regions represent $95\%$ confidence intervals. Plots are smoothed uniformly for visual clarity.}
    \label{fig:pixel_results_supp}
\end{figure}

Figure \ref{fig:pixel_results_supp} shows full results obtained on 16 DeepMind Control Suite tasks with pixel observations to supplement results of Section \ref{sec:results_pixels}. Domains that require excessive exploration and large number of time steps (e.g., acrobot, swimmer, and humanoid) and domains with visually small targets (e.g., reacher hard and finger turn hard) are not included in this benchmark. In each plot, the solid lines present algorithms with image augmentation and dashed lines present algorithms without image augmentation.

\begin{figure}[h!]
    \centering
    \begin{subfigure}[b]{0.45\textwidth}
        \includegraphics[width=\textwidth]{figures/pixels_rliable/pixels_performance_profiles_500k.pdf}
        \caption{500k step benchmark.}
    \end{subfigure}
    \hfill
    \begin{subfigure}[b]{0.45\textwidth}
        \includegraphics[width=\textwidth]{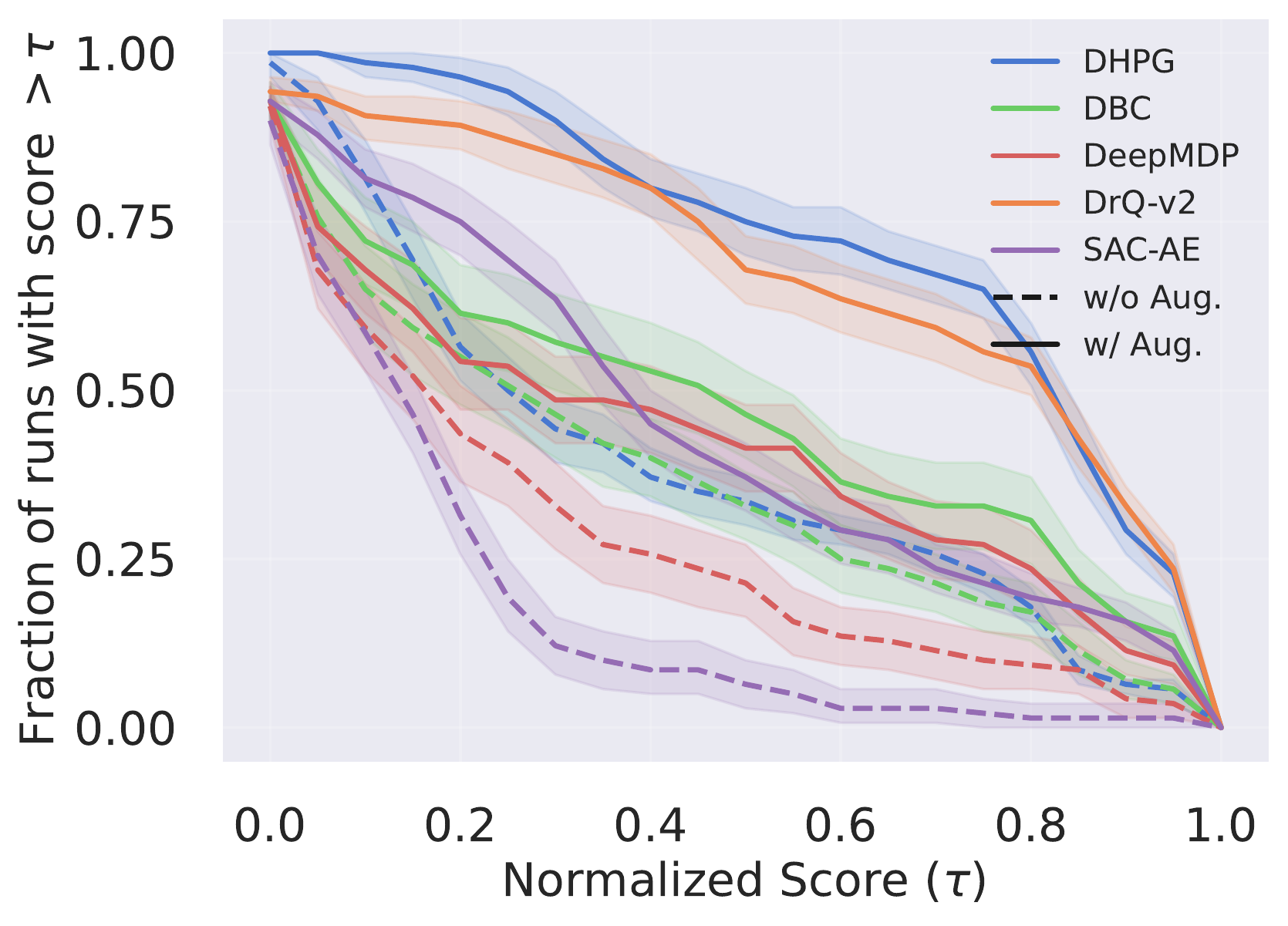}
        \caption{1m step benchmark.}
    \end{subfigure}
    \caption{Performance profiles for \textbf{pixel observations} based on 14 tasks over 10 seeds, at 500k steps \textbf{(a)}, and at 1m steps \textbf{(b)}. Shaded regions represent $95\%$ confidence intervals.}    
    \label{fig:pixel_results_performance_profiles}
\end{figure}

\begin{figure}[h!]
    \centering
    %
    \begin{subfigure}[b]{0.95\textwidth}
        \includegraphics[width=\textwidth]{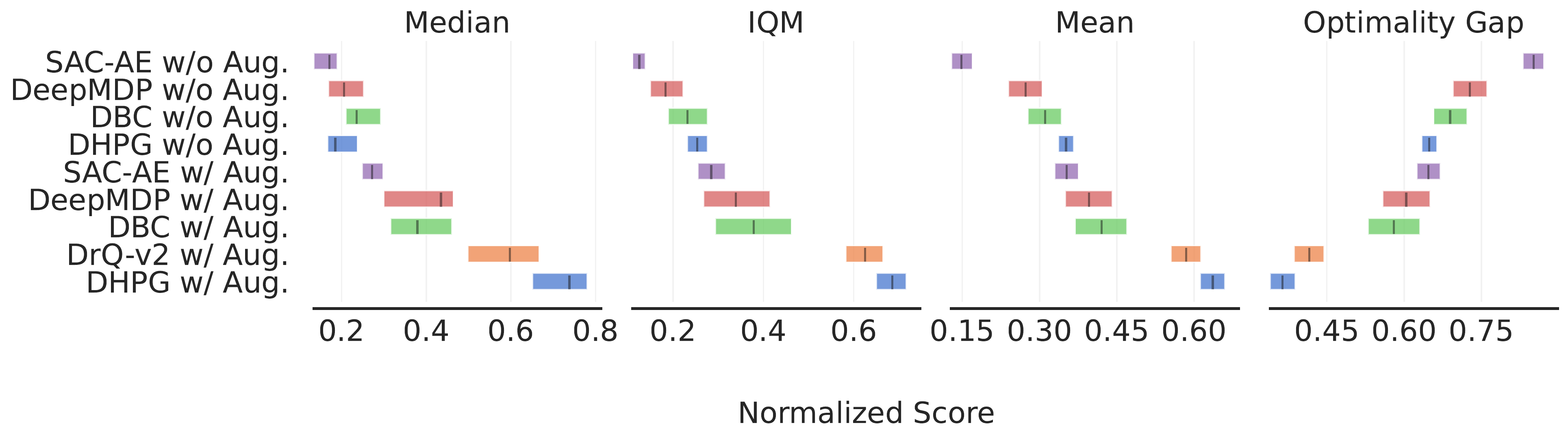}
        \caption{500k step benchmark.}
    \end{subfigure}
    
    \begin{subfigure}[b]{0.95\textwidth}
        \includegraphics[width=\textwidth]{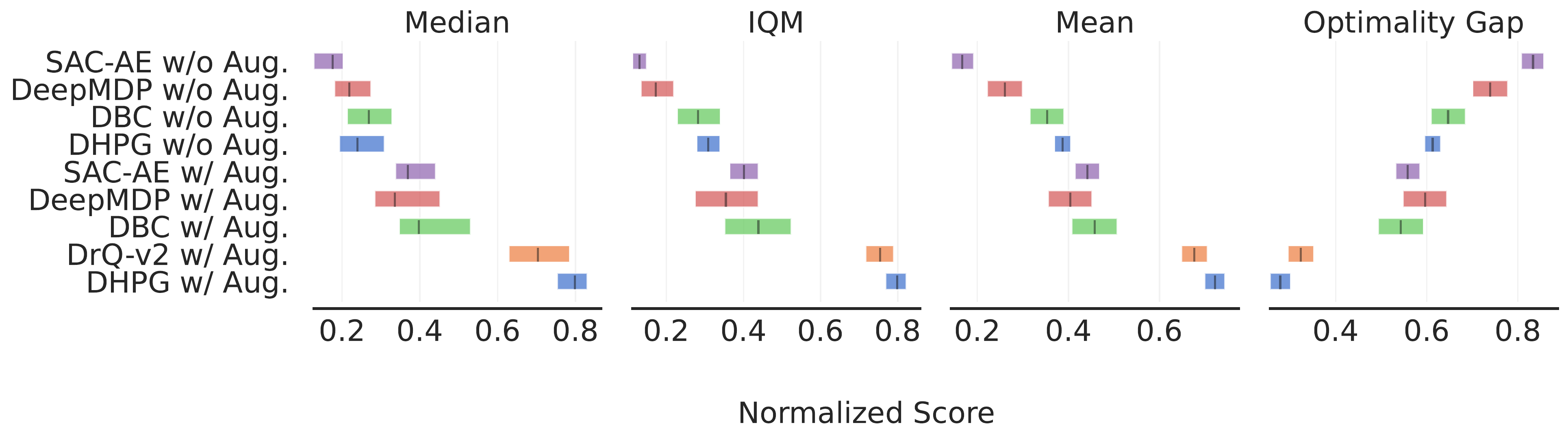}
        \caption{1m step benchmark.}
    \end{subfigure}
    \caption{Aggregate metrics for \textbf{pixel observations} with $95\%$ confidence intervals based on 14 tasks over 10 seeds, at 500k steps \textbf{(a)}, and at 1m steps \textbf{(b)}.}    
    \label{fig:pixel_results_aggregate_metrics}
\end{figure}

Figures \ref{fig:pixel_results_performance_profiles} and \ref{fig:pixel_results_aggregate_metrics} respectively show performance profiles and aggregate metrics \cite{agarwal2021deep} on 14 tasks; hopper hop and walker run are removed from RLiable evaluation as none of the algorithms have acquired reasonable performance in 1 million steps.
\label{sec:additional_results_pixels}

\clearpage

\clearpage

\subsection{Transfer Learning Experiments}
\label{sec:transfer_supp}
\begin{figure}[h!]
     \centering
     \begin{subfigure}[b]{0.32\textwidth}
         \centering
         \includegraphics[width=\textwidth]{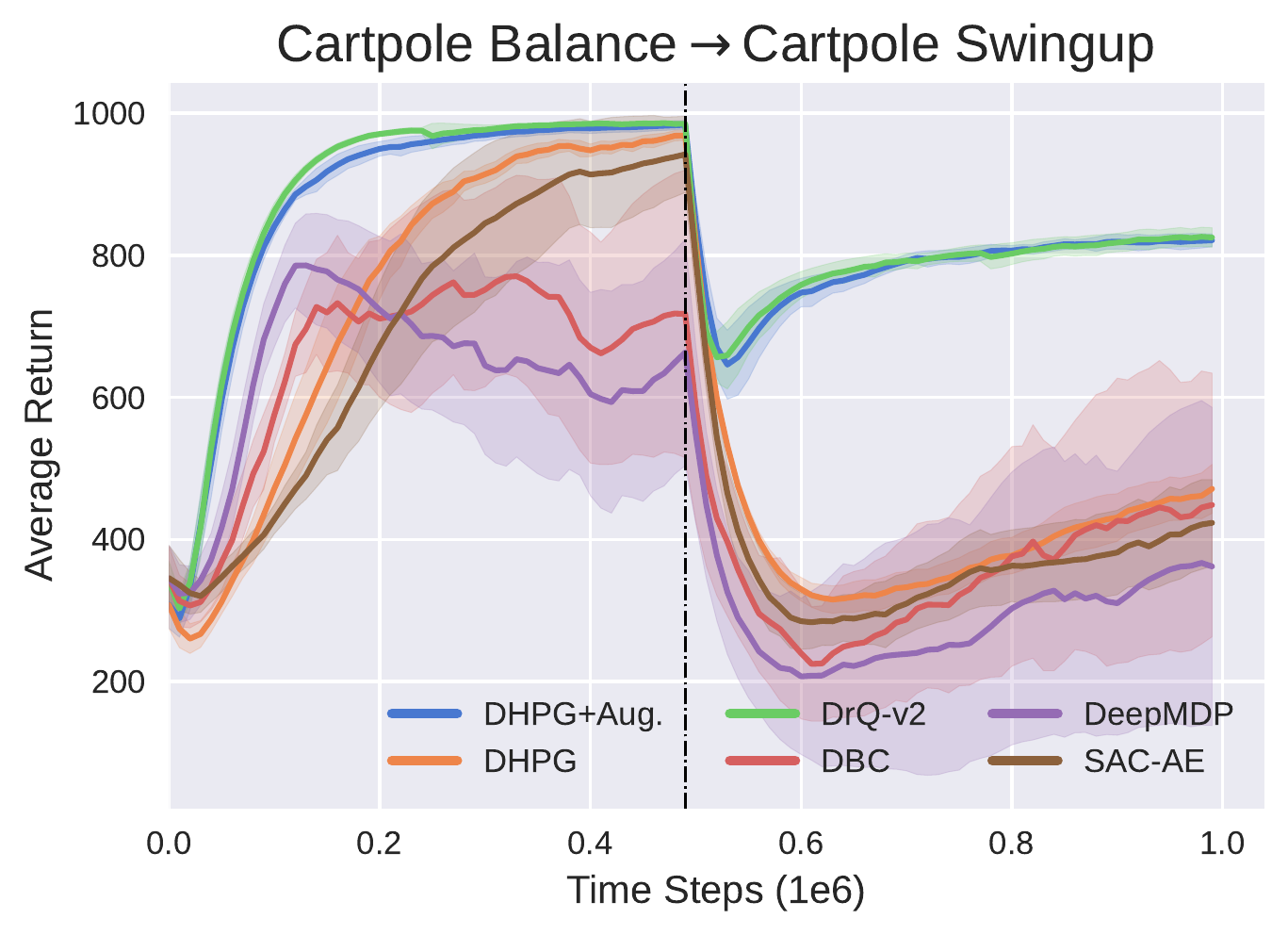}
     \end{subfigure}
     \hfill
     \begin{subfigure}[b]{0.32\textwidth}
         \centering
         \includegraphics[width=\textwidth]{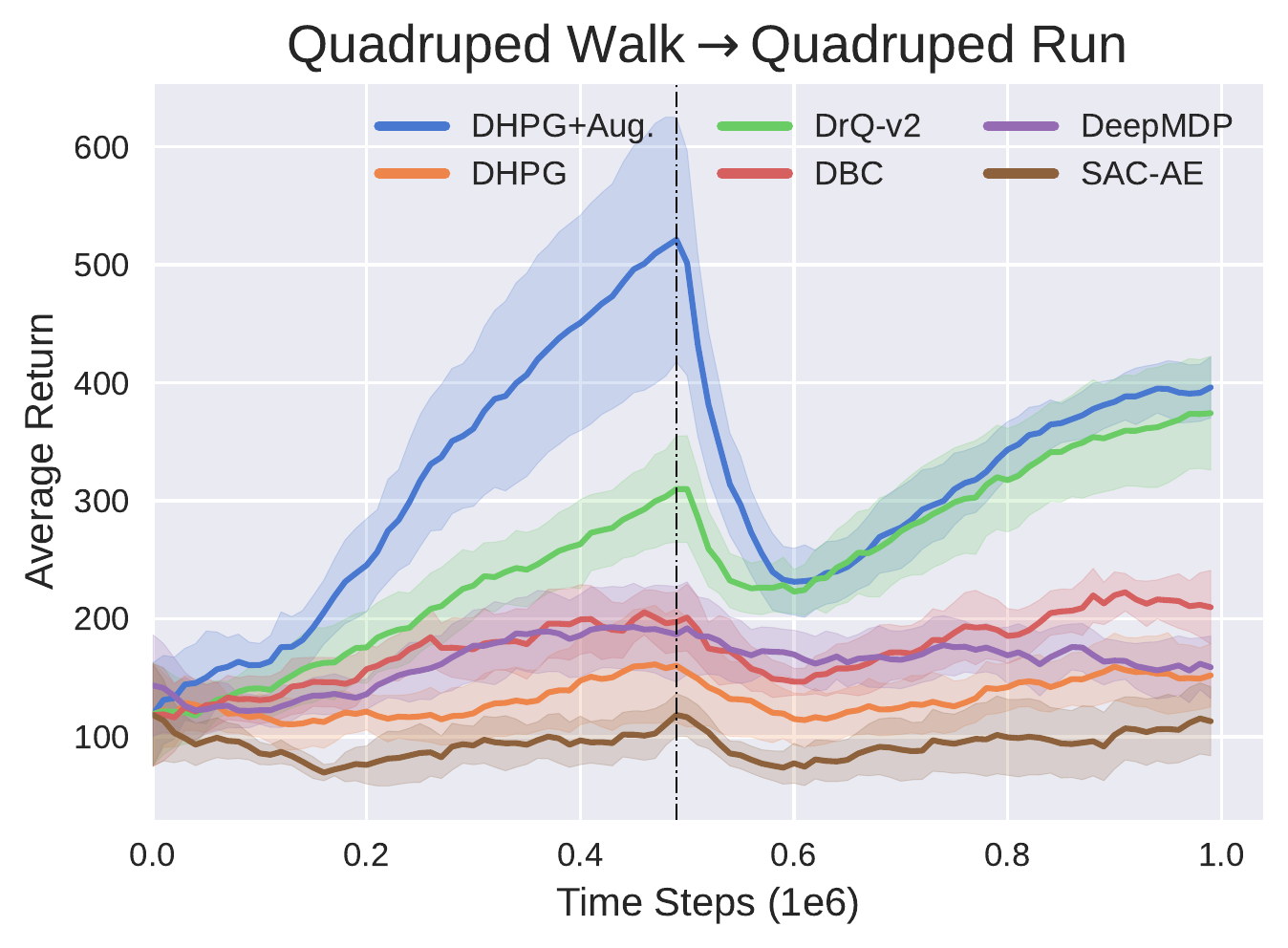}
     \end{subfigure}
     \hfill
     \begin{subfigure}[b]{0.32\textwidth}
         \centering
         \includegraphics[width=\textwidth]{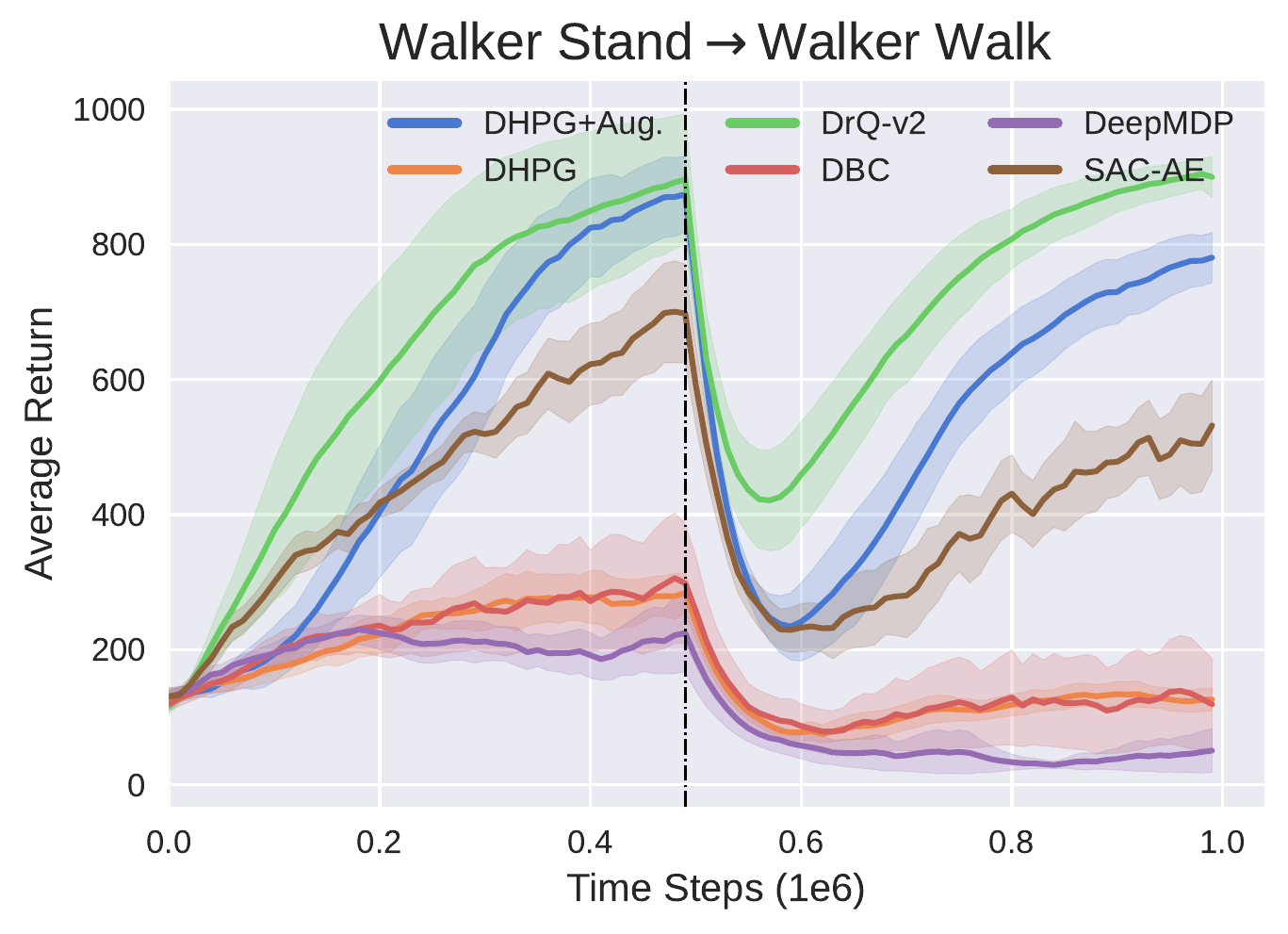}
     \end{subfigure}
     \hfill
    \caption{Learning curves for transfer experiments with \textbf{pixel observations}. At 500k time step mark, all components
are transferred to a new task on the same domain. Mean performance is obtained over 10 seeds and shaded regions represent $95\%$ confidence intervals. Plots are smoothed uniformly for visual clarity.}
    \label{fig:pixels_transfer_supp}
\end{figure}

As discussed in Section \ref{sec:results_pixels}, the purpose of transfer experiments is to ensure that using MDP homomorphisms does not compromise transfer abilities. Figure \ref{fig:pixels_transfer_supp} shows learning curves for a series of transfer scenarios in which the critic, actor, and representations are transferred to a new task within the same domain. DHPG matches the same transfer abilities of other methods. 

\subsection{Value Equivalence Property in Practice}
\label{sec:value_equiv_supp}
\begin{figure}[h!]
    \centering
    \includegraphics[width=0.7\textwidth]{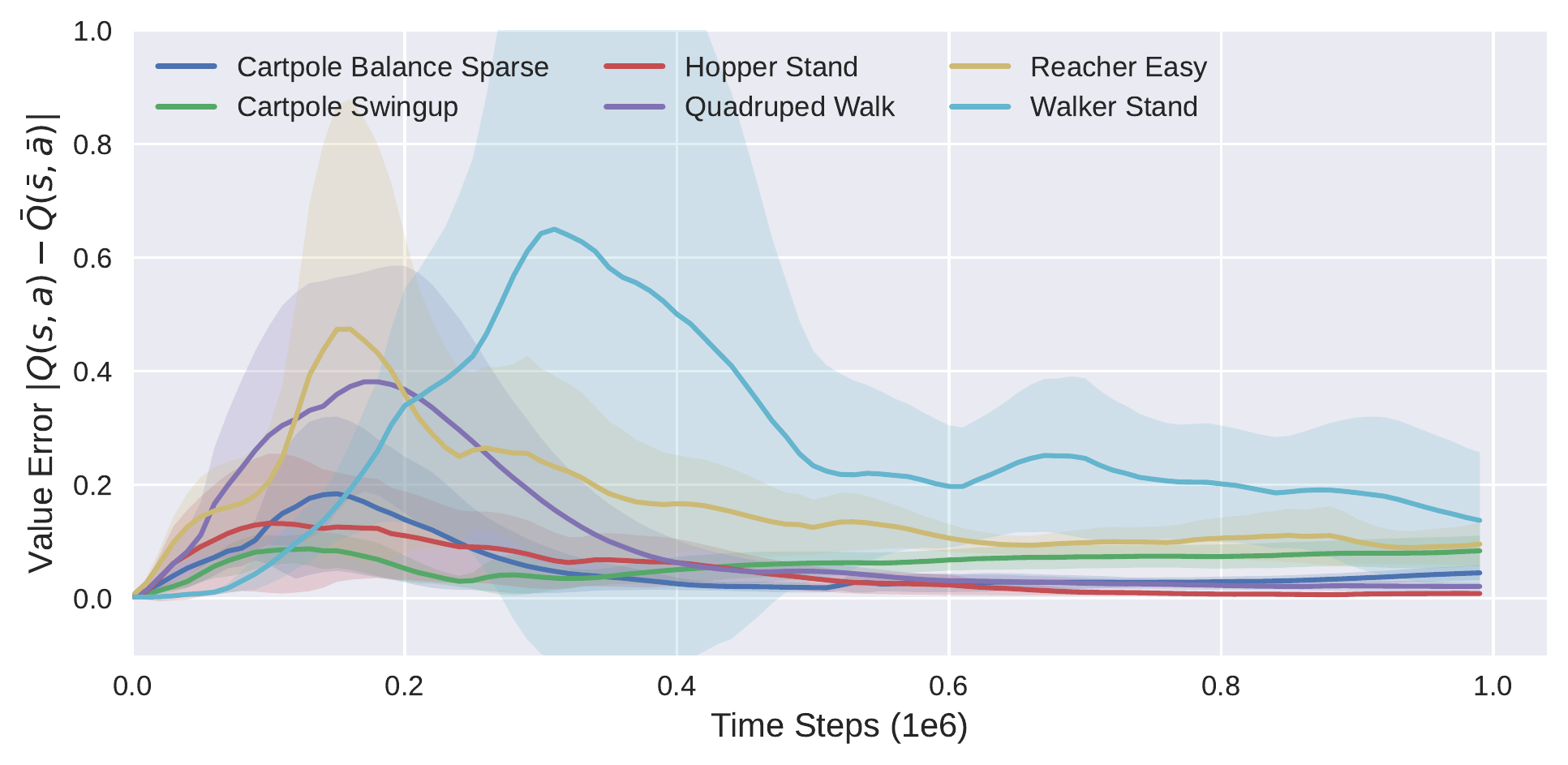}
    \caption{Normalized mean absolute error $|Q(s, a) - \overline{Q}(\overline{s}, \overline{a})|$ as a measure for the value equivalence property during training of different tasks from \textbf{pixel observations}. The error is measured on samples from the replay buffer and is normalzied by the range of the value function. The error is averaged over 10 seeds and shaded regions represent $95\%$ confidence intervals.}
    \label{fig:value_equivalence}
\end{figure}

We can use the value equivalence between the critics of the actual and abstract MDPs as a measure for the quality of learned MDP homomorphismsm, since the two critics are not directly trained to minimize this distance, instead they have equivalent values through the learned MDP homomorphism map. Figure \ref{fig:value_equivalence} shows the normalized mean absolute error of $|Q(s, a) \!-\! \overline{Q}(\overline{s}, \overline{a})|$ during training, indicating the property is holding in practice. Expectedly, for lower-dimensional tasks with easily learnable homomorphism maps (e.g., cartpole) the error is reduced earlier than more complicated tasks (e.g., quadruped and walker). But importantly, in all cases the error decreases over time and is at a reasonable range towards the end of the training, meaning the continuous MDP homomorphisms is adhering to conditions of Definition \ref{def:cont_mdp_homo}.

\clearpage

\subsection{Ablation Study on the Combination of HPG with DPG}
\label{sec:ablation_dhpg_variants}
\begin{figure}[h!]
     \centering
     \begin{subfigure}[b]{0.24\textwidth}
         \centering
         \includegraphics[width=\textwidth]{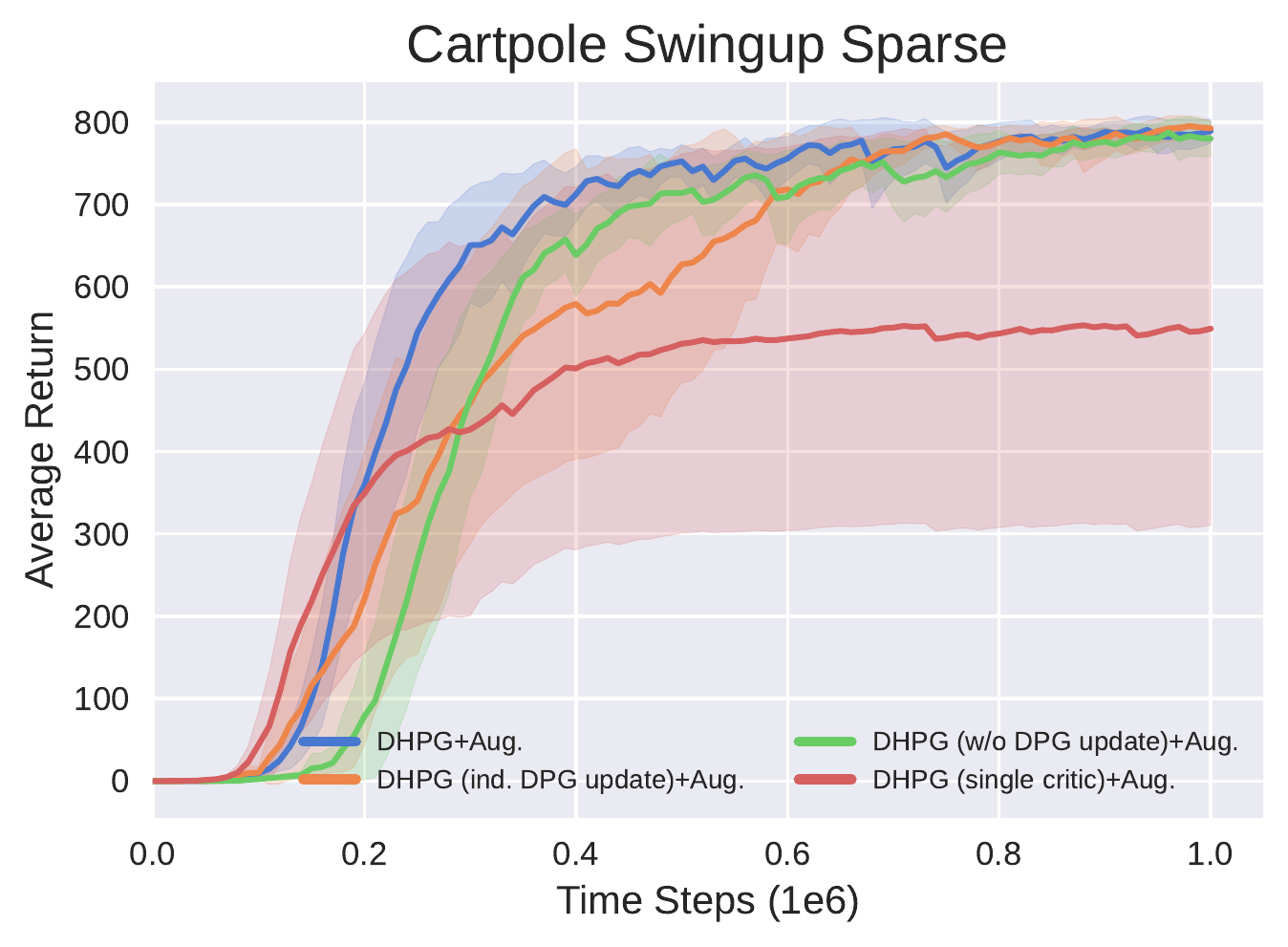}
     \end{subfigure}
     \hfill
     \begin{subfigure}[b]{0.24\textwidth}
         \centering
         \includegraphics[width=\textwidth]{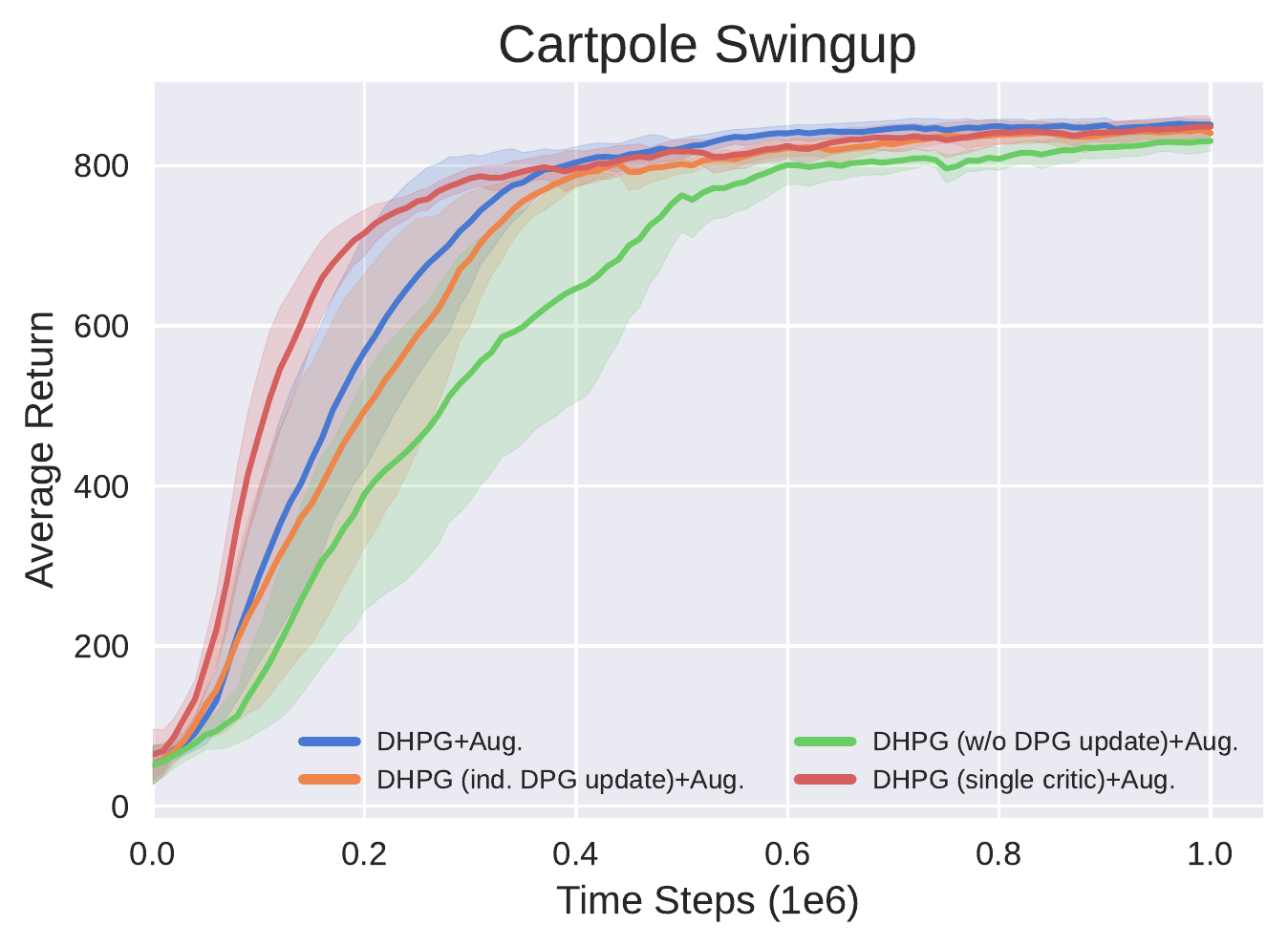}
     \end{subfigure}
     \hfill
     \begin{subfigure}[b]{0.24\textwidth}
         \centering
         \includegraphics[width=\textwidth]{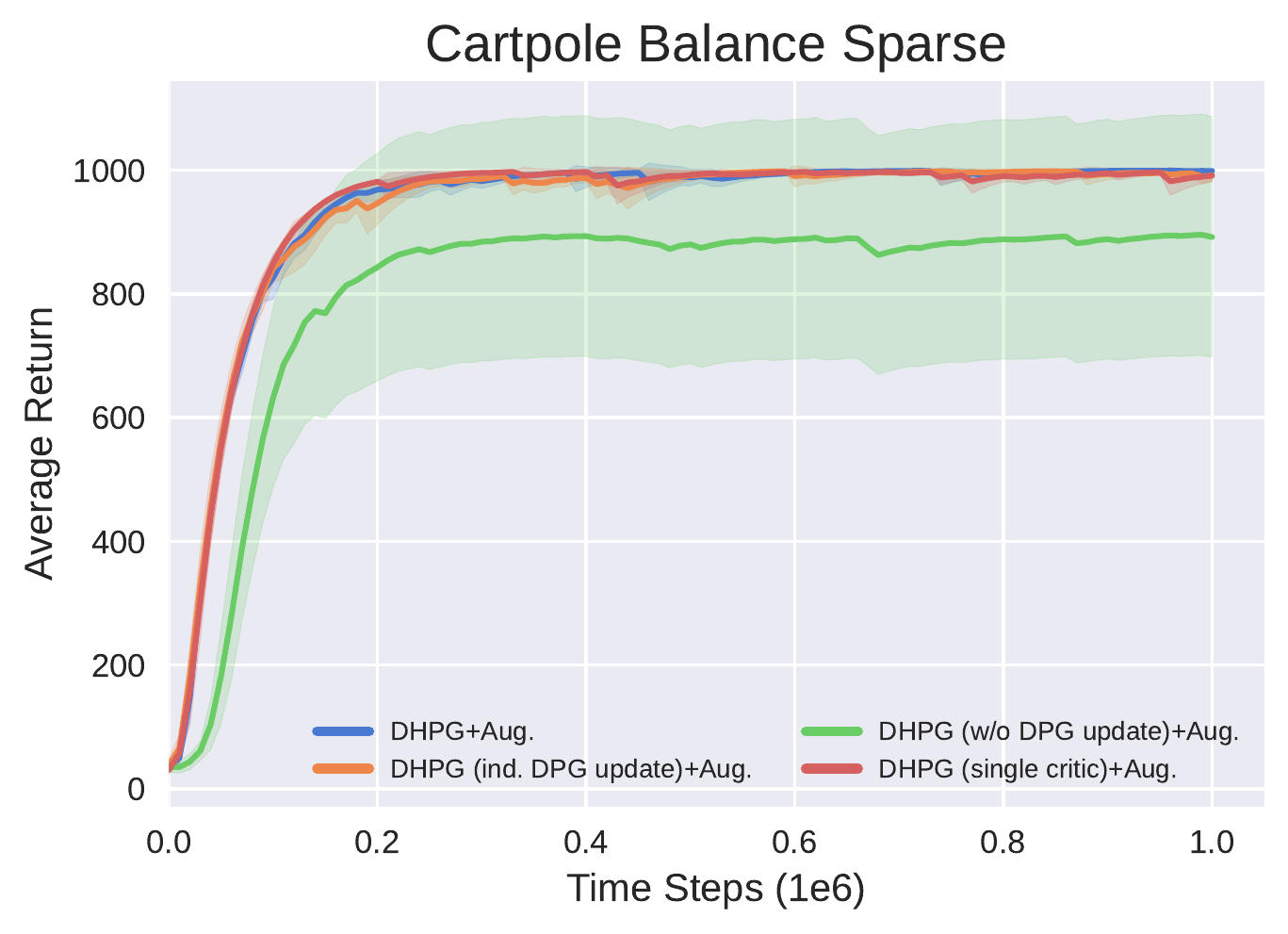}
     \end{subfigure}
     \hfill
     \begin{subfigure}[b]{0.24\textwidth}
         \centering
         \includegraphics[width=\textwidth]{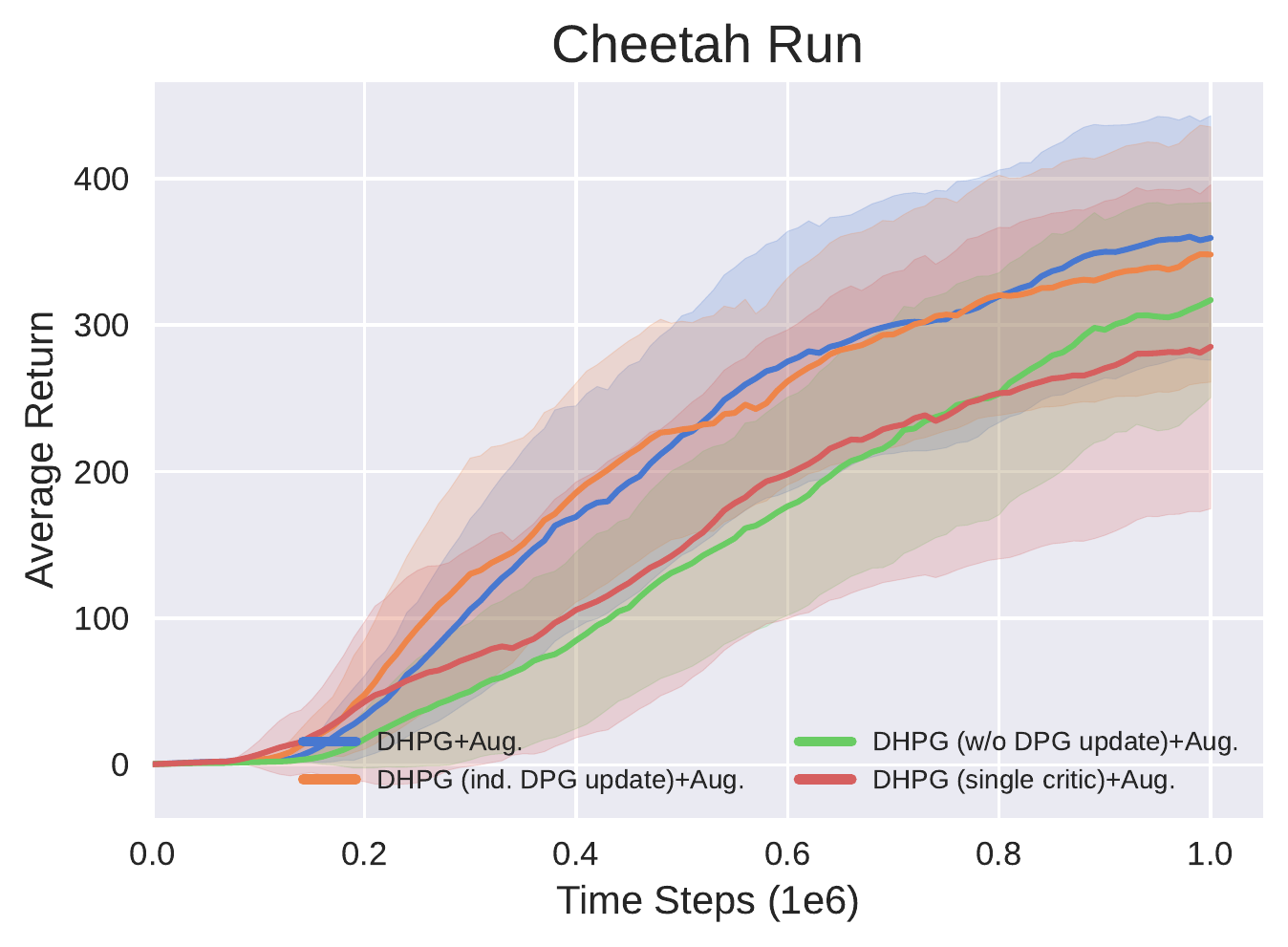}
     \end{subfigure}
     \hfill
     
     \begin{subfigure}[b]{0.24\textwidth}
         \centering
         \includegraphics[width=\textwidth]{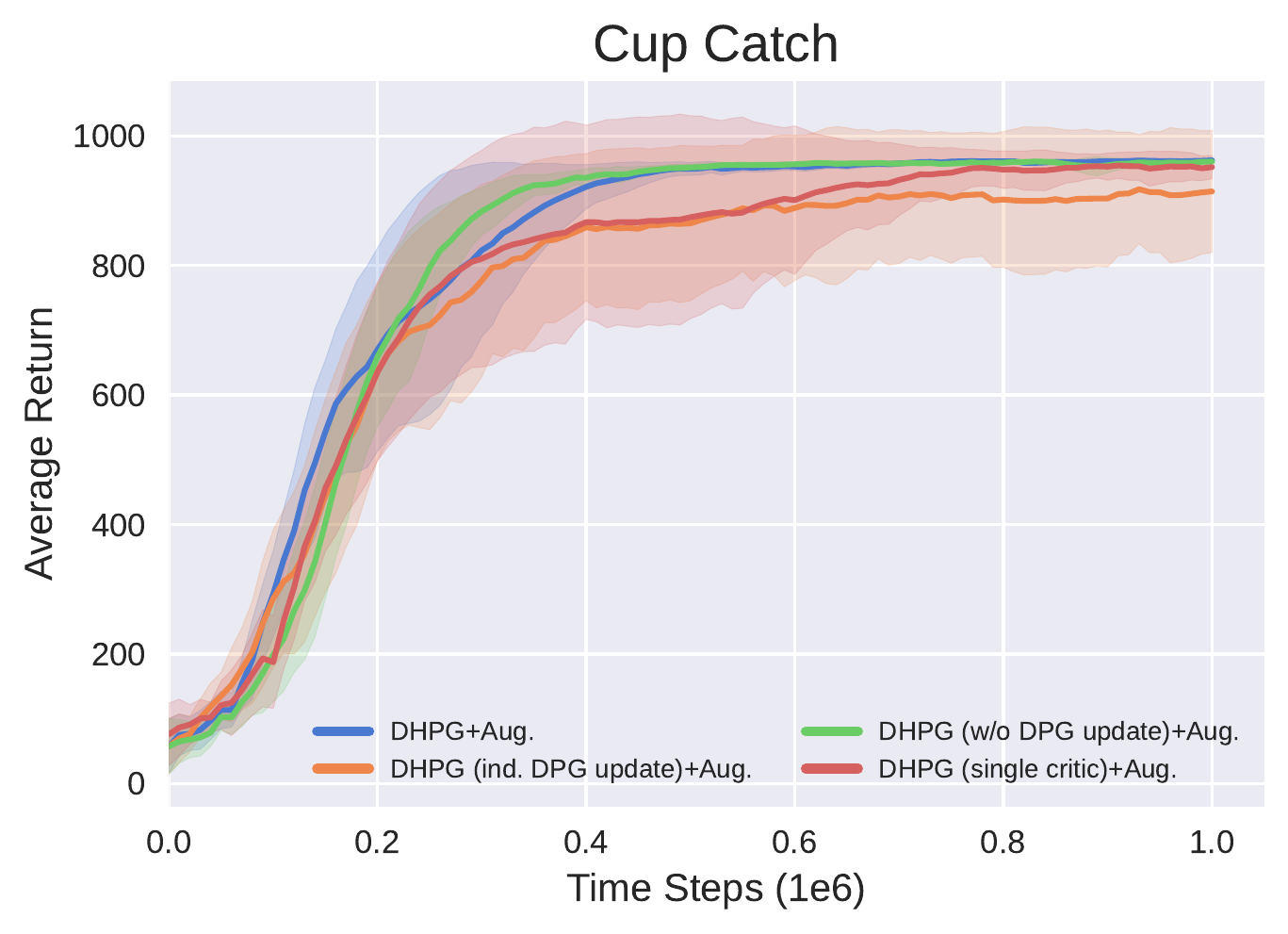}
     \end{subfigure}
     \hfill
     \begin{subfigure}[b]{0.24\textwidth}
         \centering
         \includegraphics[width=\textwidth]{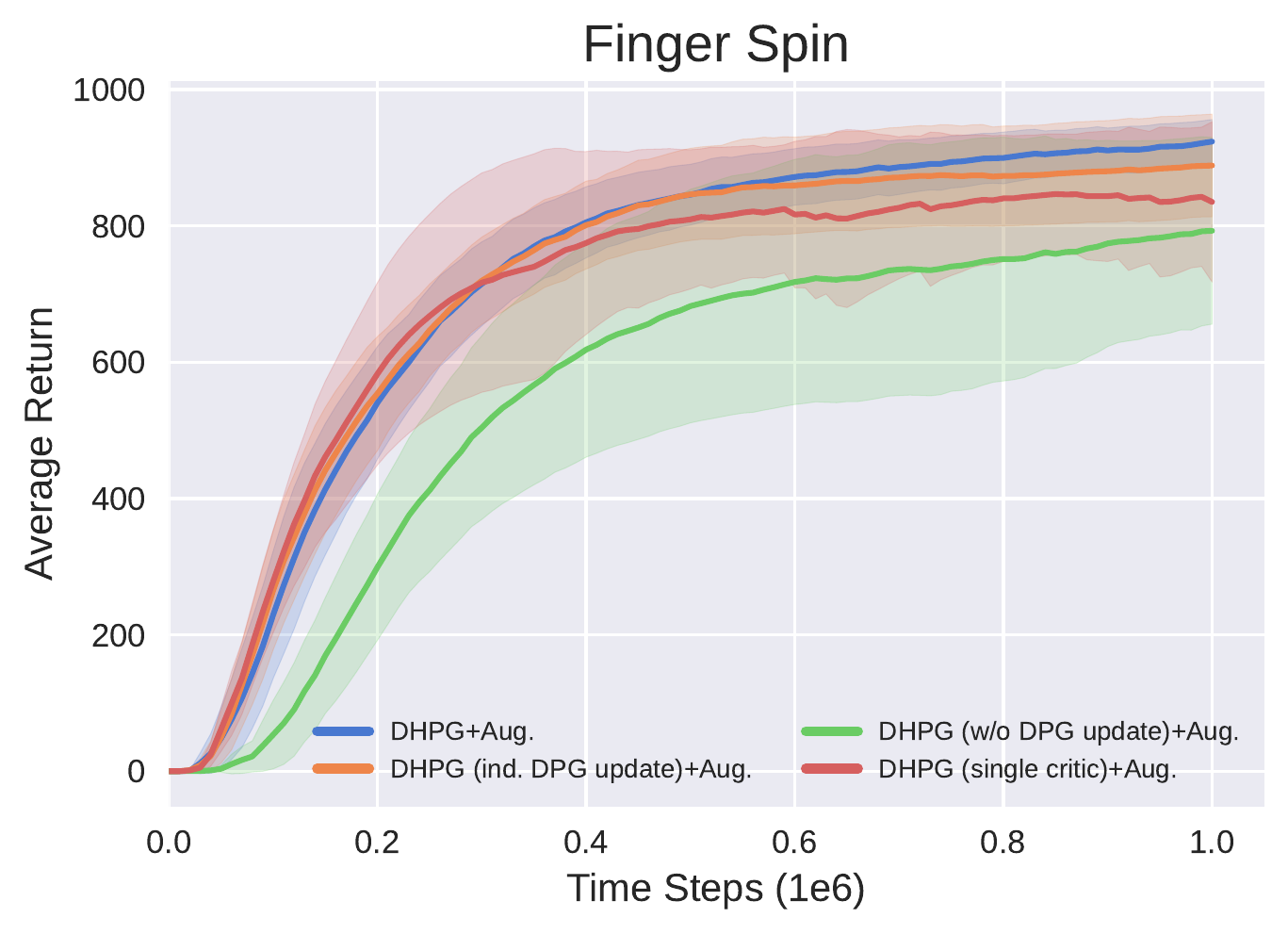}
     \end{subfigure}
     \hfill
     \begin{subfigure}[b]{0.24\textwidth}
         \centering
         \includegraphics[width=\textwidth]{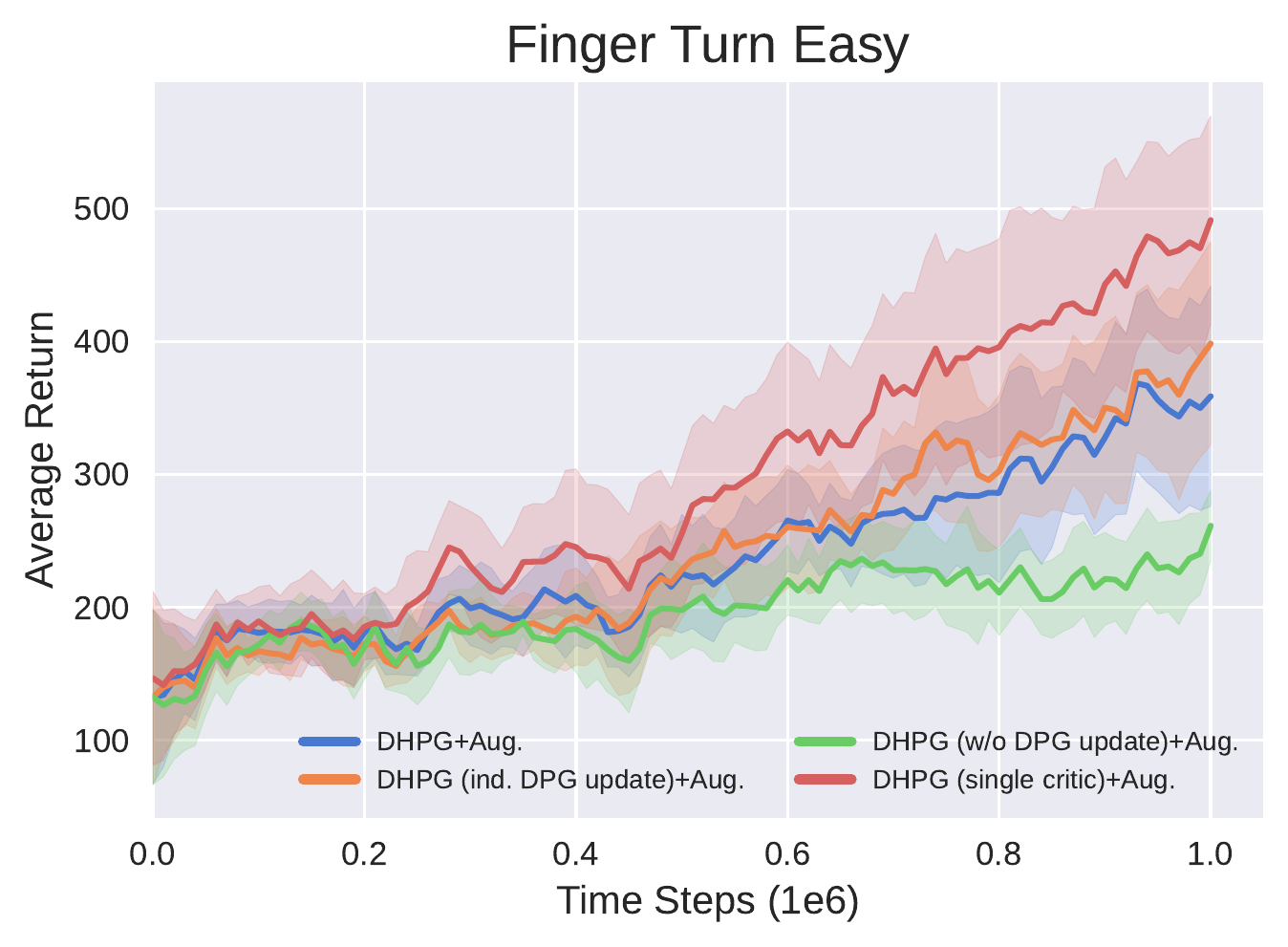}
     \end{subfigure}
     \hfill
     \begin{subfigure}[b]{0.24\textwidth}
         \centering
         \includegraphics[width=\textwidth]{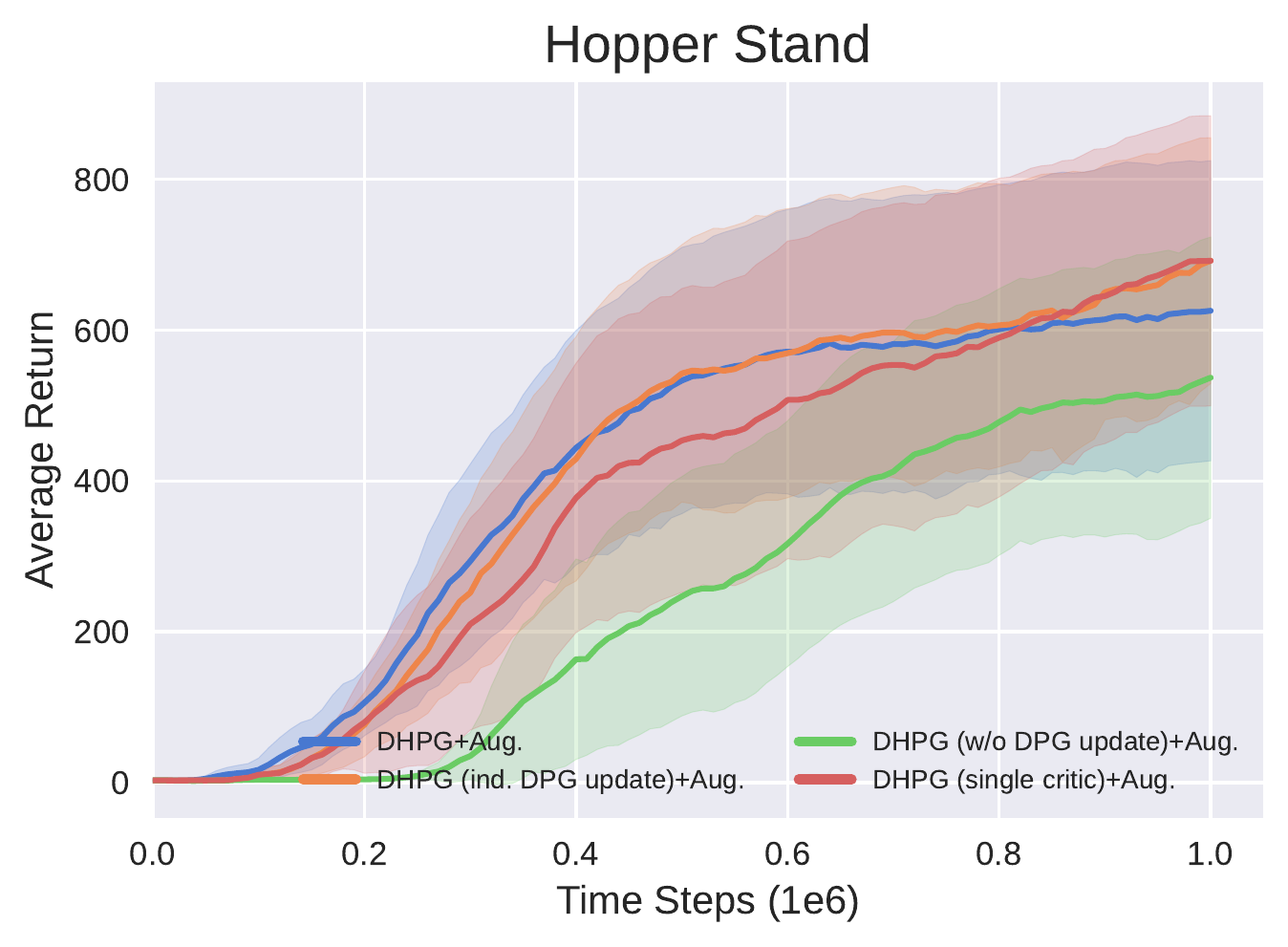}
     \end{subfigure}
     \hfill
     
     \begin{subfigure}[b]{0.24\textwidth}
         \centering
         \includegraphics[width=\textwidth]{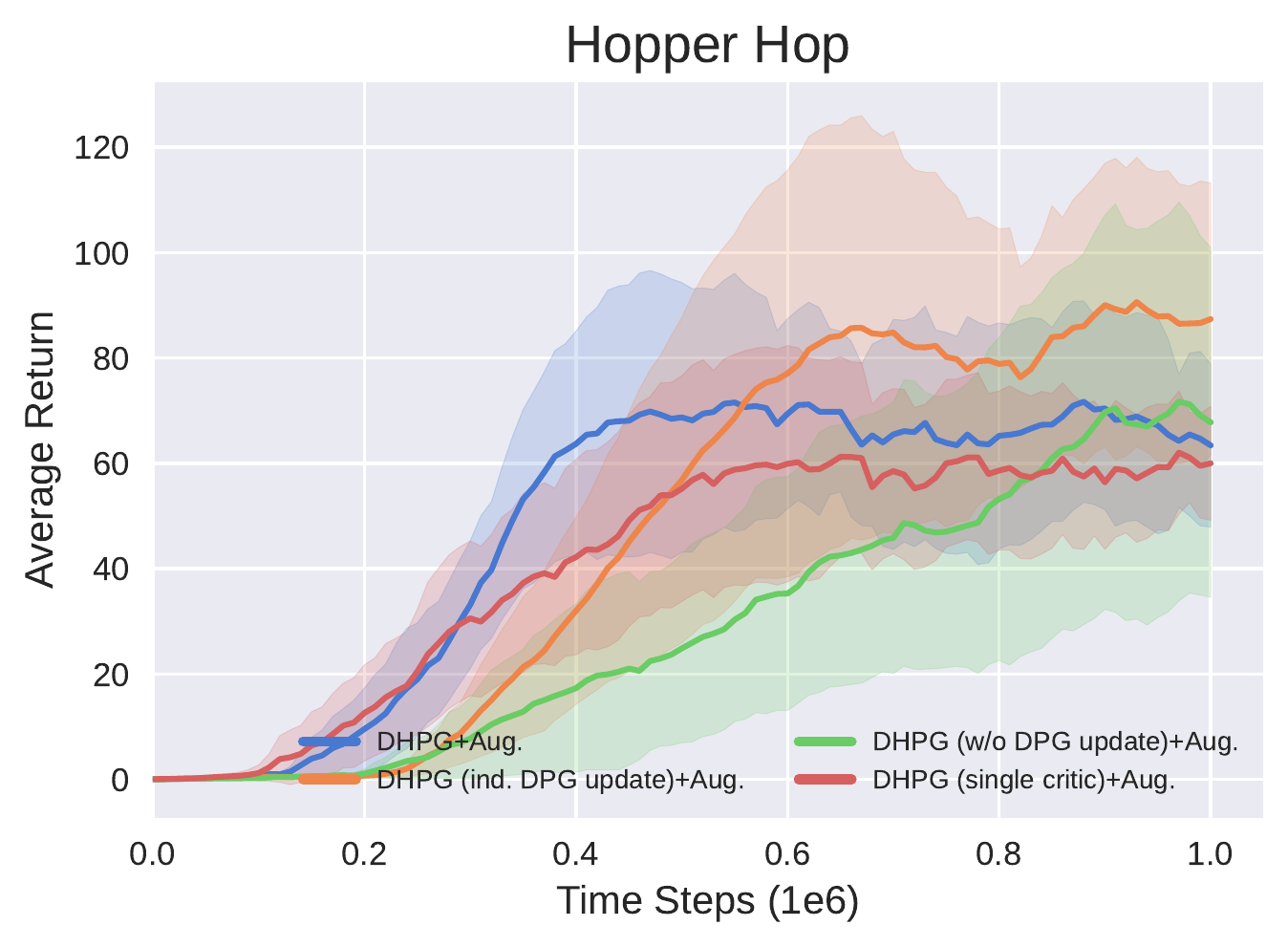}
     \end{subfigure}
     \hfill
     \begin{subfigure}[b]{0.24\textwidth}
         \centering
         \includegraphics[width=\textwidth]{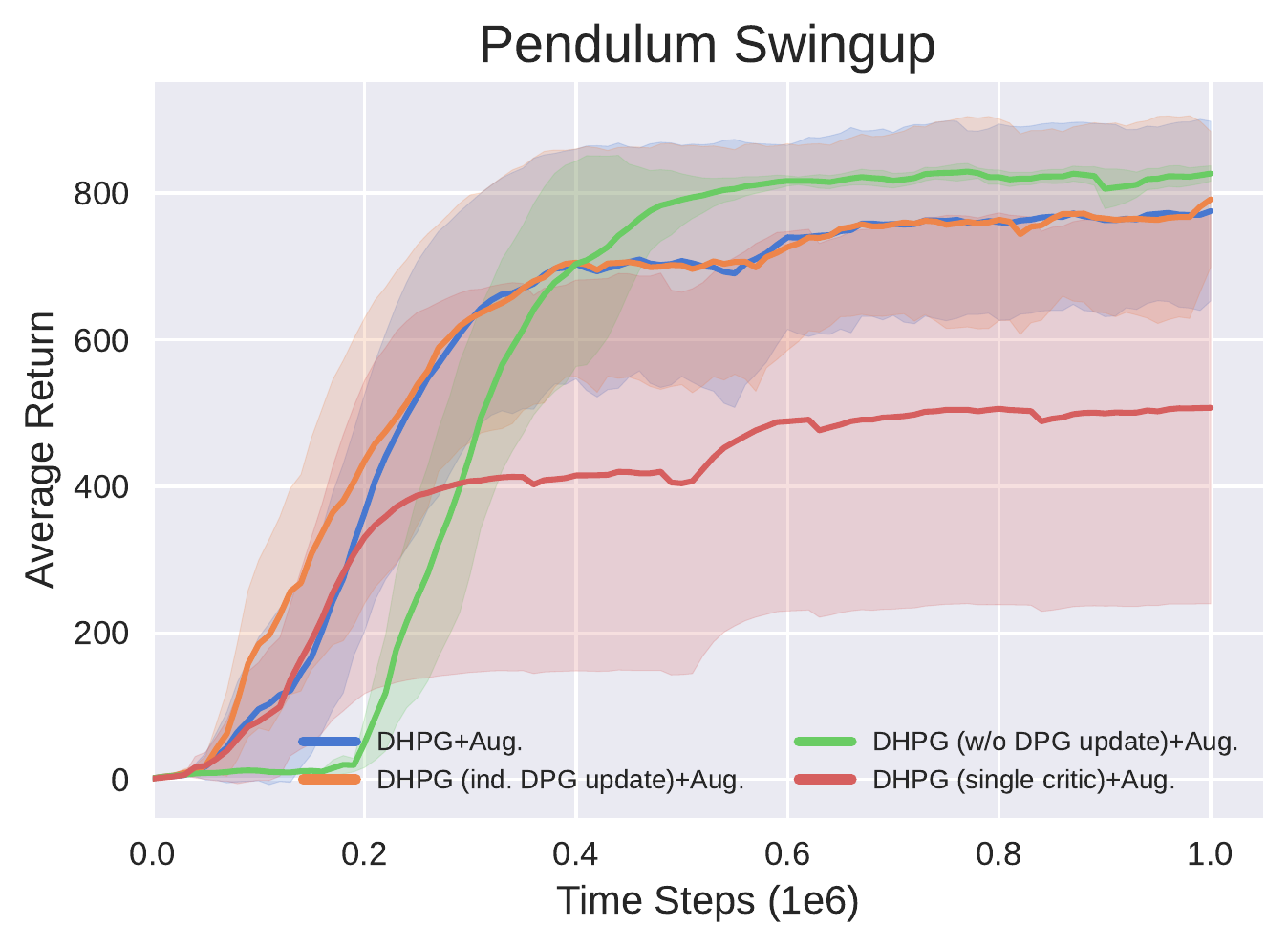}
     \end{subfigure}
     \hfill
     \begin{subfigure}[b]{0.24\textwidth}
         \centering
         \includegraphics[width=\textwidth]{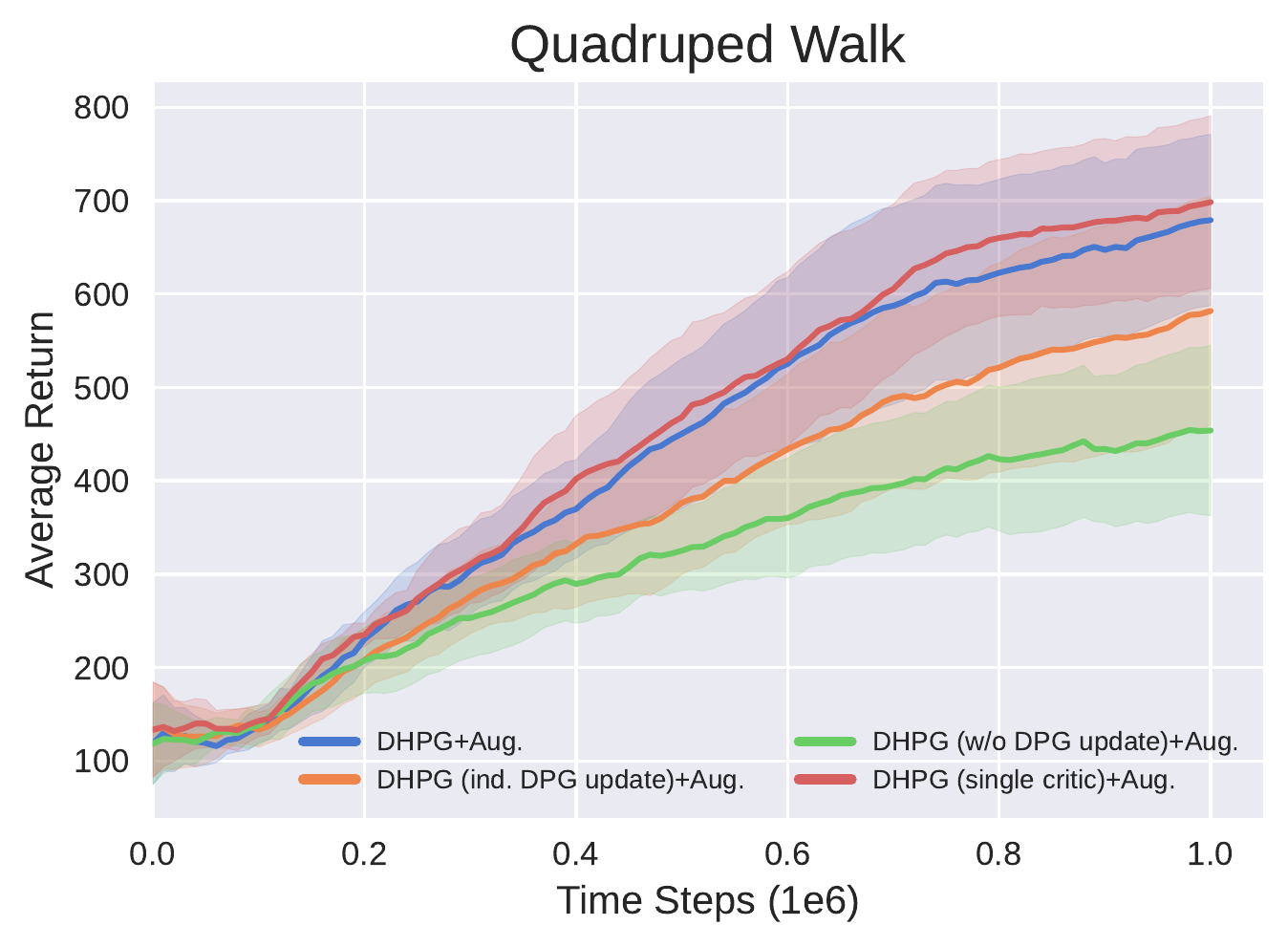}
     \end{subfigure}
     \hfill
     \begin{subfigure}[b]{0.24\textwidth}
         \centering
         \includegraphics[width=\textwidth]{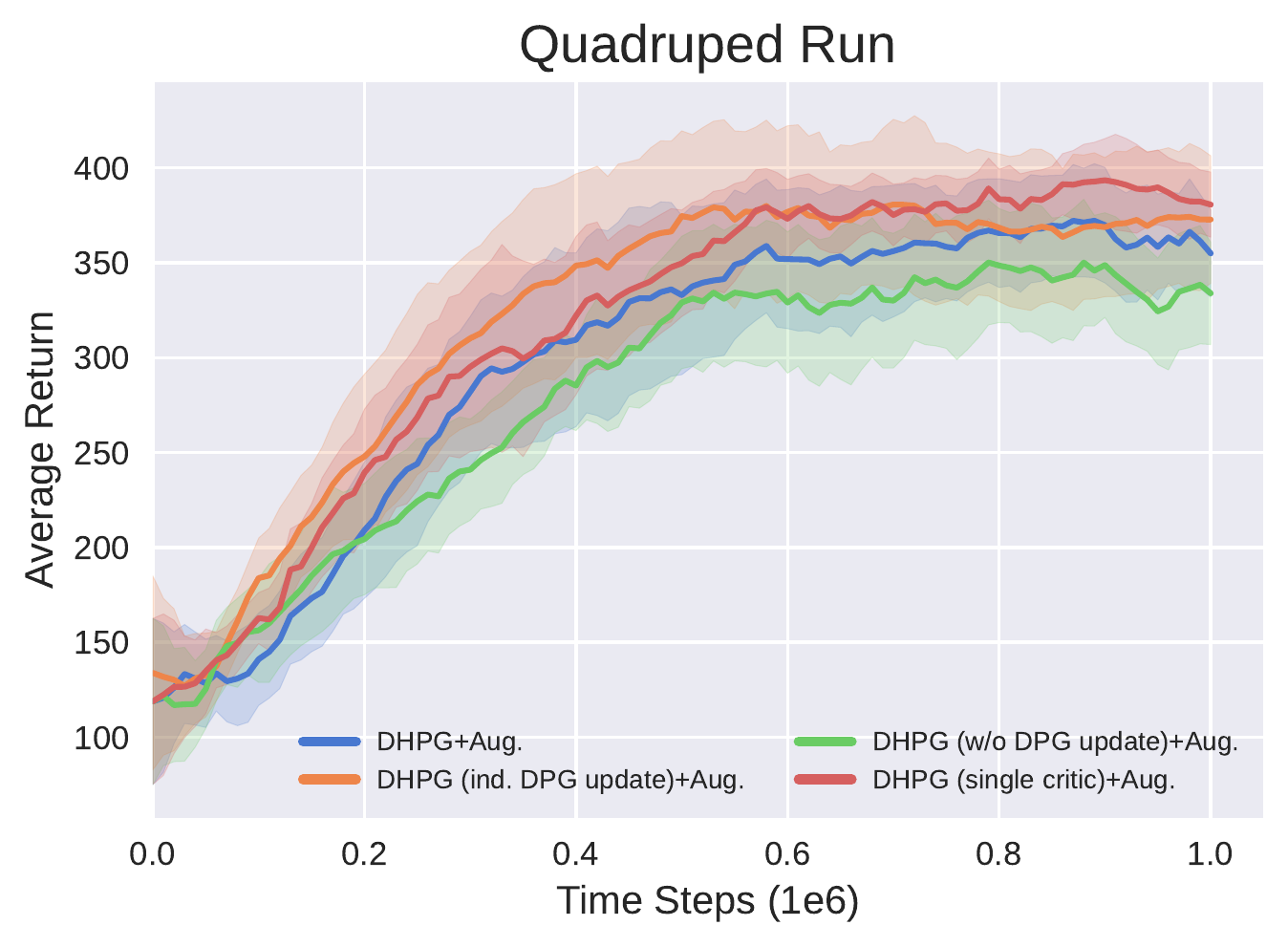}
     \end{subfigure}
     \hfill
     
     \begin{subfigure}[b]{0.24\textwidth}
         \centering
         \includegraphics[width=\textwidth]{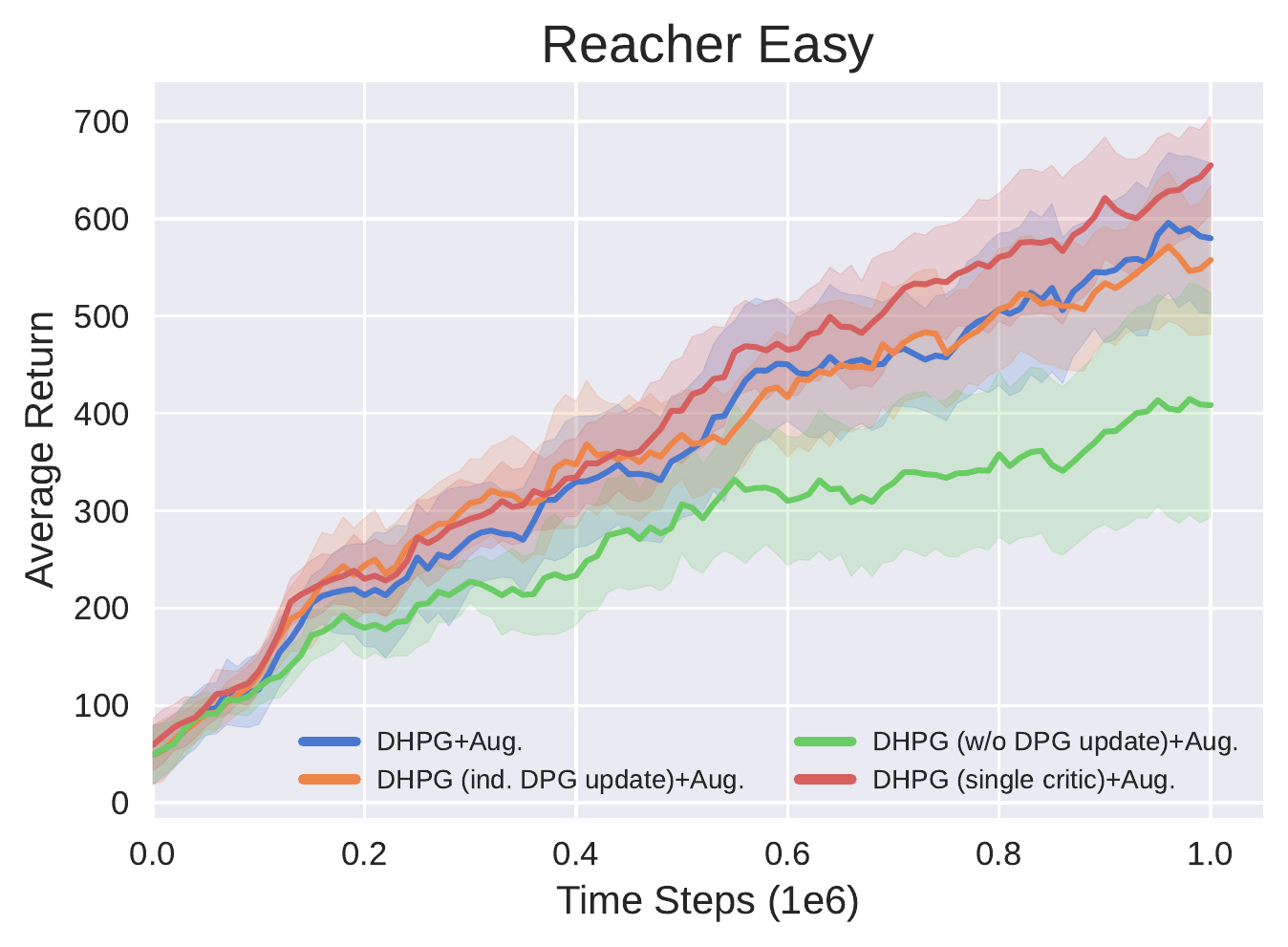}
     \end{subfigure}
     \hfill
     \begin{subfigure}[b]{0.24\textwidth}
         \centering
         \includegraphics[width=\textwidth]{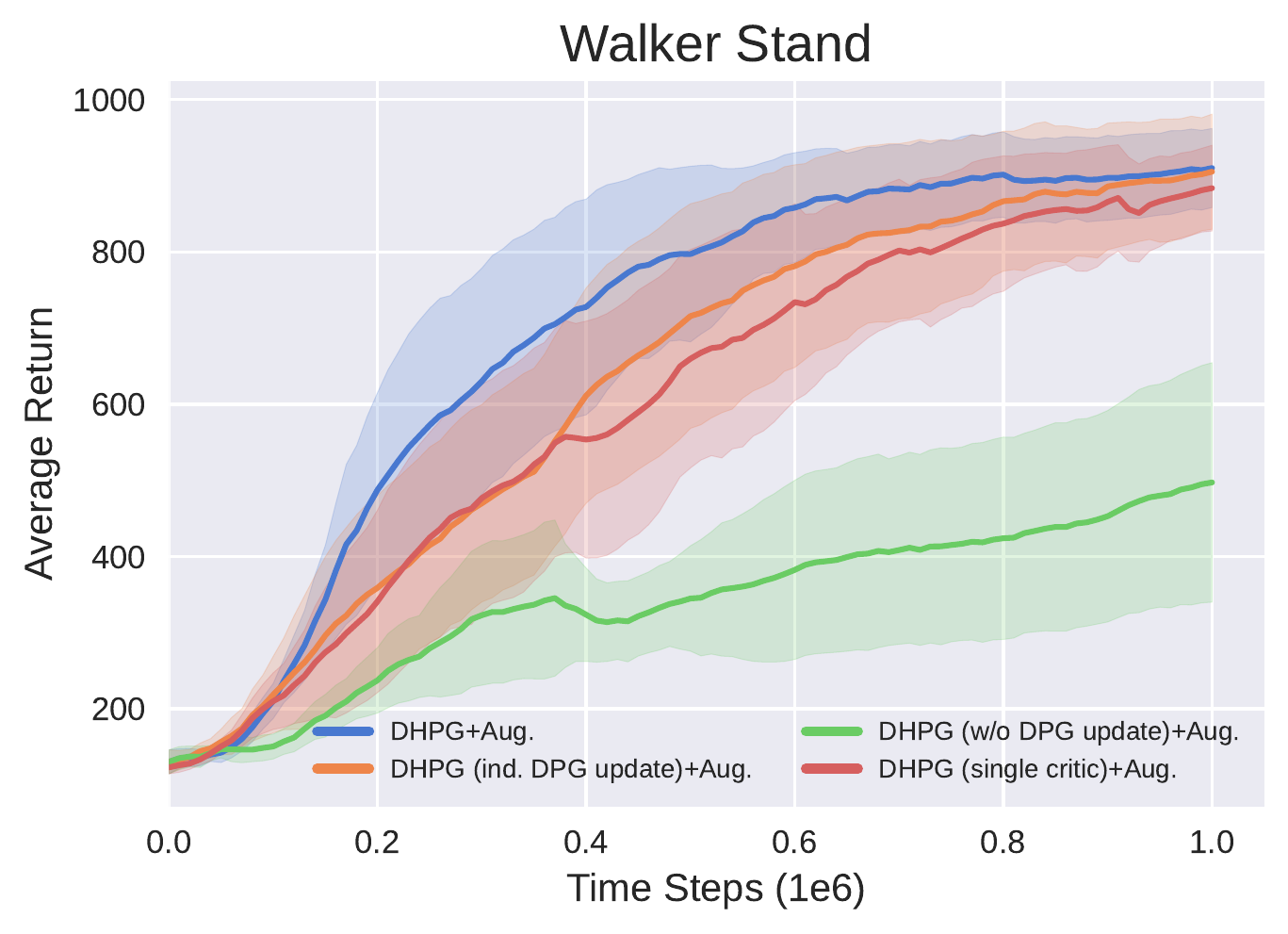}
     \end{subfigure}
     \begin{subfigure}[b]{0.24\textwidth}
         \centering
         \includegraphics[width=\textwidth]{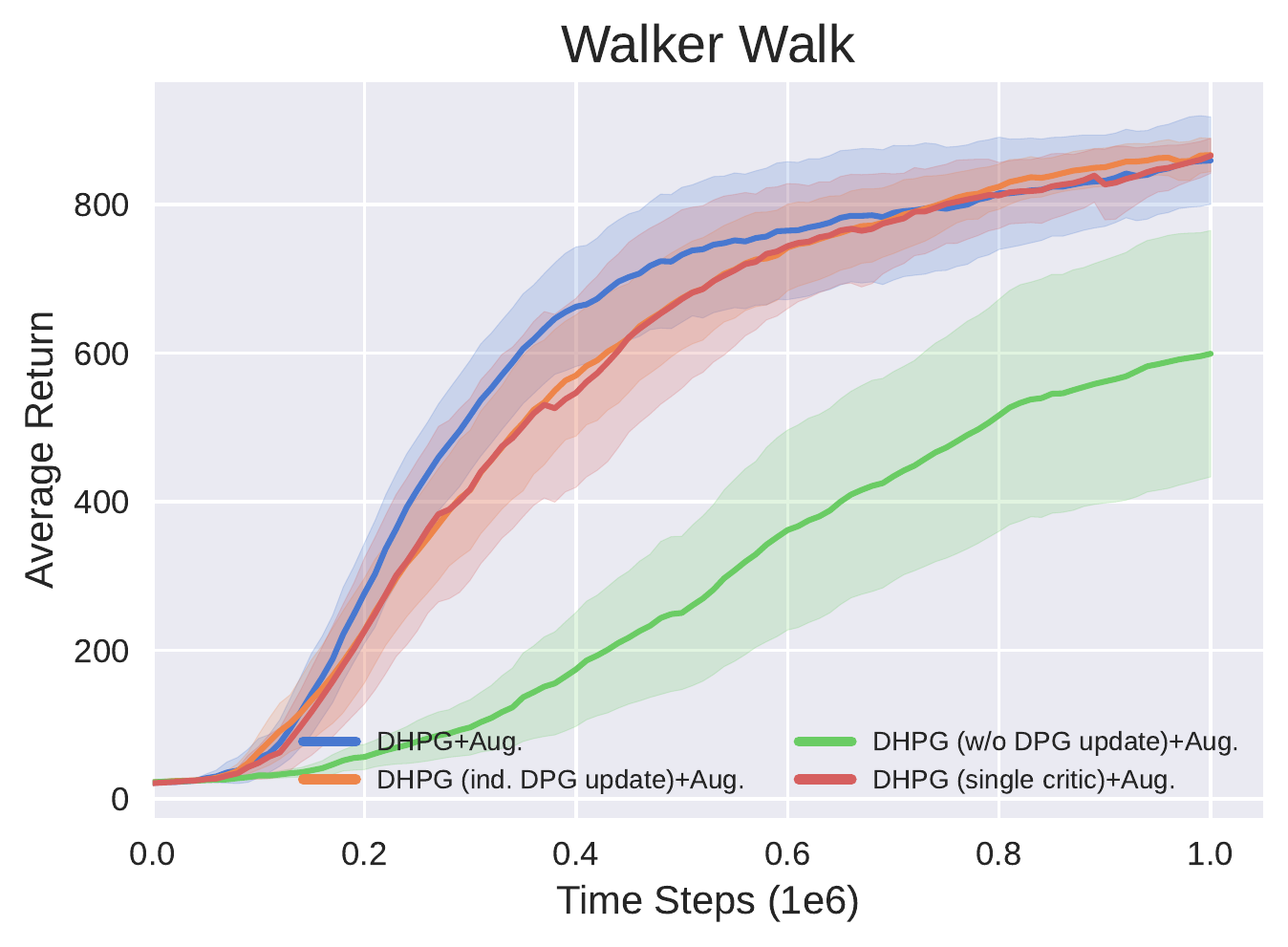}
     \end{subfigure}
     \hfill
     \begin{subfigure}[b]{0.24\textwidth}
         \centering
         \includegraphics[width=\textwidth]{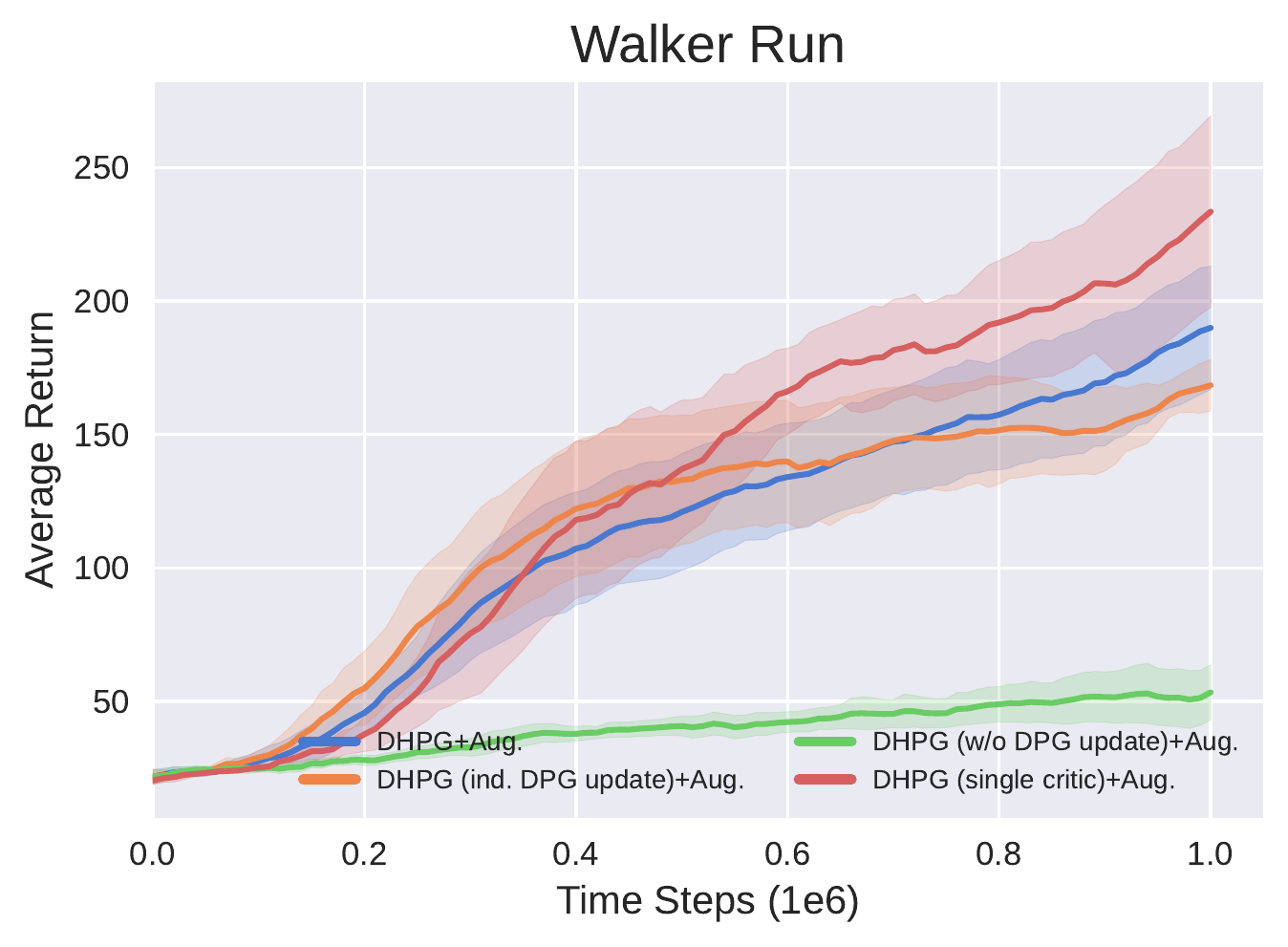}
     \end{subfigure}
     \hfill
    \caption{Ablation study on the combination of HPG and DPG. Learning curves for 16 DM control tasks with \textbf{pixel observations}. Mean performance is obtained over 10 seeds and shaded regions represent $95\%$ confidence intervals. Plots are smoothed uniformly for visual clarity.}
    \label{fig:ablation_dhpg_variants}
\end{figure}

\begin{figure}[h!]
    \centering
    \begin{subfigure}[b]{0.9\textwidth}
         \centering
         \includegraphics[width=\textwidth]{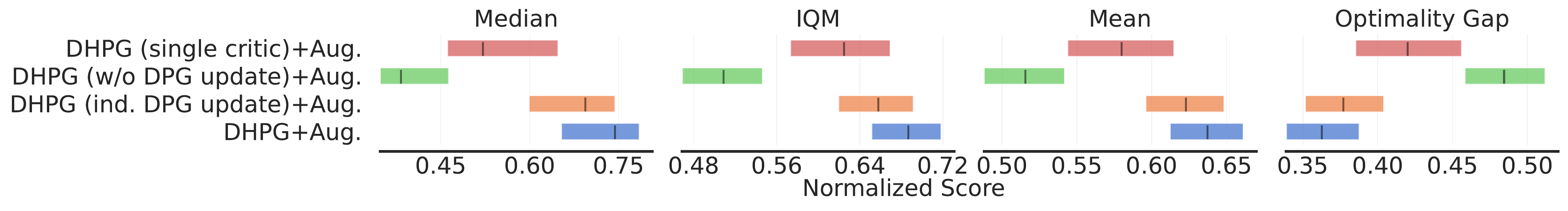}
         \caption{Aggregate metrics at 500k steps.}
    \end{subfigure}
    
    \begin{subfigure}[b]{0.32\textwidth}
         \centering
         \includegraphics[width=\textwidth]{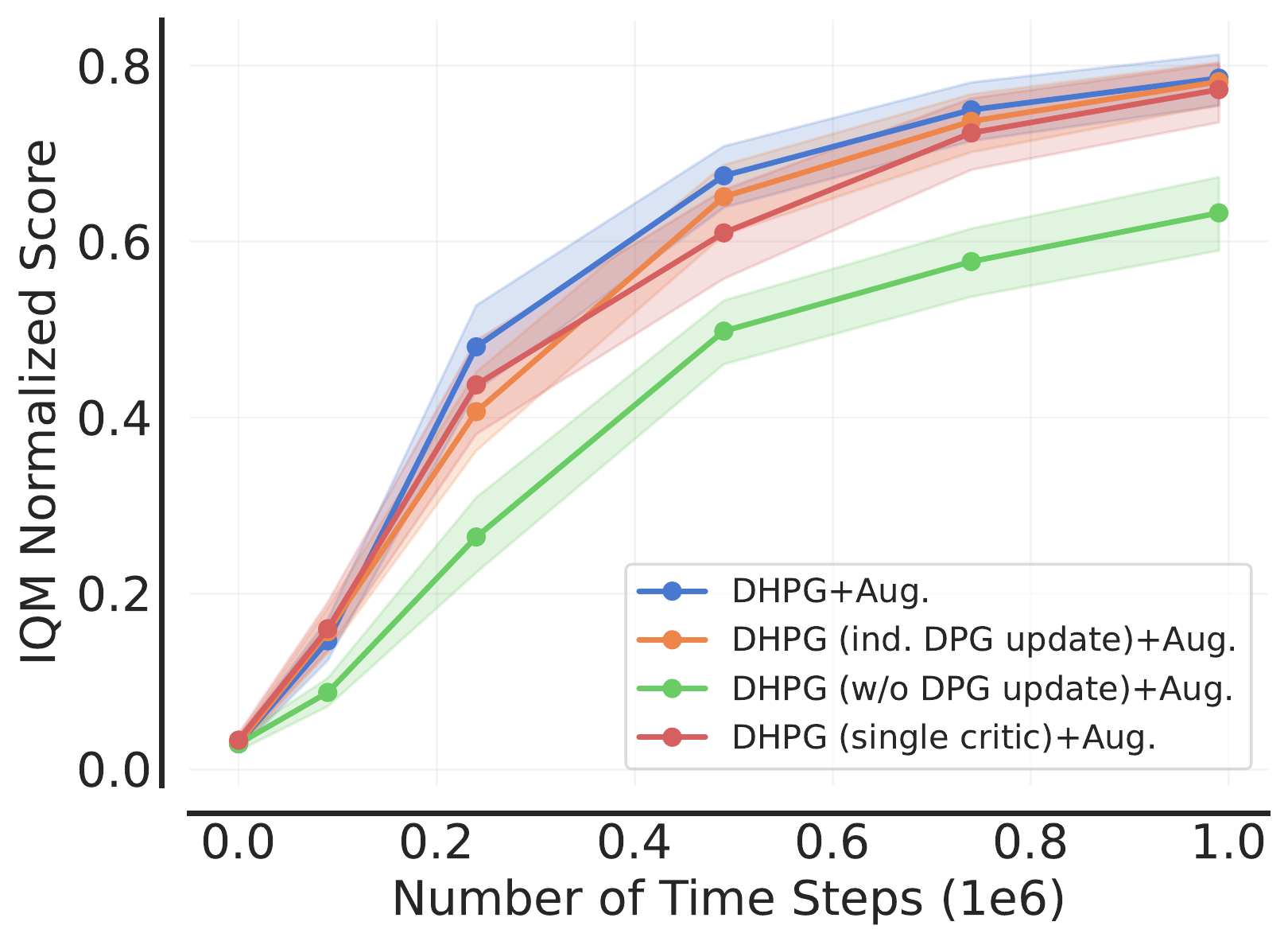}
         \caption{Sample efficiency.}
    \end{subfigure}
    \hfill
    \begin{subfigure}[b]{0.32\textwidth}
         \centering
         \includegraphics[width=\textwidth]{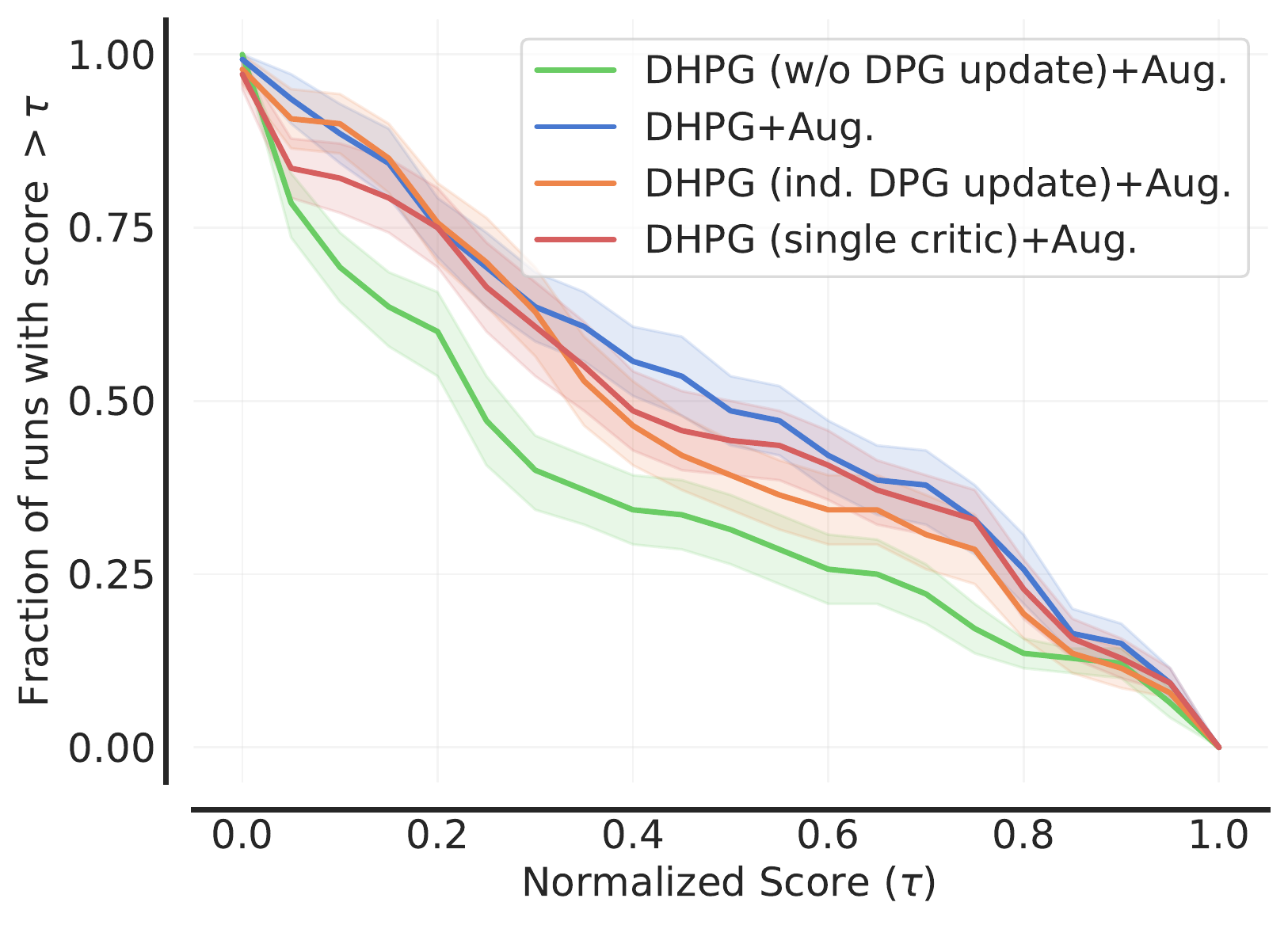}
         \caption{Performance profiles at 250k steps.}
    \end{subfigure}
    \hfill
    \begin{subfigure}[b]{0.32\textwidth}
         \centering
         \includegraphics[width=\textwidth]{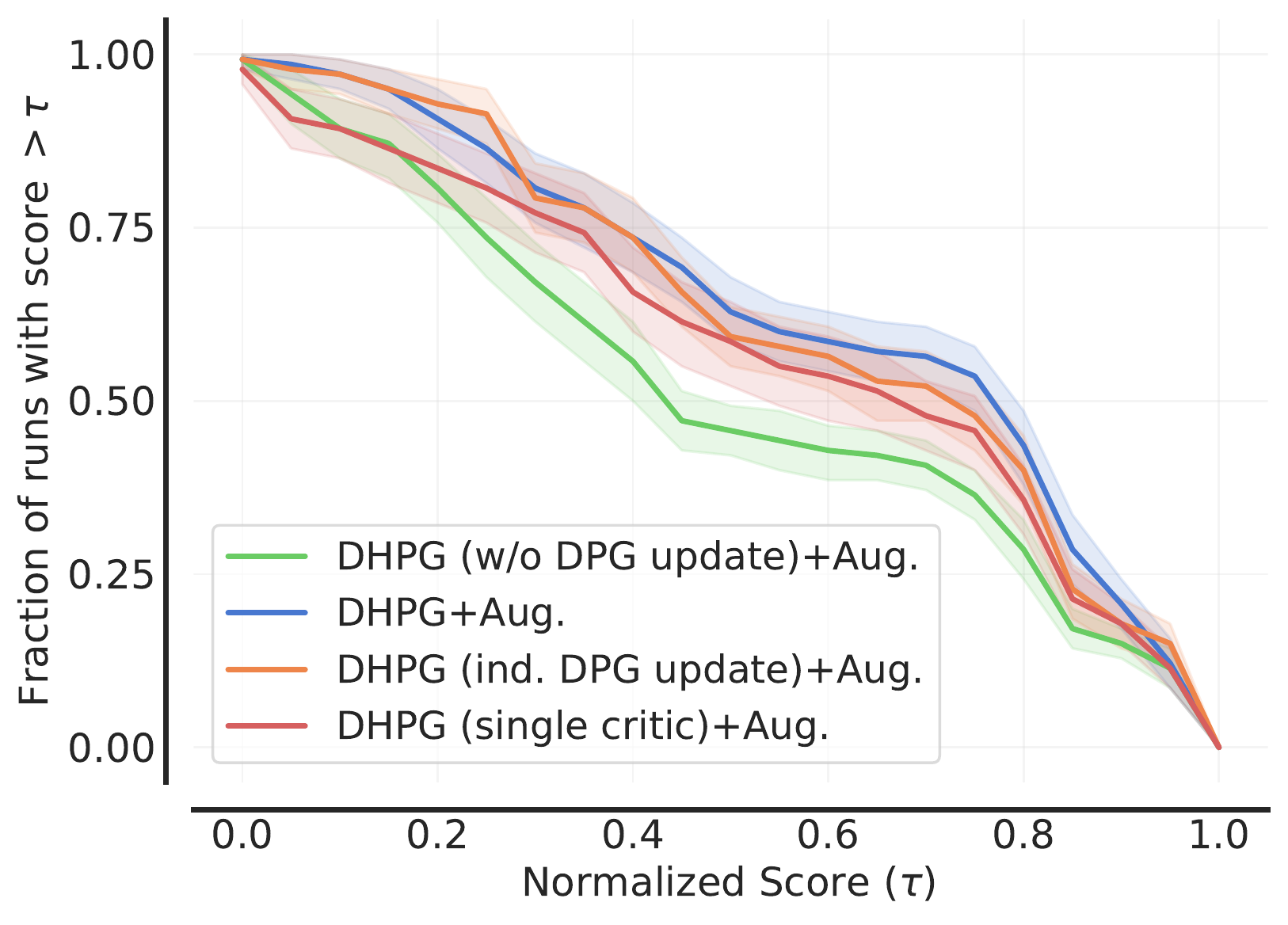}
         \caption{Performance profiles at 500k steps.}
    \end{subfigure}
    \caption{Ablation study on the combination of HPG and DPG. RLiable evaluation metrics for \textbf{pixel observations} averaged on 14 tasks over 10 seeds. Aggregate metrics at 500k steps \textbf{(a)}, IQM scores as a function of number of steps for comparing sample efficiency \textbf{(b)}, performance profiles at 250k steps \textbf{(c)}, performance profiles at 500k steps \textbf{(d)}. Shaded regions represent $95\%$ confidence intervals.}
    \label{fig:ablation_dhpg_variants_rliable}
\end{figure}

We carry out an ablation study on the combination of HPG with DPG for actor updates as indicated discussed in Section \ref{sec:hac}. To that end, we evaluate the performance of four variants of DHPG (all using image augmentation) on pixel observations:
\begin{enumerate}[noitemsep,nosep]
    \item \textbf{DHPG:} Gradients of HPG and DPG are added together and a single actor update is done based on the sum of gradients. This is the standard DHPG algorithm that is used throughout the paper.
    \item \textbf{DHPG with independent DPG update:} Gradients of HPG and DPG are independently used to update the actor. 
    \item \textbf{DHPG without DPG update:} Only HPG is used to update the actor. 
    \item \textbf{DHPG with single critic:} A single critic network is trained for learning values of both the actual and abstract MDP. Consequently, HPG and DPG are used to update the actor. 
\end{enumerate}
Figure \ref{fig:ablation_dhpg_variants} shows learning curves obtained on 16 DeepMind Control Suite tasks with pixel observations, and Figure \ref{fig:ablation_dhpg_variants_rliable} shows RLiable \cite{agarwal2021deep} evaluation metrics. In general, summing the gradients of HPG and DPG (variant 1) results in lower variance of gradient estimates compared to independent HPG and DPG updates (variant 2).
Interestingly, the variant of DHPG without DPG (variant 3) performs reasonably well or even outperforms other variants in simple tasks where learning MDP homomorphisms is easy (e.g., cartpole and pendulum), indicating the effectiveness of our method in using \textbf{only} the abstract MDP to update the policy of the actual MDP. However, in the case of more complicated tasks (e.g., walker), DPG is required to additionally use the actual MDP for policy optimization. Finally, using a single critic for both the actual and abstract MDPs (variant 4) can improve sample efficiency in symmetrical MDPs, but may result in performance drops in non-symmetrical MDPs due to the large error bound between the two MDPs, $\| Q^{\pi^\uparrow}\!(s,a) \!-\! Q^{\overline{\pi}}\!(\overline{s}, \overline{a}) \|$ \cite{taylor2008bounding}.

\clearpage

\subsection{Ablation Study on n-step Return}
\begin{figure}[h!]
    \centering
    \begin{subfigure}[b]{0.9\textwidth}
         \centering
         \includegraphics[width=\textwidth]{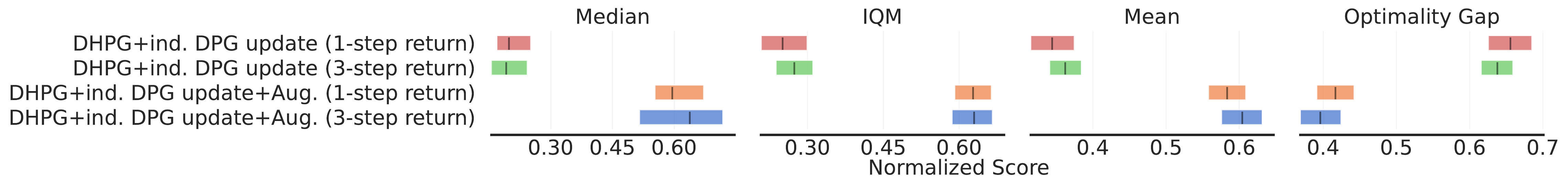}
         \caption{Aggregate metrics at 500k steps.}
    \end{subfigure}
    
    \begin{subfigure}[b]{0.32\textwidth}
         \centering
         \includegraphics[width=\textwidth]{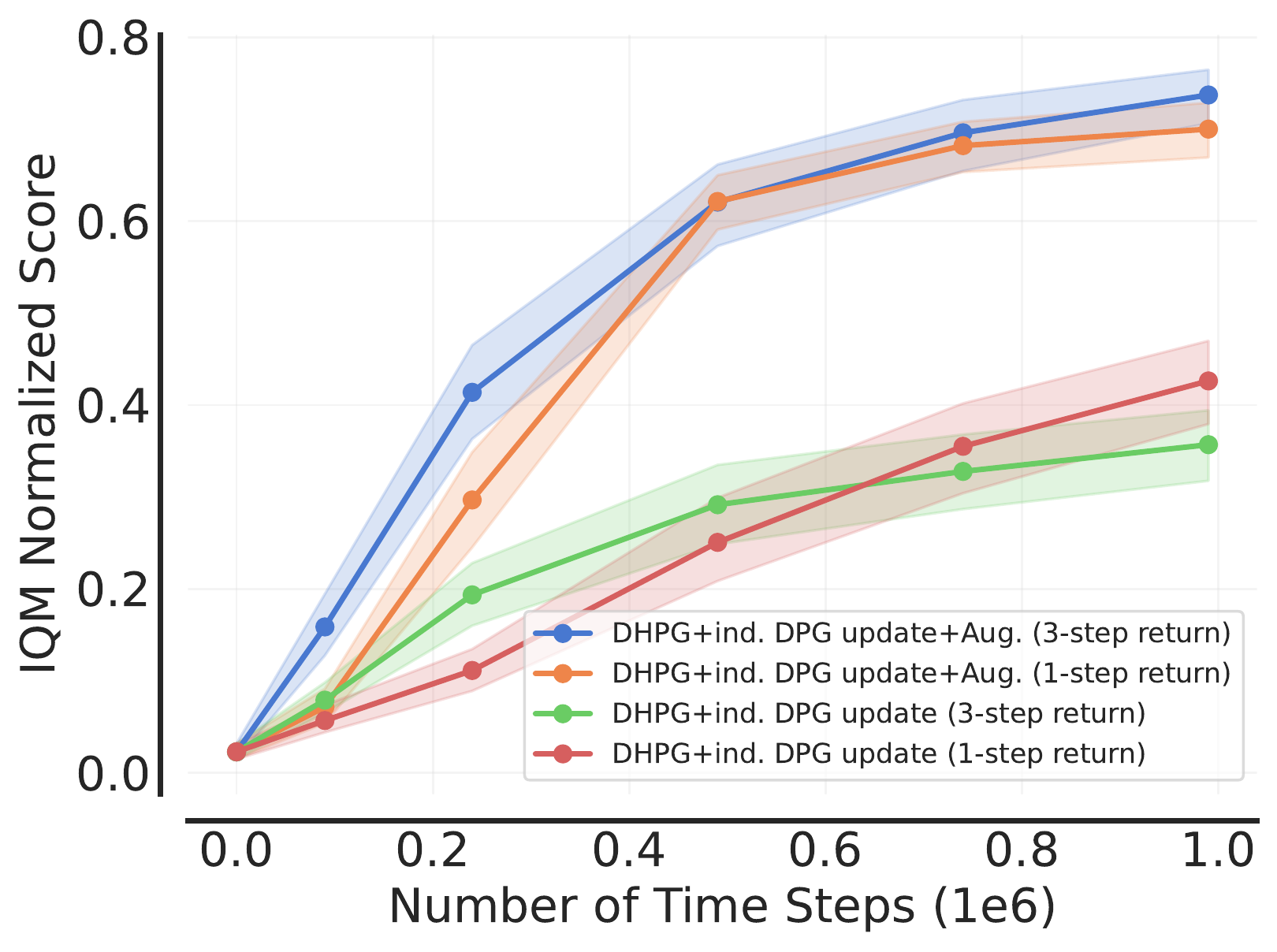}
         \caption{Sample efficiency.}
    \end{subfigure}
    \hfill
    \begin{subfigure}[b]{0.32\textwidth}
         \centering
         \includegraphics[width=\textwidth]{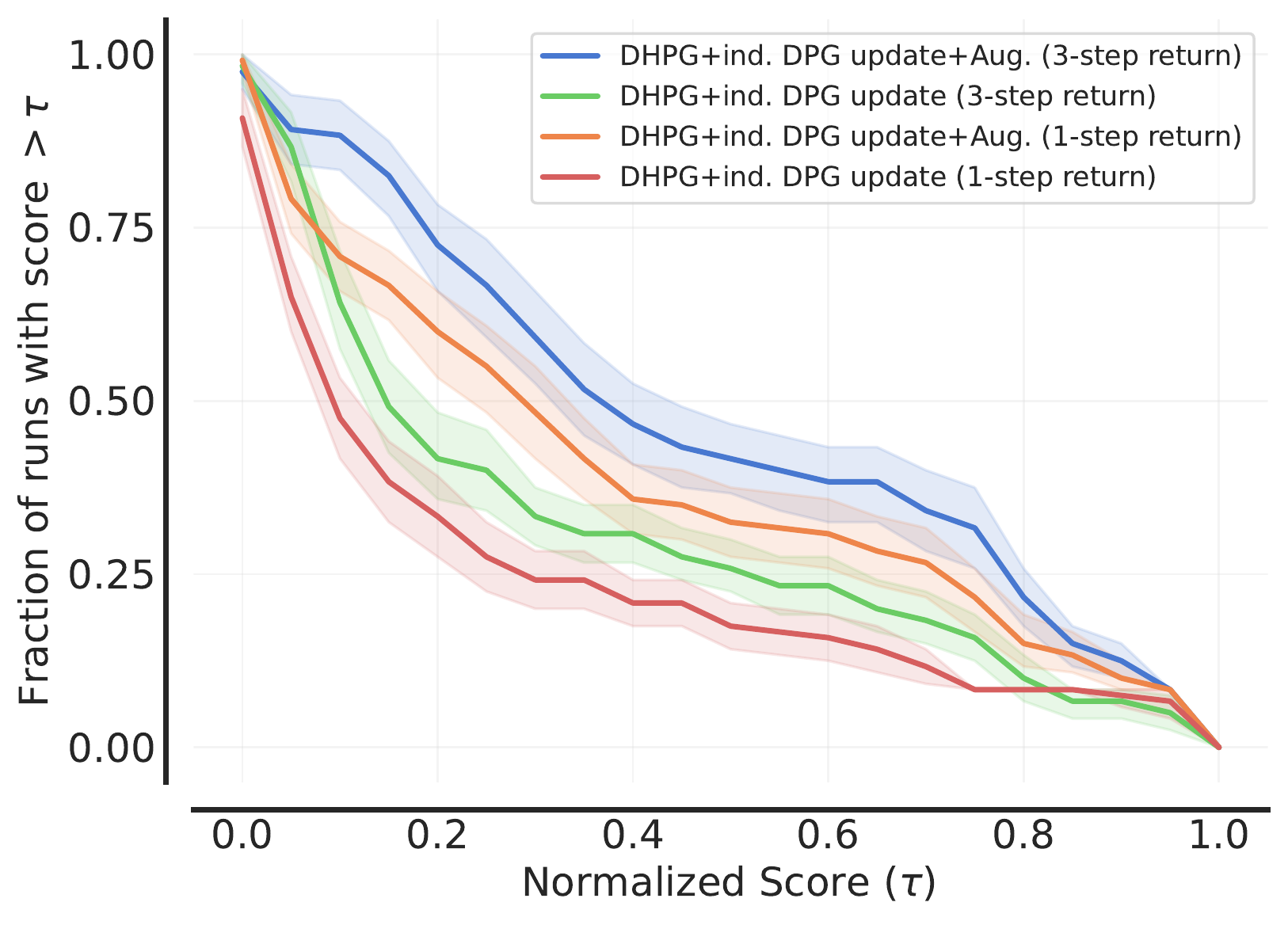}
         \caption{Performance profiles at 250k steps.}
    \end{subfigure}
    \hfill
    \begin{subfigure}[b]{0.32\textwidth}
         \centering
         \includegraphics[width=\textwidth]{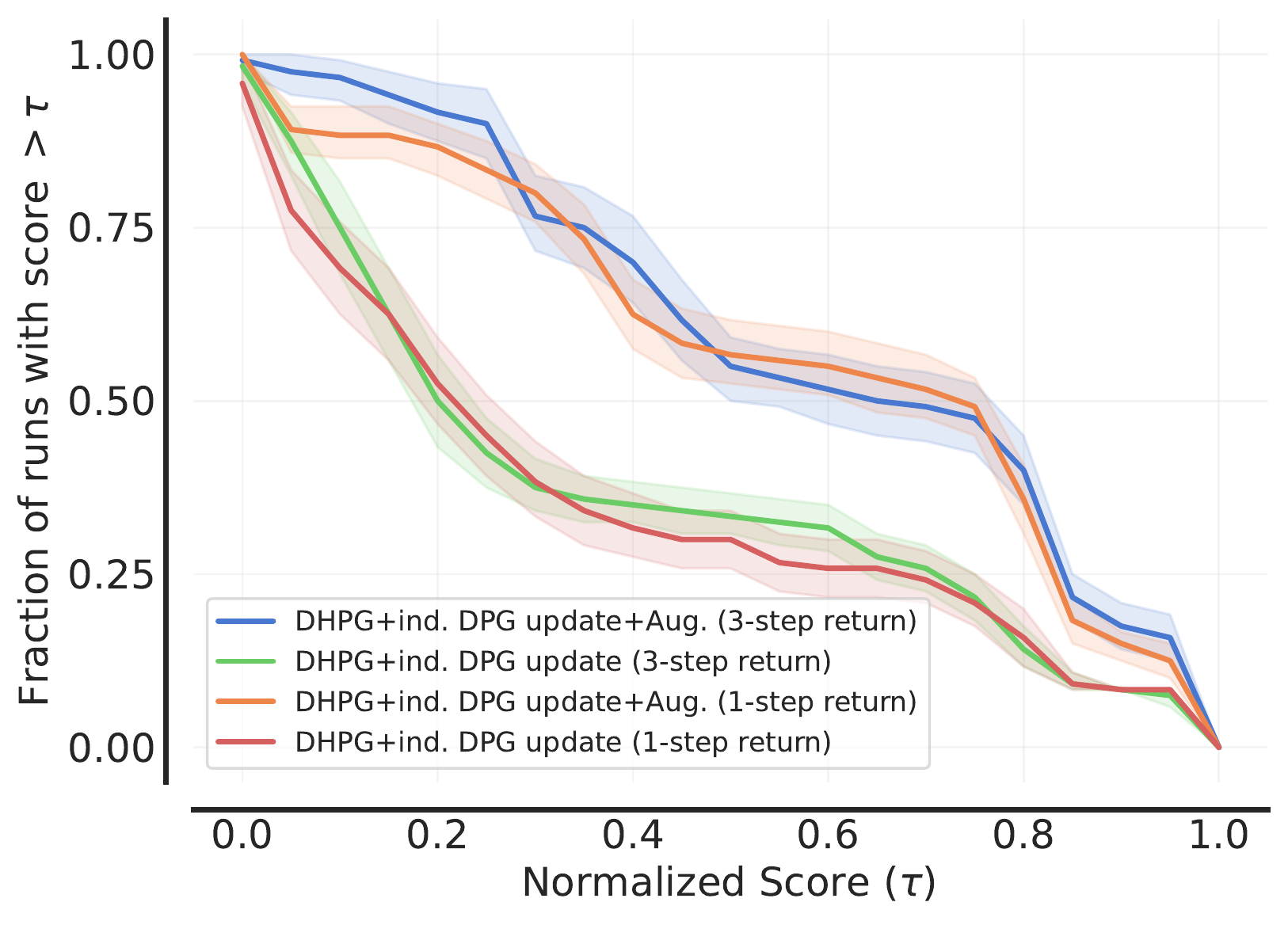}
         \caption{Performance profiles at 500k steps.}
    \end{subfigure}
    \caption{Ablation study on $n$-step return. RLiable evaluation metrics for \textbf{pixel observations} averaged on 12 tasks over 10 seeds. Aggregate metrics at 1m steps \textbf{(a)}, IQM scores as a function of number of steps for comparing sample efficiency \textbf{(b)}, performance profiles at 250k steps \textbf{(c)}, and performance profiles at 500k steps \textbf{(d)}. Shaded regions represent $95\%$ confidence intervals.}
    \label{fig:ablation_nstep_rliable}
\end{figure}

We carry out an ablation study on the choice of $n$-step return for DHPG. Figure \ref{fig:ablation_nstep_rliable} shows RLiable \cite{agarwal2021deep} evaluation metrics for DHPG with $1$-step and $3$-step returns for pixel observations. We show the impact of $n$-step return on DHPG with and without image augmentation. Overall, $n$-step return appears to improve the early stages of training. In the case of DHPG without image augmentation, the final performance of $1$-step return is better than $3$-step return, perhaps indicating that using $n$-step return can render learning MDP homomorphisms more difficult.
\label{sec:ablation_n_step}
\clearpage

\begin{figure}[h!]
     \centering
     \begin{subfigure}[b]{0.24\textwidth}
         \centering
         \includegraphics[width=\textwidth]{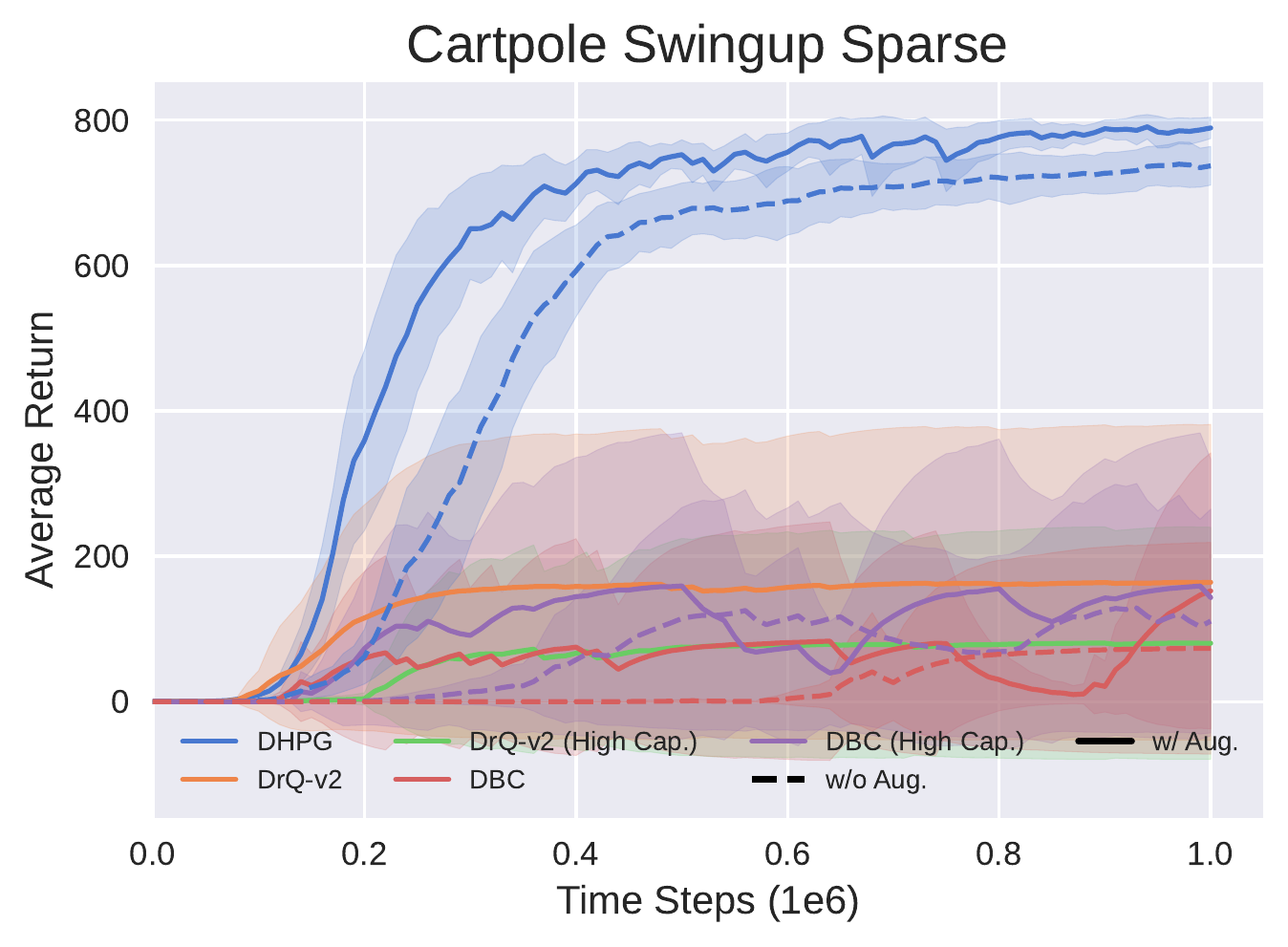}
     \end{subfigure}
     \hfill
     \begin{subfigure}[b]{0.24\textwidth}
         \centering
         \includegraphics[width=\textwidth]{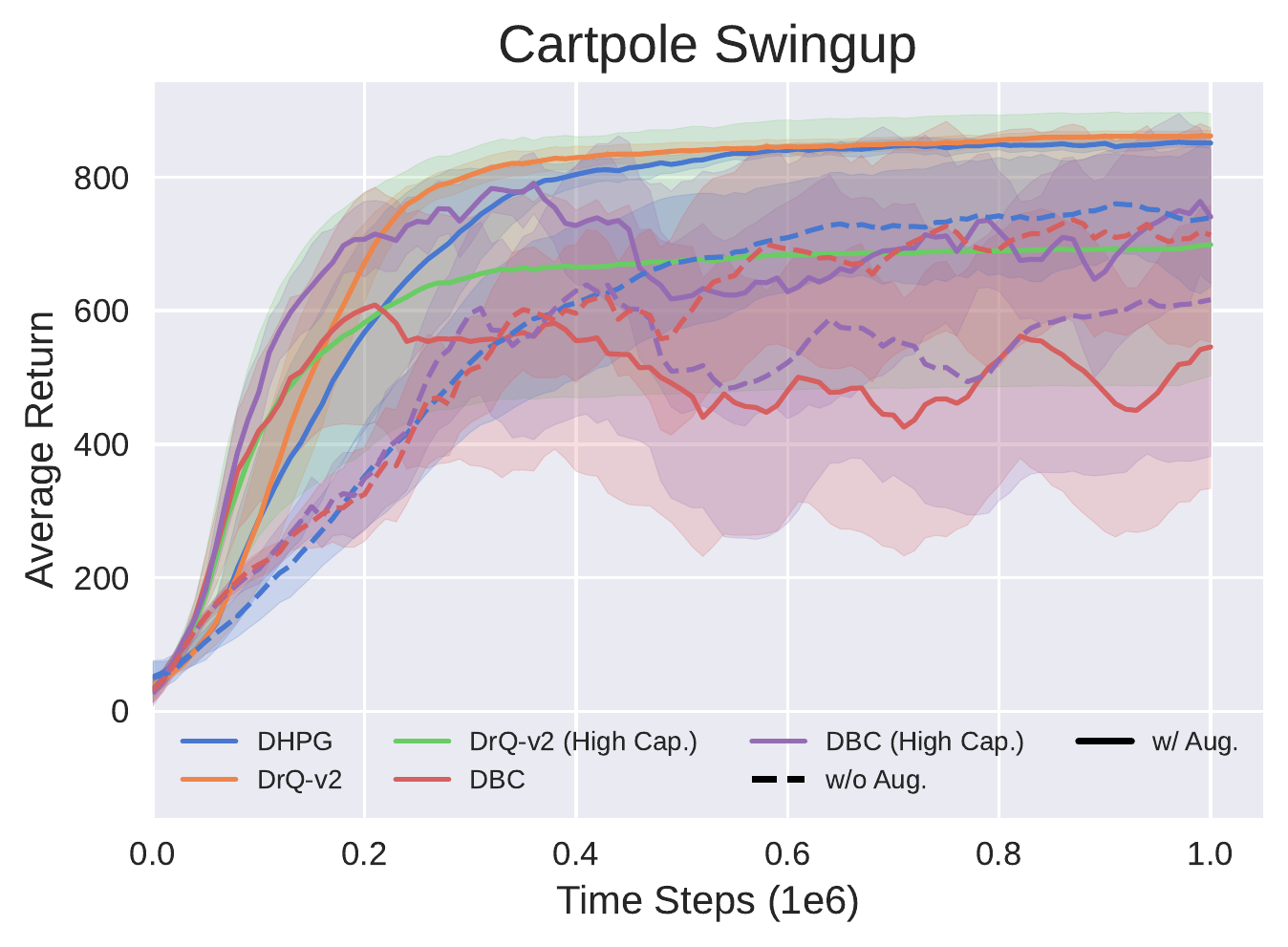}
     \end{subfigure}
     \hfill
     \begin{subfigure}[b]{0.24\textwidth}
         \centering
         \includegraphics[width=\textwidth]{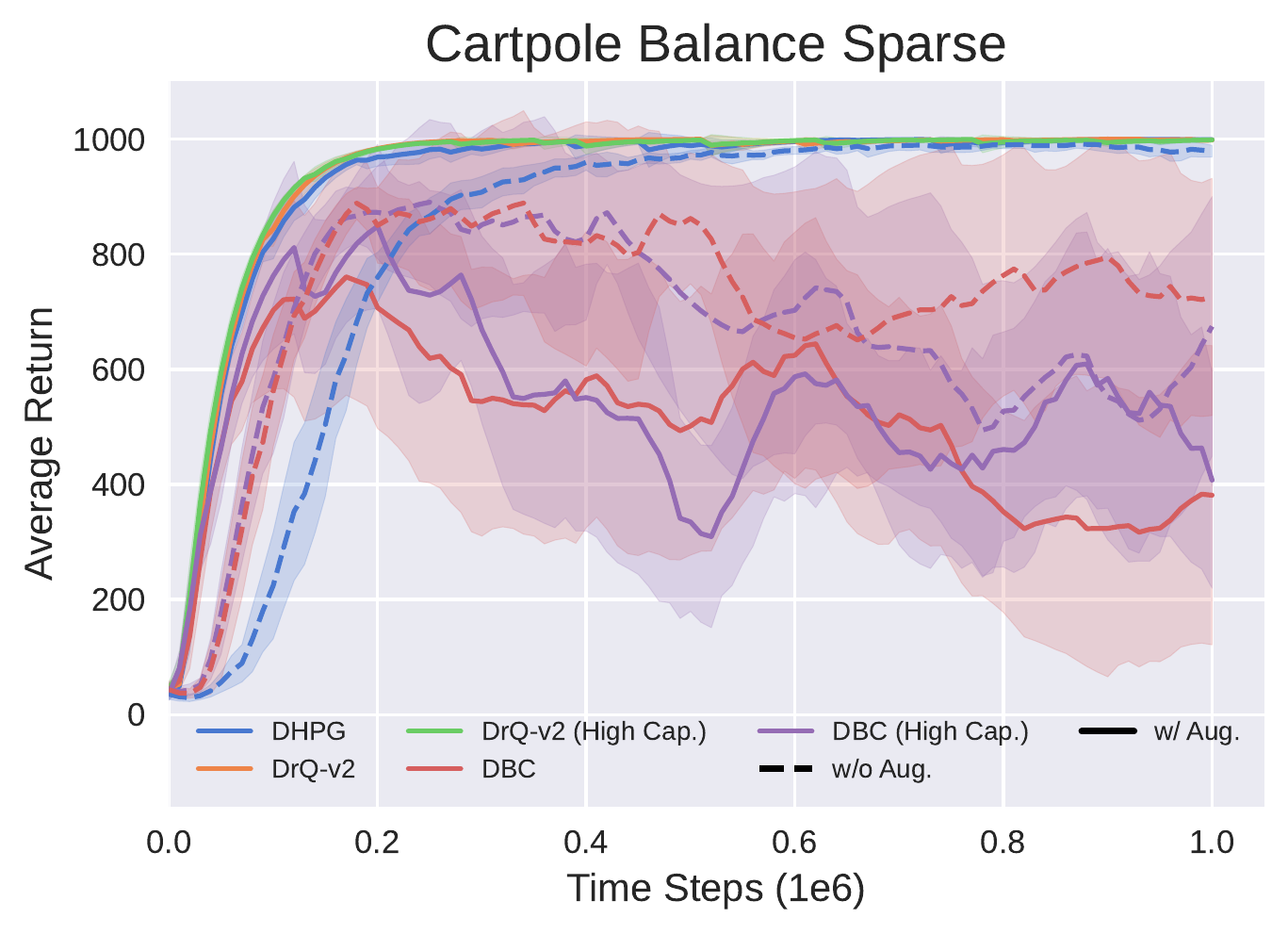}
     \end{subfigure}
     \hfill
     \begin{subfigure}[b]{0.24\textwidth}
         \centering
         \includegraphics[width=\textwidth]{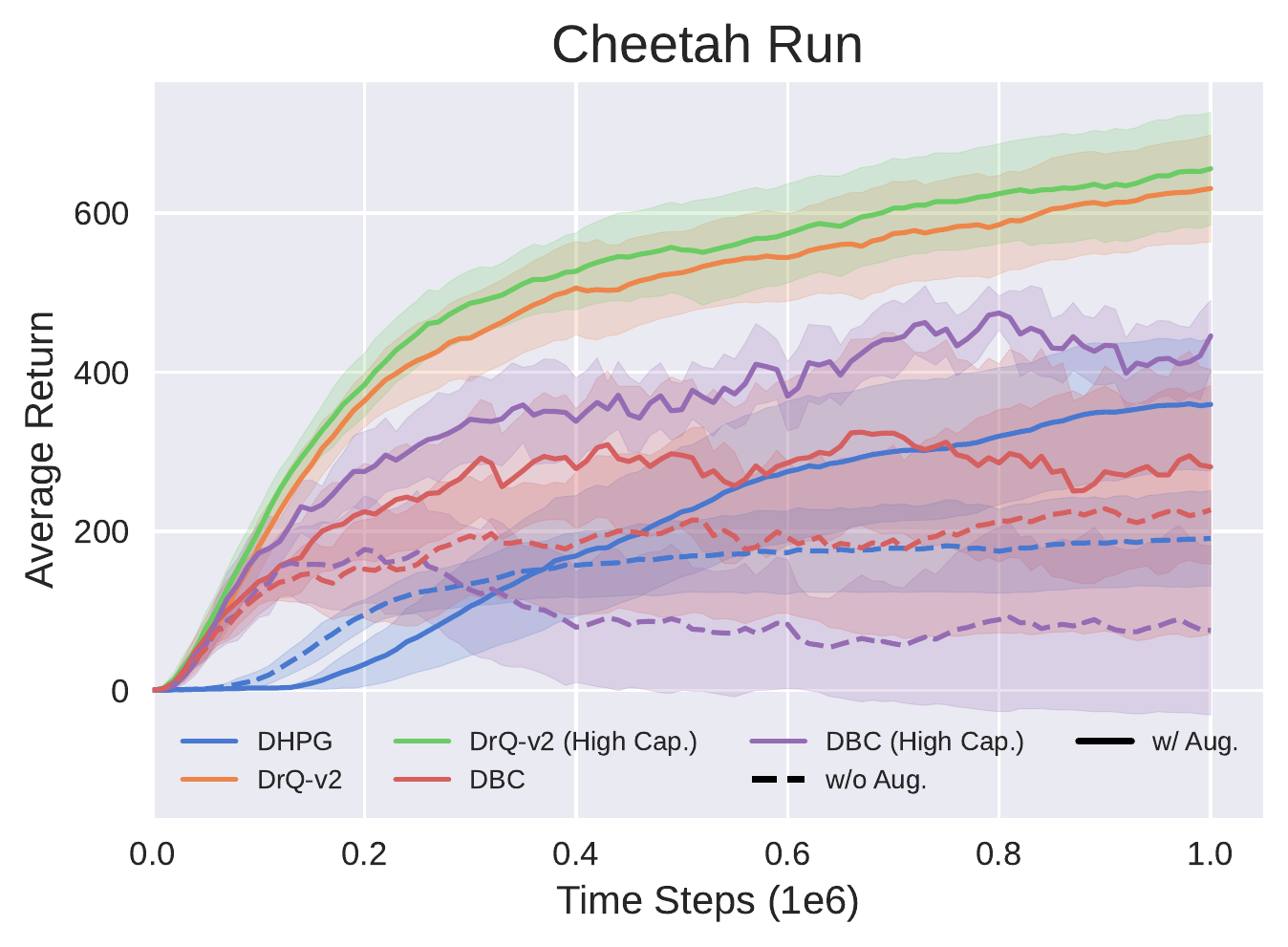}
     \end{subfigure}
     \hfill
     
     \begin{subfigure}[b]{0.24\textwidth}
         \centering
         \includegraphics[width=\textwidth]{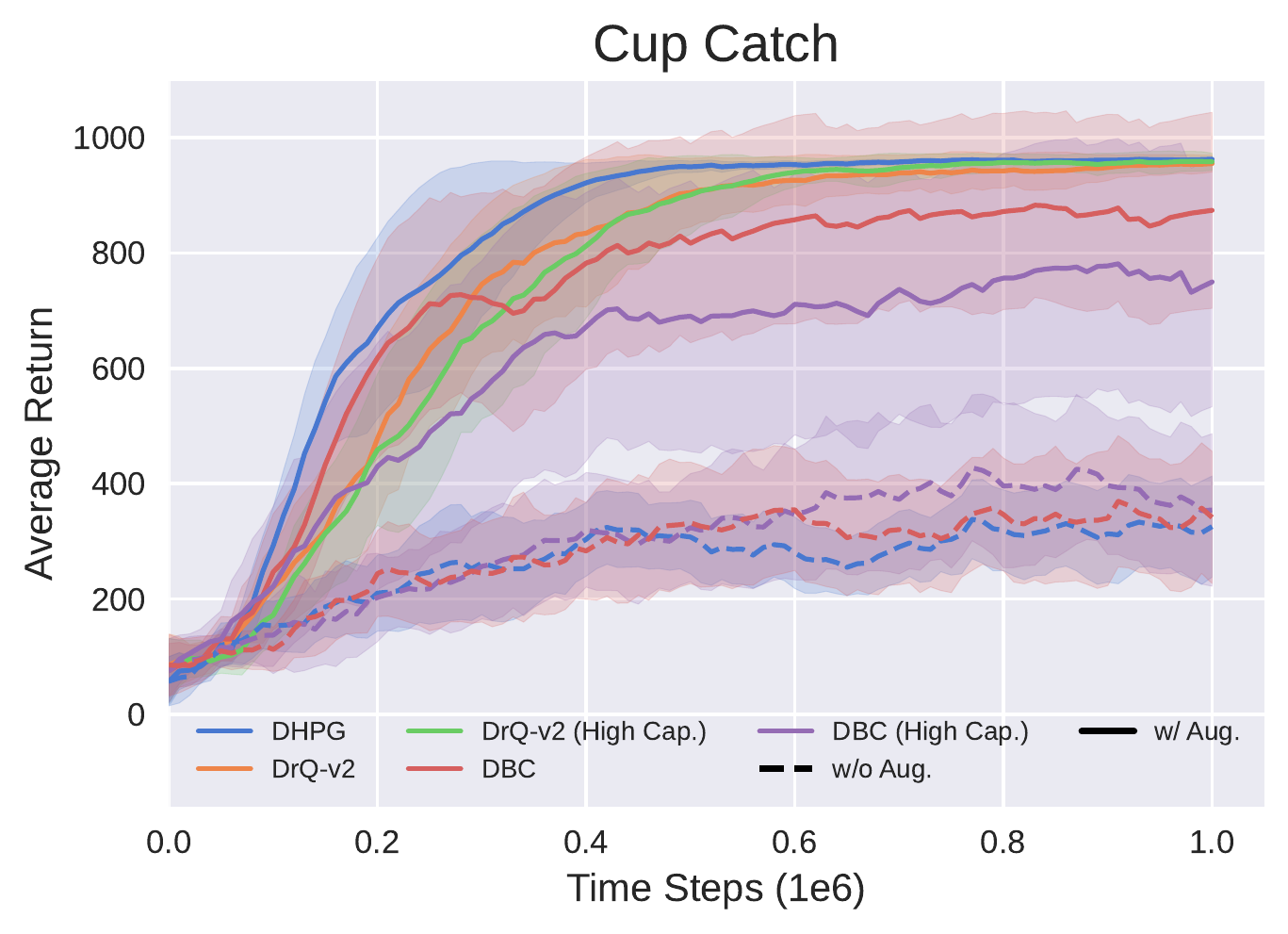}
     \end{subfigure}
     \hfill
     \begin{subfigure}[b]{0.24\textwidth}
         \centering
         \includegraphics[width=\textwidth]{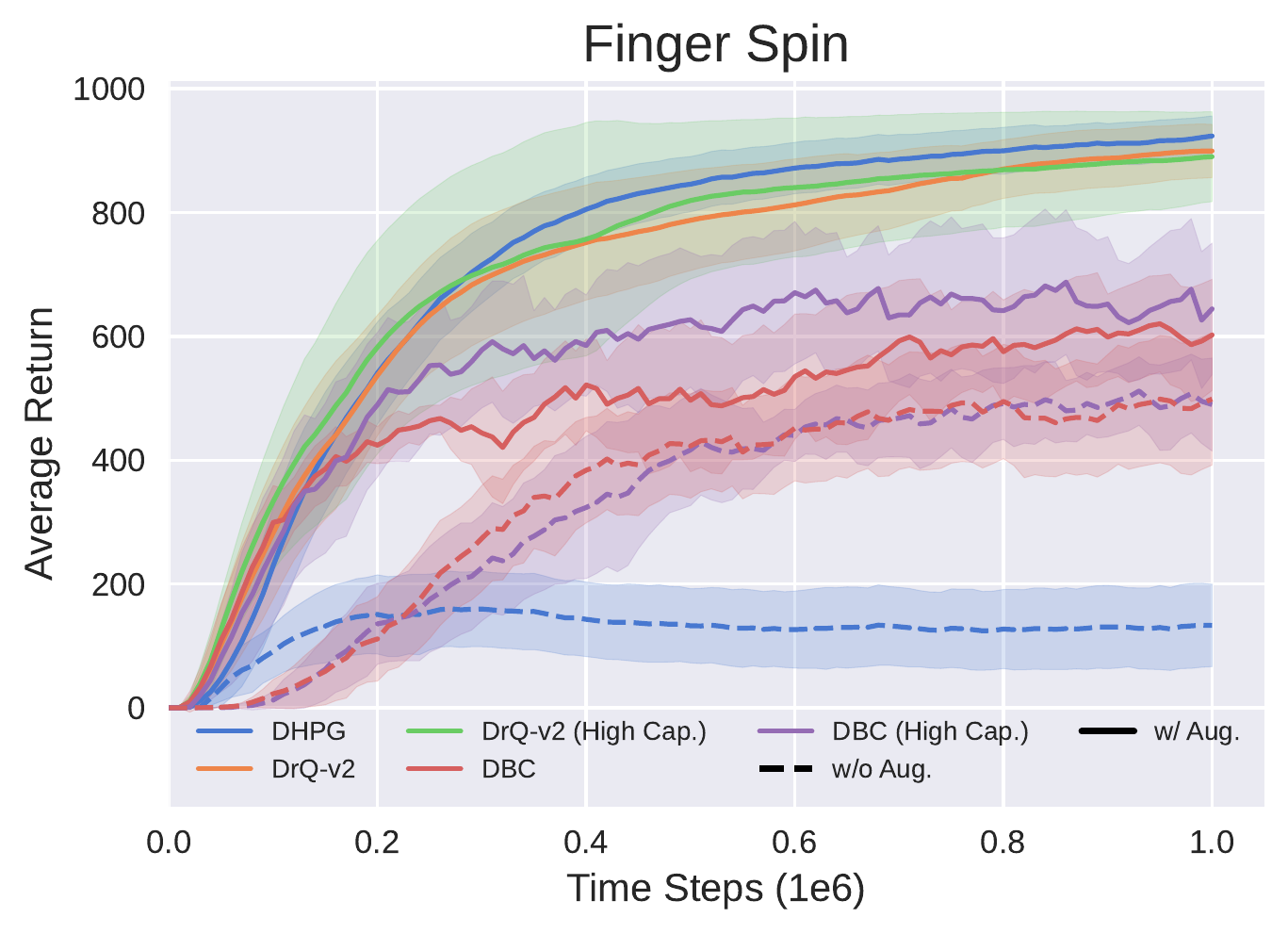}
     \end{subfigure}
     \hfill
     \begin{subfigure}[b]{0.24\textwidth}
         \centering
         \includegraphics[width=\textwidth]{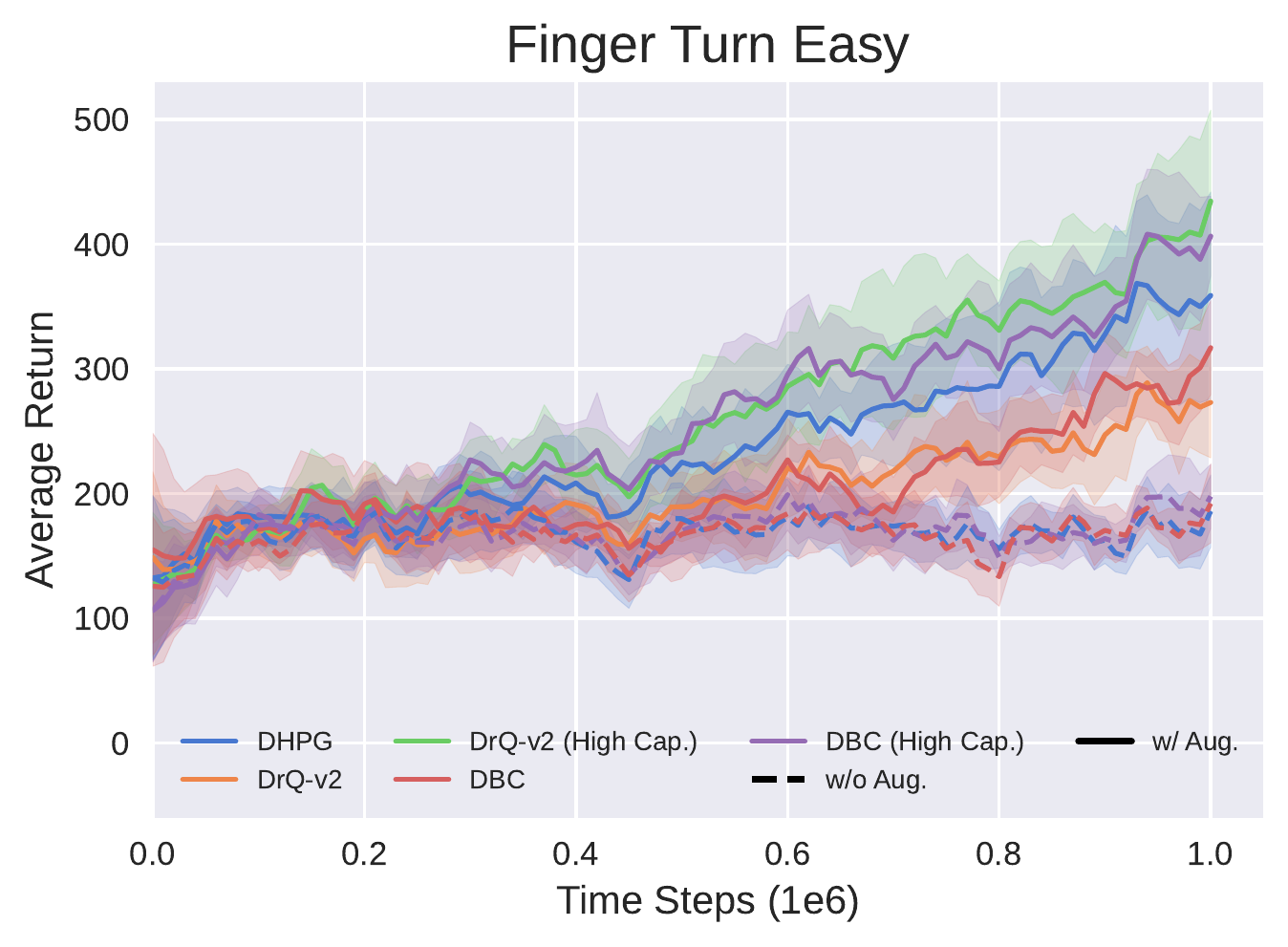}
     \end{subfigure}
     \hfill
     \begin{subfigure}[b]{0.24\textwidth}
         \centering
         \includegraphics[width=\textwidth]{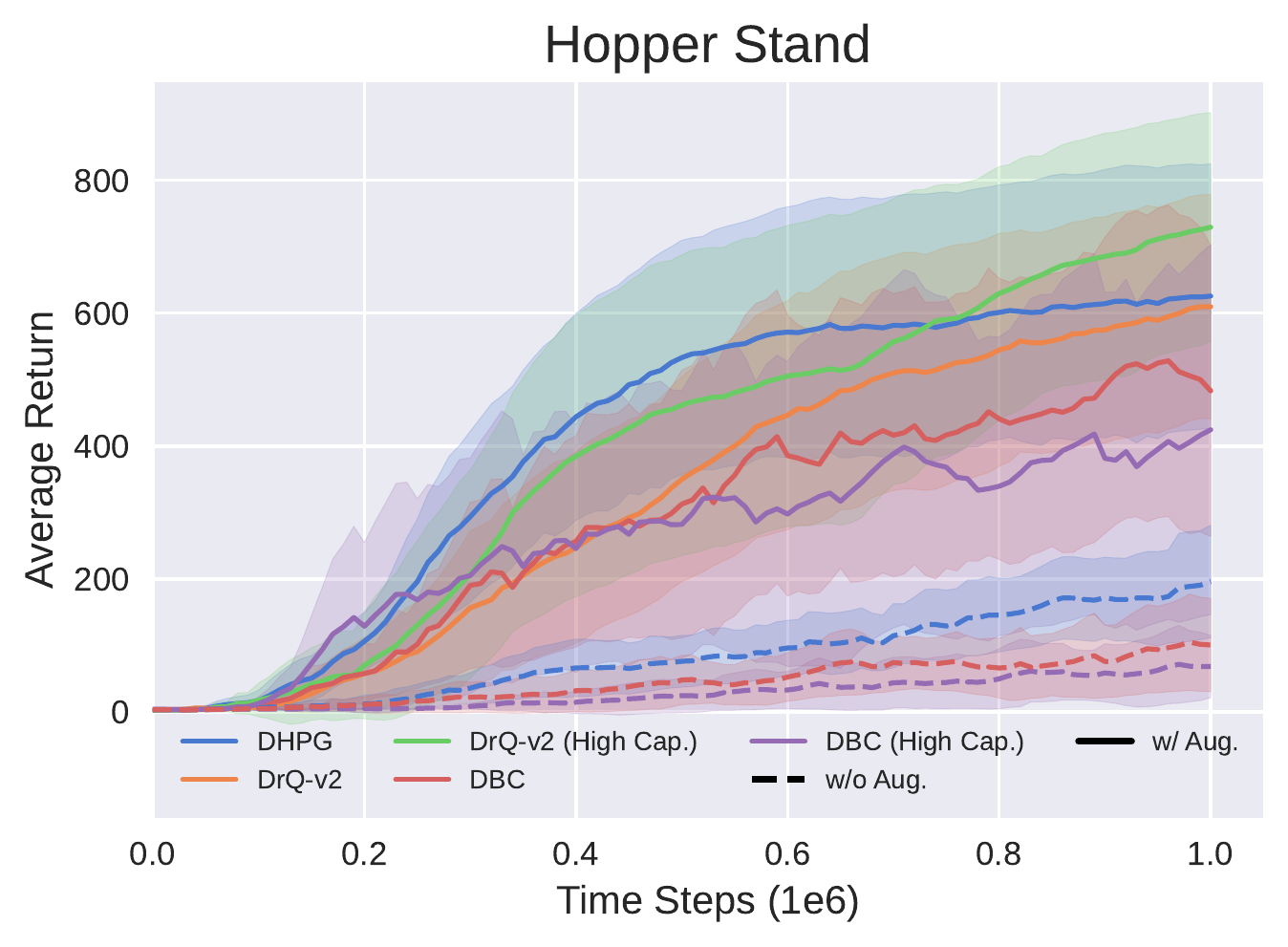}
     \end{subfigure}
     \hfill
     
     \begin{subfigure}[b]{0.24\textwidth}
         \centering
         \includegraphics[width=\textwidth]{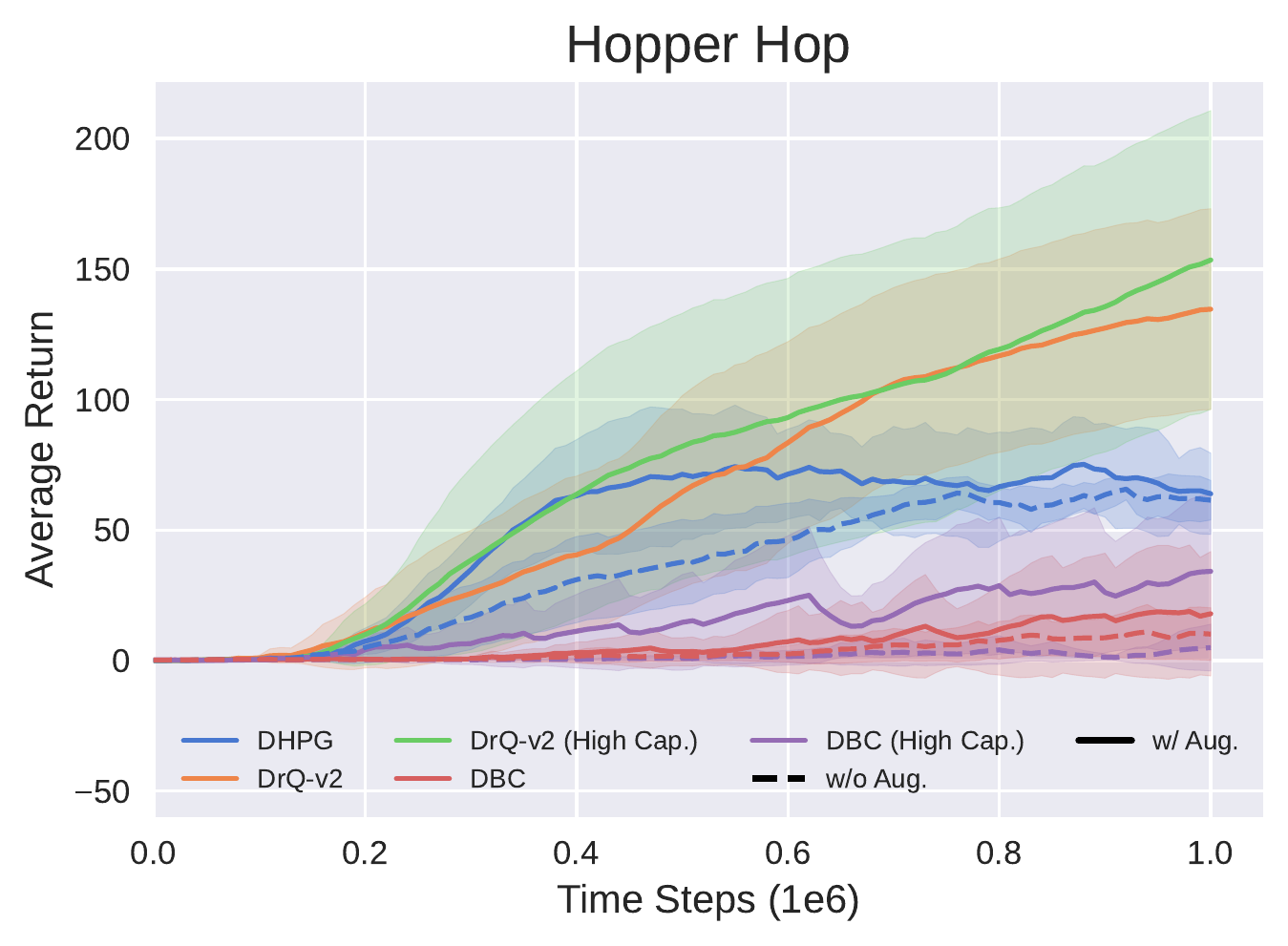}
     \end{subfigure}
     \hfill
     \begin{subfigure}[b]{0.24\textwidth}
         \centering
         \includegraphics[width=\textwidth]{figures/high_capacity_main/rebuttal_pendulum_swingup_episode_reward_eval.pdf}
     \end{subfigure}
     \hfill
     \begin{subfigure}[b]{0.24\textwidth}
         \centering
         \includegraphics[width=\textwidth]{figures/high_capacity_main/rebuttal_quadruped_walk_episode_reward_eval.pdf}
     \end{subfigure}
     \hfill
     \begin{subfigure}[b]{0.24\textwidth}
         \centering
         \includegraphics[width=\textwidth]{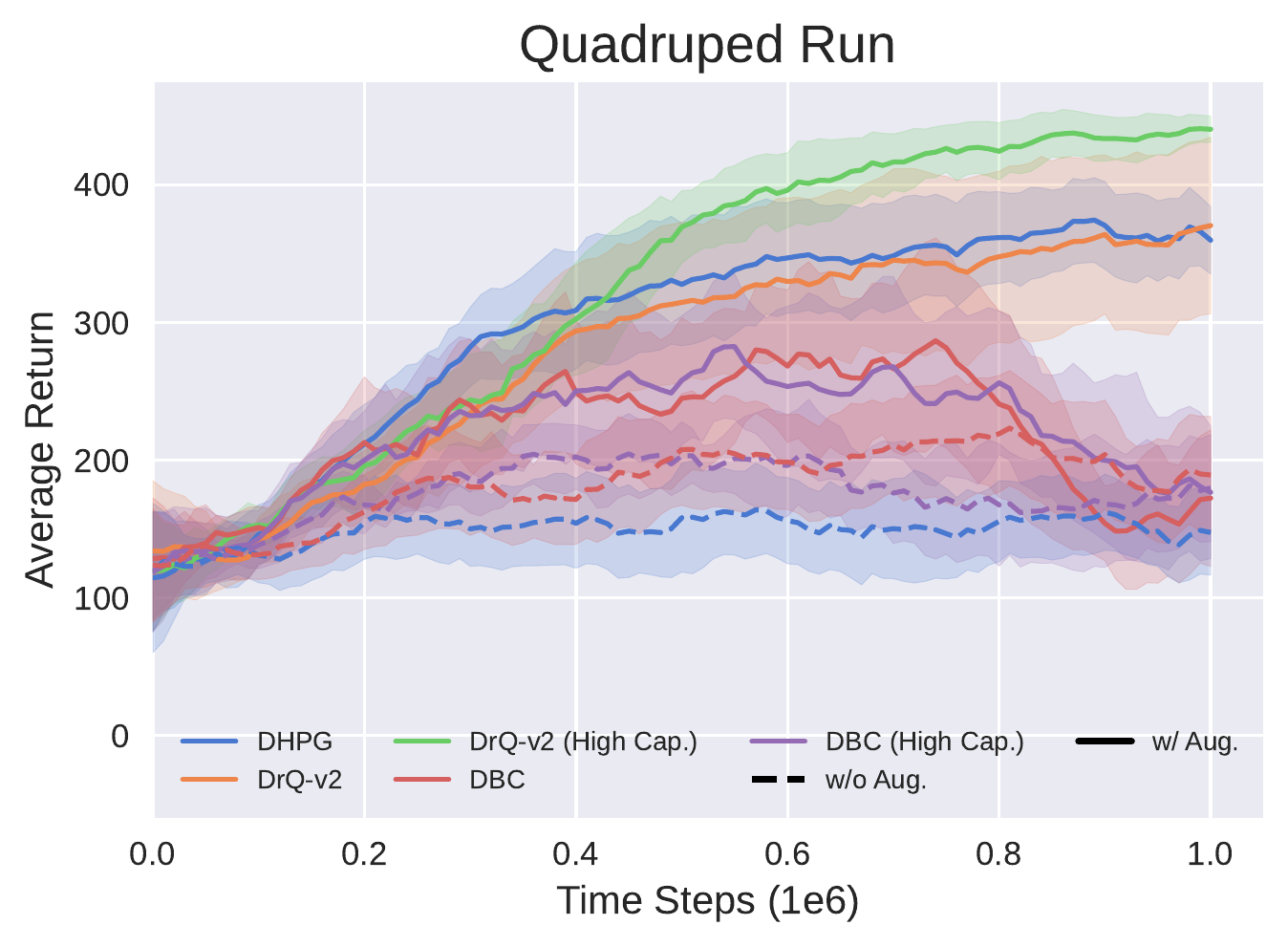}
     \end{subfigure}
     \hfill
     
     \begin{subfigure}[b]{0.24\textwidth}
         \centering
         \includegraphics[width=\textwidth]{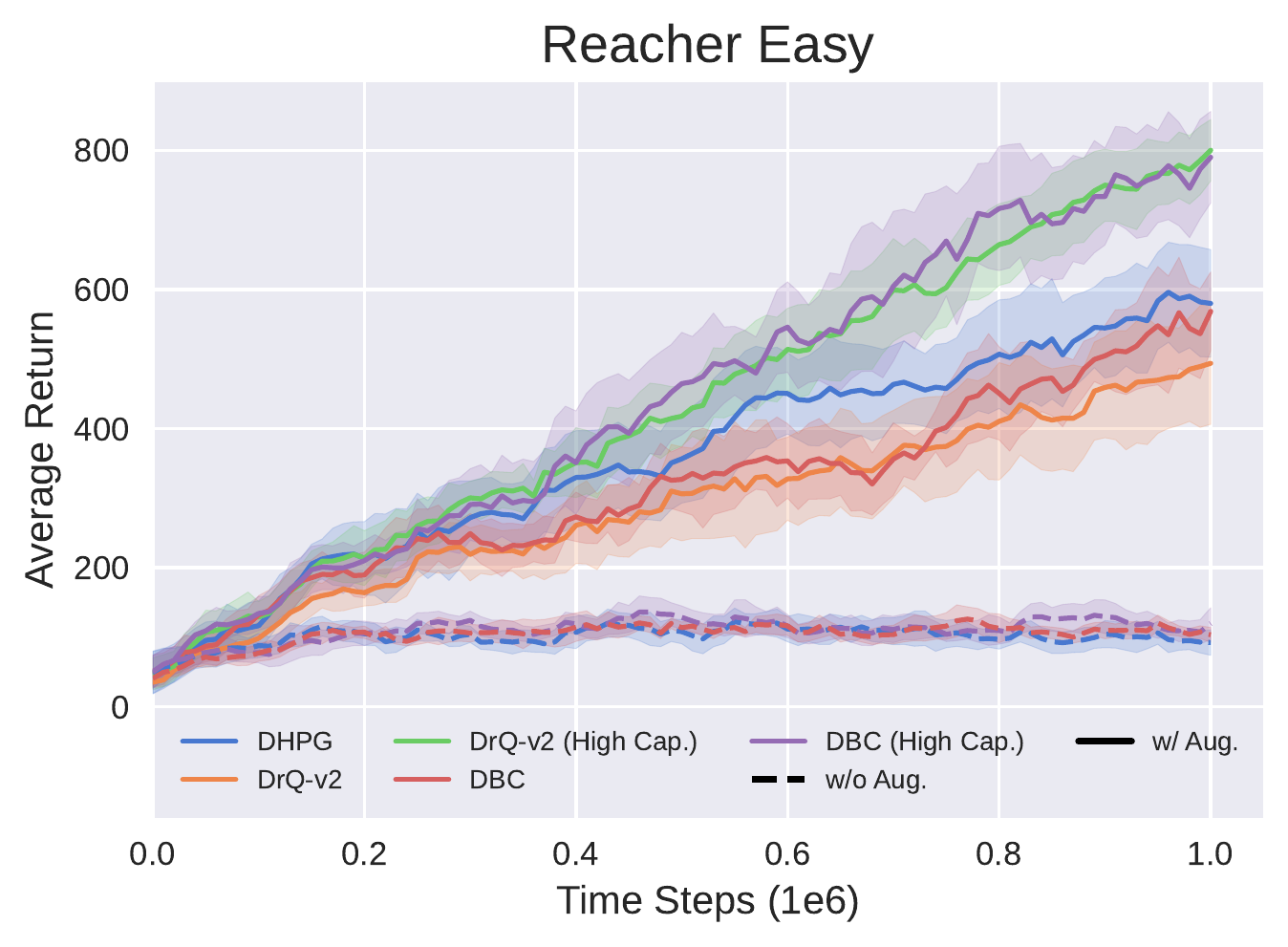}
     \end{subfigure}
     \hfill
     \begin{subfigure}[b]{0.24\textwidth}
         \centering
         \includegraphics[width=\textwidth]{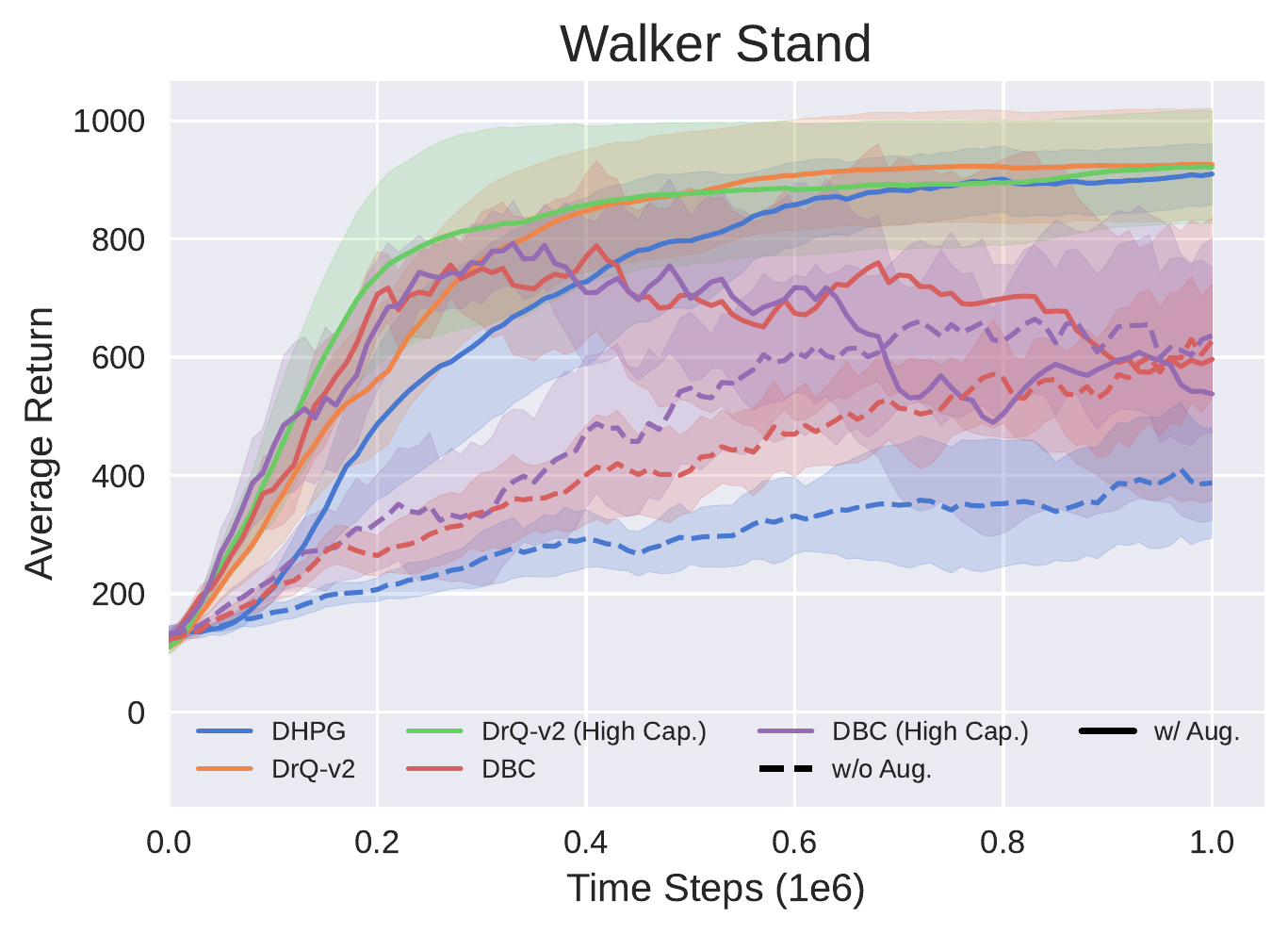}
     \end{subfigure}
     \begin{subfigure}[b]{0.24\textwidth}
         \centering
         \includegraphics[width=\textwidth]{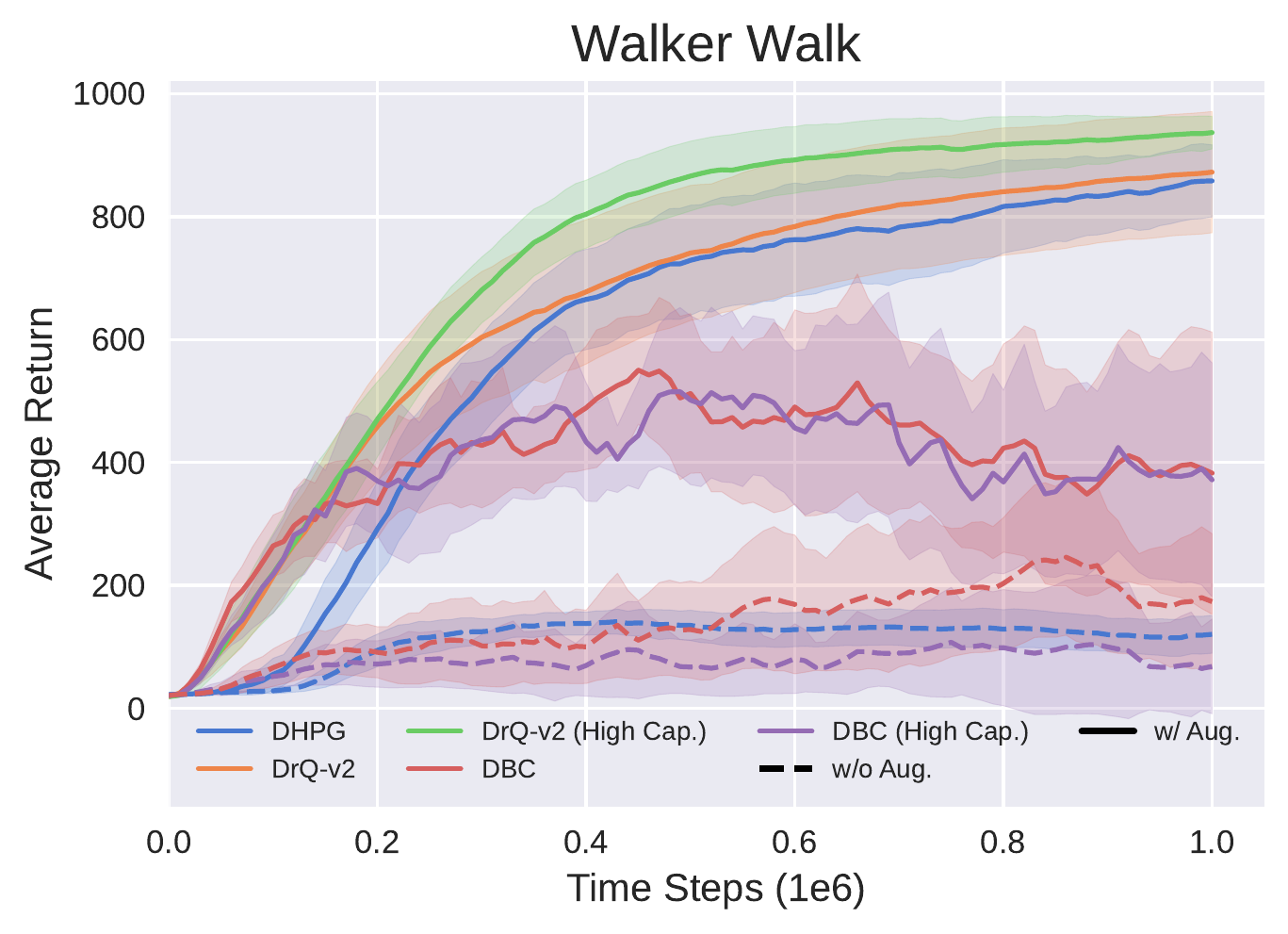}
     \end{subfigure}
     \hfill
     \begin{subfigure}[b]{0.24\textwidth}
         \centering
         \includegraphics[width=\textwidth]{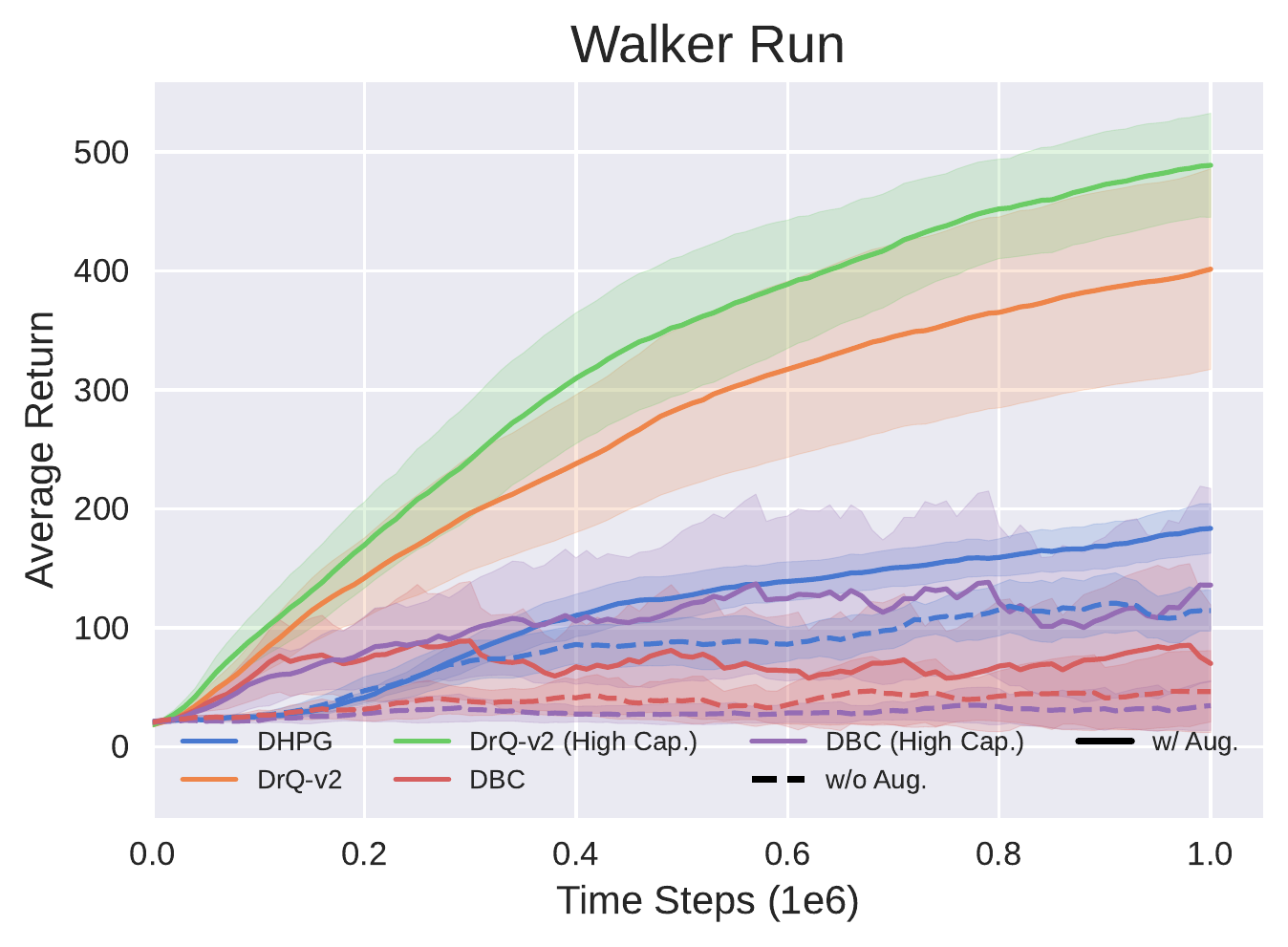}
     \end{subfigure}
     \hfill
    \caption{\rebuttal{Learning curves for 16 DM control tasks with \textbf{pixel observations} for \textbf{higher-capacity variants} of DBC and DrQ-v2. Mean performance is obtained over 10 seeds and shaded regions represent $95\%$ confidence intervals. Plots are smoothed uniformly for visual clarity.}}
    \label{fig:high_capacity_results_supp}
\end{figure}

\begin{figure}[h!]
    \centering
    \begin{subfigure}[b]{0.45\textwidth}
        \includegraphics[width=\textwidth]{figures/high_capacity_rliable/rebuttal_performance_profiles_500k.pdf}
        \caption{500k step benchmark.}
    \end{subfigure}
    \hfill
    \begin{subfigure}[b]{0.45\textwidth}
        \includegraphics[width=\textwidth]{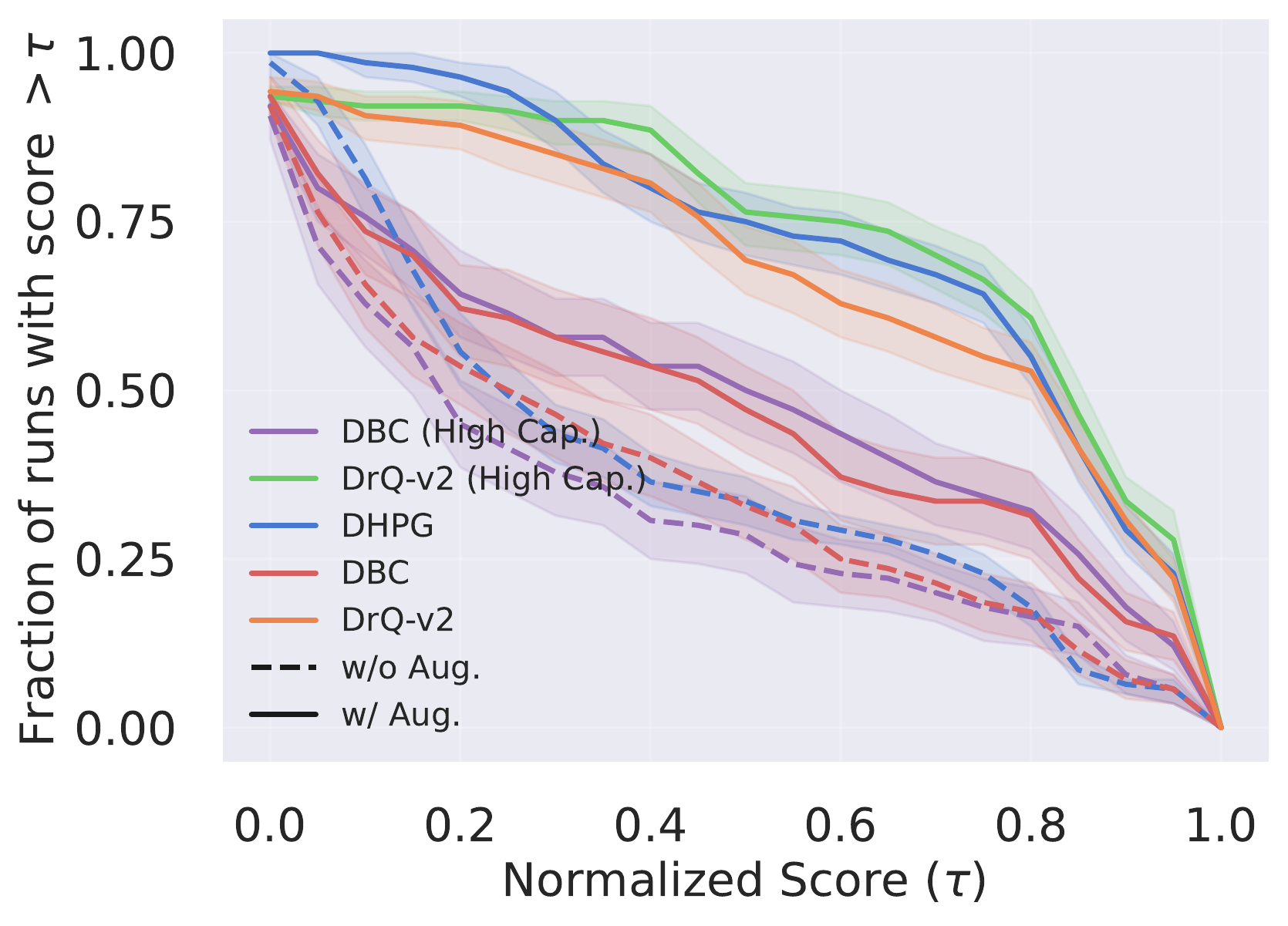}
        \caption{1m step benchmark.}
    \end{subfigure}
    \caption{\rebuttal{Performance profiles for \textbf{pixel observations} for \textbf{higher-capacity variants} of DBC and DrQ-v2 based on 14 tasks over 10 seeds, at 500k steps \textbf{(a)}, and at 1m steps \textbf{(b)}. Shaded regions represent $95\%$ confidence intervals.}}
    \label{fig:high_capacity_results_performance_profiles}
\end{figure}

\begin{figure}[h!]
    \centering
    %
    \begin{subfigure}[b]{0.95\textwidth}
        \includegraphics[width=\textwidth]{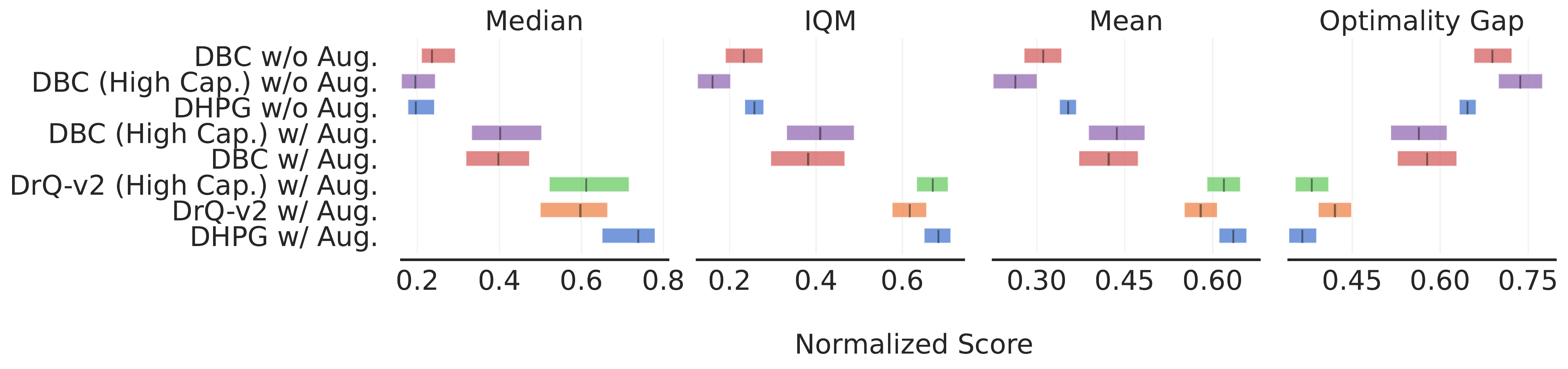}
        \caption{500k step benchmark.}
    \end{subfigure}
    
    \begin{subfigure}[b]{0.95\textwidth}
        \includegraphics[width=\textwidth]{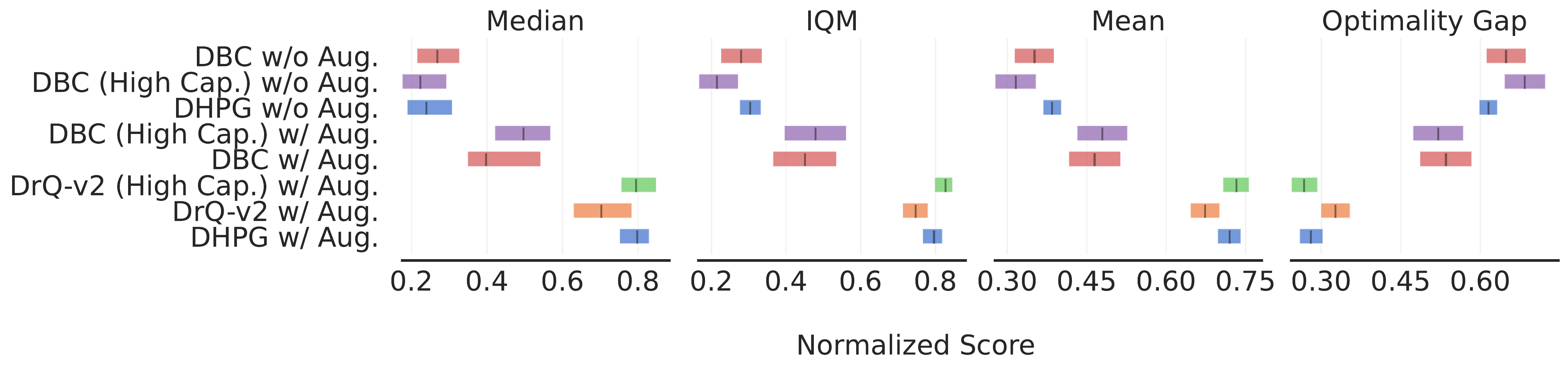}
        \caption{1m step benchmark.}
    \end{subfigure}
    \caption{\rebuttal{Aggregate metrics for \textbf{pixel observations} for \textbf{higher-capacity variants} of DBC and DrQ-v2 with $95\%$ confidence intervals based on 14 tasks over 10 seeds, at 500k steps \textbf{(a)}, and at 1m steps \textbf{(b)}.}}    
    \label{fig:high_capacity_results_aggregate_metrics}
\end{figure}

\rebuttal{
\subsection{Comparison Against Higher-Capacity Baselines}
\label{sec:high_capacity_supp}
The DHPG algorithm contains additional networks, such as the parameterized MDP homomorphism map and the abstract critic, thus it may have a higher network capacity compared to the baselines. To control for the effect of the network capacity and for a fair evaluation, we compare DHPG with higher-capacity variants of DBC and DrQ-v2 that have a larger critic networks. First, we provide a detailed list of network parameters based on the architecture described in Appendix \ref{sec:hyperparams}:
\begin{enumerate}[noitemsep]
    \item DHPG: image encoder (1,990,518) + actor (79,105) + critic (79,361) + dynamics model (117,348) + reward model (79,105) + abstract critic (91,905) + f (91,698) + g (91,954) = 2,620,994
    \item DBC: image encoder (1,990,518) + actor (79,362) + critic (158,722) + dynamics model (104,804) + reward model (79,105) = 2,412,511
    \item DrQ-v2: image encoder (1,990,518) + actor (79,105) + critic (158,722) = 2,228,602
\end{enumerate}

To account for the parameter increase, we present variations of DBC and DrQ with a larger critic (512 hidden dim compared to the initial 256). Consequently, the new total number of parameters for DBC and DrQ are respectively 2,833,375 and 2,649,466. Figure \ref{fig:high_capacity_results_supp} shows full results obtained on 16 DeepMind Control Suite tasks with pixel observations for higher-capacity variants of DBC and DrQ-v2 to supplement results of Section \ref{sec:results_pixels}. Domains that require excessive exploration and large number of time steps (e.g., acrobot, swimmer, and humanoid) and domains with visually small targets (e.g., reacher hard and finger turn hard) are not included in this benchmark. In each plot, the solid lines present algorithms with image augmentation and dashed lines present algorithms without image augmentation.

Figures \ref{fig:high_capacity_results_performance_profiles} and \ref{fig:high_capacity_results_aggregate_metrics} respectively show performance profiles and aggregate metrics \cite{agarwal2021deep} on 14 tasks; hopper hop and walker run are removed from RLiable evaluation as none of the algorithms have acquired reasonable performance in 1 million steps.
}

\clearpage

\clearpage

\clearpage
\section{Implementation Details}
\label{sec:implementation}
\subsection{Pseudo-code}
\label{sec:pseudocode}
Algorithm \ref{alg:homomorphic_ddpg} presents the details of the Deep Homomorphic Policy Gradient (DHPG) for pixel observations. This is the main variant used throughout the paper, in which policy gradients obtained from DPG and HPG are added together before updating the actor. For clarity, here the TD error is estimated with $1$-step returns. 

In the image augmentation version of DHPG, as well as all the baselines, we use image augmentation of DrQ \cite{yarats2020image} that simply applies random shifts to pixel observations. First, $84 \times 84$ images are padded by $4$ pixels (by repeating boundary pixels), and then a random $84 \times 84$ crop is selected, rendering the original image shifted by $\pm 4$ pixels. Similarly to Yarats et al. \cite{yarats2021mastering}, we also apply bilinear interpolation on top of the shifted image by replacing each pixel value with the average of four nearest pixel values. 

In order to use DHPG for state observations, Lines 8-11 should be simply removed.

\begin{algorithm}[h!]
    \caption{Deep Homomorphic Policy Gradient (DHPG) for Pixel Observations}
    \label{alg:homomorphic_ddpg}
\begin{algorithmic}[1]
    \State \textbf{Hyperparameters:}\newline Target network update weight $\alpha$, actor update delay $d$, clipped noise parameters $c$ and $\sigma$.
    \State \textbf{Inputs:}\newline Policy $\pi_\theta(s, a)$, actual critic $Q_{\psi}(s, a)$, abstract critic $\overline{Q}_{\overline{\psi}}(\overline{s}, \overline{a})$, MDP homomorphism map $h_{\phi, \eta} = (f_\phi(s), g_\eta(s, a))$, reward predictor $\overline{R}_\rho(\overline{s})$, transition model $\tau_\nu(\overline{s}' | \overline{s}, \overline{a})$, CNN image encoder $E_\mu$, and replay buffer $\mathcal{B}$.
    \State Initialize target networks $\psi' \leftarrow \psi$, $\overline{\psi'} \leftarrow \overline{\psi}$, $\theta' \leftarrow \theta$.
    \For{$t=1$ {\bfseries to} $T$}
        \State Select action with exploration noise $a \sim \pi_\theta(E_\mu(s)) + \epsilon$, where $\epsilon \sim {N}(0, \sigma)$ 
        \State Store transition $( s, a, r, s')$ in $\mathcal{B}$
        \State Sample mini-batch $B_i \sim \mathcal{B}$ \newline
        \If {using image augmentation}
            \State $s \leftarrow \text{aug}(s), \; s' \leftarrow \text{aug}(s')$
        \EndIf
        \State Encode pixel observations: $\; s \leftarrow E_\mu(s), \; s' \leftarrow E_\mu(s')$ \newline
        \State \textbf{Critic and MDP Homomorphism Update:}
        \State Compute MDP homomorphism loss: $\; \mathcal{L}_\text{lax}(\phi, \eta, \mu) + \mathcal{L}_\text{h}(\phi, \eta, \rho, \nu, \mu)$ \Comment{Equations (\ref{eq:lax_bisim_loss}-\ref{eq:hom_loss})}
        \State Add clipped noise: $\; a' \leftarrow \pi_{\theta'}(s') + \epsilon$, where $\epsilon \sim \text{clip}({N}(0, \sigma), -c, c)$ \Comment{TD3 \cite{fujimoto2018addressing}}
        \State Compute critic loss: $\; \mathcal{L}_\text{actual critic} (\psi) + \mathcal{L}_\text{abstract critic}(\overline{\psi}, \phi, \eta)$ \Comment{Equations (\ref{eq:critic_loss_1}-\ref{eq:critic_loss_2})}
        \State Update:  $\psi, \overline{\psi}, \phi, \eta, \rho, \nu, \mu \leftarrow \argmin_{\psi, \overline{\psi}, \phi, \eta, \rho, \nu, \mu} \mathcal{L}_\text{lax} + \mathcal{L}_\text{h} + \mathcal{L}_\text{actual critic} + \mathcal{L}_\text{abstract critic} $ 
        \State 
        \State \textbf{Actor update:}
        \If{$t \mod d$}
            \State Freeze $Q_\psi, \overline{Q}_{\overline{\psi}}, f_\phi, g_\eta$, and $E_\mu$
            \State Compute policy loss using DPG and HPG: $\; \mathcal{L}_\text{actor}(\theta)$ \Comment{Equation \eqref{eq:hpg_actor_update}}
            \State Update policy: $\; \theta \leftarrow \argmin \mathcal{L}_\text{actor}(\theta)$
            \State Update target networks $\quad \psi' \leftarrow \alpha \psi + (1 - \alpha) \psi', \; \overline{\psi}' \leftarrow \alpha \overline{\psi} + (1 - \alpha) \overline{\psi}', \; \theta' \leftarrow \alpha \theta + (1 - \alpha) \theta'$
        \EndIf
    \EndFor
\end{algorithmic}
\end{algorithm}

\subsection{Hyperparameters}
\label{sec:hyperparams}
Our code is submitted in the suplemental material. 

We implemented our method in PyTorch \cite{paszke2019pytorch} and results were obtained using Python v3.8.10, PyTorch v1.10.0, CUDA 11.4, and Mujoco 2.1.1 \cite{todorov2012mujoco} on A100 GPUs on a cloud computing service. Tables \ref{tab:hyperparams}-\ref{tab:hyperparams_pixels} present the hyperparameters used in our experiments. The hyperparameters are all adapted from DrQ-v2 \cite{yarats2021mastering} \emph{without any further hyperparameter tuning}. We have kept the same set of hyperparameters across all algorithms and tasks, except for the walker domain which similarly to DrQ-v2 \cite{yarats2021mastering}, we used $n$-step return of $n=1$ and mini-batch size of $512$.

The core RL components (actor and critic networks), as well as the components of DHPG (state and action encoders, transition and reward models) are all MLP networks with the ReLU activation function and one hidden layer with dimension of $256$. 

In the case of state observations, the abstract MDP has the same state and action dimensions as the actual MDP. In the case of pixel observations, the image encoder is based on the architecture of DrQ-v2 which is itself based on SAC-AE \cite{yarats2021improving} and consists of four convolutional layers of $32 \times 3 \times 3$ with ReLU as their activation functions, followed by a one-layer fully-connected neural network with layer normalization \cite{ba2016layer} and tanh activation function. The stride of the convolutional layers are $1$, except for the first layer which has stride $2$. The image decoder of the baseline models with image reconstruction is based on SAC-AE \cite{yarats2021improving} and has a single-layer fully connected neural network followed by four transpose convolutional layers of $32 \times 32 \times 3$ with ReLU activation function. The stride of the transpose convolutional layers are $1$, except for the last layer which has stride $2$.

\begin{table}[h!]
\centering
\caption{Hyperparameters used in our experiments.}
\label{tab:hyperparams}
\begin{tabular}{cc}
\hline
\textbf{Hyperparameter}                       & \textbf{Setting}         \\
\hline
Learning rate                            & $1$e$-4$                   \\
Optimizer                                & Adam                     \\
$n$-step return                          & $3$                        \\
Mini-batch size                          & $256$                      \\
Actor update frequency $d$               & $2$                        \\
Target networks update frequency         & $2$                        \\
Target networks soft-update $\tau$       & $0.01$                     \\
Target policy smoothing stddev. clip $c$ & 0.3                      \\
Hidden dim.                              & $256$                      \\
Replay buffer capacity                   & $10^6$                   \\
Discount $\gamma$                        & $0.99$                     \\
Seed frames                              & $4000$                     \\
Exploration steps                        & $2000$                     \\
Exploration stddev. schedule             & linear$(1.0, 0.1, 1$e$6)$ \\
\hline
\end{tabular}
\end{table}

\begin{table}[h!]
\centering
\caption{Hyperparameters specific to state observations.}
\label{tab:hyperparams_states}
\begin{tabular}{cc}
\hline
\textbf{Hyperparameter}                       & \textbf{Setting}         \\
\hline
Feature dim.                             & Same as the state dim. of the task                     \\
Action repeat                            & $1$                      \\
Frame stack                              & N/A                      \\
\hline
\end{tabular}
\end{table}

\begin{table}[h!]
\centering
\caption{Hyperparameters specific to pixel observations.}
\label{tab:hyperparams_pixels}
\begin{tabular}{cc}
\hline
\textbf{Hyperparameter}                       & \textbf{Setting}         \\
\hline
Feature dim.                             & $50$                     \\
Action repeat                            & $2$                      \\
Frame stack                              & $3$                      \\
\hline
\end{tabular}
\end{table}

\subsection{Baseline Implementations}
\label{sec:baseline_impl}
All of the baselines are submitted in the supplemental material.
We use the official implementations of DBC, SAC-AE, and TD3. DeepMDP does not have a publicly available code, and we use the implementation available in the official DBC code-base. The official DDPG implementation is in TensorFlow, thus we used the implementation available in the official TD3 code-base with additional improvements detailed in Section \ref{sec:results_states}. Similarly, the official SAC implementation is in TensorFlow, thus we used the SAC implementation available in the official SAC-AE code-base. As discussed in Section \ref{sec:results_pixels}, we have run two versions of the baselines, with and without image augmentation. The image augmented variants, use the same image augmentation method of DrQ-v2 described in Appendix \ref{sec:pseudocode}.


\end{document}